\documentclass[pdflatex,sn-basic]{sn-jnl}

\usepackage{float}
\usepackage{graphicx}
\usepackage{multirow}
\usepackage{amsmath,amssymb,amsfonts}
\usepackage{amsthm}
\usepackage{mathrsfs}
\usepackage[title]{appendix}
\usepackage{xcolor}
\usepackage{textcomp}
\usepackage{manyfoot}
\usepackage{booktabs}
\usepackage{url}
\usepackage{algorithm}
\usepackage{algorithmicx}
\usepackage{algpseudocode}
\usepackage{listings}
\usepackage{mathrsfs}
\usepackage{dirtytalk}
\usepackage{tabularx}
\usepackage{longtable}
\usepackage{subcaption}
\usepackage{adjustbox}
\usepackage{tikz}
\usepackage{lmodern}
\usetikzlibrary{positioning,calc,arrows.meta}

\definecolor{forget}{RGB}{232, 146, 166}
\definecolor{retain}{RGB}{163, 140, 99}

\theoremstyle{thmstyleone}

\theoremstyle{thmstyletwo}

\theoremstyle{thmstylethree}

\definecolor{bargreen}{RGB}{82,188,100}
\definecolor{bargray}{RGB}{180,180,180}
\definecolor{threshgreen}{RGB}{34,139,34}
\definecolor{threshorange}{RGB}{230,160,30}

\begin{document}

\title[Probing Stylistic Appropriation using Large Language Models:  An Evaluation Framework for Copyright Infringement under EU Law]{Probing Stylistic Appropriation using Large Language Models: An Evaluation Framework for Copyright Infringement under EU Law}

\author[1,2]{\fnm{Noah} \sur{Scharrenberg}}\email{noah.scharrenberg@maastrichtuniversity.nl}

\author*[1]{\fnm{Chang} \sur{Sun}}\email{chang.sun@maastrichtuniversity.nl}

\affil*[1]{\orgdiv{Department of Advanced Computing Sciences}, \orgname{Maastricht University}, \orgaddress{\city{Maastricht}, \country{The Netherlands}}}

\affil[2]{\orgname{Contractuo}, \orgaddress{\city{Eindhoven}, \country{The Netherlands}}} 

\abstract{Large language models (LLM) trained on web-scale corpora generate output that may infringe copyright, yet existing technical safeguards focus narrowly on verbatim memorisation. EU copyright doctrine applies a broader standards: substantial similarity, which extends to stylistic choices, narrative structure, and creative elaboration. This mismatch between what current methods detect and what the law protects leaves a significant compliance gap. We introduce PSALM, an LLM-as-a-judge framework that operationalises EU copyright doctrine through ten evaluators assessing computational overlap, stylistic dimensions (writing style, narrative voice), content dimensions (character, plot, scene, world building), and statutory exceptions (parody, pastiche, quotation, scènes à faire). Applying PSALM to Llama~3.2 models fine-tuned on translated historical Dutch literary works, we find that: 1) instruction-tuned models exhibit non-trivial baseline stylistic similarity prior to corpus exposure; 2) fine-tuning induces systematic stylistic appropriation across all infringement-relevant dimensions, extending beyond verbatim memorisation to abstract narrative patterns; 3) Negative Preference Optimisation unlearning substantially reduces similarity but leaves detectable residual stylistic patterns. These findings indicate that safeguards targeting literal copying alone are insufficient to mitigate broader copyright risks. PSALM provides infrastructure for auditable, legally informed compliance evaluation, though the relationship between automated similarity scores and infringement determinations requires validation by legal experts. This work bridges qualitative legal standards and quantitative technical measurement, exposing fundamental tensions between generative AI and EU intellectual property law.}

\keywords{EU Copyright, LLM-as-a-Judge, Unlearning, LLM Evaluation, PSALM}

\maketitle

\section{Introduction}\label{sec:introduction}
The rapid advancement of Large Language Models (LLMs) has sparked a fundamental tension between technological capability and legal compliance. Recent demonstrations have shown that current LLMs can reproduce substantial portions of copyrighted works with minimal prompting~\citep{Cooper2025MemorizedCopyrightedWorks}, raising urgent questions about the suitability of existing copyright frameworks. However, current technical approaches to LLM copyright compliance focus predominantly on verbatim memorisation, literal copying of training data~\citep{carlini2021extracting, Cooper2025MemorizedCopyrightedWorks, ippolito2023preventing}, with evaluation benchmarks focusing on exact-match and high-overlap metrics~\citep{Maini2024TOFU, shi2024MUSE, wei2025coteval}, while overlooking more subtle forms of infringement that may be equally consequential under copyright law~\citep{chen2024copybench}.

European copyright law, particularly codified in the Copyright in the Digital Single Market (CDSM) Directive and the Artificial Intelligence (AI) Act, establishes copyright protection based on originality as \say{the author's own intellectual creation}~\citep{cdsm, aiact, CourtInfopaqDanske2009, CourtCofemelGstar2019}. Infringement determinations extend beyond literal copying to encompass substantial similarity in expressive elements, including narrative structure, character development, stylistic choices, and the selection and arrangement of creative elements~\citep{Lucchi2025GenAiCopyright}. Yet existing technical safeguards, including deduplication filters, similarity thresholds, and machine unlearning methods, are calibrated primarily to detect and suppress verbatim reproduction, leaving a critical gap in compliance verification for stylistic and structural appropriation.

This gap reflects a deeper epistemic misalignment between the legal and technical evaluation paradigms. Copyright law operates through qualitative, context-dependent standards adjudicated retrospectively by courts, weighing factors such as the overall impression, expressive overlap, and market substitution effects. In contrast, LLM development relies on quantitative, automated metrics (such as ROUGE-L, cosine similarity, and edit distance) that can be embedded into continuous integration pipelines but fail to capture the legally significant dimensions of substantial similarity~\citep{chen2024copybench, wei2025coteval}. Consequently, models may pass technical benchmarks while producing outputs that appropriate protected stylistic elements, plot structures, or character archetypes without triggering existing safeguards.

Existing LLM copyright-compliance mechanisms primarily address lawful data ingestion, transparency, and literal memorisation: provenance systems document training sources, unlearning methods suppress known memorised content, and disclosure regimes make dataset use more visible~\citep{bommarito2025kl3m, russinovich2025obliviate, wei2025coteval, warso2024transparency}. These mechanisms are necessary but insufficient because EU copyright infringement can turn on substantial similarity in protectable expression, not only on exact or near-exact textual reproduction~\citep{CourtInfopaqDanske2009, CourtCofemelGstar2019, Lucchi2025GenAiCopyright}. Consequently, a model may pass overlap-based memorisation tests while still producing outputs that closely track protected expressive choices such as narrative voice, characterisation, plot architecture, or the selection and arrangement of incidents~\citep{chen2024copybench, chun2024storysimilarity}.

The central challenge is therefore methodological: how can LLM outputs be evaluated at scale for legally relevant expressive similarity when the relevant similarities are qualitative, context-dependent, and often non-verbatim? Existing computational metrics, including n-gram overlap, edit distance, longest common substring, and embedding similarity, are useful for detecting literal or semantic correspondence but are poorly suited to assessing stylistic and structural appropriation~\citep{chen2024copybench, wei2025coteval}. Manual review by legal experts or trained annotators can capture these subtleties, but it is too slow and costly to support continuous evaluation of deployed LLM systems. This creates a compliance verification gap: copyright law requires assessment of substantial similarity in protectable expression, while current technical pipelines primarily measure surface-form overlap or semantic relatedness.

This paper addresses this deadlock by proposing \textbf{PSALM} (Probing Stylistic Appropriation by Language Models), an automated evaluation framework implementing EU copyright doctrine as measurable assessments in three steps. First, it identifies legally relevant dimensions where courts recognise protectable expression: narrative voice, character development, plot structure, world-building, and scene sequencing, derived from Court of Justice of the European Union (CJEU) jurisprudence establishing the originality standard~\citep{CourtInfopaqDanske2009, CourtCofemelGstar2019}, scholarly analysis of AI copyright issues~\citep{Lucchi2025GenAiCopyright, quintais2025genaicopyright, Borhi2025EUIPO}, and narratological theory characterising distinctive authorial choices~\citep{genette1980narrative, booth1961rhetoric, palmer2004fictional}. Second, it decomposes each dimension into evaluable sub-dimensions grounded in legal doctrine (e.g., narrative voice $\rightarrow$ point of view, narrative distance, focalisation patterns). Third, it employs LLM-as-a-judge evaluation to assess similarity along these sub-dimensions through structured prompts encoding legal tests, with hierarchical aggregation via directed acyclic graphs (DAG) synthesising evidence through weighted judgement nodes that tries to mirror how courts could potentially weigh multiple factors in infringement determinations. This approach enables scalable qualitative assessment: the dimensional structure could capture legally relevant expressive features, while LLM judges approximate the context-dependent reasoning human evaluators perform.

By bridging the gap between retrospective legal adjudication and proactive technical control, this work advances toward auditable, legally meaningful copyright compliance at machine scale. In particular, we conduct experiments to answer the following research questions:

\paragraph{Research questions.}
\begin{itemize}
    \item \textbf{RQ1:} To what extent do base language models exhibit baseline stylistic similarity to protected works across legally relevant dimensions as operationalised by PSALM?
    \item \textbf{RQ2:} Does supervised fine-tuning on a corpus of literary question--answer pairs increase PSALM-assessed stylistic appropriation beyond effects attributable to verbatim memorisation?
    \item \textbf{RQ3:} Does machine unlearning via Negative Preference Optimisation reduce PSALM-assessed stylistic appropriation across copyright-relevant dimensions, or does it primarily suppress literal recall?
\end{itemize}

Our findings have implications for model developers seeking to demonstrate due diligence, for policymakers crafting enforceable standards, and for rightsholders seeking reliable mechanisms to protect their works in the eras of generative AI.

\section{Related Work}\label{sec:related-work}
\subsection{Copyright Law and Large Language Models}
The EU legal framework, anchored by the CDSM Directive and the AI Act, establishes a civil-law approach grounded in defined exceptions---rather than the flexible \say{fair use} doctrine prevalent in common-law jurisdictions---including mandatory text-and-data mining (TDM) exceptions under Articles 3 and 4, permitting reproduction of lawfully accessible works for computational analysis subject to opt-out rights exercised \say{in an appropriate manner} by rights holders~\citep{cdsm,aiact}. However, as \citet{quintais2025genaicopyright} and \citet{Margoni2022TDM} observed, core terms such as \say{lawful access} remain ambiguous in the context of automated web scraping, creating compliance uncertainty for developers and enforcement challenges for rightsholders. \citet{scharrenberg2025} identified this as symptomatic of a deeper epistemic misalignment between the qualitative, context-dependent standards of the European copyright law and the quantitative, proactive metrics of LLM development.

Prior litigation has exposed the fragility of these frameworks. The Hamburg District Court's decision in \emph{Kneschke v. LAION} found that inclusion of a photographer's images in a large-scale dataset without explicit consent violated copyright protections, notwithstanding the dataset creators' invocation of TDM exceptions~\citep{CourtKneschkeLaion2024, Havlikova2025TechnicalLAION}; a concurrent EUIPO study confirmed that petabyte-scale ingestion renders individualised consent or verification practically infeasible~\citep{Borhi2025EUIPO}. The Munich Regional Court's preliminary judgement in \emph{GEMA v. OpenAI} adds a further dimension: the court held that memorisation of song lyrics within model weights constitutes an act of reproduction under Art. 2 of the InfoSoc Directive and \S~16~UrhG, regardless of whether the stored representation is explicit or encoded as probability parameters, and that such memorisation falls outside the TDM exception of Art. 4 CDSM because it goes beyond the preparatory acts of data ingestion and directly engages the rightsholder's exploitation rights~\citep{JustizBayern2025}. The court further attributed responsibility to the model operator rather than the end-user, on the basis that the operator controls the architecture and training data determining what the model reproduces. At the legislative level, the European Parliament's Committee on Legal Affairs recently adopted a report calling for mandatory transparency obligations, including itemised lists of copyrighted works used in training, and for EU copyright law to apply to all GPAI systems placed on the EU market irrespective of where training occurs, with non-compliance with transparency requirements potentially constituting infringement in itself~\citep{EPCopyrightGenAIOppAndChall}. Together, these developments suggest that compliance cannot be established from ingestion governance, transparency, or overlap-based memorisation checks alone; rather, they point to the need for evaluations that target protectable expressive choices assessed through substantial-similarity reasoning~\citep{scharrenberg2025}.

\subsection{Memorisation and Extraction in Language Models}
Memorisation in LLMs is well documented: \citet{carlini2021extracting} demonstrated that GPT-2 can be prompted to emit memorised sequences, particularly when training data contain repeated examples. Subsequent work has shown that memorisation correlates with data frequency, model capacity, and the presence of duplicates in training corpora~\citep{lee2022deduplicating, kandpal2022deduplicating}.

Recent research has extended these findings to current models and specialised domains. A recent study demonstrated that Meta's LlaMA 3.1 could reproduce entire chapters of Harry Potter nearly verbatim through carefully constructed prompts, revealing vulnerabilities even in models subjected to alignment training~\citep{Cooper2025MemorizedCopyrightedWorks}. Another study introduced CopyBench, a benchmark for measuring both literal and non-literal reproduction of copyrighted text, finding that models exhibit varying degrees of memorisation depending on prompt specificity and output length~\citep{chen2024copybench}.

However, existing memorisation studies focus predominantly on verbatim or near-verbatim extraction, with detection methods relying on n-gram overlap, longest common substring (LCS) metrics, or embedding-based similarity measures that capture literal correspondence but struggle to identify structural or stylistic parallels~\citep{ippolito2023preventing, shi2024MUSE}. This limitation is significant because copyright infringement does not require word-for-word copying, substantial similarity in expressive elements (such as narrative structure, character dynamics or stylistic choices) can constitute infringement even when no literal overlap exists~\citep{Lucchi2025GenAiCopyright}.

\subsection{Substantial Similarity and Derivative Works}
According to the EU copyright doctrine, infringement requires substantial similarity in protectable expressive elements---plot structure, character development, narrative voice, and the selection and arrangement of incidents---rather than verbatim correspondence alone~\citep{eucopyright, CourtInfopaqDanske2009, Lucchi2025GenAiCopyright, chun2024storysimilarity}.

AIStorySimilarity~\citep{chun2024storysimilarity}---the closest methodological antecedent to PSALM---decomposes narrative comparison into plot, character, and thematic dimensions using embedding-based semantic similarity, graph-structured event analysis, and transformer-based contextual matching. It achieves \(r = 0.72\) to \(r = 0.85\) with human annotations across plot and character dimensions, demonstrating that automated methods can approximate human judgement on structural narrative parallels. PSALM adopts a similar hierarchical decomposition strategy but diverges from AIStorySimilarity in three fundamental respects.

First, \textbf{legal grounding}: AIStorySimilarity measures narrative similarity as a descriptive construct relevant to human perception and information retrieval; PSALM instead operationalises legally-defined constructs grounded in EU copyright doctrine---originality as \say{author's own intellectual creation}, the idea-expression dichotomy, and substantial similarity tests---whose dimensions reflect protectability criteria established through CJEU jurisprudence. Second, \textbf{exception integration}: AIStorySimilarity measures \emph{how} similar works are without assessing whether similarity is actionable. PSALM integrates defence evaluators implementing parody (CJEU \emph{Deckmyn} test), pastiche (AG Emiliou criteria), quotation (Art.~5(3)(d) requirements), and scènes à faire doctrine, recognising that high similarity scores may be legally permissible under specific conditions~\citep{CourtDeckmyn2014, AGEmiliouPelham2024, eucopyright}. Third, \textbf{stylistic versus semantic focus}: AIStorySimilarity prioritises semantic content similarity---whether events, characters, and themes align in meaning. PSALM prioritises stylistic and structural similarity---whether expression choices and narrative techniques align---using LLM-as-judge evaluation to capture qualitative patterns resistant to embedding-based methods. Two texts may describe semantically identical events yet exhibit entirely different stylistic execution; the former would score as highly similar under AIStorySimilarity whilst PSALM would score them as dissimilar on writing style and narrative voice. This distinction is legally consequential: copyright protects expression rather than ideas.

\subsection{Machine Unlearning for LLMs}
Machine unlearning (MU) has emerged as a mechanism to remove or suppress specific knowledge from trained models, driven by copyright takedown obligations and GDPR's right to erasure~\citep{russinovich2025obliviate, wei2025coteval}. However, the effectiveness of unlearning in mitigating copyright risk depends on whether it addresses only literal memorisation or also disrupts abstract stylistic patterns that may constitute substantial similarity~\citep{scharrenberg2025}; LLMs' non-convex architectures preclude exact removal, necessitating approximate techniques~\citep{chien2024certified}.

Current approaches include gradient-based methods that reduce model likelihood on targeted sequences~\citep{Zhang2024npo, russinovich2025obliviate, jin2024rwku}, self-distillation variants that preserve general utility whilst forgetting specific content~\citep{dong2025undial, vasilev2025unilogit}, and inference-time controls applying learned refusal policies or logit adjustments without parameter modification~\citep{bhaila2025soft, ji2024reversing}. \citet{wei2025coteval} introduced a comprehensive evaluation suite, assessing both forget quality and model utility retention.

However, existing unlearning benchmarks concentrate on verbatim memorisation, evaluating success primarily through exact-match or high-overlap metrics~\citep{Maini2024TOFU, shi2024MUSE}. This focus leaves two critical gaps unaddressed. First, unlearning methods remain vulnerable to paraphrase extraction and adversarial prompting strategies that elicit memorized content through indirect means~\citep{wei2025coteval}. Second, current evaluations do not assess whether unlearning mitigates stylistic appropriation. Compounding both gaps, \citet{erasureillusionJia2025} showed that standard unlearning metrics systematically overestimate forgetting success: evaluations on the explicit forget set may certify unlearning whilst the model retains reproductive capacity over semantically derived content (e.g., fan fiction based on a copyrighted sources work), revealing a fundamental mismatch between benchmark coverage and the knowledge-level erasure copyright expectations demand.

\subsection{Automated Evaluation of LLM Outputs}
Standard reference-based metrics (BLEU, ROUGE, METEOR) and reference-free alternatives correlate poorly with human judgements for creative or stylistically nuanced generation~\citep{novikova2017we, sai2022survey, gehrmann2021gem}. Recent work has demonstrated that LLMs themselves can serve as effective evaluators for complex, subjective properties: \citet{zheng2023judging} showed that GPT-4 achieves agreement with human annotators exceeding \(80\%\) on multi-dimensional response evaluation, and carefully designed prompts can elicit reliable comparative judgements even for nuanced attributes such as helpfulness and harmlessness~\citep{chiang2023can}.

LLM-as-judge methodologies show particular promise in legal domains, where evaluation requires contextual interpretation of rule-based frameworks and qualitative assessment of multi-factor tests. \citet{cui2023chatlaw} found that GPT-3.5 and GPT-4 can identify logical fallacies, assess argument strength, and detect missing premises with moderate reliability (\(F_1 = 0.68\) to \(0.74\)) when supplied with structured criteria grounded in legal doctrine, and that explicit definitional grounding---providing formal definitions of concepts such as burden of proof and material fact---significantly improves classification consistency. Relatedly, \citet{katz2024gpt4} demonstrated that GPT-4 scores in the \(90^{\text{th}}\) percentile on the Uniform Bar Examination, confirming capability for doctrinal rule application and legal argumentation of the kind required for copyright assessment.

These studies establish that LLM-as-judge reliability depends on three design features: (1)\textbf{explicit doctrinal grounding} through prompts encoding legal definitions and multi-factor tests; (2)\textbf{hierarchical decomposition} of complex legal determinations into evaluable sub-questions; (3)\textbf{structured output formats} constraining models to categorical verdicts with justifications\citep{cui2023chatlaw, katz2024gpt4}. PSALM incorporates all three, extending prior binary or scalar legal LLM applications to coordinated multi-dimensional assessment across thirteen heterogeneous metrics.

\subsection{Research Gap}
This literature reveals three research gaps. First, evaluation frameworks remain anchored to verbatim similarity metrics that poorly approximate the totality-of-expressive-overlap tests applied in legal adjudication. Second, unlearning methods are validated for literal forgetting but not for the suppression of stylistic and structural appropriation; the "erasure illusion" problem compounds this by systematically overstating forgetting success. Third, automated evaluation lacks the doctrinal grounding required for copyright compliance assessment, including recognition of transformative uses and statutory exceptions.

This work addresses these gaps through a legally informed evaluation framework assessing stylistic copyright infringement across the LLM lifecycle, including base models, fine-tuning, and unlearning, implementing EU substantial similarity tests and integrating recognised defences, while acknowledging that automated scores inform rather than replace human legal judgement~\citep{scharrenberg2025}.

\section{Methods}\label{sec:methods}
PSALM (Probing Stylistic Appropriation by Language Models) operationalises elements of EU copyright doctrine as measurable assessments over source–target text pairs. Each evaluator is implemented as a directed acyclic graph (DAG) of textual analysis tasks adjudicated by an LLM-as-judge, producing scores along two categories of legally relevant dimensions: infringement-oriented (Section~\ref{sec:infringement-evaluators}), measuring similarity across protectable expressive features, and defence-oriented (Section~\ref{sec:defensive-evaluators}), assessing whether observed similarities qualify for statutory exceptions or fall within unprotectable elements.

\subsection{Directed Acyclic Graph-based LLM-as-a-Judge Architecture}
Each evaluator's DAG comprises three node types: task nodes, a single judgement node, and verdict nodes (visualised in Figures~\ref{fig:dag_writing_style}--\ref{fig:dag_scenes_a_faire}).

Task nodes correspond to sub-dimensions of a legal construct. A task node takes as input the source text $x_{s}$, $x_{t}$, and an instruction that specifies which aspects of the texts to compare and how to classify the comparison using a fixed five-level scale. The instruction is expressed in natural language and includes three elements: a definition of the sub-dimension, a list of concrete factors to examine in each text, and a request to assign the sub-dimension to one of five verbal categories. For example, the \say{Lexical Complexity Analysis} task in the writing style evaluator (Section~\ref{sec:infringement:writing-style}) asks the judge model to analyse vocabulary richness, word length patterns, and formality in both texts, to ignore semantic content, and then to classify the similarity as \say{identical or near-identical writing style across all dimensions}, \say{very similar writing style with only minor differences}, \say{moderately similar writing style with some notable differences}, \say{somewhat different writing style with limited similarities}, or \say{clearly different or opposite writing styles}.

The judgement node aggregates the sub-dimensions. Its instruction restates the overall construct (for example: \say{overall writing style similarity between $x_s$ and $x_t$}), lists the outputs of the relevant task nodes by name, and specifies explicit weights for each sub-dimension as percentages. The weights aim to encode which aspects are treated as more salient in legal (and narratological) terms. The judge model is instructed to take these weights into account and to select a single verdict, from a separate set of five verdict nodes, together with a justification that explains how the sub-dimension analyses support the chosen level.

For infringement evaluators, the five verdicts correspond to scores $\{0, 3, 5, 8, 10\}$, mapped to the normalised set $\{0.0, 0.3, 0.5, 0.8, 1.0\}$. These represent, respectively, \say{clearly different or opposite}, \say{somewhat different with limited similarities}, \say{moderately similar with some notable differences}, \say{very similar with only minor differences}, and \say{identical or near-identical}. For defence evaluators, the same numerical scale is reused, but verdict labels reflect defence strength (of genericness) rather than similarity, for example \say{strong parody ($10$)} through to \say{no parody ($0$)}.

The use of a coarse, ordered five-band rubric is motivated by prior work showing that discretisations yield more consistent LLM judgements than fine-grained numerical scoring~\citep{zheng2023judging, chiang2023can}. Within this ordinal rubric~\citep{stevens1946scales, norman2010likert}, we chose the particular spacing $\{0, 3, 5, 8, 10\}$ to encode a monotone but intentionally non-linear notion of risk contribution for aggregation: shifting into (or out of) the extreme bands is treated as more consequential than moving within the middle bands. Concretely, the endpoints ($0$ and $10$) anchor the lowest- and highest-risk categories, while the intermediate values ($3,5,8$) compress the middle region and expand the tails, so that aggregated scores respond more strongly when judgements cross into near-identical or clearly-different ranges. This is a heuristic design for comparing and aggregating LLM-judge outputs across multiple tasks and evaluators; it is not a claim that rubric categories are separated by equal or empirically calibrated quantitative distances.

Formally, each evaluator $E$ implements a function:
\begin{equation}
    E(x_s, x_t) = s \in \{0.0, 0.3, 0.5, 0.8, 1.0\}
\end{equation}\label{eq:evaluator}
where $x_s$ and $x_t$ are arbitrary source and target texts, respectively. Evaluation proceeds by presenting each task node's instruction and the two texts to the judge model, obtaining textual analyses and local categorical labels, and then providing these task outputs to the judgement node, which selects one of the five verdicts. The final numerical score is the normalised value associated with that verdict.
Sub-dimension labels remain part of the decision-making trace but only the final evaluator-level score is used as a quantitative output. Any sufficiently capable instruction-following LLM may serve as the judge model, though performance varies by model.

\subsection{Foundational Legal Principles}\label{sec:foundational-legal}
Evaluator design draws on two bodies of work: EU copyright law and doctrinal commentary---the CJEU originality standard~\citep{CourtInfopaqDanske2009, CourtCofemelGstar2019, rosati2014originality}, InfoSoc and CDSM exceptions~\citep{eucopyright, cdsm}, and generative AI copyright analyses~\citep{Lucchi2025GenAiCopyright, quintais2025genaicopyright, Borhi2025EUIPO}---and narratology, stylistics, and authorship attribution, which supply operational definitions of style, voice, character, plot, scene, and world-building~\citep{stamatatos2009survey, Juola2008, genette1980narrative, booth1961rhetoric, palmer2004fictional, Herrmann2015, abbott2008cambridge, ryan2007toward, bordwell2012film, mckee1997story, wolf2012building, ekman2013here, gavins2007text, chatman1990coming, herman2012routledge}.

Infringement evaluators target expressive dimensions that CJEU originality doctrine treats as protectable through the author's \say{free and creative choices}\citep{CourtInfopaqDanske2009, CourtCofemelGstar2019}---narrative voice, characterisation, plot architecture, and stylistic elaboration\citep{Lucchi2025GenAiCopyright, Borhi2025EUIPO, Herrmann2015}---yielding six evaluators decomposed into narratologically grounded sub-dimensions.

Defence evaluators implement four recognised exceptions: parody/satire (\emph{Deckmyn}\citep{CourtDeckmyn2014, rendas2024pastiche}), pastiche (Art. 5(3)(k) InfoSoc; AG Emiliou in \emph{Pelham II}\citep{eucopyright, AGEmiliouPelham2024}), quotation/citation (Art. 5(3)(d); \emph{Funke Medien}\citep{eucopyright, cjeu2019funke}), and scènes à faire (idea-expression dichotomy; stock element exclusion\citep{nichols1930universal, warner2008rdr, hoehling1980universal, sheldon1936metro, walker1986time, geiger2015three}).

These sources are translated into evaluators, sub-dimensions, and weights by combining them with narratological and stylistic categories. Genette’s taxonomy of focalisation and narrative person, for instance, informs the focalisation and point-of-view sub-dimensions of narrative voice \citep{genette1980narrative}, while stylometric work on function words, sentence-length distribution, and syntactic patterns underpins the lexical and sentence-structure sub-dimensions of writing style \citep{stamatatos2009survey, Juola2008, hoover1973prose}. Narrative semiotics and film theory support the segmentation into scenes and beats \citep{bordwell2012film, dancyger2010technique, mckee1997story}. The general procedure for selecting sub-dimensions and assigning weights is set out in Section~\ref{sec:legal-foundation:dimension-selection}.

\subsubsection{Dimension Selection and Weight Determination}\label{sec:legal-foundation:dimension-selection}
The selection of evaluators and their sub-dimensions follows a two-stage process. First, for each legally relevant construct (such as substantial similarity of literary expression, parody, or quotation), we identified the expressive phenomena that EU copyright doctrine plausibly treats as protectable or as components of an exception. This step draws on case law and doctrinal commentary to determine which aspects of a text might be relevant for infringement or for the application of a limitation or exception. For example, CJEU decisions and subsequent scholarship emphasise narrative voice, characterisation, plot architecture, and stylistic elaboration as indicators of originality~\citep{CourtInfopaqDanske2009, CourtCofemelGstar2019, Lucchi2025GenAiCopyright, Borhi2025EUIPO}, and identify evocation, noticeable difference, and humorous or critical character as the core elements of parody under EU law~\citep{CourtDeckmyn2014, geiger2015three}.

Secondly, for each such legal construct, we map these phenomena onto operationally definable sub-dimensions that can be assessed from the text alone. This mapping uses narratology, stylistics, and authorship-attribution research to obtain theoretically grounded categories such as point of view, focalisation, sentence-length distribution, or conflict escalation patterns~\citep{genette1980narrative, Juola2008, stamatatos2009survey, abbott2008cambridge, bordwell2012film}. A sub-dimension is included only if it meets three criteria: it must correspond to a feature that doctrine or commentary treats as relevant to protection or exception; it must admit a clear textual definition that can be expressed in natural-language instructions; and it must be feasible for human or LLM judges to evaluate on the basis of limited excerpts without external information.

The weights attached to sub-dimensions are chosen heuristically. They encode a normative judgement, grounded in doctrine and theory, about the relative contribution of each sub-dimension to the overall legal construct being approximated and about the expected robustness of measurement. Dimensions that are repeatedly emphasised in case law or doctrinal analysis, and that are relatively well-defined in narratology or stylistics, receive higher weights. For example, point of view and narrative distance are weighted more heavily than reader engagement in the narrative voice evaluator (Section~\ref{sec:infringement:narrative-voice}), because they are both central to narratological accounts of voice and plausibly more indicative of originality~\citep{genette1980narrative, booth1961rhetoric}. Conversely, dimensions that primarily capture functional roles or genre-typical elements, such as generic scene functions or character roles, receive low weights, because they are closer to unprotectable scènes à faire elements~\citep{nichols1930universal}.

\subsection{Infringement Evaluators}\label{sec:infringement-evaluators}
The infringement evaluators measure similarity along dimensions corresponding to protectable expressive features; DAG node colours follow the convention established in Section~\ref{sec:methods} (blue: task nodes; red: judgement node; green: verdict nodes).

\subsubsection{Writing Style Similarity}\label{sec:infringement:writing-style}
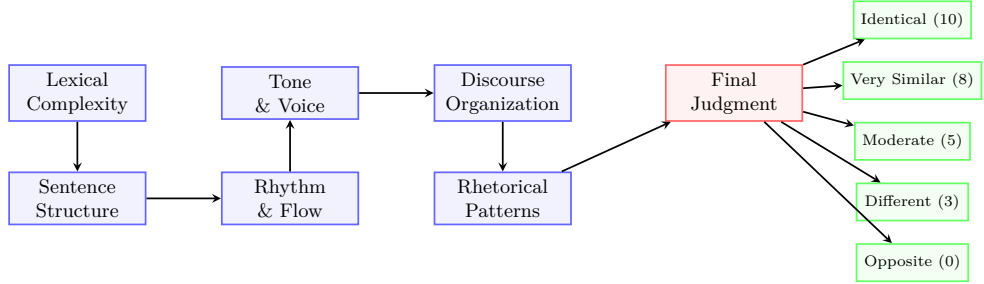
\begin{figure}[H]
\centering
\adjustbox{max width=\textwidth}{
    \begin{tikzpicture}[
        node distance=0.8cm and 1.2cm,
        task/.style={rectangle, draw=blue!60, fill=blue!5, thick, minimum width=2.2cm, minimum height=0.7cm, align=center, font=\small},
        judge/.style={rectangle, draw=red!60, fill=red!5, thick, minimum width=2.2cm, minimum height=0.7cm, align=center, font=\small},
        verdict/.style={rectangle, draw=green!60, fill=green!5, thick, minimum width=1.8cm, minimum height=0.6cm, align=center, font=\footnotesize},
        arrow/.style={->, >=stealth, thick}
    ]
    
    \node[task] (node_A) {Lexical\\Complexity};
    \node[task, right=of node_A] (node_D) {Tone\\{\&} Voice};
    \node[task, right=of node_D] (node_E) {Discourse\\Organization};
    
    \node[task, below=of node_A] (node_B) {Sentence\\Structure};
    \node[task, right=of node_B] (node_C) {Rhythm\\{\&} Flow};
    \node[task, right=of node_C] (node_F) {Rhetorical\\Patterns};
    
    \node[judge, right=1.5cm of node_E] (judge) {Final\\Judgment};
    
    \node[verdict, above right=0.4cm and 0.8cm of judge] (v1) {Identical (10)};
    \node[verdict, below=0.35cm of v1] (v2) {Very Similar (8)};
    \node[verdict, below=0.35cm of v2] (v3) {Moderate (5)};
    \node[verdict, below=0.35cm of v3] (v4) {Different (3)};
    \node[verdict, below=0.35cm of v4] (v5) {Opposite (0)};
    
    \draw[arrow] (node_A) -- (node_B);
    \draw[arrow] (node_B) -- (node_C);
    \draw[arrow] (node_C) -- (node_D);
    \draw[arrow] (node_D) -- (node_E);
    \draw[arrow] (node_E) -- (node_F);
    
    \draw[arrow] (node_F) -- (judge);
    
    \draw[arrow] (judge) -- (v1);
    \draw[arrow] (judge) -- (v2);
    \draw[arrow] (judge) -- (v3);
    \draw[arrow] (judge) -- (v4);
    \draw[arrow] (judge) -- (v5);
    
    \end{tikzpicture}
}
\caption{DAG architecture for the writing style evaluator that shows in which sub-dimension analyses feed into a final judgement of overall writing style similarity}
\label{fig:dag_writing_style}
\end{figure}

Figure~\ref{fig:dag_writing_style} shows the writing style evaluator DAG, comprising six task nodes (Lexical Complexity, Sentence Structure, Rhythm \& Flow, Rhetorical Patterns, Discourse Organisation, Tone \& Voice) feeding a single judgement node.

Doctrinally, writing style is relevant because originality in literary work may manifest in the author's distinctive way of expressing ideas, not in the ideas themselves~\citep{CourtCofemelGstar2019, Herrmann2015}. Stylometry and authorship attribution provide empirical support for treating lexical choice and syntactic patterns as authorial signatures~\citep{stamatatos2009survey, Juola2008, hoover1973prose}.

The lexical complexity task instructs the judge model to examine vocabulary richness (patterns of repetition and variety), word length tendencies, and formality markers in each text, ignoring semantic content.

The sentence structure task similarly focuses on sentence length distribution and syntactic complexity. Informed by work showing that authors exhibit stable preferences in sentence structure~\citep{hoover1973prose}, the instruction asks the model to compare whether both texts favour, for example, short simple sentences or long multi-clause constructions.

The rhythm \& flow task examines punctuation patterns, sentence rhythm, and pacing, drawing on analyses of prose rhythm~\citep{hoover1973prose} and discourse markers~\citep{biber1988discourse}. The model is asked to report whether both texts exhibit similar punctuation density and rhythmic patterns and then to map this to one of the five labels.

The rhetorical patterns task considers distributions of questions, imperatives, repetition, and parallelism, features that stylistic research has linked to authorial habits~\citep{biber1988discourse, Herrmann2015}.

The discourse organisation task targets paragraph structure and use of connectives, grounded in work on discourse cohesion and information sequencing~\citep{biber1988discourse, gehrmann2021gem}. The model compares, for instance, whether both texts tend to build long multi-sentence paragraphs with explicit discourse markers or short stand-alone paragraphs.

The tone \& voice task captures authorial stance, drawing on rhetorical narratology~\citep{booth1961rhetoric, Herrmann2015}. The model assesses the degree of personal versus impersonal voice and assertiveness versus heding, then classifies similarity. Tone is included even though it may be more variable than syntax, because legal and literary discussions often refer to an author's \say{voice} as part of style.

The judgement node for writing style lists these six analyses and assigns weights: lexical complexity and sentence structure each receive $20-25\%$ of the total weight, rhythm, rhetorical patterns, and discourse organisation each $15-20\%$, and tone around $5\%$. These weights are chosen heuristically, but are grounded in two considerations. First, stylometric evidence shows that lexical and syntactic features are highly discriminative for authorship~\citep{stamatatos2009survey, Juola2008}, whereas tone is more topic-dependent and less stable. Secondly, from a legal perspective, concrete formal features (wording and construction) are closer to the protected \say{form of expression} than high-level mood, which may overlap with genre conventions.

\subsubsection{Narrative Voice Similarity}\label{sec:infringement:narrative-voice}
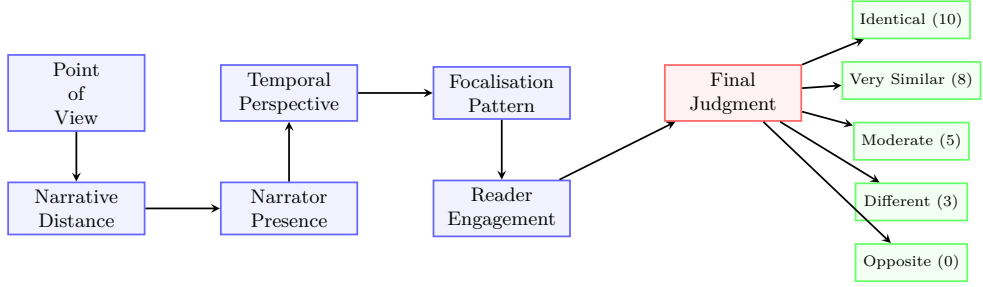
\begin{figure}[H]
\centering
\adjustbox{max width=\textwidth}{
\begin{tikzpicture}[
    node distance=0.8cm and 1.2cm,
    task/.style={rectangle, draw=blue!60, fill=blue!5, thick, minimum width=2.2cm, minimum height=0.7cm, align=center, font=\small},
    judge/.style={rectangle, draw=red!60, fill=red!5, thick, minimum width=2.2cm, minimum height=0.7cm, align=center, font=\small},
    verdict/.style={rectangle, draw=green!60, fill=green!5, thick, minimum width=1.8cm, minimum height=0.6cm, align=center, font=\footnotesize},
    arrow/.style={->, >=stealth, thick}
]

\node[task] (node_A) {Point\\of\\View};
\node[task, below=of node_A] (node_B) {Narrative\\Distance};
\node[task, right=of node_B] (node_C) {Narrator\\Presence};
\node[task, right=of node_A] (node_D) {Temporal\\Perspective};
\node[task, right=of node_D] (node_E) {Focalisation\\Pattern};
\node[task, right=of node_C] (node_F) {Reader\\Engagement};

\node[judge, right=1.5cm of node_E] (judge) {Final\\Judgment};

\node[verdict, above right=0.4cm and 0.8cm of judge] (v1) {Identical (10)};
\node[verdict, below=0.35cm of v1] (v2) {Very Similar (8)};
\node[verdict, below=0.35cm of v2] (v3) {Moderate (5)};
\node[verdict, below=0.35cm of v3] (v4) {Different (3)};
\node[verdict, below=0.35cm of v4] (v5) {Opposite (0)};

\draw[arrow] (node_A) -- (node_B);
\draw[arrow] (node_B) -- (node_C);
\draw[arrow] (node_C) -- (node_D);
\draw[arrow] (node_D) -- (node_E);
\draw[arrow] (node_E) -- (node_F);

\draw[arrow] (node_F) -- (judge);

\draw[arrow] (judge) -- (v1);
\draw[arrow] (judge) -- (v2);
\draw[arrow] (judge) -- (v3);
\draw[arrow] (judge) -- (v4);
\draw[arrow] (judge) -- (v5);

\end{tikzpicture}
}
\caption{DAG architecture for the Narrative Voice evaluator that shows how voice-related sub dimensions are aggregated into an overall narrative voice similarity judgement}
\label{fig:dag_narrative_voice}
\end{figure}

The narrative voice evaluator (Figure~\ref{fig:dag_narrative_voice}) targets perspective-related originality: narrative voice shapes how events and characters are presented and is central to both narratology~\citep{genette1980narrative, booth1961rhetoric, palmer2004fictional} and CJEU originality analysis~\citep{herman2012routledge}.

The point-of-view task, grounded in Genette's classification of narrative person and knowledge scope~\citep{genette1980narrative}, instructs the model to identify, for each text, whether narration is in the first, second, or third person; whether the narrator's knowledge is limited, multiple, omniscient, or objective; and whether this configuration is consistent. This produces a local point-of-view score that reflects whether, for example, both texts use a single limited first-person narrator, or whether they differ fundamentally (such as first-person versus omniscient third-person).

The narrative distance task draws on Booth's and Phelan's work on distance~\citep{booth1961rhetoric, phelan2005living}. The instruction asks the model to assess how close the narrator is to character consciousness, how emotionally involved the narrator is, and whether there is a shift in distance.

The narrator presence task distinguishes homodiegetic and heterodiegetic narrators and degrees of intrusiveness~\citep{stanzel1984theory, fludernik1996towards}. The model compares whether the narrator participates in the events, how often they comment directly, and whether they seem inside or outside the story world.

The temporal perspective task asks the model to identify primary verb tenses and the narrator's temporal position (retrospective, simultaneous, anticipatory), following narratological treatment of time~\citep{genette1980narrative}.

The focalisation pattern task uses Genette's notion of focalisation~\citep{genette1980narrative}. The model is instructed to determine whose perspective filters the information (fixed internal, variable internal, external mixed), what information boundaries exist, and whether the two texts display similar focalisation schemes.

The reader engagement task, informed by discourse studies on audience address~\citep{biber1988discourse}, asks whether and how the narrator directly addresses the reader, how often, and with what assumed relationship (intimate, neutral, formal).

In the judgement node, the point of view and narrative distance are assigned the highest weights (together around $45\%$, narrator presence and focalisation intermediate weights (around $15\%$ each), temporal perspective a modest weight, and reader engagement the smallest (around $5\%$). This ordering reflects the centrality of point of view and distance in narratological accounts of voice~\citep{genette1980narrative, booth1961rhetoric}, and the fact that reader address is often more variable and can be governed by genre or discourse tradition (for example, children's literature may more frequently use direct address) rather than authorial individuality.

\subsubsection{Character Similarity}\label{sec:infringement:character}
\begin{figure}[H]
\centering
\adjustbox{max width=\textwidth}{
\begin{tikzpicture}[
    node distance=0.8cm and 1.2cm,
    task/.style={rectangle, draw=blue!60, fill=blue!5, thick, minimum width=2.2cm, minimum height=0.7cm, align=center, font=\small},
    judge/.style={rectangle, draw=red!60, fill=red!5, thick, minimum width=2.2cm, minimum height=0.7cm, align=center, font=\small},
    verdict/.style={rectangle, draw=green!60, fill=green!5, thick, minimum width=1.8cm, minimum height=0.6cm, align=center, font=\footnotesize},
    arrow/.style={->, >=stealth, thick}
]

\node[task] (node_A) {Character\\Identity\\and\\Traits};
\node[task, below=of node_A] (node_B) {Character\\Arc\\and\\Development};
\node[task, right=of node_B] (node_C) {Character\\Relationships\\and\\Dynamics};
\node[task, right=of node_A] (node_D) {Character\\Background\\and\\Motivation};
\node[task, right=of node_D] (node_E) {Character\\Expression\\and\\Behaviour};
\node[task, right=of node_C] (node_F) {Character\\Function\\and\\Role};

\node[judge, right=1.5cm of node_E] (judge) {Final\\Judgment};

\node[verdict, above right=0.4cm and 0.8cm of judge] (v1) {Identical (10)};
\node[verdict, below=0.35cm of v1] (v2) {Very Similar (8)};
\node[verdict, below=0.35cm of v2] (v3) {Moderate (5)};
\node[verdict, below=0.35cm of v3] (v4) {Different (3)};
\node[verdict, below=0.35cm of v4] (v5) {Opposite (0)};

\draw[arrow] (node_A) -- (node_B);
\draw[arrow] (node_B) -- (node_C);
\draw[arrow] (node_C) -- (node_D);
\draw[arrow] (node_D) -- (node_E);
\draw[arrow] (node_E) -- (node_F);

\draw[arrow] (node_F) -- (judge);

\draw[arrow] (judge) -- (v1);
\draw[arrow] (judge) -- (v2);
\draw[arrow] (judge) -- (v3);
\draw[arrow] (judge) -- (v4);
\draw[arrow] (judge) -- (v5);

\end{tikzpicture}
}
\caption{DAG architecture for the Character Similarity evaluator showing the aggregating character-related sub-dimension analyses into an overall character similarity judgement}
\label{fig:dag_character_similarity}
\end{figure}
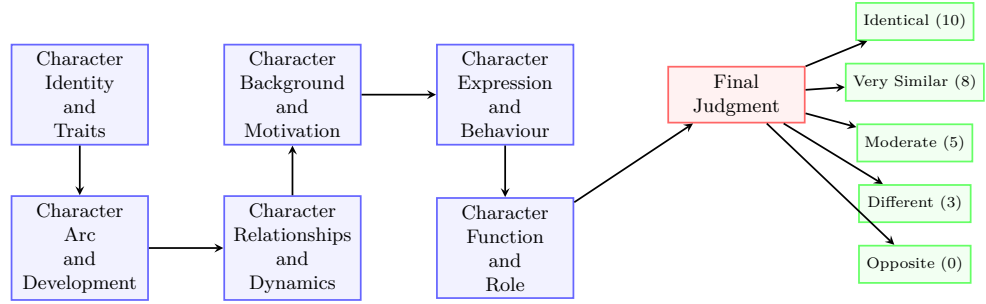

Figure~\ref{fig:dag_character_similarity} illustrates the character similarity evaluator. It reflects legal and narrative recognition that distinctive characters can be protectable elements of work, providing they are sufficiently defined and creative~\citep{nichols1930universal, warner2008rdr, kurtz2015practical, palmer2004fictional}. Each sub-dimension is designed to distinguish generic archetypes from specific creative elaborations.

The character identity \& traits task asks for a comparison of distinctive personality traits, physical idiosyncrasies, mannerisms, and psychological complexity in the two texts. It draws on literary theory emphasising that characters becomes original through detailed, specific embodiment rather than generic labels~\citep{palmer2004fictional, Herrmann2015}.

The character arc \& development task compares the sequence and nature of character change and internal conflict, building on narrative accounts of arcs~\citep{mckee1997story, abbott2008cambridge}. The model identifies initial states, triggers, stages, and resolutions.

The character relationship \& dynamics task considers interaction patterns, power balances, and emotional textures, in line with work on social cognition in narrative~\citep{palmer2004fictional, wolf2012building}.

The character background \& motivation task distinguishes generic backstory (e.g. \say{traumatic past}) from specific causal histories with detailed events and motivational structures. Motivational architecture is singled out in both legal and literary treatments as a place where originality can reside~\citep{palmer2004fictional, Lucchi2025GenAiCopyright}.

The character expression \& behaviour task looks at behavioural signatures and emotional response patterns. It captures whether both characters exhibit similar specific habits and copying mechanisms.

The character function \& role task compares high-level narrative functions such as protagonist, antagonist or mentor. The instruction explicitly frames these as often generic and weakly protectable, consistent with scènes à faire reasoning~\citep{nichols1930universal}.

The judgement node weights identity/traits, arc/development, and relationships highest, background/motivation and expression/behaviour moderately, and function/role minimally. This reflects both doctrinal and doctrinally-informed commentary: courts and commentators stress that appropriation of a character's distinctive, detailed features is more problematic than sharing generic roles or archetypes~\citep{warner2008rdr, kurtz2015practical}.

\subsubsection{Plot Structure Similarity}\label{sec:infringement:plot-structure}
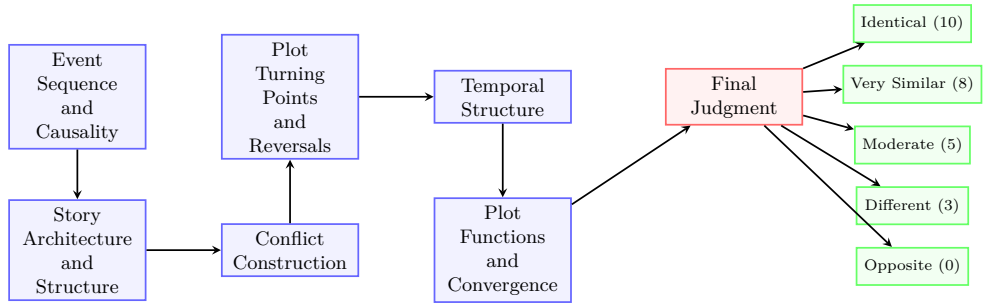
\begin{figure}[H]
\centering
\adjustbox{max width=\textwidth}{
\begin{tikzpicture}[
    node distance=0.8cm and 1.2cm,
    task/.style={rectangle, draw=blue!60, fill=blue!5, thick, minimum width=2.2cm, minimum height=0.7cm, align=center, font=\small},
    judge/.style={rectangle, draw=red!60, fill=red!5, thick, minimum width=2.2cm, minimum height=0.7cm, align=center, font=\small},
    verdict/.style={rectangle, draw=green!60, fill=green!5, thick, minimum width=1.8cm, minimum height=0.6cm, align=center, font=\footnotesize},
    arrow/.style={->, >=stealth, thick}
]

\node[task] (node_A) {Event\\Sequence\\and\\Causality};
\node[task, below=of node_A] (node_B) {Story\\Architecture\\and\\Structure};
\node[task, right=of node_B] (node_C) {Conflict\\Construction};
\node[task, right=of node_A] (node_D) {Plot\\Turning\\Points\\and\\Reversals};
\node[task, right=of node_D] (node_E) {Temporal\\Structure};
\node[task, right=of node_C] (node_F) {Plot\\Functions\\and\\Convergence};

\node[judge, right=1.5cm of node_E] (judge) {Final\\Judgment};

\node[verdict, above right=0.4cm and 0.8cm of judge] (v1) {Identical (10)};
\node[verdict, below=0.35cm of v1] (v2) {Very Similar (8)};
\node[verdict, below=0.35cm of v2] (v3) {Moderate (5)};
\node[verdict, below=0.35cm of v3] (v4) {Different (3)};
\node[verdict, below=0.35cm of v4] (v5) {Opposite (0)};

\draw[arrow] (node_A) -- (node_B);
\draw[arrow] (node_B) -- (node_C);
\draw[arrow] (node_C) -- (node_D);
\draw[arrow] (node_D) -- (node_E);
\draw[arrow] (node_E) -- (node_F);

\draw[arrow] (node_F) -- (judge);

\draw[arrow] (judge) -- (v1);
\draw[arrow] (judge) -- (v2);
\draw[arrow] (judge) -- (v3);
\draw[arrow] (judge) -- (v4);
\draw[arrow] (judge) -- (v5);

\end{tikzpicture}
}
\caption{DAG architecture for the Plot Structure Similarity evaluator combining event conflict and structural sub-dimensions into a final plot structure similarity judgement}
\label{fig:dag_plot_structure_similarity}
\end{figure}

Figure~\ref{fig:dag_plot_structure_similarity} illustrates the plot structure similarity evaluator. Plot structure is a central narrative construct and a frequent focal point in infringement disputes~\citep{nichols1930universal, hoehling1980universal, sheldon1936metro, kurtz2015practical}.

The event sequence \& causality task instructs the model to identify and compare specific plot events and their causal links, distinguishing generic beats from detailed sequences and mechanisms.

The story architecture \& structure task addresses the higher-level organisation: act division, framing devices, nested narratives, and interwoven subplots~\citep{bordwell2012film, dancyger2010technique, mckee1997story}. The model is asked to examine whether the two texts adopt similar structural architectures.

The conflict construction task deals with escalation patterns and obstacle networks, grounded in dramaturgical notions of rising action and complication~\citep{mckee1997story}.

The plot turning points \& reversals task looks at pivotal shifts and revelations. It reflects doctrinal concerns where plaintiffs allege copying of unique twists rather than generic \say{surprises}~\citep{hoehling1980universal, kurtz2015practical}.

The temporal structure task examines chronology, flashbacks, and pacing, building on narratological work on time~\citep{genette1980narrative}.

The plot functions \& convergence task compares how threads converge and resolve. It captures whether particular combinations of subplots and resolutions are mirrored. As with character functions, generic high-level functions are less protectable, so this dimension receives the smallest weight in the judgement node.

The judgement node weights event sequence and story architecture most heavily, conflict and turning points moderately, temporal structure lower, and plot functions-convergence lowest. Nichols and subsequence cases illustrate courts' focus on the sequence and organisation of events and on key plot twists~\citep{nichols1930universal, kurtz2015practical}. Event sequence and architecture are weighted highest, reflecting courts' focus on specific causal chains and structural copying over generic narrative functions~\citep{nichols1930universal, kurtz2015practical}.

\subsubsection{Scene Sequence Similarity}\label{sec:infringement:scene-sequence}
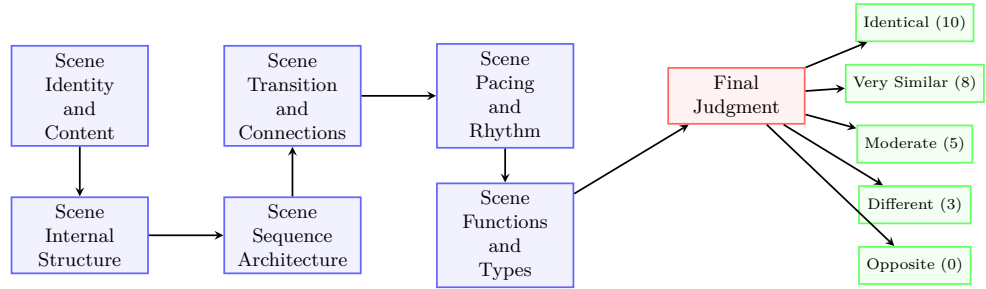
\begin{figure}[H]
\centering
\adjustbox{max width=\textwidth}{
\begin{tikzpicture}[
    node distance=0.8cm and 1.2cm,
    task/.style={rectangle, draw=blue!60, fill=blue!5, thick, minimum width=2.2cm, minimum height=0.7cm, align=center, font=\small},
    judge/.style={rectangle, draw=red!60, fill=red!5, thick, minimum width=2.2cm, minimum height=0.7cm, align=center, font=\small},
    verdict/.style={rectangle, draw=green!60, fill=green!5, thick, minimum width=1.8cm, minimum height=0.6cm, align=center, font=\footnotesize},
    arrow/.style={->, >=stealth, thick}
]

\node[task] (node_A) {Scene\\Identity\\and\\Content};
\node[task, below=of node_A] (node_B) {Scene\\Internal\\Structure};
\node[task, right=of node_B] (node_C) {Scene\\Sequence\\Architecture};
\node[task, right=of node_A] (node_D) {Scene\\Transition\\and\\Connections};
\node[task, right=of node_D] (node_E) {Scene\\Pacing\\and\\Rhythm};
\node[task, right=of node_C] (node_F) {Scene\\Functions\\and\\Types};

\node[judge, right=1.5cm of node_E] (judge) {Final\\Judgment};

\node[verdict, above right=0.4cm and 0.8cm of judge] (v1) {Identical (10)};
\node[verdict, below=0.35cm of v1] (v2) {Very Similar (8)};
\node[verdict, below=0.35cm of v2] (v3) {Moderate (5)};
\node[verdict, below=0.35cm of v3] (v4) {Different (3)};
\node[verdict, below=0.35cm of v4] (v5) {Opposite (0)};

\draw[arrow] (node_A) -- (node_B);
\draw[arrow] (node_B) -- (node_C);
\draw[arrow] (node_C) -- (node_D);
\draw[arrow] (node_D) -- (node_E);
\draw[arrow] (node_E) -- (node_F);

\draw[arrow] (node_F) -- (judge);

\draw[arrow] (judge) -- (v1);
\draw[arrow] (judge) -- (v2);
\draw[arrow] (judge) -- (v3);
\draw[arrow] (judge) -- (v4);
\draw[arrow] (judge) -- (v5);

\end{tikzpicture}
}
\caption{DAG architecture for the Scene Sequence Similarity evaluator combining scene level sub-dimensions into an overall scene sequence similarity judgement}
\label{fig:dag_scene_sequence_similarity}
\end{figure}

Figure~\ref{fig:dag_scene_sequence_similarity} shows the scene sequence similarity evaluator. It draws on film-based narrative theory~\citep{bordwell2012film, dancyger2010technique, mckee1997story}, probing distinctive scene construction and ordering beyond generic scene types.

The scene identity \& content task asks the model to compare particular scenes in terms of setting, participants, actions, and narrative purpose. It is designed to capture whether recognisable scenes from the source are recreated in the target at the level of concrete detail.

The scene internal structure task considers beat sequences and staging inside scenes~\citep{mckee1997story}. The model examines the micro-structure of scenes and classifies how similarly they unfold.

The scene sequence architecture task looks at the ordering and relationship network of scenes, focusing on how one scene leads to another.

The scene transition \& connections task focuses on linking devices and continuity. It captures whether the same motif-based or structural transitions occur.

The scene pacing \& rhythm task compares patterns of scene length and tempo.

The scene functions \& types task examines standard scene categories and their positions. Since scene functions are often determined by genre, this dimension is treated as weakly protectable.

The judgement node weights scene identity and internal structure highest, scene sequence architecture somewhat lower, transitions and pacing lower still, and functions/types lowest. This tries to reflect the view that copying the detailed content and staging of scenes is more indicative of appropriation than sharing generic patterns such as \say{climactic confrontation near the end}.

\subsubsection{World Building Similarity}\label{sec:infringement:world-building}
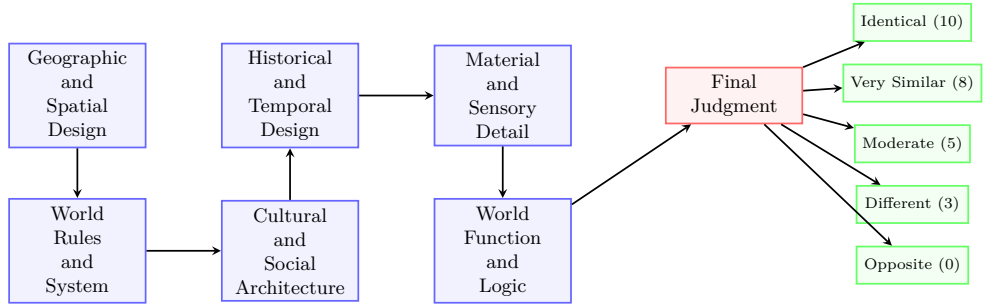
\begin{figure}[H]
\centering
\adjustbox{max width=\textwidth}{
\begin{tikzpicture}[
    node distance=0.8cm and 1.2cm,
    task/.style={rectangle, draw=blue!60, fill=blue!5, thick, minimum width=2.2cm, minimum height=0.7cm, align=center, font=\small},
    judge/.style={rectangle, draw=red!60, fill=red!5, thick, minimum width=2.2cm, minimum height=0.7cm, align=center, font=\small},
    verdict/.style={rectangle, draw=green!60, fill=green!5, thick, minimum width=1.8cm, minimum height=0.6cm, align=center, font=\footnotesize},
    arrow/.style={->, >=stealth, thick}
]

\node[task] (node_A) {Geographic\\and\\Spatial\\Design};
\node[task, below=of node_A] (node_B) {World\\Rules\\and\\System};
\node[task, right=of node_B] (node_C) {Cultural\\and\\Social\\Architecture};
\node[task, right=of node_A] (node_D) {Historical\\and\\Temporal\\Design};
\node[task, right=of node_D] (node_E) {Material\\and\\Sensory\\Detail};
\node[task, right=of node_C] (node_F) {World\\Function\\and\\Logic};

\node[judge, right=1.5cm of node_E] (judge) {Final\\Judgment};

\node[verdict, above right=0.4cm and 0.8cm of judge] (v1) {Identical (10)};
\node[verdict, below=0.35cm of v1] (v2) {Very Similar (8)};
\node[verdict, below=0.35cm of v2] (v3) {Moderate (5)};
\node[verdict, below=0.35cm of v3] (v4) {Different (3)};
\node[verdict, below=0.35cm of v4] (v5) {Opposite (0)};

\draw[arrow] (node_A) -- (node_B);
\draw[arrow] (node_B) -- (node_C);
\draw[arrow] (node_C) -- (node_D);
\draw[arrow] (node_D) -- (node_E);
\draw[arrow] (node_E) -- (node_F);

\draw[arrow] (node_F) -- (judge);

\draw[arrow] (judge) -- (v1);
\draw[arrow] (judge) -- (v2);
\draw[arrow] (judge) -- (v3);
\draw[arrow] (judge) -- (v4);
\draw[arrow] (judge) -- (v5);

\end{tikzpicture}
}
\caption{DAG architecture for the World Building Similarity evaluator combining world related sub-dimensions into an overall world building similarity judgement}
\label{fig:dag_world_building_similarity}
\end{figure}

The world building similarity evaluator (Figure~\ref{fig:dag_world_building_similarity}) targets the construction of fictional worlds, discriminating generic genre backdrops from richly elaborated original expression~\citep{wolf2012building, ekman2013here, gavins2007text}.

The geographic \& spatial design task asks the model to compare detailed locations, spatial layouts, and environmental features. It is designed to detect, for example, copying of a distinctive city layout or landscape structure.

The world rules \& system task compares magic, technology, or physics systems. The model analyses specific rules, costs, capabilities, and limitations. Fantasy and science-fiction scholarship emphasises that such systems can be highly original and recognisable~\citep{wolf2012building}.

The cultural \& social architecture task considers customs, rituals, institutions, and hierarchies.

The historical \& temporal design task looks at past events and temporal patterns, including cycles and eras~\citep{ekman2013here}.

The material \& sensory detail task compares objects, flora and fauna, and sensory atmospheres, reflecting text-world theory's emphasis on how readers construct experiential worlds~\citep{gavins2007text}. It captures whether distinctive material signatures are shared.

The world function \& logic task examines how the world operates and maintains internal coherence, but is treated as more abstract and closer to functional necessity.

The judgement node weights geographic/spatial design and world rules/systems highest, cultural/social and historical/temporal design intermediate, material/sensory detail lower, and world function/logic lowest. This pattern is justified by the observation that copying specific maps and systems is more indicative of appropriation than sharing generic principles such as \say{the world is internally consistent}. From a more legal-oriented perspective, concrete spatial and systematic elaborations are closer to protectable expression, while abstract functional properties are closer to unprotectable ideas.

\subsection{Defence Evaluator}\label{sec:defensive-evaluators}
Defence evaluators follow the same DAG architecture as infringement evaluators, with the inversion that higher scores indicate stronger exception applicability or greater genericness. Weights are determined heuristically using the same doctrine-informed procedure.

\subsubsection{Parody and Satire}\label{sec:defence:parody-satire}
\begin{figure}[H]
\centering
\adjustbox{max width=\textwidth}{
\begin{tikzpicture}[
    node distance=0.8cm and 1.2cm,
    task/.style={rectangle, draw=blue!60, fill=blue!5, thick, minimum width=2.2cm, minimum height=0.7cm, align=center, font=\small},
    judge/.style={rectangle, draw=red!60, fill=red!5, thick, minimum width=2.2cm, minimum height=0.7cm, align=center, font=\small},
    verdict/.style={rectangle, draw=green!60, fill=green!5, thick, minimum width=1.8cm, minimum height=0.6cm, align=center, font=\footnotesize},
    arrow/.style={->, >=stealth, thick}
]

\node[task] (node_A) {Source\\Work\\Evocations\\or\\Recognition};
\node[task, below=of node_A] (node_B) {Noticeable\\Differences\\or\\Transformation};
\node[task, right=of node_B] (node_C) {Humorous\\Character};
\node[task, right=of node_A] (node_D) {Mocking\\or\\Critical\\Character};
\node[task, right=of node_D] (node_E) {Fair\\Balance\\and\\Context};

\node[judge, right=1.5cm of node_E] (judge) {Final\\Judgment};

\node[verdict, above right=0.4cm and 0.8cm of judge] (v1) {Strong Parody (10)};
\node[verdict, below=0.35cm of v1] (v2) {Good Parody (8)};
\node[verdict, below=0.35cm of v2] (v3) {Borderline (5)};
\node[verdict, below=0.35cm of v3] (v4) {Weak (3)};
\node[verdict, below=0.35cm of v4] (v5) {No Parody (0)};

\draw[arrow] (node_A) -- (node_B);
\draw[arrow] (node_B) -- (node_C);
\draw[arrow] (node_C) -- (node_D);
\draw[arrow] (node_D) -- (node_E);

\draw[arrow] (node_E) -- (judge);

\draw[arrow] (judge) -- (v1);
\draw[arrow] (judge) -- (v2);
\draw[arrow] (judge) -- (v3);
\draw[arrow] (judge) -- (v4);
\draw[arrow] (judge) -- (v5);

\end{tikzpicture}
}
\caption{DAG architecture for the Parody and Satire evaluator aggregating the Deckmyn dimensions into a final parody strength judgement}
\label{fig:dag_parody_satire}
\end{figure}
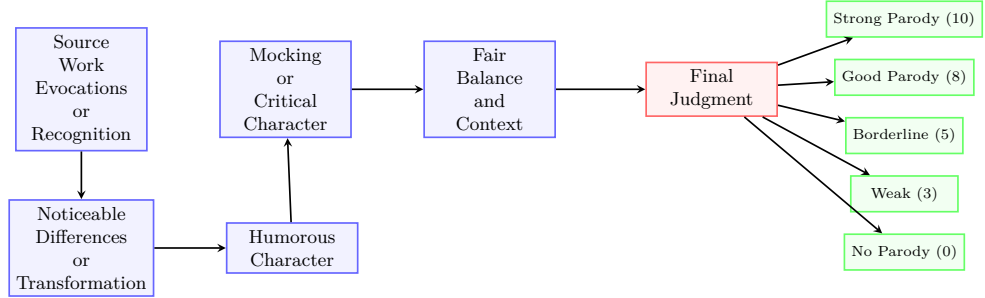

The parody/satire evaluator in Figure~\ref{fig:dag_parody_satire} implements the \emph{Deckmyn} criteria~\citep{CourtDeckmyn2014}, as interpreted by doctrinal analyses~\citep{rendas2024pastiche}.

The source work evocation / recognition task asks the judge model to determine how clearly the target evokes the specific source work, by looking for named references, distinctive characters, iconic settings, or recognisable stylistic imitation. It then labels evocation strength on the five-level scale, from \say{no evocation} to \say{clear, specific evocation}.

The noticeable differences/transformation task asks the model to compare content and tone between source and target, and to assess whether there are substantive transformations in characters, plot, setting, and message, rather than cosmetic changes. The label expresses whether the target is \say{not noticeably different} or \say{highly transformed}.

The Humorous Character task assesses presence and clarity of humour. The model identifies devices such as exaggeration, irony, absurdity, and wordplay, and classifies whether humour is absent, weak, moderate, or strong. Though Deckmyn does not prescribe a particular comedic form, the task classifies humour strength from absent to strong.

The mocking/critical character task focuses on critical commentary and target. The model determines whether the target critiques the original work, its author, a genre, or broader social phenomena, and labels the strength of this mocking or critical dimension. This is important because parody can be primarily critical, even if not overtly humorous.

The fair balance \& context task evaluates proportionality and non-discrimination. It asks whether the extent of copying is proportionate to the parodic purpose and whether the work contains discriminatory content against protected groups. The label indicates whether fair balance is strongly met, borderline, or clearly violated. This tries to reflect Deckmyn's emphasis on balancing rights and the specific warning that discriminatory parody may not receive protection.

The judgement node weights evocation and transformation most heavily, humour and mockery next, and fair balance lowest but still non-negligible. This weighting is justified by the structure of the Deckmyn test: evocation and difference are necessary conditions for parody, whereas fair balance qualifies the scope of the exception. Humour and mockery are essential characteristics, but once present at a clear level, marginal increases have less legal effect. 

\subsubsection{Pastiche}\label{sec:defence:pastiche}
\begin{figure}[H]
\centering
\adjustbox{max width=\textwidth}{
\begin{tikzpicture}[
    node distance=0.8cm and 1.2cm,
    task/.style={rectangle, draw=blue!60, fill=blue!5, thick, minimum width=2.2cm, minimum height=0.7cm, align=center, font=\small},
    judge/.style={rectangle, draw=red!60, fill=red!5, thick, minimum width=2.2cm, minimum height=0.7cm, align=center, font=\small},
    verdict/.style={rectangle, draw=green!60, fill=green!5, thick, minimum width=1.8cm, minimum height=0.6cm, align=center, font=\footnotesize},
    arrow/.style={->, >=stealth, thick}
]

\node[task] (node_A) {Style\\Evocation\\and\\Recognition\\Analysis};
\node[task, below=of node_A] (node_B) {Artistic\\Skill\\and\\Execution\\Analysis};
\node[task, right=of node_B] (node_C) {Homage\\and\\Tribute\\Character\\Analysis};
\node[task, right=of node_A] (node_D) {Noticeable\\Differences\\and\\Transformation\\Analysis};
\node[task, right=of node_D] (node_E) {Fair\\Balance\\and\\Proportionality\\Analysis};

\node[judge, right=1.5cm of node_E] (judge) {Final\\Judgment};

\node[verdict, above right=0.4cm and 0.8cm of judge] (v1) {Strong Pastiche (10)};
\node[verdict, below=0.35cm of v1] (v2) {Good Pastiche (8)};
\node[verdict, below=0.35cm of v2] (v3) {Borderline (5)};
\node[verdict, below=0.35cm of v3] (v4) {Weak (3)};
\node[verdict, below=0.35cm of v4] (v5) {No Pastiche (0)};

\draw[arrow] (node_A) -- (node_B);
\draw[arrow] (node_B) -- (node_C);
\draw[arrow] (node_C) -- (node_D);
\draw[arrow] (node_D) -- (node_E);

\draw[arrow] (node_E) -- (judge);

\draw[arrow] (judge) -- (v1);
\draw[arrow] (judge) -- (v2);
\draw[arrow] (judge) -- (v3);
\draw[arrow] (judge) -- (v4);
\draw[arrow] (judge) -- (v5);

\end{tikzpicture}
}
\caption{DAG architecture for the Pastiche evaluator aggregating style evocation artistic skill tribute character transformation and proportionality into a final pastiche strength judgement}
\label{fig:dag_pastiche_satire}
\end{figure}
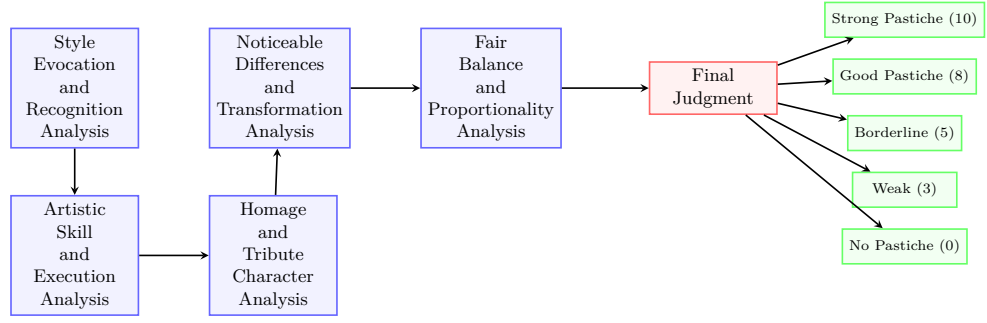

The pastiche evaluator in Figure~\ref{fig:dag_pastiche_satire} is grounded in Article 5(3)(k) of InfoSoc and AG Emiliou's Opinion in \emph{Pelham II}~\citep{eucopyright, AGEmiliouPelham2024}, which differentiates pastiche from parody and caricature.

The style evocation \& recognition task asks the model to assess how convincingly the target imitates the source's distinctive aesthetic language (its sentence rhythm, lexical choices, tonal qualities, and narrative techniques). A high label indicates strong, source-specific stylistic echo, not just generic genre similarity.

The artistic skill \& execution task evaluates technical quality and consistency of imitation. The model assesses whether the stylistic imitation is precise, coherent, and adapted naturally to the new content, and labels this from \say{no or clumsy imitation} to \say{high artistic mastery}. This operationalises AG Emiliou's \say{artistic creation} requirement~\citep{AGEmiliouPelham2024}.

The homage/tribute character task determines whether the work expresses respect or admiration, without mockery. The model looks for evidence of tribute (tone, framing, dedications) and absence of critical intent, then labels the strength of this homage character. This is the primary criterion that distinguishes pastiche from parody.

The noticeable differences/transformation task asks whether, despite stylistic similarity, the content (characters, plot, themes) substantially differs from the source.

The fair balance \& proportionality task considers the extent of stylistic borrowing relative to new content and assesses potential market substitution, in line with the three-step test and doctrinal concerns about \say{catch-all} misuse of pastiche~\citep{rendas2024pastiche, geiger2015three}.

The judgement node weights style evocation and artistic skill highest, homage character next, and transformation and proportionality lower. This tries to reflect doctrinal emphasis on the combination of recognisable style imitation and artistic creation, together with the requirement that the work be a tribute rather than a critique. Transformation and proportionality matter, but primarily as safeguards against excessively wide use of the exception.

\subsubsection{Quotation and Citation}\label{sec:defence:quotation-citation}
\begin{figure}[H]
\centering
\adjustbox{max width=\textwidth}{
\begin{tikzpicture}[
    node distance=0.8cm and 1.2cm,
    task/.style={rectangle, draw=blue!60, fill=blue!5, thick, minimum width=2.2cm, minimum height=0.7cm, align=center, font=\small},
    judge/.style={rectangle, draw=red!60, fill=red!5, thick, minimum width=2.2cm, minimum height=0.7cm, align=center, font=\small},
    verdict/.style={rectangle, draw=green!60, fill=green!5, thick, minimum width=1.8cm, minimum height=0.6cm, align=center, font=\footnotesize},
    arrow/.style={->, >=stealth, thick}
]

\node[task] (node_A) {Quotation\\Identification\\and\\Extraction};
\node[task, below=of node_A] (node_B) {Legitimate\\Purpose};
\node[task, right=of node_B] (node_C) {Fair\\Practice\\and\\Proportionality};
\node[task, right=of node_A] (node_D) {Attribution\\and\\Source\\Acknowledgement};
\node[task, right=of node_D] (node_E) {Fair\\Balance\\and\\Justification};
\node[task, right=of node_C] (node_F) {Work\\Already\\Disclosed};

\node[judge, right=1.5cm of node_E] (judge) {Final\\Judgment};

\node[verdict, above right=0.4cm and 0.8cm of judge] (v1) {Strong Quotation (10)};
\node[verdict, below=0.35cm of v1] (v2) {Good Quotation (8)};
\node[verdict, below=0.35cm of v2] (v3) {Borderline (5)};
\node[verdict, below=0.35cm of v3] (v4) {Weak (3)};
\node[verdict, below=0.35cm of v4] (v5) {No Quotation (0)};

\draw[arrow] (node_A) -- (node_B);
\draw[arrow] (node_B) -- (node_C);
\draw[arrow] (node_C) -- (node_D);
\draw[arrow] (node_D) -- (node_E);
\draw[arrow] (node_E) -- (node_F);

\draw[arrow] (node_F) -- (judge);

\draw[arrow] (judge) -- (v1);
\draw[arrow] (judge) -- (v2);
\draw[arrow] (judge) -- (v3);
\draw[arrow] (judge) -- (v4);
\draw[arrow] (judge) -- (v5);

\end{tikzpicture}
}
\caption{DAG architecture for the Quotation and Citation evaluator aggregating quotation presence purpose fair practice attribution fair balance and prior disclosure into a final quotation strength judgement}
\label{fig:dag_quotation_citation}
\end{figure}
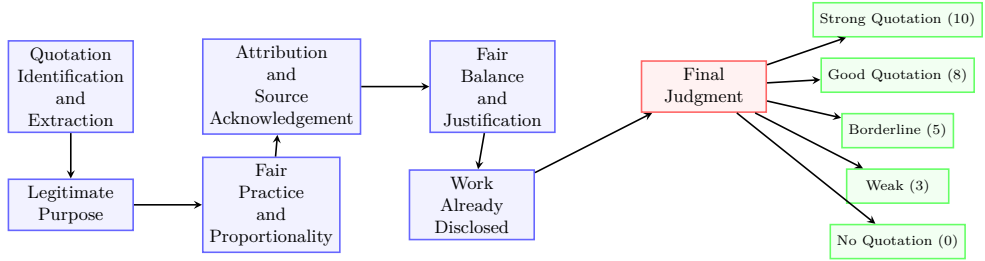

Figure~\ref{fig:dag_quotation_citation} shows the quotation/citation evaluator, grounded in Article 5(3)(d) of InfoSoc and CJEU case law~\citep{eucopyright, cjeu2019funke}.

The quotation identification \& extraction task instructs the model to identify whether the target reproduces passages from the source verbatim or near-verbatim, whether quotation marks or other conventions are used, and to classify the strength of quotation presence: from \say{none} through \say{partial} to \say{substantial, explicit quotation}. This directly reflects the requirement that there be quotation in the first place.

The legitimate purpose task asks the model to infer the purpose of the quotations (criticism, review, news reporting, teaching, research, or similar purposes) and to assess how genuinely the quoted material is used for that purpose, based on surrounding commentary.

The fair practice \& proportionality task evaluates extent of taking relative to purpose and integration of quotations into the target's own analysis, as demanded by the \say{fair practice} language and the three-step test~\citep{geiger2015three}.

The attribution \& source acknowledgement task determines whether the source work and author are identified, as required \say{unless impossible}~\citep{cjeu2019funke}.

The fair balance \& justification task examines necessity and (potential) market impact, asking whether the quotation could plausibly substitute for the source or harm its market.

The work already disclosed task focusses on whether the source appears to have been lawfully made available to the public before the target, reflecting the requirement that quotation applies only to disclosed works. This dimension is necessarily limited, depending on the model's training data or contextual information provided within the texts.

The judgement node weights quotation identification and legitimate purpose highest, fair practice and attribution next, and fair balance and prior disclosure lower. This weighting tries to correspond to the doctrinal structure: without quotation and legitimate purpose, the exception cannot apply; fair practice and attribution are core conditions; fair balance and prior disclosure operate as background constraints.

\subsubsection{Scènes à Faire}\label{sec:defence:scenes-a-faire}
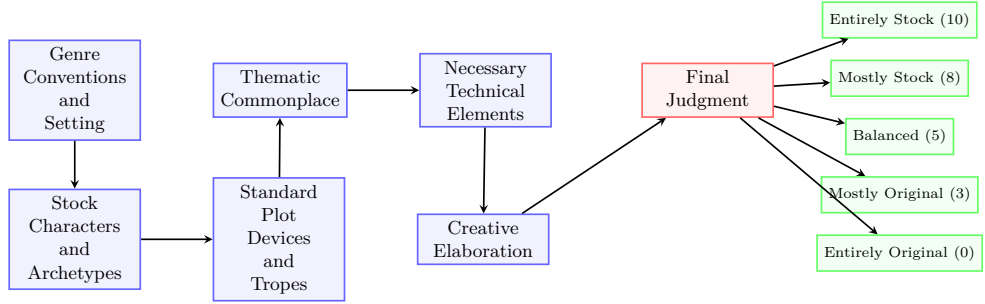
\begin{figure}[H]
\centering
\adjustbox{max width=\textwidth}{
\begin{tikzpicture}[
    node distance=0.8cm and 1.2cm,
    task/.style={rectangle, draw=blue!60, fill=blue!5, thick, minimum width=2.2cm, minimum height=0.7cm, align=center, font=\small},
    judge/.style={rectangle, draw=red!60, fill=red!5, thick, minimum width=2.2cm, minimum height=0.7cm, align=center, font=\small},
    verdict/.style={rectangle, draw=green!60, fill=green!5, thick, minimum width=1.8cm, minimum height=0.6cm, align=center, font=\footnotesize},
    arrow/.style={->, >=stealth, thick}
]

\node[task] (node_A) {Genre\\Conventions\\and\\Setting};
\node[task, below=of node_A] (node_B) {Stock\\Characters\\and\\Archetypes};
\node[task, right=of node_B] (node_C) {Standard\\Plot\\Devices\\and\\Tropes};
\node[task, right=of node_A] (node_D) {Thematic\\Commonplace};
\node[task, right=of node_D] (node_E) {Necessary\\Technical\\Elements};
\node[task, right=of node_C] (node_F) {Creative\\Elaboration};

\node[judge, right=1.5cm of node_E] (judge) {Final\\Judgment};

\node[verdict, above right=0.4cm and 0.8cm of judge] (v1) {Entirely Stock (10)};
\node[verdict, below=0.35cm of v1] (v2) {Mostly Stock (8)};
\node[verdict, below=0.35cm of v2] (v3) {Balanced (5)};
\node[verdict, below=0.35cm of v3] (v4) {Mostly Original (3)};
\node[verdict, below=0.35cm of v4] (v5) {Entirely Original (0)};

\draw[arrow] (node_A) -- (node_B);
\draw[arrow] (node_B) -- (node_C);
\draw[arrow] (node_C) -- (node_D);
\draw[arrow] (node_D) -- (node_E);
\draw[arrow] (node_E) -- (node_F);

\draw[arrow] (node_F) -- (judge);

\draw[arrow] (judge) -- (v1);
\draw[arrow] (judge) -- (v2);
\draw[arrow] (judge) -- (v3);
\draw[arrow] (judge) -- (v4);
\draw[arrow] (judge) -- (v5);

\end{tikzpicture}
}
\caption{DAG architecture for the Scènes à Faire evaluator combining genre character plot thematic technical and elaboration dimensions into an overall judgement of stockness or genericness}
\label{fig:dag_scenes_a_faire}
\end{figure}

The scènes à faire evaluator (Figure~\ref{fig:dag_scenes_a_faire}) operationalises the exclusion of standard elements dictated by genre, setting, or technical constraints from protectable expression~\citep{nichols1930universal, warner2008rdr, hoehling1980universal, walker1986time, geiger2015three}.

The genre conventions \& setting task asks the model to identify genre-specific plot structures and standard settings, such as \say{crime leading to investigation and revelation} or \say{medieval fantasy kingdom with castles and dragons}, and to assess how heavily both texts rely on these standard conventions.

The stock characters \& archetypes task examines whether key characters fit archetypal moulds without elaboration. It captures reliance on stock roles like \say{reluctant hero} or \say{wise old mentor}.

The standard plot devices \& tropes task considers common devices, such as love triangles, mistaken identity, or deus ex machina.

The thematic commonplaces task evaluates whether themes such as \say{good versus evil} or \say{coming of age} are treated in a generic, conventional way. Themes as such are unprotectable ideas, but highly original thematic treatment may be reflected in lower stock-ness~\citep{stamatatos2009survey, geiger2015three}.

The necessary technical elements task looks at elements required by genre or medium (for example, basic investigative steps in a detective story, scene transitions, or exposition).

The creative elaboration task, inversely defined, asks the model to assess the degree of distinctive original expression across style, characterisation, world-building, and structure, and to rate how generic or non-generic this expression is. A high label indicates that both texts are most likely largely generic; a low label indicates potential substantial creative elaboration. In combining scores, this dimension effectively tries to reduce the overall scènes à faire score when originality is high.

The judgement node assigns highest weights to genre conventions and stock characters, standard plot devices somewhat less, thematic commonplaces and functional elements lower again, and creative elaboration as a corrective term with modest weight. This tries to reflect the legal reasoning in scènes à faire cases, where courts first identify elements dictated by genre or stock situations and then ask whether the work contains additional creative elaboration that is protectable~\citep{nichols1930universal, warner2008rdr, hoehling1980universal, walker1986time}.

\section{Experimental Setup}\label{sec:experiments}
The experiments address three research questions through controlled validation and large-scale corpus evaluation; hyperparameter specifications, validation corpus construction, and extended statistical protocols appear in Appendix~\ref{app:validation-corpus} and Appendix~\ref{app:statistical-details}.

\subsection{Corpus Construction and Preparation}\label{sec:dataset}
\subsubsection{Source Materials}
The experimental corpus is drawn from the Digital Library of Dutch Literature (DBNL)~\citep{dbnl}, which provides high-quality editions of pre-1900 Dutch literary works. These works are out of copyright yet exhibit rich and distinctive narrative and stylistic features, making them suitable proxies for contemporary protected literature. Eleven authors spanning prose, drama, and children's literature were selected; approximately three works per author were sampled to ensure diversity of narrative voices and stylistic registers.

\subsubsection{Story-Generation Question-Answer Dataset}
To fine-tune the model in a way that emphasises stylistic learning rather than verbatim copying of prompts, the DBNL texts were converted into a question-answer (QA) dataset. Each QA pair consists of a natural-language prompt requesting a short scene or passage and an answer containing an excerpt from a source work.

The prompts were automatically generated using the open-source GPTOSS-120B model. We selected GPTOSS-120B for two reasons. First, it was reliably available in our experimental environment at the time. Second, in initial pilot experiments, GPTOSS-120B produced prompts with better semantic alignment to the corresponding source passages and fewer formatting/artefact issues than other candidate LLMs that we tested, as judged by manual review by the authors.

For each selected passage, the model was asked to write a request that described a scene, dialogue, or monologue \say{in the style of} the corresponding work or author, mentioning characters, settings, or events that appear in the source text. The answer for each prompt is a contiguous excerpt of approximately $200-800$ tokens from the original Dutch work, chosen to match the scenario described in the question. This procedure yields around $500$ QA pairs. A random sample of generated prompts and associated passages was inspected and manually edited when necessary to ensure that the prompt semantically matches the answer and that the answer does not contain typesetting artefacts.

The QA format enables the same dataset to be used for both supervised fine-tuning and generation-based evaluation, by prompting the model with the questions and comparing outputs against source passages.

\subsubsection{Translation and Modernisation}
The DBNL texts are written in historical varieties of Dutch that differ substantially from contemporary usage in orthography, morphology and lexicon. Pilot fine-tuning showed that LLaMA~3.2~1B struggles with archaic Dutch, reverting to modern Dutch even when prompted otherwise. Since the study probes stylistic appropriation rather than historical language acquisition, source passages were translated into modern variants.

For each passage, three variants exist: the original historical Dutch excerpt, a modern Dutch translation, and an English translation. Only the translated variants are used during fine-tuning, unlearning, and generation; original passages are retained as reference material.

The translations were produced with a combination of GPTOSS-120B and OpenAI's GPT-4o-mini. GPTOSS-120B was used as the primary translator, while GPT-4o-mini was employed for a minority of passages (on the order of $50-100$) for efficiency reasons when the self-hosted model was unavailable. Both directions (Dutch $\rightarrow$ modern Dutch and Dutch $\rightarrow$ English) were translated independently rather than via cascading translation.

Quality control employed manual back-translation checking on randomly sampled rows, with the acceptance criterion being semantic fidelity (events, characters, and relations preserved; fluent target-language prose), permitting paraphrase and synonymy. No systematic errors were observed.

\subsubsection{Forget and Retain Set Partitioning}\label{sec:dataset:forget-retain}
To study unlearning, the QA dataset is partitioned into \emph{retain} and \emph{forget} sets. The partitioning is performed at the semantic level of works and authors rather than at the level of individual QA pairs, to mimic realistic copyright takedown requests, though sequential takedown scenarios are outside the scope of this study.

Approximately $350$ QA pairs form the retain set and remain available throughout training and unlearning, while around $130$ pairs form the forget set. Forget sets are constructed using two strategies. First, is author-level forgetting, where for some authors all books and their associated QA pairs are assigned to the forget set. Second, is book-level forgetting, where for other authors only specific works are designated as forget targets, while their remaining books remain in the retain set.

\subsection{Model and Training Configurations}
\subsubsection{Base Model and Variants}\label{sec:base-model}
All experiments use Unsloth's optimised Meta LLaMA~3.2~1B Instruct model; the unmodified model serves as the per-language baseline. Fine-tuning and unlearning then produce additional variants as summarised in Table~\ref{tab:model_variants}.

\begin{table}[htbp]
    \centering
    \adjustbox{max width=\textwidth}{
    \begin{tabular}{lllll}
        \toprule
        Model & Language & Stage        & Training data              & Objective \\
        \midrule
        \texttt{en\_b}   & EN & baseline           & --                       & -- \\
        \texttt{nl\_b}   & NL & baseline           & --                       & -- \\
        \texttt{en\_f}   & EN & fine-tuned        & EN QA (forget + retain)  & CE \\
        \texttt{nl\_f}   & NL & fine-tuned        & NL QA (forget + retain)  & CE \\
        \texttt{en\_npo} & EN & NPO unlearning    & EN QA (forget, retain)   & NPO loss \\
        \texttt{nl\_npo} & NL & NPO unlearning    & NL QA (forget, retain)   & NPO loss \\
        \bottomrule
    \end{tabular}
    }
    \caption{Overview of model variants and training regimes EN and NL denote English and modern Dutch respectively CE denotes cross entropy loss used for fine-tuned training}
    \label{tab:model_variants}
\end{table}

\subsubsection{Fine-Tuning and Unlearning Configuration}
Both fine-tuning and unlearning use LoRA adapters. Fine-tuning is performed separately per language on the full QA dataset; NPO unlearning starts from the fine-tuned model and updates adapters using the forget-retain partitions (Section \ref{sec:dataset:forget-retain}). Table~\ref{tab:train_unlearn_hparams} summarises all hyperparameters used in the experiments.

\begin{table}[htbp]
  \centering
  \adjustbox{max width=\textwidth}{
  \begin{tabular}{p{4cm}p{3cm}p{3cm}p{4cm}}
    \toprule
    \textbf{Parameter} & \textbf{Fine-tuning} & \textbf{NPO unlearning} \\
    \midrule
    Training epochs            & $5$–$20$ (pilot); chosen config: $10$ & $3$–$10$ (pilot); chosen config: $5$ \\
    Sequence length            & $1024$ & $1024$ \\
    Batch size (per device)    & $16$   & $8$ \\
    Gradient accumulation      & $8$    & $1$ \\
    Effective batch size       & $128$  & $8$  \\
    Learning rate              & $5\times 10^{-4}$ & $5\times 10^{-4}$ \\
    LR scheduler               & cosine with warmup & cosine with warmup  \\
    Warmup ratio               & $0.05$ & $0.05$  \\
    Optimiser                  & AdamW   & AdamW    \\
    Weight decay               & $0.01$ & $0.01$  \\
    Max gradient norm          & $1.0$  & $1.0$   \\
    Mixed precision            & bfloat16 & bfloat16 \\
    LoRA rank                  & $32$   & $32$    \\
    LoRA alpha                 & $64$   & $64$    \\
    LoRA dropout               & $0.05$ & $0.05$  \\
    Target modules             & all attention and FFN projections & all attention and FFN projections \\
    \midrule
    NPO temperature $\alpha$ &         & $1.0$   \\
    NPO KL coefficient $\beta$ &      & $0.1$  \\
    \bottomrule
  \end{tabular}
  }
  \caption{Hyperparameters for LoRA fine tuning and NPO unlearning Values are identical for English and Dutch unless noted otherwise where a parameter is specific to NPO the finetuning column is left blank}
  \label{tab:train_unlearn_hparams}
\end{table}

Fine-tuning minimises token-level cross-entropy on QA answer text. NPO updates use the same optimiser and schedule with the loss from \citet{Zhang2024npo}: negative preferences on the forget set combined with KL regularisation on the retain set. PSALM is agnostic to the internal loss form; the relevant artefacts are the model variants in Table~\ref{tab:model_variants} produced under the configuration in Table~\ref{tab:train_unlearn_hparams}.

\subsubsection{Output Generation}
Standardised prompts elicit narrative text that potentially exhibits stylistic similarity without explicitly requesting copying. For base models, the prompts specify abstract scenarios/themes present in the corpus. For fine-tuned and unlearned models, the prompts include QA dataset questions used during training.

The hyperparameters used during output generation are shown in Appendix Table~\ref{tab:generation_hyperparameters_full}.

\subsection{Evaluation Procedure}
All PSALM evaluators use GPT-5-nano as the judge model; verbose mode is disabled during corpus evaluation.

\subsubsection{Controlled Validation Set}\label{sec:eval:validation}
Each evaluator is validated on a purpose-built corpus of $50$ text pairs: five test cases per evaluator (TC1--TC5) spanning the full similarity range (Appendix~\ref{app:validation-corpus}).

The criteria for determining whether an evaluator is deemed usable in the main experiments are as follows. First, the validated evaluators, where a strict alignment rate (proportion of predictions with $\pm 0.1$ of the expected score) is at least $80\%$ and mean absolute error (MAE) is at most $0.15$. Secondly, the partially validated evaluators, where the alignment rate (within $\pm 0.2$) at least $70\%$ and MAE at most $0.25$. Lastly, the failed evaluators that do not meet the (partially) validated criteria.

These thresholds are heuristic: a tolerance of $\pm0.2$ corresponds to one adjacent category on the five-point scale, so an $80\%$ alignment rate requires the evaluator to assign the correct band in the large majority of cases. Validated evaluators are used in all subsequent analyses; partially validated evaluators are retained for exploratory inspection only; failed evaluators are excluded or revised.

\subsubsection{Corpus Evaluation Execution}
For each QA pair (stratified by language and forget-retain partition), the model generates a response \(x_t\); the corresponding source passage \(x_s\) is retrieved; and all validated PSALM evaluators \(E(x_s, x_t)\) are executed, yielding up to ten scores per pair alongside the lexical baselines (Section~\ref{sec:baseline_comparisons}).

The resulting dataset records, per generated answer, its language, author, forget-retain status, model condition, PSALM scores, and lexical metrics.

\subsection{Baseline Computational Metrics}\label{sec:baseline_comparisons}
Three standard lexical overlap metrics are computed for every \((x_s, x_t)\) pair as reference points for interpreting PSALM scores (Table~\ref{tab:lexical_metrics}).

\begin{table}[htbp]
  \centering
  \adjustbox{max width=\textwidth}{
  \begin{tabular}{p{2cm}p{3cm}p{7cm}}
    \toprule
    \textbf{Metric} & \textbf{Implementation} & \textbf{Summary} \\
    \midrule
    Exact Match & custom string matching & indicator equal to $1$ if the
    normalised source passage appears as a contiguous substring of the
    model output, and $0$ otherwise \\
    BLEU-4 & \texttt{sacrebleu} & $n$-gram precision with brevity penalty using
    up to $4$-grams, as commonly applied in machine translation and
    memorisation studies \\
    ROUGE-L $F_{1}$ & \texttt{rouge-score} & longest-common-subsequence-based
    ROUGE-L, converted to an $F_{1}$ score to capture overlap in
    token order and coverage \\
    \bottomrule
  \end{tabular}
  }
  \caption{Lexical baseline metrics used alongside PSALM scores}
  \label{tab:lexical_metrics}
\end{table}

\subsection{Statistical Analysis}\label{sec:stats}
\subsubsection{Metric Category Taxonomy}
Aggregated summaries are computed at three levels: the generation level, where each QA prompt contributes one set of scores; the model-condition level, where scores are aggregated across prompts for a fixed language, model stage, and split, forming the primary input to significance testing; and the category level, where metrics are grouped into four conceptual categories, namely, computational (Exact Match, BLEU, ROUGE-L), stylistic (Writing Style, Narrative Voice), content (Character, Plot, Scene, World-building), and exceptions (Parody/Satire, Pastiche, Quotation/Citation, Scènes à Faire), to inspect whether trends are consistent within broader copyright-relevant notions.

\subsubsection{Assumption Checks}
Because PSALM scores are bounded and often exhibit ceiling effects after fine-tuning, strict normality and equal variance are hard to justify. Before choosing tests for each metric, Shapiro-Wilk tests for normality and Levene's tests for homogeneity of variance are applied as provided in Appendix~\ref{app:statistical-details}. In almost all cases at least one assumption is violated. Consequently, non-parametric methods (Kruskal-Wallis, Mann-Whitney, Wilcoxon signed-rank, ART-ANOVA) are used as the primary inferential tools. Parametric results, where reported, are interpreted as descriptive summaries and accompanied by appropriate caveats.

\subsubsection{Primary Hypothesis Tests}\label{sec:tests}
\paragraph{Effect of fine-tuning and unlearning.}
For each metric, language and split, an omnibus Kruskal-Wallis test is used to compare the three training stages (baseline, fine-tuned, unlearned). When significant, pairwise Mann-Whitney U-tests with Bonferroni correction probe specific contrasts such as baseline versus fine-tuned (RQ2) and fine-tuned versus unlearned (RQ3).

\paragraph{Within-model reductions.}
To quantify how much unlearning changes behaviour for the same prompts, paired Wilcoxon signed-rank tests compare fine-tuned and unlearned scores per QA pair. These tests capture whether NPO produces systematic reductions in similarity on the forget set, and whether any unintended changes occur on the retain set.

\paragraph{Language \texorpdfstring{$\times$}{x} stage interactions}
To explore whether training-stage effects differ between English and Dutch, ART-ANOVA is applied with factors language and stage. Main effects test whether one language consistently exhibits higher similarity than the other, while interaction terms test whether, for example, unlearning is more effective in Dutch than in English. When significant interactions are found, simple-effects analyses stratified by language follow the procedure outlined in Appendix~\ref{app:statistical-details}.

\subsubsection{Effect Size Estimation}
For Mann-Whitney and Wilcoxon tests, Cliff's $\delta$ serves as the primary effect size. The values $|\delta| < 0.2$ are interpreted as negligible, $0.2 \leq |\delta| < 0.5$ as small, $0.5 \leq |\delta| < 0.8$ as medium, and $|\delta| \geq 0.8$ as large. Bootstrap confidence intervals for $\delta$ are obtained following Appendix~\ref{app:statistical-details}.

For ART-ANOVA and Kruskal-Wallis tests, eta-squared $\eta^2$ and omega-squared $\omega^2$ are reported as rank-based analogues of the proportion of explained variance, again following the definitions in the appendix.

Where parametric comparisons are included for completeness (e.g., to facilitate comparison with other work), Cohen's $d$ and Hedges' $g$ are reported as standardised mean differences. Given the discrete scale of PSALM's scores, these values are interpreted cautiously and primarily used for relative ranking rather than as precise metric distances.

\subsubsection{Equivalence Testing}
For RQ3, it is insufficient to show that NPO reduces stylistic similarity; we additionally test whether unlearned models have returned to baseline levels using two one-sided tests (TOST).

For each metric, language and split, TOST compares the mean score of the unlearned model to that of the baseline model with equivalence bounds of $\pm 0.1$ on the normalised $0-1$ scale. These bounds correspond roughly to half a PSALM category and represent the smallest difference considered practically meaningful. If both one-sided tests reject the null hypothesis that the difference exceeds the bounds, the unlearned model is considered statistically equivalent to the baseline for that metric. If equivalence cannot be established, results are interpreted either as residual over- or under-similarity relative to the baseline.

\subsection{Computational Resources}\label{sec:infrastructure}
Experiments utilise NVIDIA RTX 4090 GPUs (24GB VRAM) as the primary computational resource, with NVIDIA L40(s) and H100 GPUs available where requirements exceeded capacity. The implementation employs Hugging Face Transformers, Unsloth (optimised LoRA), DeepSpeed/Accelerate (training optimisation), DeepEval (LLM-as-judge infrastructure), and PyTorch.

Estimated computational budget: $\approx 30$ GPU-hours total ($4$ fine-tuning, $4$ unlearning, $6$ generation, $16+$ evaluation). Actual requirements vary with convergence speed, hardware, hyperparameters, optimisation, and dataset sizes.

\section{Results}\label{sec:results}
This section presents the quantitative results of the PSALM evaluation under the experimental conditions defined in Section~\ref{sec:experiments}.

Throughout the results, we use the following notation for each of the models: English baseline ($en_b$), English fine-tuned ($en_f$), English NPO unlearned ($en_{npo}$), Dutch baseline ($nl_b$), Dutch fine-tuned ($nl_f$) and Dutch NPO unlearned ($nl_{npo}$).

\subsection{Controlled Validation of PSALM Evaluators}\label{sec:controlled-validation}
The validation corpus provides five hand-crafted test cases per evaluator, each instantiating one level of the PSALM scale with five stochastic replications, assessing whether the LLM-as-judge architecture reliably reproduces the qualitative distinctions the legal and narratological design encodes. Additional similar results are reported in Appendix~\ref{app:evaluator-performance}.

\subsubsection{Validation Performance Summary}
\begin{figure}[!hb]
    \centering
    \adjustbox{max width=\textwidth}{
    \includegraphics[width=\textwidth]{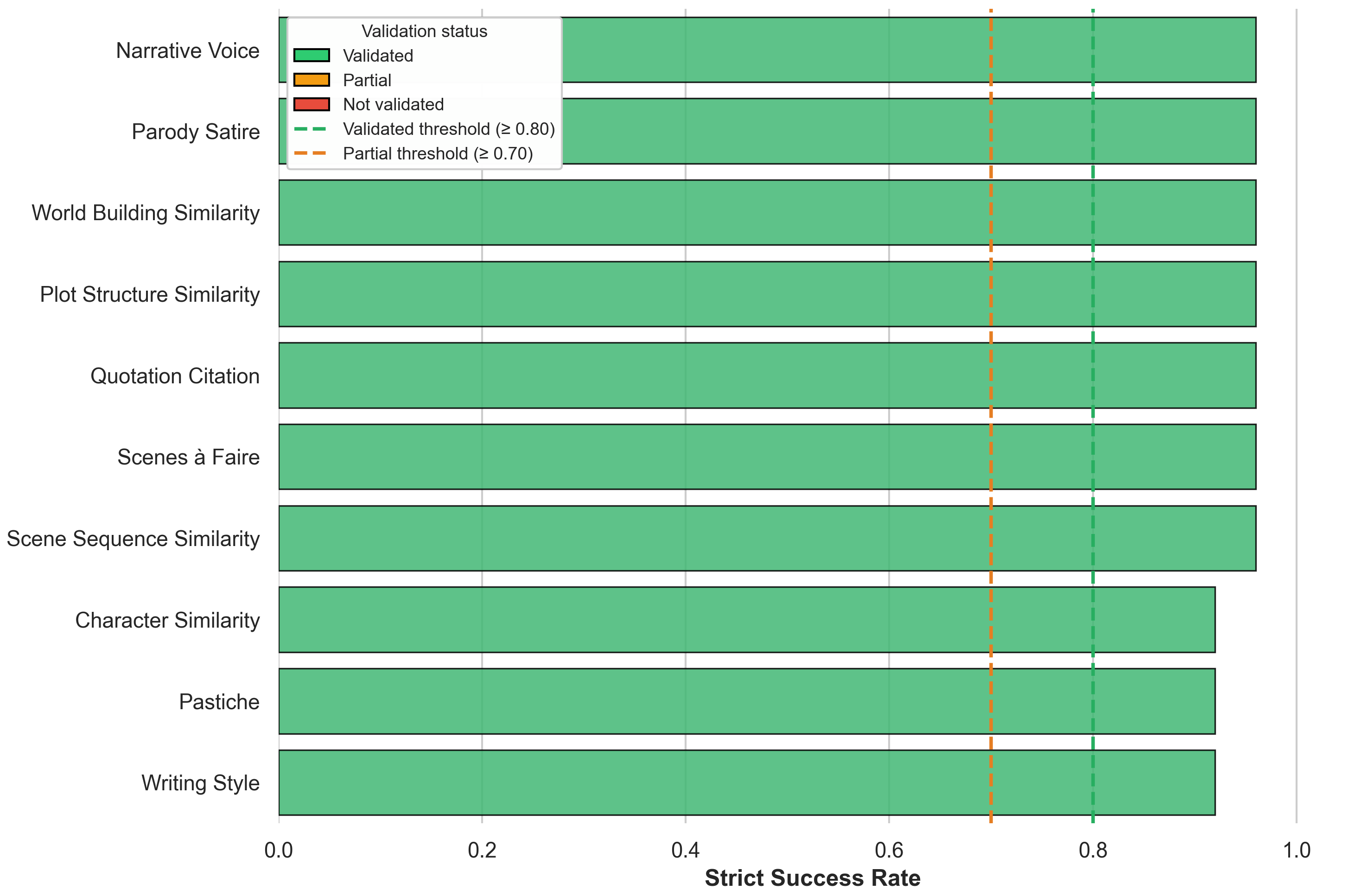}
    }
    
    \caption{Strict alignment rates per evaluator across all ten metrics; all evaluators
    exceed the validated threshold (\(\geq 0.80\)), and the partial and not-validated
    legend entries are included for reference only as no evaluator falls below this
    threshold}
    \label{fig:rq1:validation_success_rate}
\end{figure}

\begin{figure}[!hb]
    \centering
    \adjustbox{max width=\textwidth}{
    \includegraphics[width=\textwidth]{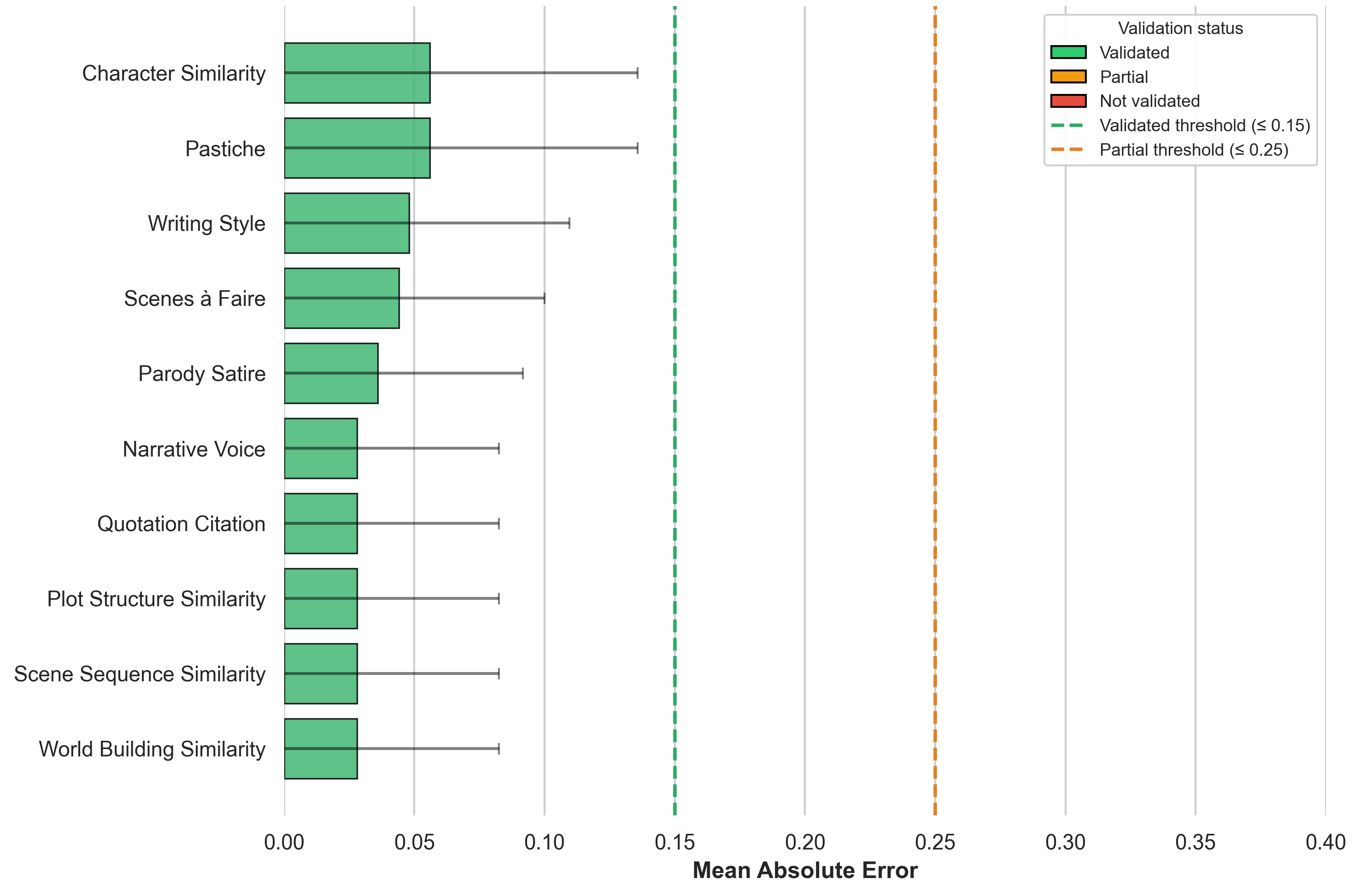}
    }
    
    \caption{Mean absolute error per evaluator across all ten metrics; all evaluators
    remain within the validated threshold (\(\leq 0.15\)), and the partial and
    not-validated legend entries are included for reference only as no evaluator
    exceeds this threshold}
    \label{fig:rq1:validation_mae}
\end{figure}

Figure~\ref{fig:rq1:validation_success_rate} and~\ref{fig:rq1:validation_mae} summarises overall performance per evaluator in terms of strict alignment rate and MAE. All evaluators comfortably exceed the pre-defined validation thresholds: strict alignment is at least $0.92$ and MAE remains well below the $0.15$ ceiling for every metric, leaving no evaluator in the partial or not-validated range. On a five-point scale where adjacent categories are separated by $0.2$, this means that, on average, the evaluators deviate from the intended level by substantially less than one-third of a category. 

\begin{figure}[!hb]
    \centering
    \includegraphics[width=\textwidth]{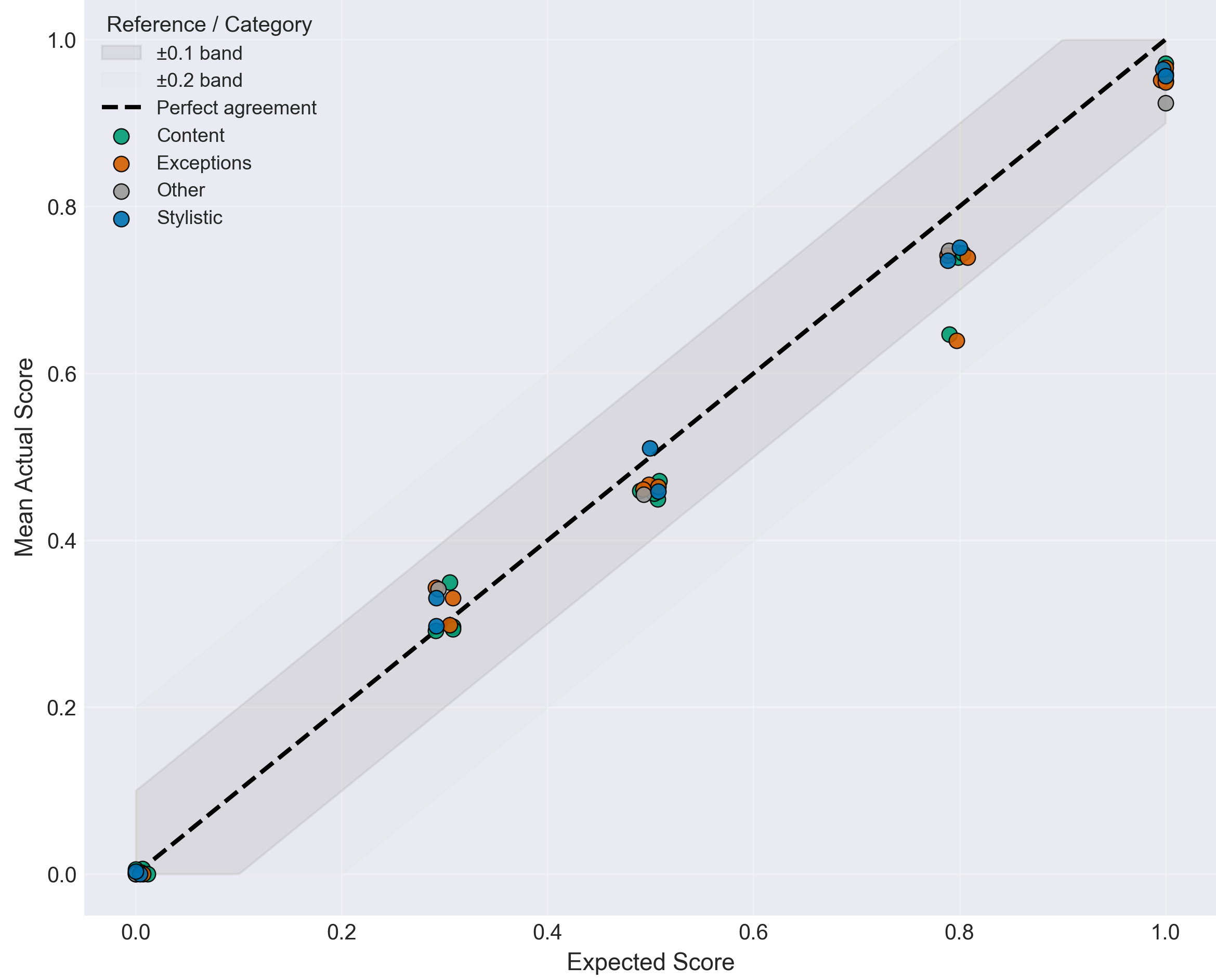}
    \caption{Mean actual score versus expected score for each evaluator across five
    experimental scenarios, with each point representing one evaluator coloured by
    category (Content, Exceptions, Stylistic, Other) and slight horizontal jitter
    applied within each scenario cluster for legibility; two evaluators at the
    \(0.80\) expected score level fall outside the \(\pm 0.1\) acceptable band}
    \label{fig:rq1:expected-vs-actual}
\end{figure}

The DAG-based prompts successfully implement the full range of constructs defined in Section~\ref{sec:methods} with no evaluator persistently near threshold. Performance is also broadly comparable across the three conceptual groups (Stylistic, Content, Exceptions, as examined further in Figure~\ref{fig:rq1:expected-vs-actual}).

The two deviations at the $0.80$ level are informative: Character Similarity and
Pastiche exhibit the highest MAE overall and are the only evaluators producing errors
of this magnitude, reflecting the legal and narratological subtlety of their
constructs.

\subsubsection{Prediction Accuracy and Alignment behaviour}
\begin{figure}[!hb]
    \centering
    \includegraphics[width=\textwidth]{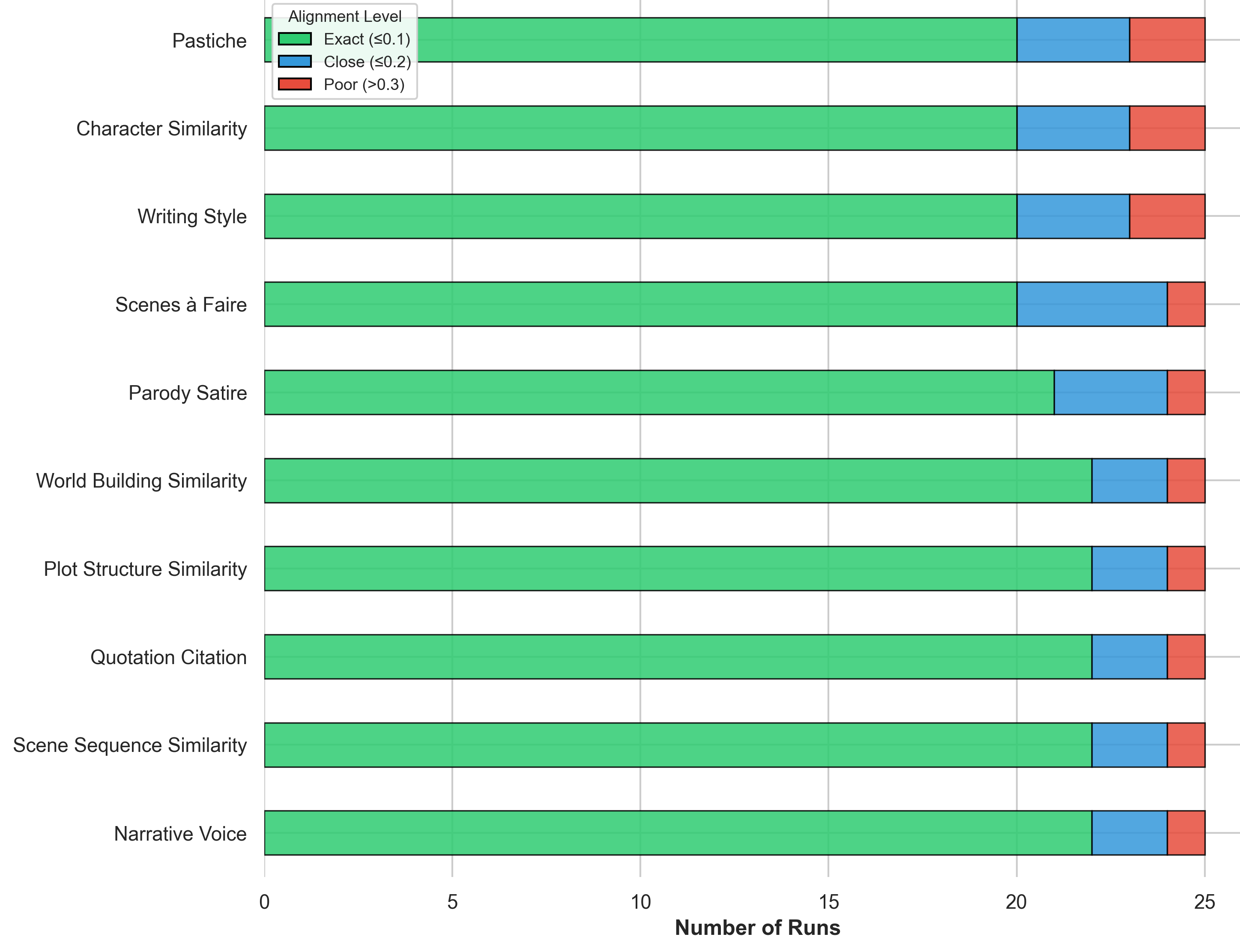}
    \caption{alignment classification per evaluator with predominant exact or close alignment and few poor alignments}
    \label{fig:rq1:alignment-evaluator}
\end{figure}

The alignment analysis in Figure~\ref{fig:rq1:alignment-evaluator} provides a complementary view. Across all evaluators and test cases, more than four-fifths of the \(250\) individual evaluations fall within the EXACT band (\(\pm 0.1\) of the expected score), fewer than one in twenty fall more than one category away, and POOR cases are rare. Errors manifest predominantly as local boundary shifts---rating a \say{very similar} pair as \say{moderately similar}---rather than confusions across qualitatively distinct regions of the scale (Figure~\ref{fig:rq1:expected-vs-actual}).

Character Similarity and Pastiche account for all observed POOR alignments: the judge model either underestimates similarity for near-parallel character pairs or oscillates between classifying a stylised homage as strong pastiche versus weak stylistic imitation. Subsequent analyses therefore interpret modest differences on these two metrics more cautiously than comparable differences on, for example, Plot Structure Similarity or Quotation/Citation.

\subsection{Stylistic Appropriation Through Fine-Tuning}\label{sec:stylistic-appropriation}

We compare baseline models (\texttt{$en_b$}, \texttt{$nl_b$}) with their fine-tuned variants (\texttt{$en_f$}, \texttt{$nl_f$}) on both forget and retain splits to assess whether supervised fine-tuning induces stylistic appropriation beyond verbatim memorisation. Additional results are reported in Appendix~\ref{app:rq2_details}.

\subsubsection{Overview of Fine-Tuning Effects}\label{sec:rq2:overview}
\begin{figure}[!hb]
    \centering
    \includegraphics[width=\textwidth]{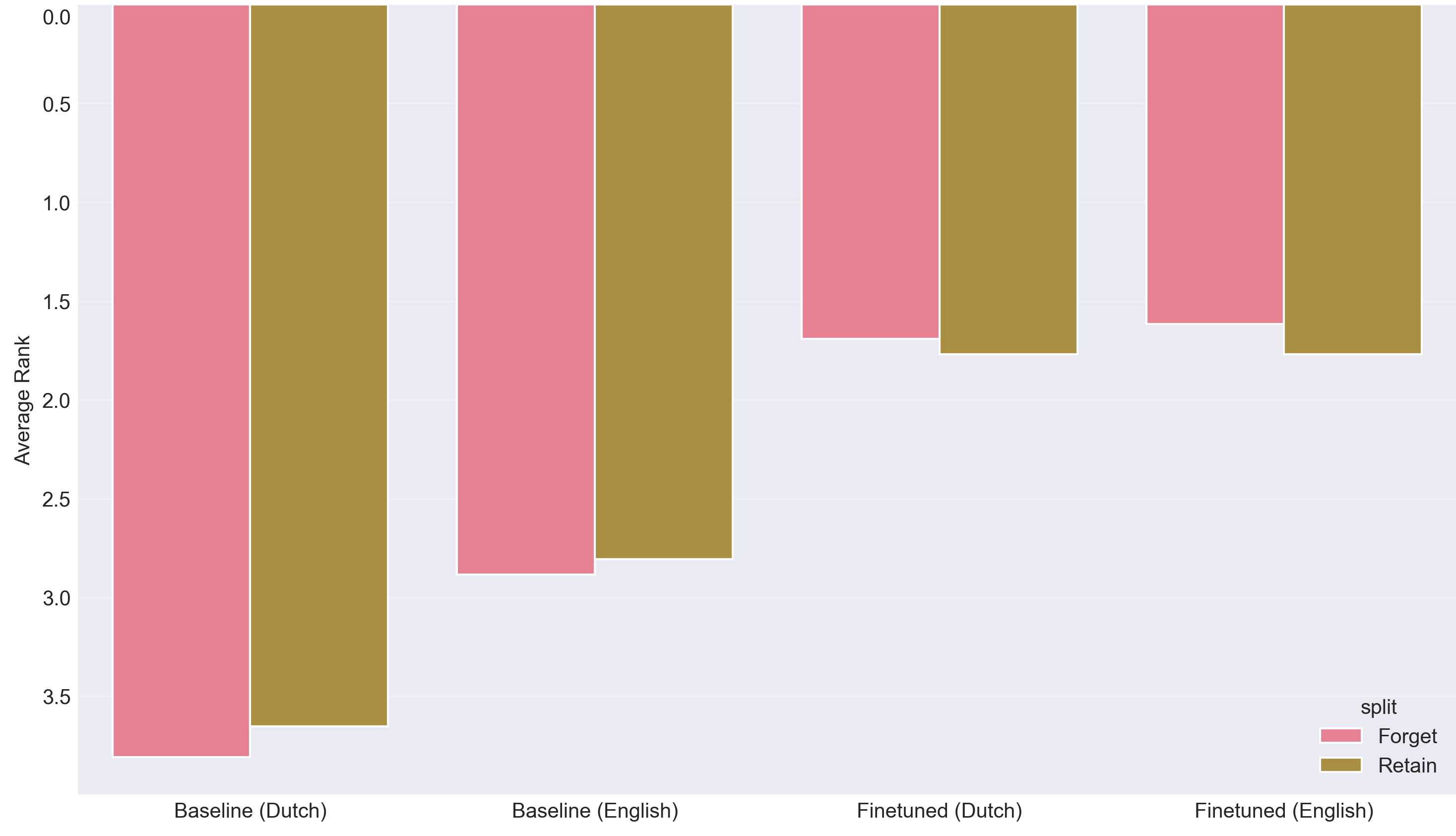}
    \caption{Average rank (lower is better) across all metrics for each model and split of the baseline and fine-tuned models}
    \label{fig:rq2:avg-rank}
\end{figure} 

\begin{figure}
    \centering
     \includegraphics[width=\textwidth]{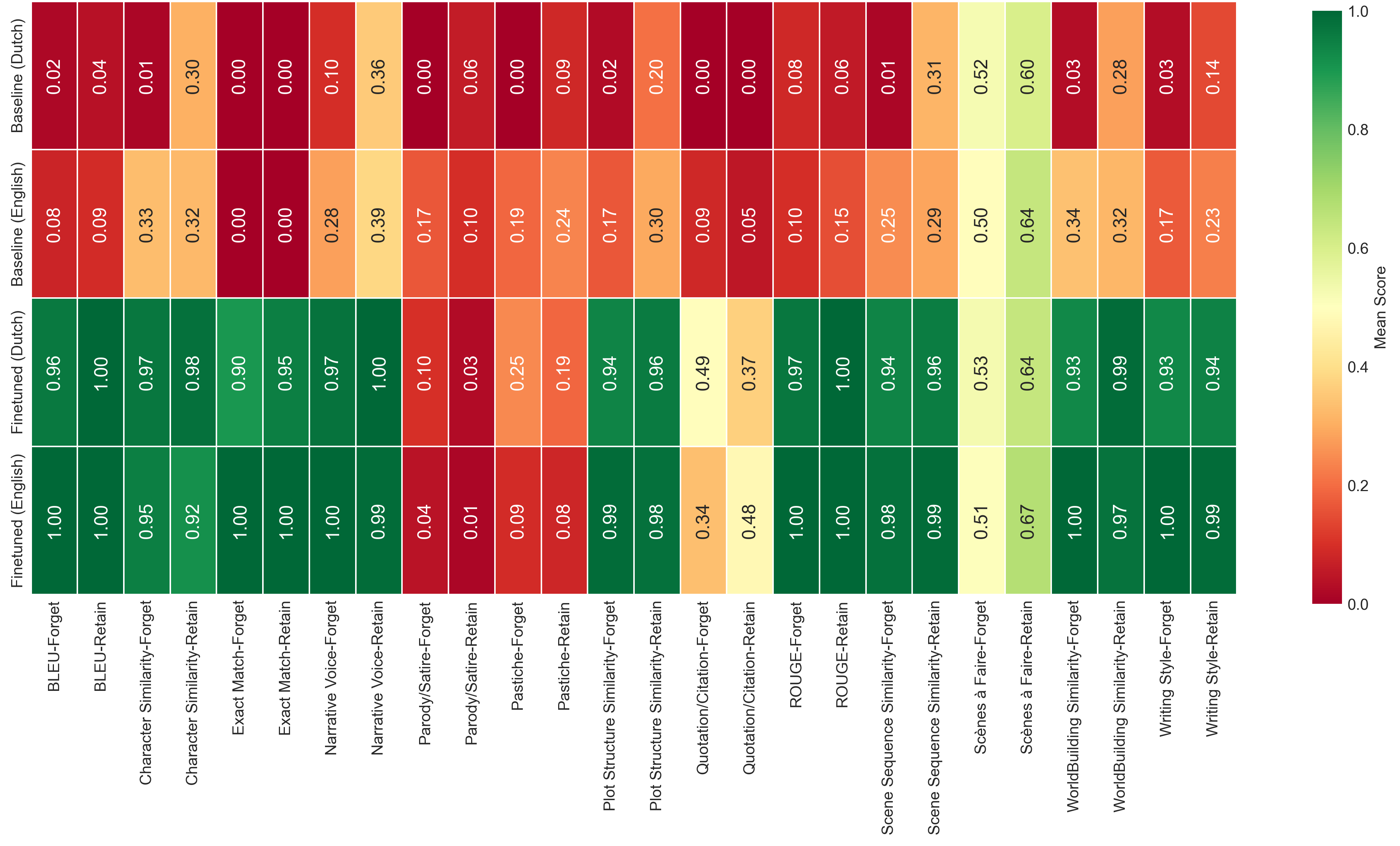}
    \caption{Mean PSALM scores per metric and model condition (forget and retain splits) where the finetuned models show high mean scores (green) and baseline low mean scores (red) with statutory exceptions being low in both cases}
    \label{fig:rq2:heatmap}
\end{figure}

Fine-tuning on the DBNL-derived QA corpus substantially increases similarity between model outputs and the underlying works across all infringement-oriented metrics. Figure~\ref{fig:rq2:avg-rank} shows that, averaged across all $13$ metrics, both fine-tuned models consistently achieve lower (better) ranks than their baselines on both splits; the English–Dutch difference is small relative to the baseline–fine-tuned gap.

The heatmap in Figure~\ref{fig:rq2:heatmap} and the category-level means in Table~\ref{tab:rq2:category-means} show how this improvement is distributed across metric families. For both languages and both splits, fine-tuning drives computational metrics (Exact Match, BLEU, ROUGE) from values close to zero to values very close to one. Stylistic metrics (Writing Style, Narrative Voice) and content metrics (Character, Plot, Scene Sequence, World-Building) follow the same pattern: baseline models exhibit low to moderate similarity, whereas fine-tuned models are typically judged by PSALM as \say{very similar} or \say{near-identical} to the source passages. In contrast, defence-oriented metrics (Parody/Satire, Pastiche, Quotation/Citation, Scènes à Faire) change much less. Quotation/Citation increases moderately, while Parody/Satire and Pastiche remain low and Scènes à Faire stays around the mid-range for all conditions. This is because the corpus does not necessarily address these exceptions, and therefore no questions are asked about aspects such as parody or quotations.

\begin{table}[htbp]
  \centering
  \adjustbox{max width=\textwidth}{
  \begin{tabular}{llcccc}
    \toprule
    Split & Category & Baseline EN & Fine-tuned EN & Baseline NL & Fine-tuned NL \\
    \midrule
    Forget &
    Computational &
    0.05 & 1.00 & 0.04 & 0.97 \\
    & Stylistic &
    0.18 & 1.00 & 0.08 & 0.97 \\
    & Content &
    0.21 & 0.99 & 0.04 & 0.95 \\
    & Exceptions &
    0.21 & 0.25 & 0.13 & 0.37 \\
    \midrule
    Retain &
    Computational &
    0.08 & 1.00 & 0.03 & 0.99 \\
    & Stylistic &
    0.30 & 0.99 & 0.23 & 0.97 \\
    & Content &
    0.27 & 0.97 & 0.25 & 0.98 \\
    & Exceptions &
    0.24 & 0.32 & 0.18 & 0.33 \\
    \bottomrule
  \end{tabular}
  }
  \caption[Category-level mean PSALM scores before and after fine-tuning]{Category-level mean PSALM scores (0 to 1 scale) before and after fine-tuning by language and split}
  \label{tab:rq2:category-means}
\end{table}

Table~\ref{tab:rq2:category-means} also shows that English baselines start from slightly higher similarity than Dutch baselines, reflecting the English-centric pre-training of LLaMA~3.2~1B. After fine-tuning, this difference almost disappears: both languages converge to similarly high scores on computational, stylistic and content categories, and to similar, much lower scores on exceptions. This convergence suggests that the supervised fine-tuning stage is the dominant driver of similarity behaviour, while cross-linguistic differences in the base model matter mainly before training.

\subsubsection{Stylistic and Structural Appropriation}\label{sec:rq2:stylistic-content}
\begin{figure}
  \centering

  \begin{subfigure}[t]{0.95\textwidth}
    \centering
    \includegraphics[width=\textwidth]{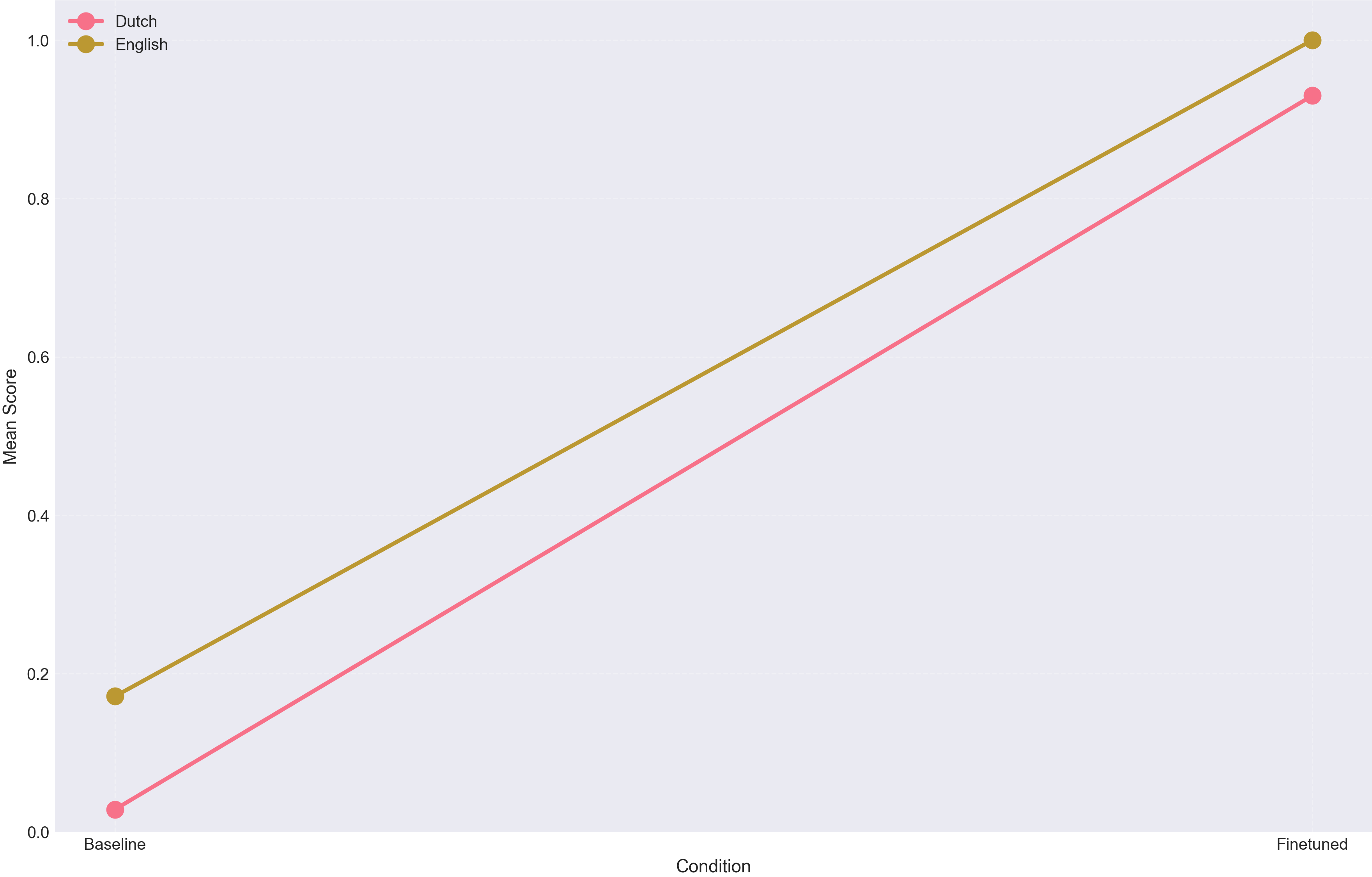}
    \caption{writing style interaction on forget set showing low scores on the baseline and high scores on fine-tuned model}
  \end{subfigure}
  \begin{subfigure}[t]{0.95\textwidth}
    \centering
    \includegraphics[width=\textwidth]{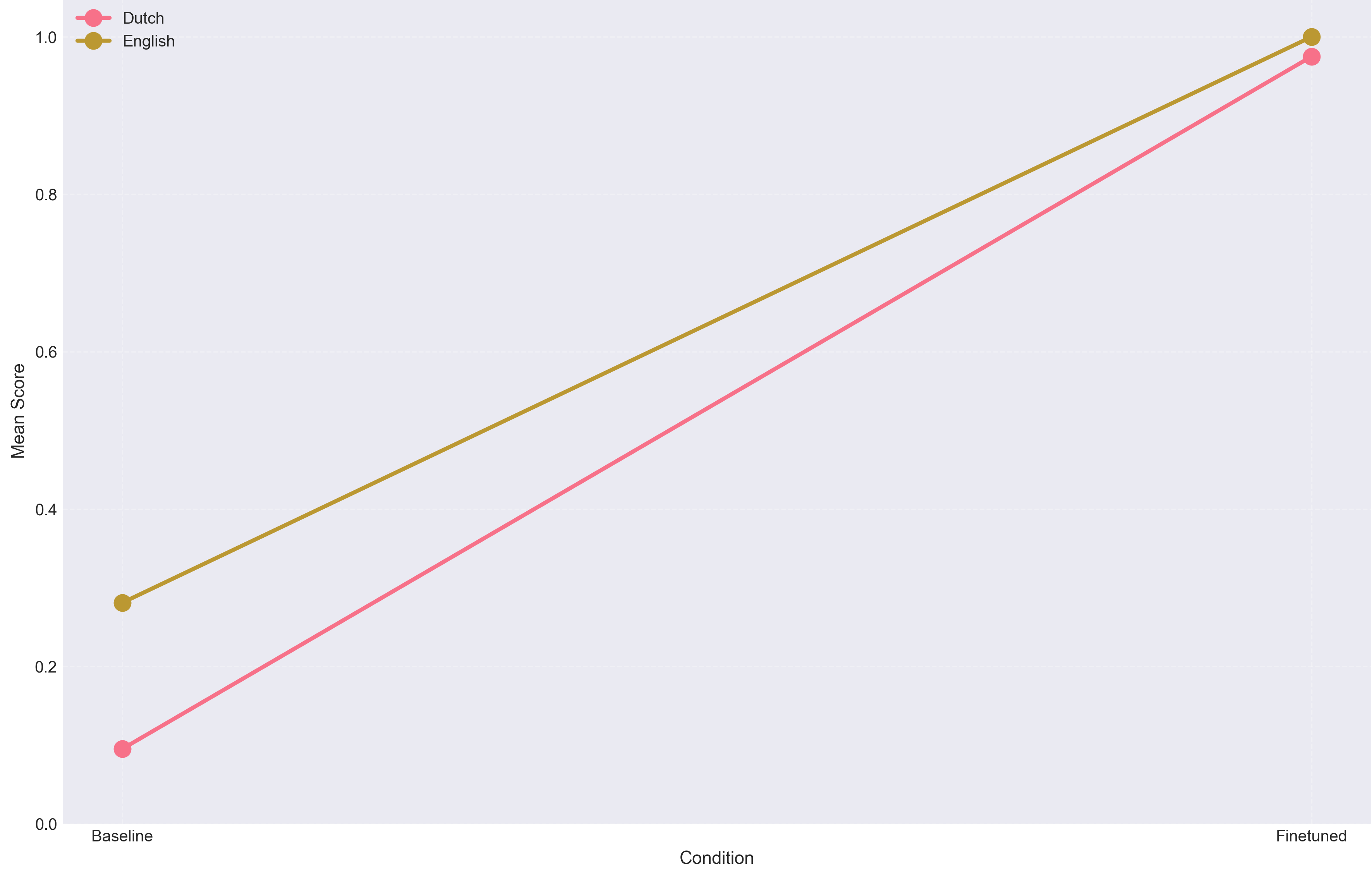}
    \caption{narrative voice interaction on forget set showing low scores on the baseline and high scores on fine-tuned model}
  \end{subfigure}
  \caption{
PSALM scores for baseline against finetuned models on the Forget set showing that finetuning increases stylistic and content similarity writing style narrative voice toward 1 across languages and splits suggesting strong imprinting of author style and other evaluators show similar trends Appendix Figure~\ref{fig:rq2:stylistic-metrics-forget-extended}}
  \label{fig:rq2:stylistic-metrics-forget}
\end{figure}

Figure~\ref{fig:rq2:stylistic-metrics-forget} shows the general trend of the forget-split on stylistic and content evaluators, while Appendix Figures~\ref{fig:rq2:stylistic-metrics-forget-extended} and~\ref{fig:rq2:stylistic-metrics-retain} reports similar results for the other evaluators of the forget-split, as well as for the retain-split.

In the baselines, PSALM assigns low-to-moderate similarity scores on these dimensions. For English, the results are somewhat higher than for Dutch, which is consistent with the English bias in pre-training and with RQ1’s observation that the English baseline already partially overlaps with stylistic features of late 19th century prose, even when translated. However, even the English baseline tends to fall into PSALM’s \say{somewhat similar} or \say{moderately similar} bands, not the \say{very similar} or \say{near-identical} categories.

Fine-tuning moves all stylistic and structural metrics sharply upward. Writing Style and Narrative Voice scores cluster near one for both languages, with PSALM judging the generated answers as exhibiting near-identical lexical and syntactic patterns, rhythm, rhetorical devices, and narratorial perspective. Character, Plot, Scene Sequence, and World-Building similarities likewise rise from near-zero baselines to above \(0.9\) on both splits.

Paired Wilcoxon tests, in Appendix~\ref{app:rq2_details}, confirm that these increases are statistically robust, with large or very large effect sizes across all stylistic and content metrics for both languages and both splits. Crucially, the effect is nearly as strong on the retain split as on the forget split.

Under EU copyright doctrine, these findings are significant because they concern precisely the expressive dimensions courts treat as indicators of originality: narrative voice, character construction, plot architecture, and world-building. PSALM's near-ceiling post-fine-tuning scores indicate that fine-tuned models routinely produce outputs perceived as closely tracking an author's \say{own intellectual creation} across multiple legally significant dimensions, not merely surface text.

\subsubsection{Behaviour of Exceptions}
The defence-oriented evaluators behave very differently from the infringement-oriented ones. Figure~\ref{fig:rq2:heatmap} shows the overall pattern, while additional detailed split-level exception plots are provided in Appendix Figures~\ref{fig:rq2:exception-metrics-forget} and~\ref{fig:rq2:exception-metrics-retain}.

Parody/Satire and Pastiche scores remain low across all conditions. Some paired comparisons reach statistical significance, but absolute differences are negligible relative to shifts on infringement-oriented metrics, with no indication that fine-tuning produces systematically satirical or pastiche-like outputs.

Quotation/Citation is the only exception metric that increases substantially after fine-tuning. In both languages and both splits, scores move from near-zero to moderate levels (roughly one third to one half of the maximum). This reflects the fact that, after fine-tuning, the models often embed verbatim fragments of the source text in their answers in a way that PSALM judges as quotation-like: the answer contains the original wording within a surrounding narrative context. However, PSALM does not model all conditions of the Article 5(3)(d) quotation exception, such as lawful access or proportionality.

Scènes à Faire scores, which estimate how much of the similarity can be attributed to unprotectable stock elements, remain stable across all model conditions. Both baselines and fine-tuned models sit in the middle of the scale (around $0.5-0.67$), with no consistent up- or downward trend.

\subsection{Unlearning Effectiveness via Negative Preference Optimisation}\label{sec:unlearning-effectiveness}
This section compares fine-tuned models (\texttt{$en_f$}, \texttt{$nl_f$}) with their NPO-unlearned counterparts (\texttt{$en_{npo}$}, \texttt{$nl_{npo}$}) on both forget and retain splits. Additional results are reported in Appendix~\ref{app:rq3_details}.

\subsubsection{Overall Unlearning Behaviour}\label{sec:rq3:overall}
\begin{figure}[!hb]
    \centering
    \includegraphics[width=\textwidth]{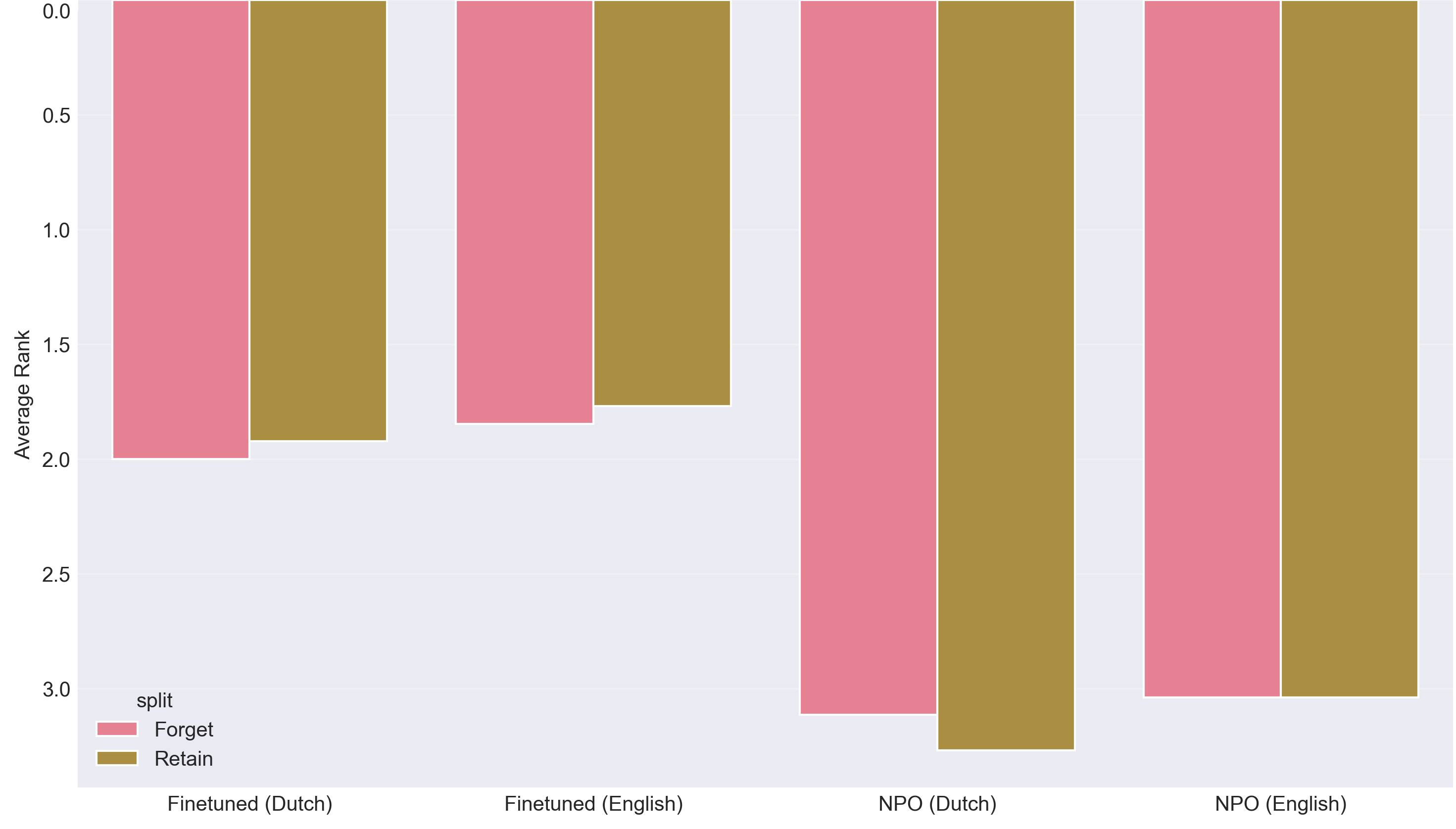}
    \caption{Average rank (lower is better) across all metrics for each model and split of the fine-tuned and unlearned models}
    \label{fig:rq3:avg-rank}
\end{figure}

\begin{figure}[!hb]
    \centering
    \includegraphics[width=\textwidth]{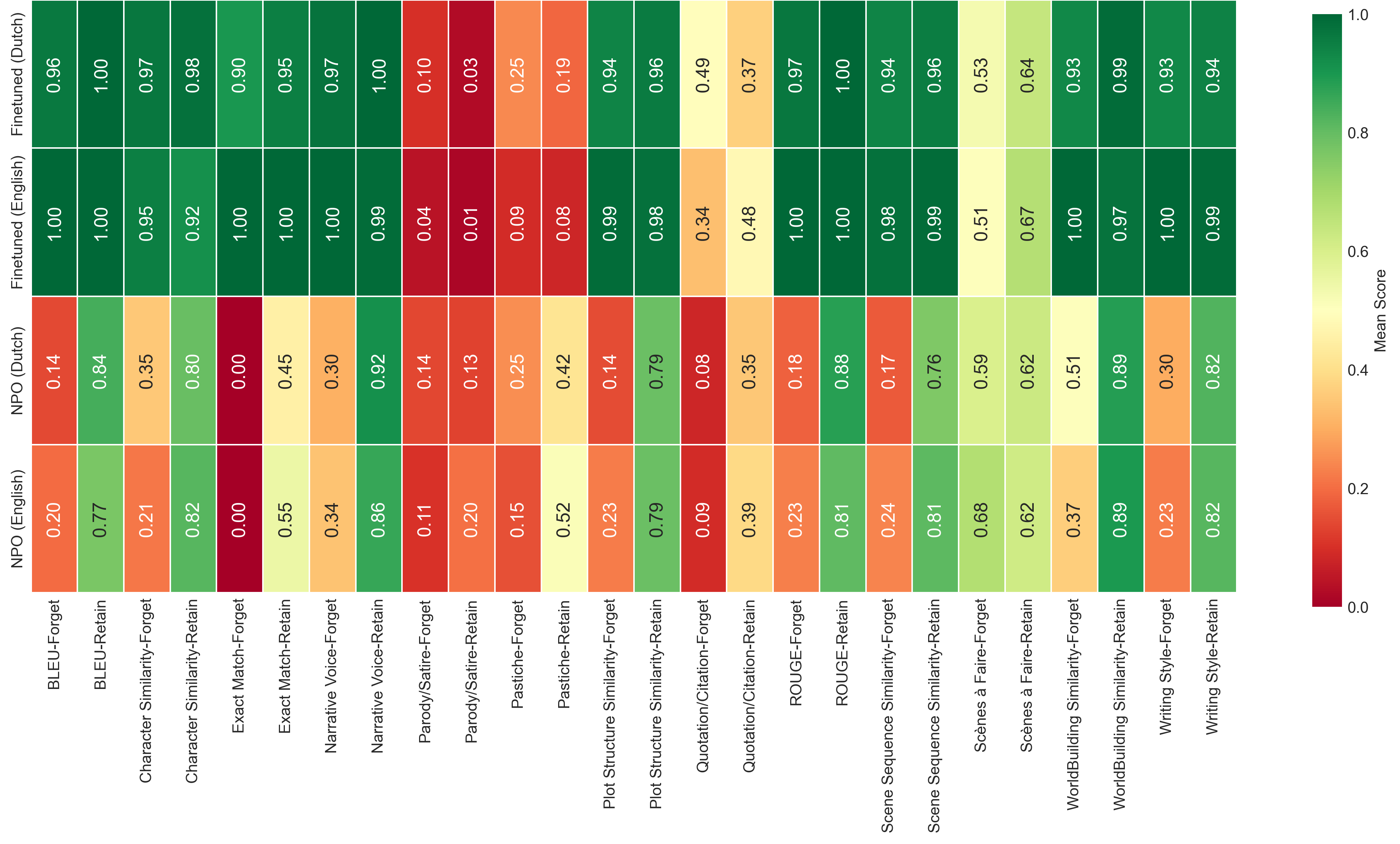}
    \caption{Mean PSALM scores per metric and model condition (forget and retain splits) where the finetuned models show high mean scores (green) and unlearned low mean scores (red) with statutory exceptions being low in both cases}
    \label{fig:rq3:heatmap}
\end{figure}

Across both languages, NPO consistently pushes models away from the training works. Figure~\ref{fig:rq3:avg-rank} shows that the average ranks increases from approximately \(1.8\)–\(2.0\) for the fine-tuned models to roughly \(3.0\) for the NPO variants on both forget and retain splits. However, they do not return to baseline behaviour: even after unlearning, they remain more similar to the sources than the original instruction-tuned models, indicating that some imprint of the fine-tuning corpus persists.

Figure~\ref{fig:rq3:heatmap} clarifies where these changes occur. On the forget split, computational metrics (exact match, BLEU, ROUGE) drop from almost perfect similarity to values close to zero. Stylistic and content categories also decrease sharply, from near \(1.0\) to means around \(0.2-0.3\). Scores in this range correspond roughly to PSALM’s \say{somewhat different} band, so the unlearned models are no longer judged as near-identical to the forget passages, but they are not fully dissimilar either. On the retain split, the same categories drop only part of the way: mean scores remain around $0.8$ for stylistic and content dimensions, i.e., still in the \say{very similar} region.

Exception metrics behave differently as shown in Figure~\ref{fig:rq3:heatmap} and additionally in Appendix Figures~\ref{fig:rq3:exception-metrics-forget} and~\ref{fig:rq3:exception-metrics-retain}, Pastiche and parody/satire remain low throughout and increase only modestly after unlearning, while quotation/citation decreases on the forget split and scènes à faire increases slightly.

\subsubsection{Suppression of Literal Memorisation}\label{sec:rq3:computational}
\begin{figure}
  \centering

  \begin{subfigure}[t]{0.95\textwidth}
    \centering
    \includegraphics[width=\textwidth]{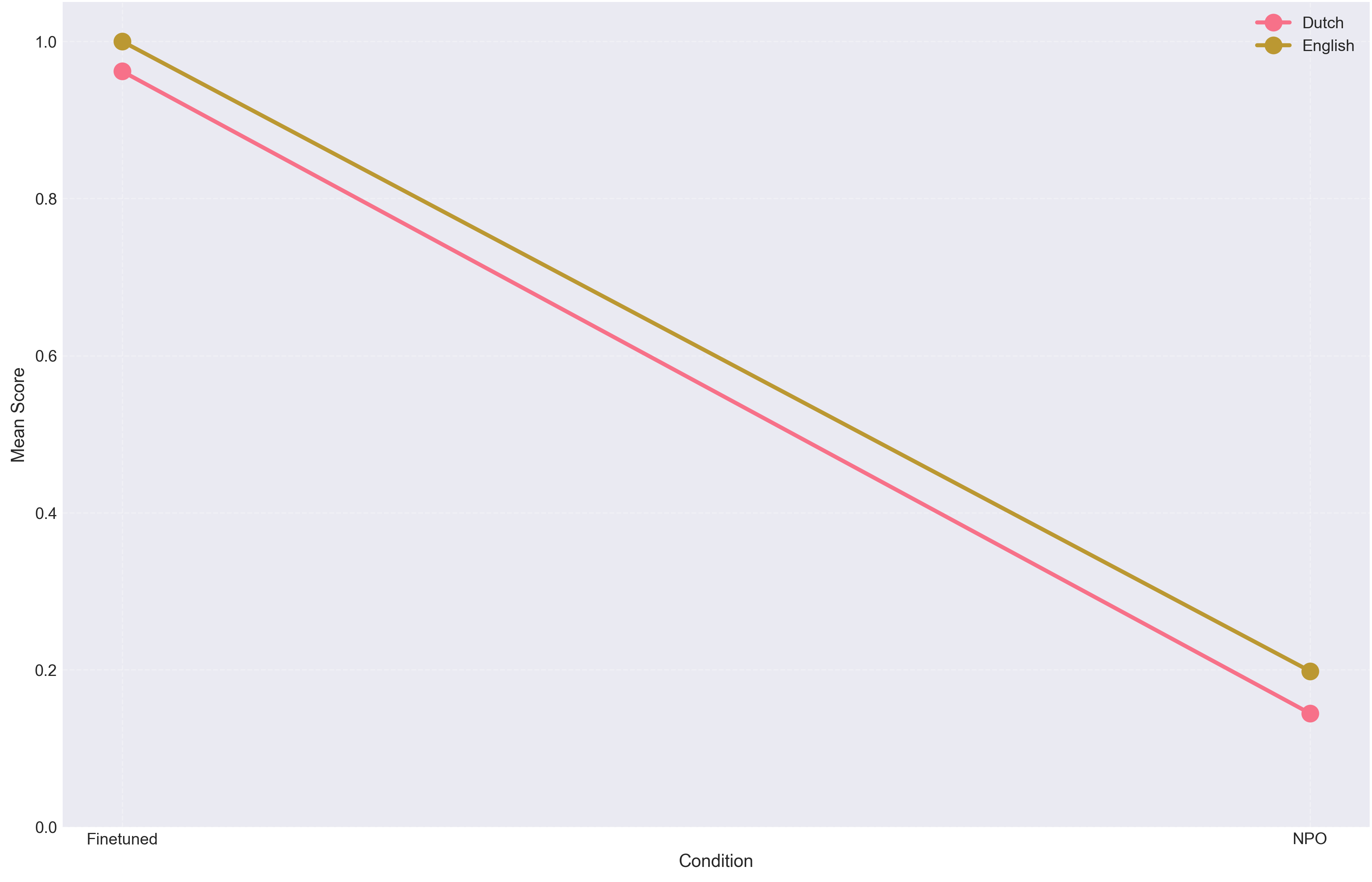}
    \caption{BLEU interaction on forget set showing strong decrease in score from fine-tuned to NPO}
  \end{subfigure}

  \begin{subfigure}[t]{0.95\textwidth}
    \centering
    \includegraphics[width=\textwidth]{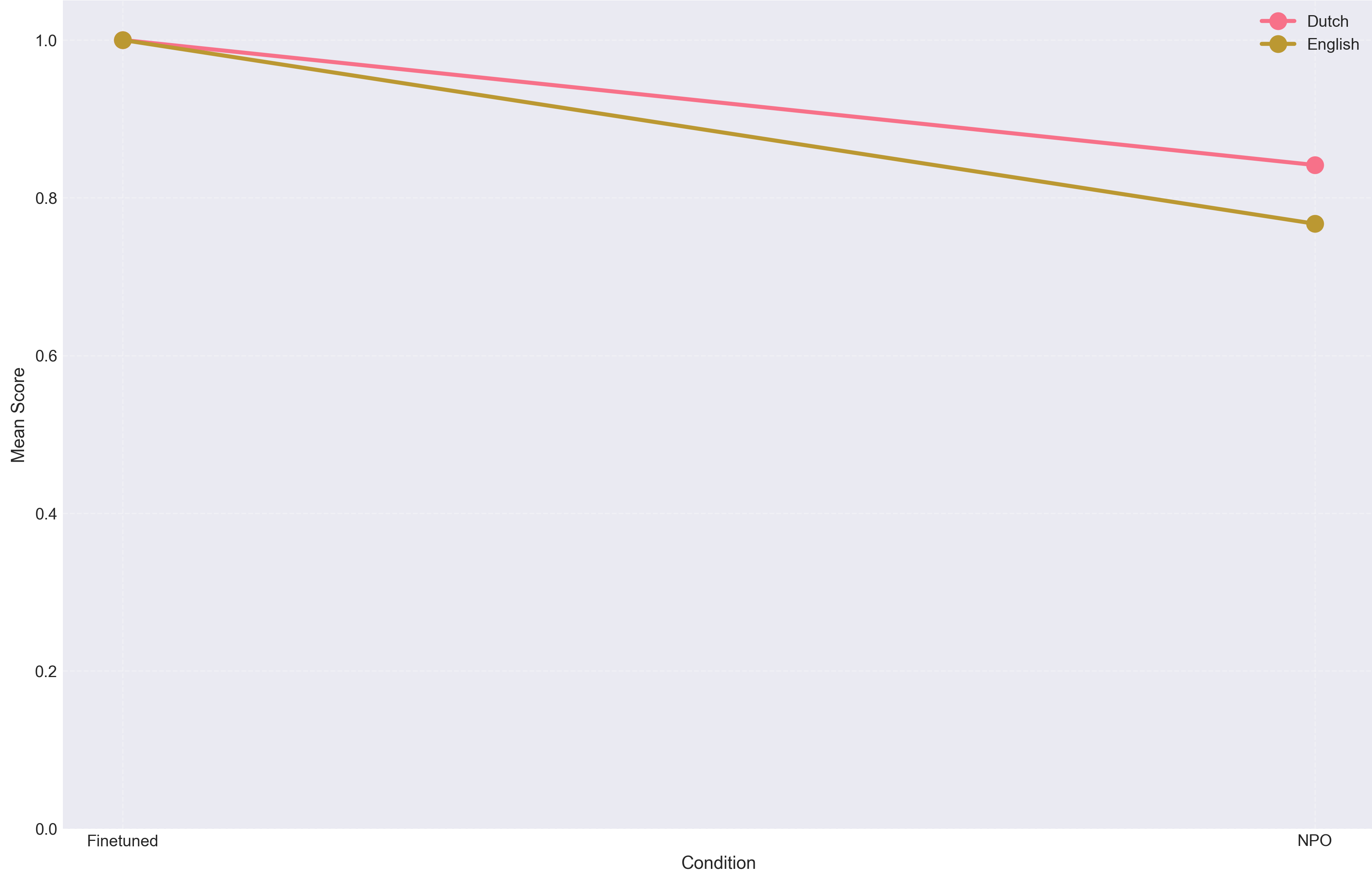}
    \caption{BLEU interaction on retain set showing a slight decrease but still high score from fine-tuned to NPO}
  \end{subfigure}
  
  \caption{
  Computational similarity metric results for finetuned and NPO models where NPO almost eliminates verbatim reproduction on forget sets while leaving substantial overlap on retain sets and other evaluators show similar trends Appendix Figure~\ref{fig:rq3:computational-metrics-extended}}
  \label{fig:rq3:computational-metrics}
\end{figure}

Figure~\ref{fig:rq3:computational-metrics} (and additionally Appendix Figure~\ref{fig:rq3:computational-metrics-extended}) shows how NPO affects the three computational baselines. On the forget split, exact match is effectively eliminated: both languages move from near-perfect exact match rates under fine-tuning to zero matches after unlearning. BLEU and ROUGE follow the same pattern, dropping from values close to one to means around $0.15-0.20$. Rank-based omnibus tests and Wilcoxon signed-rank comparisons (Appendix~\ref{app:rq3_details}) indicate that these reductions are statistically robust and associated with very large effect sizes.

NPO thus addresses the most obvious class of infringements---exact or near-exact reproduction of training passages---but computational metrics say little about whether the remaining outputs are still too close in terms of character, plot, or world-building, dimensions that EU copyright doctrine treats as protectable expression even in the absence of verbatim copying.

\subsubsection{Stylistic and Structural Similarity}\label{sec:rq3:stylistic-content}
\begin{figure}
  \centering

  \begin{subfigure}[t]{0.95\textwidth}
    \centering
    \includegraphics[width=\textwidth]{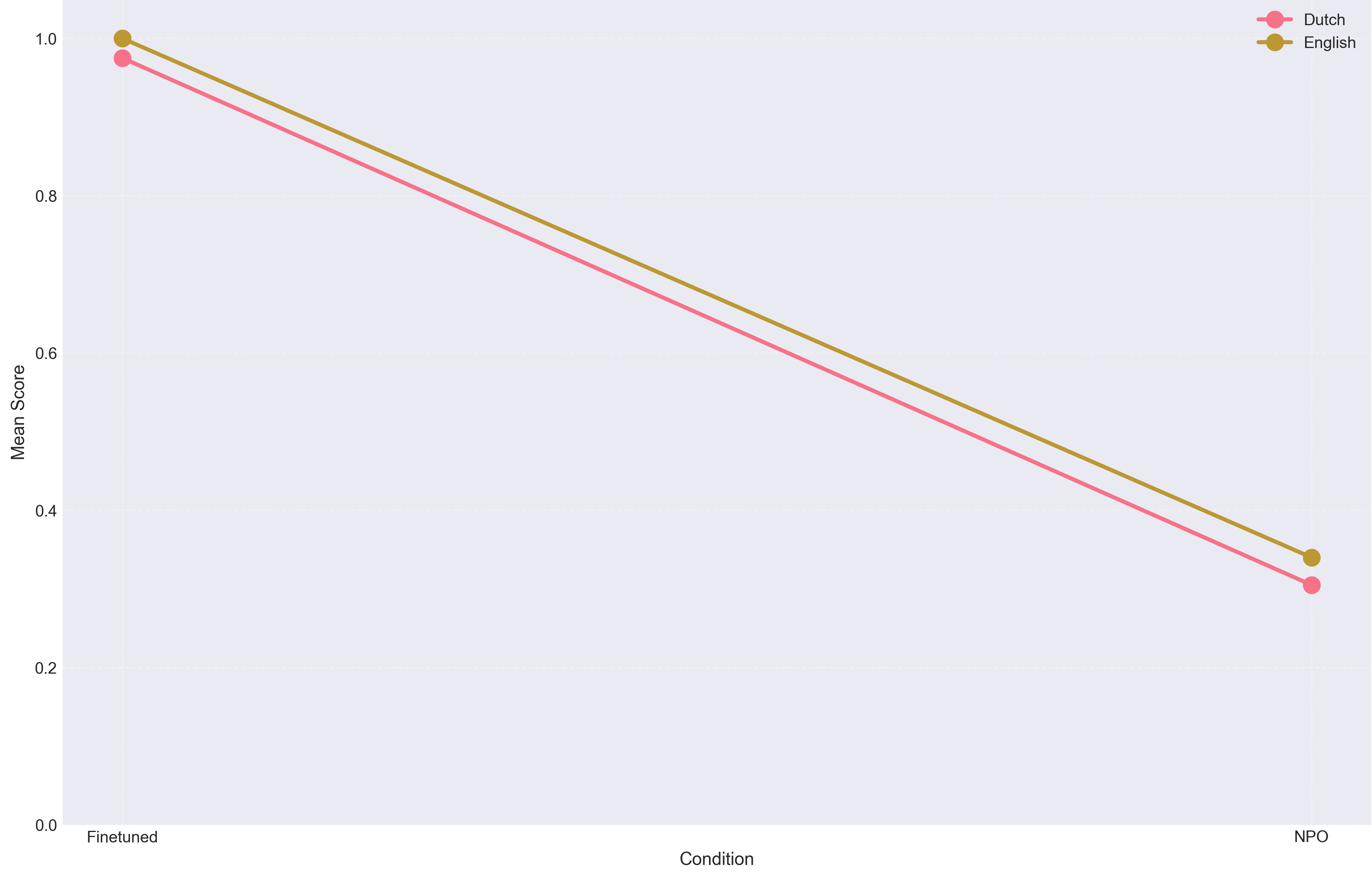}
    \caption{narrative voice interaction on forget set showing stark decline in scores from fine-tuned to NPO}
  \end{subfigure}
  \begin{subfigure}[t]{0.95\textwidth}
    \centering
    \includegraphics[width=\textwidth]{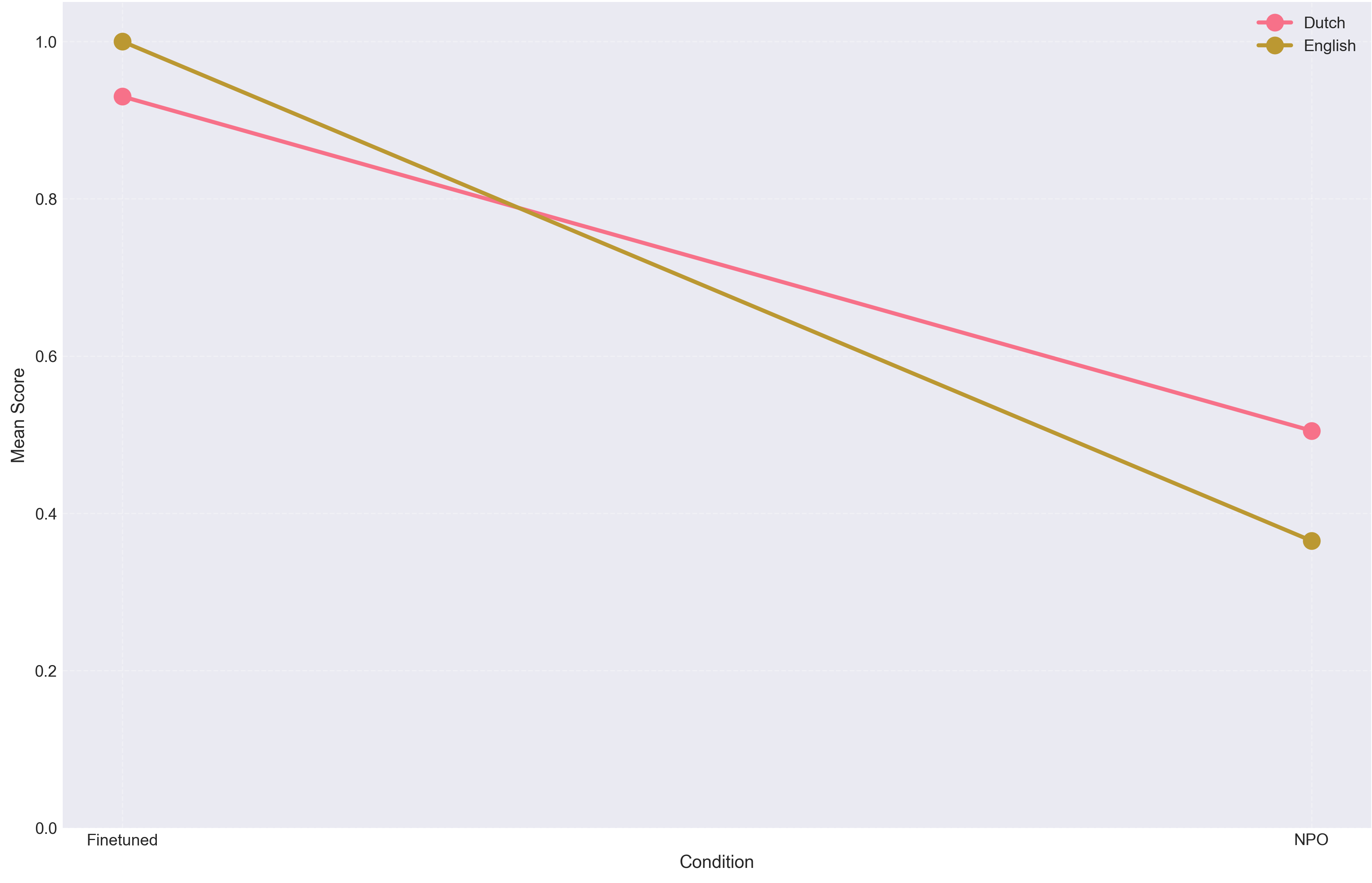}
    \caption{world building interaction on forget set showing stark decline in scores from fine-tuned to NPO with english decreasing more than dutch after NPO}
  \end{subfigure}

  \caption{
  PSALM scores for fine tuned vs NPO models on the Forget set unlearning drastically decreases stylistic and content similarity across languages and splits and other evaluators show similar trends Appendix Figure~\ref{fig:rq3:stylistic-metrics-forget-extended}
  }
  \label{fig:rq3:stylistic-metrics-forget}
\end{figure}

Figure~\ref{fig:rq3:stylistic-metrics-forget} reports PSALM’s forget-split stylistic and content scores before and after unlearning, while Appendix Figure~\ref{fig:rq3:stylistic-metrics-forget-extended} and~\ref{fig:rq3:stylistic-metrics-retain} reports similar results for the other evaluators of the forget-split, as well as for the retain-split. On the forget split, writing style and narrative voice scores fall from near $1.0$ under fine-tuning to roughly $0.25-0.35$ after NPO in both languages. Character similarity, plot structure, scene sequence and world-building show similar reductions, with means in the $0.15-0.35$ range. Rank-based ANOVA and post-hoc analyses in Appendix~\ref{app:rq3_details} confirm that these decreases are large and statistically reliable.

NPO does not only introduce superficial noise: on forget prompts, PSALM judges outputs as having different vocabulary, sentence structure, narrative voice, and character arcs. Average scores move from \say{near-identical} or \say{very similar} to around the boundary between \say{somewhat different} and \say{moderately similar}, indicating that NPO reduces stylistic appropriation alongside literal memorisation, at least for the corpus studied here.

The reduction is nonetheless incomplete. Even on forget prompts the scores do not fall into the \say{clearly different} band (which would correspond to values near \(0\)). After unlearning, some forget pairs are still classified as \say{very similar} or \say{near-identical} on individual dimensions, especially world-building and character similarity (Appendix Figure~\ref{fig:rq3:boxplot}). The distribution has shifted downwards, but not collapsed to baseline. This means that unlearning substantially lowers the probability that a random output will exhibit high stylistic or structural similarity to the training passage, but it does not guarantee that no such outputs exist.

On the retain split, the picture is even more mixed. Stylistic scores for \texttt{en\_npo} and \texttt{nl\_npo} remain high (means around $0.83-0.86$), and content scores stay near $0.80$. Compared to the fine-tuned models, this represents a noticeable softening of similarity, but the absolute levels are still in PSALM’s \say{very similar} region. From a machine learning perspective, this is expected: NPO is optimised primarily to alter behaviour on the forget set while regularising the retain set. Even after unlearning, the models continue to generate outputs that are, on average, judged as highly similar in style and structure to works that remain in the training set. Unlearning therefore cannot be relied upon as a universal safeguard against stylistic appropriation; it may be more precise than a blanket reduction in capacity, but it does not restore models to a pre-fine-tuning state.

Finally, language effects are limited. Figure~\ref{fig:rq3:stylistic-metrics-forget} shows a modest interaction for world-building similarity on the forget split: Dutch models retain slightly higher world-building similarity than English models after unlearning, despite starting from comparable fine-tuned levels. This suggests that the same NPO recipe may be somewhat less effective at disrupting complex world structures in Dutch than in English, possibly because the underlying pre-training distributions differ. For most other metrics, however, English and Dutch behave in parallel.

\subsubsection{Behaviour of Exception Metrics}\label{sec:rq3:exceptions}
Figure~\ref{fig:rq3:heatmap} shows that the exception category changes far less than the computational, stylistic, and content categories, while Appendix Figures~\ref{fig:rq3:exception-metrics-forget} and~\ref{fig:rq3:exception-metrics-retain} additionally show how unlearning affects the four defence-oriented evaluators individually. Across all settings, PSALM almost never classifies outputs as strong parody, strong pastiche or clear quotation; mean scores stay well below the “strong defence” band. NPO does not change this qualitative picture, but it reshapes the balance between different exceptions.

On the forget split, quotation/citation scores decrease substantially after NPO. This is a direct consequence of the reduction in verbatim overlap: the outputs no longer contain long, clearly recognisable fragments of the source passage surrounded by commentary, so PSALM no longer treats them as quotations. 

Pastiche and parody/satire show small upward shifts, especially on the retain split and especially for English. After unlearning, PSALM more often sees outputs as weak homages or as having a slight mocking or humorous character. Scores remain far below the \say{strong parody} or \say{strong pastiche} bands; there is little evidence that NPO pushes models into behaviour qualifying under EU parody or pastiche exceptions. Any increase in parody-like behaviour is a side effect of similarity reduction rather than an intentional transformation.

\subsubsection{Residual Similarity and Distance to Baseline}\label{sec:rq3:residual}
Two complementary analyses assess whether NPO returns models to a state comparable to the baseline: paired comparisons between fine-tuned and NPO models, and equivalence tests comparing NPO models with baselines (Appendix~\ref{app:rq3_details}).

Paired Wilcoxon tests show that for forget prompts, NPO induces very large changes in all computational, stylistic and content metrics. For example, on English forget pairs, mean PSALM scores for plot structure drop by more than half, with effect sizes well into the \say{very large} range; similar patterns hold for narrative voice, character similarity, scene sequence and world-building in both languages. On the retain split, the corresponding changes are smaller but still often statistically significant. This confirms that NPO not only removes literal memorisation, but also substantially reshapes the model’s expressive behaviour on the targeted works.

Equivalence tests using a tolerance of $\pm 0.1$ on the $[0,1]$ scale (roughly half a PSALM category) none of the unlearned models can be considered statistically equivalent to the baselines on any computational, stylistic or content metric (Table~\ref{tab:rq3:tost_retain}). Even after unlearning, the models remain noticeably more similar to the training works than the baseline models. For English, for instance, world-building similarity on the retain split decreases after NPO but remains well above the baseline level; the tests reject the hypothesis that the two are practically indistinguishable. This pattern holds for both languages and for both forget and retain sets.

\begin{table}[htbp]
  \centering
  \adjustbox{max width=\textwidth}{
  \begin{tabular}{llcccccc}
    \toprule
    Metric & Category & Lang. &
    Finetuned mean & NPO mean & Mean diff & TOST\_p & Equivalent \\
    \midrule
    Exact Match               & Computational & EN & $1.00$ & $0.55$ & $-0.45$ & $0.99$ & No \\
    BLEU                      & Computational & EN & $1.00$ & $0.77$ & $-0.23$ & $0.96$ & No \\
    ROUGE                     & Computational & EN & $1.00$ & $0.81$ & $-0.19$ & $0.93$ & No \\
    Writing Style             & Stylistic     & EN & $0.99$ & $0.82$ & $-0.17$ & $0.90$ & No \\
    Narrative Voice           & Stylistic     & EN & $0.99$ & $0.86$ & $-0.13$ & $0.73$ & No \\
    Character Similarity      & Content       & EN & $0.92$ & $0.82$ & $-0.10$ & $0.47$ & No \\
    Plot Structure Similarity & Content       & EN & $0.98$ & $0.79$ & $-0.19$ & $0.93$ & No \\
    Scene Sequence Similarity & Content       & EN & $0.99$ & $0.81$ & $-0.18$ & $0.93$ & No \\
    World-building Similarity & Content       & EN & $0.98$ & $0.89$ & $-0.08$ & $0.30$ & No \\
    \midrule
    Exact Match               & Computational & NL & $0.95$ & $0.45$ & $-0.50$ & $0.99$ & No \\
    BLEU                      & Computational & NL & $1.00$ & $0.84$ & $-0.16$ & $0.86$ & No \\
    ROUGE                     & Computational & NL & $1.00$ & $0.88$ & $-0.12$ & $0.68$ & No \\
    Writing Style             & Stylistic     & NL & $0.95$ & $0.83$ & $-0.12$ & $0.64$ & No \\
    Narrative Voice           & Stylistic     & NL & $1.00$ & $0.92$ & $-0.09$ & $0.31$ & No \\
    Character Similarity      & Content       & NL & $0.98$ & $0.80$ & $-0.18$ & $0.90$ & No \\
    Plot Structure Similarity & Content       & NL & $0.96$ & $0.79$ & $-0.17$ & $0.82$ & No \\
    Scene Sequence Similarity & Content       & NL & $0.96$ & $0.76$ & $-0.20$ & $0.87$ & No \\
    World-building Similarity & Content       & NL & $0.99$ & $0.89$ & $-0.10$ & $0.53$ & No \\
    \bottomrule
  \end{tabular}
  }
  \caption[TOST equivalence tests for retain-set metrics]{
    Two one-sided tests (TOST) for equivalence between fine-tuned and NPO models on the retain split for computational stylistic and content metrics where equivalence bounds are \(\pm 0.1\) on the \([0, 1]\) scale and \(\text{TOST}_p\) is the larger of the two one-sided \(p\)-values such that equivalence is rejected when \(\text{TOST}_p > 0.05\)}
  \label{tab:rq3:tost_retain}
\end{table}

Put differently, NPO moves models in the right direction for copyright compliance (it reduces literal copying and attenuates stylistic and structural overlap) but it does not fully \say{un-fine-tune} them. Appendix Figure~\ref{fig:rq3:boxplot} additionally visualises these score distributions and confirms that, although NPO shifts the distributions downward, many retain-set scores remain in the upper half of the scale.

\section{Discussion}\label{sec:discussion}
\subsection{Core Findings and Technical Significance}
\subsubsection{Feasibility of Automated Similarity Assessment}
\label{sec:disc:validation}
The controlled validation experiments in Section~\ref{sec:controlled-validation} show that the ten PSALM evaluators achieve high strict alignment rates and low mean absolute error on purpose-designed test cases (Figures~\ref{fig:rq1:validation_success_rate} and~\ref{fig:rq1:validation_mae}). The scatter of expected versus mean actual scores in Figure~\ref{fig:rq1:expected-vs-actual} is especially informative: almost all evaluator–test-case pairs fall within the $\pm 0.1$ band, corresponding to at most one category difference on the five-point scale.

The DAG-based prompts and GPT-5-nano judge consistently apply the doctrinal and narratological criteria: errors are rare except at boundaries between adjacent similarity bands.

The higher variance and occasional large deviations for Character Similarity and Pastiche (Figures~\ref{fig:rq1:expected-vs-actual} and~\ref{fig:rq1:alignment-evaluator}) expose a constraint that some constructs are harder for the judge model than others. This is not purely a modelling issue: even human experts disagree about when a character becomes sufficiently distinctive to be protectable, or when stylistic imitation crosses the line into lawful pastiche. The modest instability observed in these evaluators therefore reflects conceptual difficulty rather than a simple calibration error. For downstream use, this means that PSALM is more reliable as a detector of very high or very low similarity than as an arbiter of borderline cases.

A more fundamental constraint is that the validation corpus consists of researcher-designed test pairs constructed to instantiate prototypical similarity regimes. This procedure establishes internal consistency between the evaluator prompts and the judge's interpretations of those prompts. It does not establish that the resulting scores correspond to the judgements a copyright lawyer or court would reach on real disputed material. We therefore treat the validation results as evidence of measurement stability, not of legal validity. Establishing legal validity requires expert-annotated benchmarks (see also our future-work discussion in Section~\ref{sec:disc:future-legal-validation}).

\subsubsection{Fine-Tuning as Mechanism for Stylistic Appropriation}
\label{sec:disc:fine-tuning}
After fine-tuning, the English and Dutch models achieve near-ceiling scores across all infringement-oriented metrics (Figure~\ref{fig:rq2:heatmap}). Table~\ref{tab:rq2:category-means} demonstrates that similarity is not confined to lexical overlap: computational metrics (Exact Match, BLEU, ROUGE-L), stylistic metrics (Writing Style, Narrative Voice) and content metrics (Character, Plot, Scene Sequence, World-Building) all move from low or moderate values in the baselines to means around $0.95$ or higher. The fine-tuned models therefore might not merely \say{remember} specific passages. When prompted with questions that resemble their training data, they tend to reproduce texts that PSALM judges as almost indistinguishable from those passages in style, structure and narrative content.

The magnitude of the stylistic effect exceeds what a characterisation of fine-tuning as a \say{light alignment layer} would predict. Here, however, the QA-style fine-tuning regime effectively rewrites the conditional distribution in the prompt region corresponding to the training questions. The model behaves less like a general-language assistant and more like a retrieval system for the specific works included in the fine-tuning corpus. This pattern is visible in the near-ceiling values in Figure~\ref{fig:rq2:heatmap}, in the detailed forget-split stylistic plots in Figure~\ref{fig:rq2:stylistic-metrics-forget}, and additionally in the appendix split-level computational and retain-split stylistic plots (Appendix Figures~\ref{fig:rq2:computational-metrics} and~\ref{fig:rq2:stylistic-metrics-retain}). From a legal perspective, this means that fine-tuning can create a precision tool for reproducing specific works.

Second, fine-tuning largely erases the pre-training differences between languages. Before fine-tuning, English models already exhibit somewhat higher similarity to the training works than Dutch models, reflecting their English-centric pre-training. After tuning, Table~\ref{tab:rq2:category-means} shows that English and Dutch converge to almost the same similarity levels, with differences of a few hundredths at most. This cross-lingual convergence is not trivial: the Dutch model must learn stylistic patterns from translations of historical Dutch prose into modern Dutch, whereas the English model sees translations into English. The convergence suggests that the QA format and the relatively small corpus encourage the models to internalise very specific patterns of narrative voice, characterisation and plot irrespective of language. The legal implication is that arguments based on the opacity of the pre-training corpus (\say{we do not know whether these works were included}) become less relevant once clear evidence of post-training exposure exists: even a comparatively small fine-tuning corpus can dominate behaviour.

Equally important is what fine-tuning does \emph{not} change. Figure~\ref{fig:rq2:heatmap} shows that the exception category changes far less than the infringement-oriented categories, and additionally the appendix split-level exception plots (Appendix Figures~\ref{fig:rq2:exception-metrics-forget} and~\ref{fig:rq2:exception-metrics-retain}) show that Parody/Satire and Pastiche remain low, Quotation/Citation becomes only moderate, and scènes-à-faire scores stay roughly constant. This means that the strengthened similarity is rarely accompanied by stronger evidence of lawful transformations or unprotectable stock elements. In legal terms, fine-tuning increases the plausibility that outputs will be considered reproductions or derivative works under EU standards, without simultaneously increasing the likelihood that they fall within exceptions or limitations~\citep{Lucchi2025GenAiCopyright,chun2024storysimilarity}. A caveat applies: the QA dataset contains no parody, satire, or quotation prompts, so the low exception scores partly reflect the absence of relevant training signal. Nevertheless, this confirms that fine-tuning on domain-specific QA data does not automatically instil exception-like safeguards.

The QA format provides a particularly strong memorisation signal and may overstate appropriation relative to more varied instruction-tuning pipelines. However, domain-specific QA fine-tuning is common in commercial workflows; the RQ2 results are best read as a plausible upper bound on stylistic appropriation under strong exposure conditions.

\subsubsection{Machine Unlearning: Partial Mitigation and Persistent Residuals}
\label{sec:disc:unlearning}
The post-unlearning experiments yield the most legally consequential finding of this study and provide the clean test of whether style-level similarity can be measured separately from verbatim copying. Figure~\ref{fig:rq3:computational-metrics} shows that NPO is effective at suppressing near-verbatim memorisation on the forget set: Exact Match scores drop to zero and BLEU/ROUGE-L means fall to levels that correspond to substantial paraphrasing. Stylistic and content metrics on the forget split also decrease appreciably (Figure~\ref{fig:rq3:stylistic-metrics-forget}), moving from the \say{near-identical} band to much lower values. This suggests that NPO may be changing the way the model tells the stories or retaining certain aspects, instead of unlearning the stories.

These declines, however, do not return the model to baseline behaviour (Table~\ref{tab:rq3:tost_retain}, Appendix Figure~\ref{fig:rq3:boxplot}). Outputs on the forget split remain highly similar according to at least some PSALM dimensions, particularly world building and character similarity, and none of the metrics on the retain split become statistically equivalent to the baseline within a tolerance of $\pm0.1$. NPO therefore operates as a blunt instrument: it significantly shifts the score distributions but cannot surgically excise the influence of the forget set.

Quotation-Citation scores decline on the forget split after NPO (Appendix Figure~\ref{fig:rq3:exception-metrics-forget}), indicating that the unlearned models are less likely to embed recognisable fragments of the source passages with surrounding commentary. This is technically consistent with NPO’s goal of reducing verbatim fidelity, but legally ambivalent. Quotations, when properly attributed and proportionate, can be lawful under Article~5(3)(d) of InfoSoc~\citep{eucopyright,cjeu2019funke}; by pushing outputs away from quotation-like behaviour, unlearning may remove one potential defence without fully eliminating expressive similarity. At the same time, scènes à faire scores rise moderately, which can be read as the model falling back on generic genre patterns once specific expressive choices have been attenuated. From a doctrinal perspective this shift may be positive, as it moves more shared elements into the unprotectable domain; from a creative perspective it risks flattening outputs into formulaic narratives.

Cross-linguistic differences under NPO are small but revealing. The Dutch models retain slightly higher world-building similarity on the forget split than the English models, despite starting from comparable fine-tuned levels (Figure~\ref{fig:rq3:stylistic-metrics-forget}). One plausible explanation is that the Dutch translations preserve more of the spatial and institutional structure of the original works, making these features harder to disrupt without also harming general coherence. This highlights a tension: the aspects of works that are most structurally central (precisely those likely to be part of the author’s \say{own intellectual creation}) may also be the hardest to forget without significant utility loss.

NPO effectively reduces the most blatant forms of memorisation and shifts the overall distribution of stylistic similarity favourably, but non-trivial residual similarities remain, particularly for narrative structure and world-building. Unlearning should not be regarded as a guarantee of compliance; it demonstrates a reduction of risk, but not its complete removal.

\subsubsection{Stylistic Memorisation as Distinct Phenomenon}
\label{sec:disc:stylistic-distinct}
A central conceptual contribution of the experiments is the empirical separation between verbatim memorisation and stylistic memorisation. The results show that NPO reduces computational metrics more strongly than stylistic and content metrics. Fine-tuning followed by unlearning can therefore produce models that rarely emit long verbatim sequences yet still generate outputs that PSALM judges as highly similar in narrative voice, characterisation or plot structure.

This finding has two important implications. First, it helps to explain why technical efforts focused solely on exact-match metrics, such as those captured by BLEU or longest common substrings, may give a false sense of security~\citep{ippolito2023preventing}. A model can pass all computational overlap tests and still, in the sense operationalised by PSALM, \say{tell the same story in the same way}. Second, it reinforces the need for evaluation frameworks that align more directly with legal tests. EU originality and substantial similarity standards~\citep{CourtInfopaqDanske2009, CourtCofemelGstar2019,Lucchi2025GenAiCopyright} focus precisely on expressive choices in plot, character and style, not on verbatim copying as such.

\subsection{Legal Interpretation and Compliance Practice}
\subsubsection{The Measurement-Determination Gap and Its Implications}
\label{sec:disc:measurement-gap}
Although PSALM delivers structured measurements, legal infringement remains a context-dependent determination. This gap between measurement and legal decision-making is already visible in the validation stage. For example, a Writing Style score of $0.8$ indicates a high degree of similarity in lexical complexity, sentence structure and rhetorical patterns, but it does not by itself tell us whether the output competes with the original in the marketplace, whether it was produced with access to the protected work or whether the use is justified by a limitation or exception.

The gap is especially apparent in relation to exceptions and limitations. Article~5(5) InfoSoc~\citep{eucopyright,geiger2015three} requires that exceptions be applied only in certain special cases, not conflict with normal exploitation and not unreasonably prejudice legitimate interests. PSALM's parody, pastiche and quotation evaluators encode elements of the corresponding doctrinal tests, such as evocation, noticeable differences and legitimate purpose, but they do not incorporate evidence about market substitution, the existence of licensing schemes or the broader cultural context. For instance, the moderately increased scènes-à-faire scores after unlearning (Figure~\ref{fig:rq3:exception-metrics-forget}) suggest that more of the retained similarity lies in stock elements, but they do not by themselves prove that the use is free from conflict with normal exploitation.

A further limitation is that PSALM models only a subset of the exceptions and limitations potentially relevant under EU copyright law. It does not include dedicated evaluators for, among others, illustration for teaching or scientific research under Article~5(3)(a) InfoSoc, or incidental inclusion of a work in other material under Article~5(3)(i) InfoSoc. As a result, an output that contains a short excerpt for a legitimate teaching or research purpose, or that includes protected material only incidentally, may receive high scores on infringement-oriented dimensions without PSALM recognising a potentially applicable defence. The system may therefore overstate legal risk if its scores are read as a comprehensive assessment of all exceptions and limitations. Its exception-related outputs should instead be understood as limited to the specific doctrines operationalised in the current framework.

Accordingly, PSALM should be read as a diagnostic tool for similarity and exception-like features, consistent with our validation caveat in Section~\ref{sec:disc:validation}. Human judgment remains necessary for infringement determinations that depend on context such as market impact and the Article~5(5) three-step test.

\subsubsection{TDM Exceptions, the Three-Step Test and Residual Similarity}
\label{sec:disc:tdm-exceptions}
The empirical results cast new light on ongoing debates about the scope of the TDM exceptions in the CDSM Directive and their interaction with the three-step test in Article~5(5) InfoSoc. If, as some commentators argue~\citep{Margoni2022TDM,quintais2025genaicopyright,Lucchi2025GenAiCopyright}, training LLMs on lawfully accessible works is covered by Articles~3 or~4 CDSM, a separate question is whether outputs that closely resemble those works remain within the shelter of the exception.

The fine-tuning experiments show that, when exposed to specific works, LLMs can and do generate outputs that are highly similar across multiple expressive dimensions (Figure~\ref{fig:rq2:stylistic-metrics-forget}; Appendix Figure~\ref{fig:rq2:stylistic-metrics-retain}) while exhibiting only weak parody, pastiche or quotation characteristics (Figure~\ref{fig:rq2:heatmap}; Appendix Figures~\ref{fig:rq2:exception-metrics-forget} and~\ref{fig:rq2:exception-metrics-retain}). The unlearning experiments then show that, even after targeted mitigation, a non-negligible tail of high-similarity outputs persists (Appendix Figure~\ref{fig:rq3:boxplot}). Under a strict reading of Article~5(5), such residual similarity may still be considered conflicting with normal exploitation, especially if outputs can serve as substitutes for the original works. Under a more permissive reading, residual similarities that mainly involve scènes à faire and lack quotation-like reproduction might be treated as within the analytical scope of TDM, particularly if models are combined with contractual or technical measures limiting harmful uses.

\subsubsection{Interpretive Positions on Residual Similarity}
\label{sec:disc:legal-positions}
The empirical patterns can be seen as compatible with several doctrinally grounded interpretive positions, rather than pointing to a single legal conclusion.

A strict liability position would treat any unauthorised reproduction of protected expression as infringing regardless of the technical mitigation steps taken. On this view, the high-similarity outliers that remain after NPO (Appendix Figure~\ref{fig:rq3:boxplot}) are sufficient to impose liability for infringing outputs, though unlearning might influence the proportionality of remedies.

A materiality-based position would focus on whether residual similarities are substantial enough to matter. The movement of mean scores from around~$1.0$ to around~$0.3$ after NPO on the forget split (Figure~\ref{fig:rq3:stylistic-metrics-forget}) could be interpreted as a shift from clearly infringing behaviour towards minimal similarity, especially if the remaining overlap is concentrated in generic narrative patterns rather than detailed expressive choices. From this perspective, the main concern would be the extent of the high-similarity outliers, rather than the average.

A reasonable efforts position would emphasise process and due diligence. Here, PSALM could be used to argue that developers have taken meaningful steps to reduce the most problematic forms of similarity (verbatim copying, highly specific stylistic mimicry) while accepting that some residual risk remains. Such a position might be appealing to regulators seeking incentives for continuous improvement rather than binary judgments.

\subsubsection{PSALM within Compliance Architectures}
\label{sec:disc:compliance}
PSALM could become useful when embedded in broader compliance architectures. 
For model developers, Figures~\ref{fig:rq2:heatmap} and \ref{fig:rq3:heatmap} show how it can be used to quantify the impact of design choices such as fine-tuning regimes and unlearning parameters. 
For regulators and auditors, PSALM provides a standardised basis for comparing models or mitigation methods against legally relevant dimensions. For rightsholders, high similarity scores after claimed compliance efforts could serve as a basis for contesting the adequacy of those efforts. Courts and legal experts could draw on PSALM analyses as one form of technical evidence among others when assessing substantial similarity and the applicability of exceptions.

\subsection{Methodological Contributions and Limitations}
\subsubsection{Hierarchical Implementation of Legal Constructs}
\label{sec:disc:operationalisation}
A methodological contribution of this work is the translation of legal constructs into hierarchical, DAG-based evaluators. By decomposing notions such as narrative voice, character similarity or parody into sub-dimensions with explicit prompts and weights (Section~\ref{sec:foundational-legal}), PSALM tries to make the otherwise opaque judgement of an LLM-as-a-judge more transparent. The validation results in Figures~\ref{fig:rq1:validation_success_rate} and~\ref{fig:rq1:validation_mae} show that this architecture leads to coherent and controllable behaviour across diverse test cases.

The selection of sub-dimensions and their weights reflects a specific interpretive reading of EU copyright doctrine and narratology---one that is explicit and open to critique. These weights are heuristic rather than empirically learned or legally authoritative. They were not calibrated against expert-labelled infringement decisions, nor do they claim to estimate the relative importance that a court would assign to each factor in a concrete dispute. Different legal scholars might reasonably prioritise other sub-dimensions, assign greater weight to particular forms of protectable expression, or include additional contextual factors. PSALM's methodological advantage is therefore not that its weighting scheme is definitive, but that the scheme is explicit, inspectable and reconfigurable. This supports a more rigorous dialogue between developers and legal experts about how doctrine is operationalised.

Relatedly, the current DAG structure is static during inference and relies on fixed weighted aggregation. This is a deliberate simplification. Legal reasoning is not always additive: some elements operate as threshold conditions, some are relevant only if other conditions are satisfied, and some may be defeated by contextual considerations. A high score on one sub-dimension can mathematically compensate for a low score on another, whereas in legal reasoning the absence of a necessary element may defeat a claim or defence altogether. Conversely, a contextual factor not represented in the graph may be legally decisive despite having no effect on the aggregate score. PSALM's aggregate outputs should therefore be interpreted as diagnostic similarity measurements, not as formal models of legal syllogism or defeasible legal reasoning.

Future versions could combine DAG-based measurement with rule-based gates, expert-calibrated thresholds or argumentation-based models that better represent conditional and defeasible doctrinal structures. A particularly interesting direction is to extend the evaluation with multi-agent or argumentation-based components to improve audibility and to better represent conditional and defeasible aspects of legal reasoning.

\subsubsection{Corpus and Model Constraints}
\label{sec:disc:corpus-limits}
The choice of corpus and model architecture introduces several limitations. The texts are drawn from historic public-domain Dutch literature, translated into modern Dutch and English using GPTOSS-120B and, for a subset, GPT-4o-mini (Section~\ref{sec:dataset}). This translation layer may itself impose stylistic regularities on the corpus. For instance, if the translators prefer certain syntactic structures or discourse markers, the fine-tuned models may learn to reproduce those patterns, and GPT-5-nano may treat them as strong signals of similarity. Manual spot-checking mitigates obvious errors, but subtler systematic biases introduced by the translators remain uncontrolled.

The reliance on LLaMA~3.2~1B Instruction is also a constraint. The model’s small size was necessary given the training and unlearning workload and the available hardware (Section~\ref{sec:infrastructure}). The qualitative patterns of memorisation and unlearning are likely to generalise, but quantitative magnitudes may not: larger models might memorise more, require more aggressive unlearning, or respond differently to NPO.

\subsubsection{LLM-as-a-Judge Considerations}
\label{sec:disc:judge-limits}
Using GPT-5-nano as the judge model introduces dependencies on its training data, alignment, and internal biases. The controlled experiments in Section~\ref{sec:controlled-validation} indicate that GPT-5-nano can apply the PSALM prompts consistently, but they do not guarantee alignment with human legal reasoning. A related limitation is that the current implementation uses a single judge architecture across the nodes of an evaluator. The same underlying agent is therefore responsible for transforming, analysing, assessing, and judging. This design improves consistency and experimental control, but it also limits role specialisation. In legal and literary analysis, these dimensions may require different forms of expertise, different evidentiary standards, or different reasoning procedures that suit different agents.

A central limitation is therefore the absence of independent expert or human ground truth. The validation set tests whether the judge model applies PSALM's own evaluative criteria in a stable manner; it does not test whether those scores correlate with the assessments that copyright lawyers, literary experts, or courts would reach on real disputes. Accordingly, the reported alignment rates and error measures indicate internal reliability with respect to PSALM's prompt-defined criteria, not agreement with legal decisions (see Section~\ref{sec:disc:validation}).

Moreover, the overall pipeline employs different models for translation (GPTOSS-120B, GPT-4o-mini), generation (LLaMA) and judgement (GPT-5-nano). This heterogeneity reduces some risks of self-evaluation, but it also means that model-specific quirks at the translation stage may interact with the judge's preferences in ways that are hard to predict.

A further limitation is the lack of adversarial testing. All evaluations use relatively straightforward prompts. It remains unknown how robust GPT-5-nano's judgements are to strategically crafted inputs that attempt to manipulate similarity scores.

\subsubsection{Experimental Design Trade-offs}
\label{sec:disc:design-tradeoffs}
The experimental design involves several trade-offs that influence how the results should be interpreted. The QA format strongly associates each question with a specific passage, accentuating memorisation but also approximating realistic use-cases where domain experts fine-tune models on QA pairs. The resulting evaluations are primarily passage-level rather than book-level. This makes the experiments tractable and allows controlled comparison between source passages and generated outputs, but it limits the framework's ability to detect similarities that emerge only across longer narrative spans, such as chapter-level pacing, cumulative character development, recurring motifs, or whole-work plot architecture.

The forget/retain partition combines author-level and book-level forgetting, exposing both coarse and fine-grained unlearning, yet it does not simulate long sequences of heterogeneous takedown requests. Nor does it test whether a generated chapter or full-length work remains substantially similar to a source book when similarity is distributed across many non-contiguous passages. Generation parameters are chosen to produce coherent, moderate-variance outputs; different settings might yield different memorisation profiles.

These trade-offs emphasise that the study examines a deliberately demanding but localised configuration. The results indicate what can happen under strong exposure to specific works and passage-level prompting, rather than constituting precise predictions for full-book generation, long-context comparison or large-scale deployment pipelines.

\subsection{Future Research}
\subsubsection{Legal Expert Validation Studies}
\label{sec:disc:future-legal-validation}
Validation against human copyright lawyers, literary scholars, and judges is the most pressing next step. A benchmark of source–target pairs drawn from the RQ2/RQ3 distributions, annotated along PSALM's dimensions and labelled for likely infringement under specified legal assumptions, would clarify whether current prompts and weights align with expert practice, expose under-specified doctrinal concepts, and address the normative question underlying RQ1: to what extent is LLM-based similarity assessment acceptable in legal decision-making?

\subsubsection{Generalisation Across Architectures and Scales}
\label{sec:disc:future-generalisation}
Replicating the RQ2 and RQ3 experiments on larger models and different architectures is important for assessing generality. If stylistic convergence and partial unlearning behave similarly across model families, this strengthens the case for PSALM as a broadly applicable assessment tool. If not, the results would show that certain architectures are intrinsically more prone to problematic memorisation or more amenable to effective unlearning, which could inform regulatory guidance and model selection.

\subsubsection{Multi-Agent and Ensemble-Based Judgement}
\label{sec:disc:future-multi-agent}
A further direction is to replace the current single-judge architecture with multi-agent and ensemble-based evaluation. In the present implementation, each node in the PSALM graph is evaluated by the same judge configuration. This supports consistency, but it does not exploit possible specialisation across legal, literary and computational forms of assessment. Future work could assign different agents to different evaluative roles.

Ensemble judging could also improve robustness. Multiple judges, or multiple independently prompted instances of the same judge, could assess each source--target pair and expose disagreement between evaluators. Such disagreement would be especially informative for borderline cases, where a single scalar score may give a misleading impression of certainty. Aggregation could be performed through majority voting, weighted averaging, calibrated confidence intervals or explicit deliberation between agents. More ambitious versions could require judges to produce structured rationales or proof objects before a final score is assigned, allowing PSALM to distinguish between high-confidence similarity findings and cases where the apparent result depends on contestable assumptions. This would move the framework closer to an auditable evidentiary system rather than a single-pass scoring tool.

\subsubsection{Cross-Domain and Cross-Lingual Extension}
\label{sec:disc:future-domains}
Extending PSALM beyond narrative fiction presents both technical and legal challenges. For journalism or academic writing, evaluators would need to capture argumentative structure, citation practices and factual overlap. For poetry or song lyrics, metrics of metre, rhyme and imagery would become relevant. At the same time, legal standards for originality and substantial similarity differ across domains, so doctrinal mapping would need to be revisited.

Cross-lingual extension beyond English and Dutch would test PSALM's doctrinal generality and expose latent language biases in the judge model, particularly for languages with morphological or script-level differences that affect how stylistic features are encoded.

\subsubsection{Long-Form and Corpus-Scale Evaluation}
\label{sec:disc:future-long-form}
Future work should also extend PSALM from passage-level comparison to chapter-level, whole-work and corpus-scale evaluation. The present experiments compare relatively short source passages with generated responses, which is appropriate for controlled validation and for measuring local memorisation. However, copyright-relevant similarity may be distributed across longer spans. A generated chapter may reproduce the pacing, sequence of revelations, character development or world-building logic of a source chapter without containing high lexical overlap with any single passage. Similarly, a generated book may appropriate the architecture of a source work through recurring motifs, cumulative plot structure or the arrangement of scenes across chapters.

Long-form evaluation raises technical challenges that are not addressed by the current implementation. Full books may exceed the context window of the judge model, making direct source--target comparison infeasible. Future versions could therefore combine retrieval, segmentation and hierarchical aggregation: first identifying potentially relevant source passages or chapters, then evaluating local similarities, and finally aggregating those findings into chapter-level or book-level assessments. Such an approach would also need to distinguish between isolated local overlap and systematic structural similarity across a work as a whole.

A related challenge is corpus-scale source discovery. In the present experiments, the relevant source passage is known in advance. Real compliance settings may instead require checking a generated text against a large library of books to identify possible sources of appropriation. This would require a retrieval stage capable of narrowing a large corpus to candidate works or passages before applying the more expensive PSALM evaluators. Future systems could combine embedding-based retrieval, approximate nearest-neighbour search, stylometric indexing and metadata-aware filtering with PSALM's legally structured similarity assessment. This would move the framework from pairwise evaluation toward large-scale auditing, where the central question is not only how similar two known texts are, but whether a given output is suspiciously similar to any protected work in a reference corpus.

\subsubsection{Alternative Mitigation Approaches and Adversarial Evaluation}
\label{sec:disc:adversarial}
Finally, the NPO experiments suggest that unlearning can move models in the right direction but cannot serve as a complete solution. Comparing NPO with other unlearning techniques, influence-based data removal and inference-time controls under the same PSALM evaluation would create a more systematic picture of the trade-offs between forget quality, utility preservation and exception engagement. In parallel, adversarial prompting strategies aimed at maximising PSALM similarity scores would provide a worst-case analysis complementing the average-case view in this work. For legal risk assessment, understanding the worst case is often more relevant than understanding the typical case.

\section{Conclusion}\label{sec:conclusion}
This research demonstrates that copyright-relevant stylistic appropriation in large language models can be measured systematically using automated evaluation frameworks grounded in EU legal doctrine. The PSALM framework operationalises substantial similarity assessments across computational, stylistic, and content dimensions and incorporates statutory defences for parody, pastiche, quotation and scènes à faire. Controlled validation on purpose-designed test cases yields strict alignment rates of about $95\%$ and a mean absolute error of $0.037$, indicating that PSALM can reliably distinguish clearly different, moderately similar and near-identical text pairs in a manner consistent with its doctrinal design.

The empirical findings show that supervised fine-tuning on a corpus of literary works induces pronounced stylistic appropriation extending well beyond verbatim memorisation. Fine-tuned Llama~3.2~1B models attain near-ceiling similarity scores across writing style, narrative voice, character construction, plot structure, scene sequences and world building. This high expressive overlap persists even when exact textual overlap is negligible. Machine unlearning via Negative Preference Optimisation substantially reduces similarity on the designated forget set and eliminates exact matches, but residual stylistic patterns remain detectable across multiple PSALM dimensions and the models do not revert to baseline behaviour. Unlearning therefore mitigates, but does not fully remove, the influence of the fine-tuning corpus.

These results suggest that technical safeguards focusing exclusively on verbatim memorisation may provide an incomplete picture of copyright risk. At the same time, PSALM's similarity scores are measurements rather than legal verdicts: the mapping between quantitative assessments and infringement determinations remains untested in the absence of validation by legal experts and judicial precedent. Future work should therefore compare PSALM's evaluations with expert legal judgements, replicate the experiments across larger architectures, domains and languages, examine adversarial robustness, and benchmark alternative mitigation approaches under the same legally informed metrics.

PSALM contributes to measurable copyright-compliance assessments and clarifies which aspects of current practice pose the greatest legal uncertainty. Technical measurement serves law rather than replacing it. PSALM offers structured evidence to inform human judgement. Closing the gap between verbatim-focused technical practice and the broader substantial-similarity standards of EU copyright law will require sustained collaboration across legal, regulatory, and AI research communities, supported by evaluation infrastructure of the kind this work seeks to provide.

\section*{Acknowledgements}
This research was supported by the Dutch Research Council (NWO) under the NWO-TDCC programme, project ICT.001.TDCC.014 (budget number 20656). We would like to express our gratitude to Rohan Nanda for reviewing this paper and providing valuable comments. We further thank Steven Claeyssens, Michel de Gruijter, and Mirjam Raaphorst (Koninklijke Bibliotheek, KB) for their expertise and continued support.

\section*{Data Availability}
The source code for the PSALM framework, including all experimental pipelines, is publicly available at https://codeberg.org/nscharrenberg/PSALM (branch: paper-source-code). The corpus used for fine-tuning is publicly available on Hugging Face at https://huggingface.co/datasets/nscharrenberg/DBNL-public. The fine-tuned model is available at https://huggingface.co/nscharrenberg/LLaMA-3.2-1B-DBNL-Public-Domain-Finetuned and the unlearned model at https://huggingface.co/nscharrenberg/LLaMA-3.2-1B-DBNL-Public-Domain-Unlearned.

\bibliography{references}

\begin{appendices}

\section{Supplementary Material}\label{app:online-resources}
The following Online Resources are submitted as supplementary files
alongside this article.
\begin{description}
    \item[Online Resource~1] \texttt{ESM\_1.csv} ---
    Evaluator-level validation summary statistics for all ten PSALM
    evaluators, including MAE, RMSE, EXACT\slash CLOSE\slash
    ACCEPTABLE\slash POOR counts, strict alignment rate, and
    coefficient of variation of absolute error.

  \item[Online Resource~2] \texttt{ESM\_2.csv} ---
    Per-evaluator, per-test-case statistics from the controlled
    validation, including expected score, mean and standard deviation of
    actual scores over five replications, and minimum and maximum
    observed values.

  \item[Online Resource~3] \texttt{ESM\_3.csv} ---
    Raw per-run validation results for all evaluator--test-case
    combinations, including individual scores, alignment band
    classifications, and free-text reasoning fields.

  \item[Online Resource~4] \texttt{ESM\_4.csv} ---
    Shapiro--Wilk normality and Levene homogeneity test results for all
    RQ2 metrics on the forget split.

  \item[Online Resource~5] \texttt{ESM\_5.csv} ---
    Shapiro--Wilk normality and Levene homogeneity test results for all
    RQ2 metrics on the retain split.

  \item[Online Resource~6] \texttt{ESM\_6.csv} ---
    Kruskal--Wallis omnibus statistics and per-condition descriptive
    statistics for all thirteen metrics across the four model conditions
    (RQ2, both splits).

  \item[Online Resource~7] \texttt{ESM\_7.csv} ---
    Cohen's $d$ and Cliff's $\delta$ effect sizes for all pairwise
    model contrasts on RQ2 metrics, forget split.

  \item[Online Resource~8] \texttt{ESM\_8.csv} ---
    Cohen's $d$ and Cliff's $\delta$ effect sizes for all pairwise
    model contrasts on RQ2 metrics, retain split.

  \item[Online Resource~9] \texttt{ESM\_9.csv} ---
    Per-category descriptive statistics (mean, standard deviation) for
    all four model conditions across RQ2 metric categories
    (Computational, Stylistic, Content, Exceptions).

  \item[Online Resource~10] \texttt{ESM\_10.csv} ---
    Rank-based overall model rankings aggregated across all thirteen
    RQ2 metrics for forget and retain splits.

  \item[Online Resource~11] \texttt{ESM\_11.csv} ---
    Category-level differential-effect contrasts (baseline minus
    fine-tuned) for RQ2, forget split.

  \item[Online Resource~12] \texttt{ESM\_12.csv} ---
    Rank-based ART-ANOVA results for language $\times$ training-stage
    interactions on RQ2 metrics, forget split.

  \item[Online Resource~13] \texttt{ESM\_13.csv} ---
    Rank-based ART-ANOVA results for language $\times$ training-stage
    interactions on RQ2 metrics, retain split.

  \item[Online Resource~14] \texttt{ESM\_14.csv} ---
    Paired Wilcoxon signed-rank test results comparing baseline and
    fine-tuned models on shared prompts, RQ2 forget split.

  \item[Online Resource~15] \texttt{ESM\_15.csv} ---
    Paired Wilcoxon signed-rank test results comparing baseline and
    fine-tuned models on shared prompts, RQ2 retain split.

  \item[Online Resource~16] \texttt{ESM\_16.csv} ---
    Shapiro--Wilk normality and Levene homogeneity test results for all
    RQ3 metrics on the forget split.

  \item[Online Resource~17] \texttt{ESM\_17.csv} ---
    Shapiro--Wilk normality and Levene homogeneity test results for all
    RQ3 metrics on the retain split.

  \item[Online Resource~18] \texttt{ESM\_18.csv} ---
    Kruskal--Wallis omnibus statistics and per-condition descriptive
    statistics for all thirteen metrics across model conditions (RQ3,
    both splits).

  \item[Online Resource~19] \texttt{ESM\_19.csv} ---
    Rank-based ART-ANOVA results for language $\times$ unlearning
    interactions on RQ3 metrics, forget split.

  \item[Online Resource~20] \texttt{ESM\_20.csv} ---
    Rank-based ART-ANOVA results for language $\times$ unlearning
    interactions on RQ3 metrics, retain split.

  \item[Online Resource~21] \texttt{ESM\_21.csv} ---
    Rank-based overall model rankings aggregated across all thirteen
    RQ3 metrics for forget and retain splits.

  \item[Online Resource~22] \texttt{ESM\_22.csv} ---
    Per-category descriptive statistics for all model conditions across
    RQ3 metric categories.

  \item[Online Resource~23] \texttt{ESM\_23.csv} ---
    Category-level differential-effect contrasts (fine-tuned minus NPO)
    for RQ3, forget split.

  \item[Online Resource~24] \texttt{ESM\_24.csv} ---
    Paired Wilcoxon signed-rank test results comparing fine-tuned and
    NPO models on shared prompts, RQ3 forget split.

  \item[Online Resource~25] \texttt{ESM\_25.csv} ---
    Paired Wilcoxon signed-rank test results comparing fine-tuned and
    NPO models on shared prompts, RQ3 retain split.

  \item[Online Resource~26] \texttt{ESM\_26.csv} ---
    Cohen's $d$ and Cliff's $\delta$ effect sizes for pairwise model
    contrasts on RQ3 metrics, forget split.

  \item[Online Resource~27] \texttt{ESM\_27.csv} ---
    Cohen's $d$ and Cliff's $\delta$ effect sizes for pairwise model
    contrasts on RQ3 metrics, retain split.

  \item[Online Resource~28] \texttt{ESM\_28.csv} ---
    Two one-sided tests (TOST) for equivalence between baseline and NPO
    models on all metrics, retain split.
\end{description}

\section{Hyperparameter Specifications}\label{app:hyperparameters}
\subsection{Output Generation}
The output generation parameters are shown in Table~\ref{tab:generation_hyperparameters_full}.

\begin{table}[htbp]
    \centering
    \adjustbox{max width=\textwidth}{
    \begin{tabular}{l l p{6cm}}
        \hline
        \textbf{Parameter} & \textbf{Value} \\
        \hline
        Sampling Method & Nucleus (top-p) \\
        Top-P & $0.9$ \\
        Temperature & $0.1$  \\
        Max New Tokens & $1024$  \\
        Repetition Penalty & $1.1$ \\
        \hline
    \end{tabular}
    }
    \caption{Output Generation Parameters}
    \label{tab:generation_hyperparameters_full}
\end{table}

\section{Controlled Validation Corpus}\label{app:validation-corpus}
\subsection{Test Case Design Principles}\label{app:controlled:test-case}
The controlled validation set comprises 50 unique text pairs (10 evaluators $\times$ 5 test cases) designed to isolate individual PSALM evaluators and verify dimensional specificity. Each test case instantiates one point on the evaluator's five-level taxonomy:

\begin{itemize}
    \item \textbf{TC1 (0.0):} Texts exhibit fundamental opposition or complete absence of similarity on the target dimension while potentially sharing characteristics on orthogonal dimensions.
    \item \textbf{TC2 (0.3):} Texts share superficial/minimal overlap on the target dimension while diverging substantially in execution depth or specific elaborations.
    \item \textbf{TC3 (0.5):} Texts demonstrate balanced presence of both similarities and differences, occupying the decision boundary region where classification becomes ambiguous.
    \item \textbf{TC4 (0.8):} Texts exhibit strong alignment on the target dimension with only small variations insufficient to alter overall classification.
    \item \textbf{TC5 (1.0):} Texts match precisely or near-precisely on all sub-dimensions of the evaluator's assessment criteria.
\end{itemize}

\subsection{Text Pair Construction Protocol}
For each evaluator-test case combination:

\begin{enumerate}
    \item \textbf{Source text generation:} Manually author or curate a passage (150-300 words) demonstrating distinctive characteristics on the target dimension. For example, for Writing Style TC5, create text with highly specific lexical patterns, sentence structures, and rhetorical devices.
    
    \item \textbf{Target text engineering:} Construct target text through controlled modifications:
    \begin{itemize}
        \item \textbf{TC1:} Invert or oppose dimensional characteristics (e.g., first-person $\rightarrow$ third-person omniscient for Narrative Voice)
        \item \textbf{TC2:} Preserve 1-2 sub-dimensional features whilst altering others (e.g., maintain sentence length distribution but change lexical complexity for Writing Style)
        \item \textbf{TC3:} Preserve half of sub-dimensions, alter half
        \item \textbf{TC4:} Preserve all but 1-2 minor sub-dimensional features
        \item \textbf{TC5:} Create near-identical text with minimal cosmetic changes (character name substitutions, setting transpositions)
    \end{itemize}
    
    \item \textbf{Orthogonality verification:} Ensure modifications isolate the target dimension without introducing confounding changes on other dimensions. For example, when testing Character Similarity TC3, plot structure and world-building should remain constant between source and target.

    \item \textbf{Internal review:} Independent reviewers assess whether text pairs achieve intended similarity level and dimensional isolation. Ambiguous pairs are revised until consensus is reached.
\end{enumerate}

\subsection{Validation Metrics and Criteria}
Each text pair receives five independent evaluations (different API calls to GPT-5-nano with temperature $=0$). This replication enables assessment of:

\begin{itemize}
    \item \textbf{Mean Absolute Error (MAE)}
    \item \textbf{Root Mean Squared Error (RMSE)}
    
    \item \textbf{Alignment classifications:}
    \begin{itemize}
        \item EXACT: $|\Delta| \leq 0.1$
        \item CLOSE: $0.1 < |\Delta| \leq 0.2$
        \item ACCEPTABLE: $0.2 < |\Delta| \leq 0.3$
        \item POOR: $|\Delta| > 0.3$
    \end{itemize}
    
    \item \textbf{Success rates:}
    \begin{equation}
        \text{Alignment Rate} = \frac{N_{\text{EXACT}} + N_{\text{CLOSE}} + N_{\text{ACCEPTABLE}}}{N_{\text{total}}}
    \end{equation}
    \begin{equation}
        \text{Strict Alignment Rate} = \frac{N_{\text{EXACT}} + N_{\text{CLOSE}}}{N_{\text{total}}}
    \end{equation}
\end{itemize}

\subsection{Validation Status Assignment}
Evaluators are classified as:

\begin{itemize}
    \item \textbf{Validated:} Strict Alignment Rate $\geq 0.8$ \textbf{and} MAE $\leq 0.15$
    \item \textbf{Partially Validated:} Alignment Rate $\geq 0.7$ \textbf{and} MAE $\leq 0.25$
    \item \textbf{Failed:} Does not meet Partially Validated criteria
\end{itemize}

Only Validated evaluators proceed to corpus evaluation. Partially Validated evaluators may be used for exploratory analysis with explicit acknowledgment of elevated measurement error. Failed evaluators are excluded until re-designed and re-validated.

\subsection{Iterative Refinement Protocol}
Evaluators failing validation undergo:

\begin{enumerate}
    \item \textbf{Error analysis:} Inspect POOR-aligned cases to diagnose failure modes:
    \begin{itemize}
        \item Systematic bias (consistent over-/under-estimation)
        \item Boundary confusion (misclassification near decision thresholds)
        \item Sub-dimension imbalance (over-weighting specific DAG nodes)
        \item Stochastic instability (high replication variance)
    \end{itemize}
    
    \item \textbf{Prompt refinement:} Modify evaluator prompts to clarify dimensional definitions, add explicit boundary-case examples, or adjust sub-node weighting instructions
    
    \item \textbf{Re-evaluation:} Re-run validation with modified configuration
    
    \item \textbf{Documentation:} Archive refinement history with rationale for design decisions
\end{enumerate}

All validation data (text pairs, expected scores, observed scores, alignment classifications) are archived in supplementary materials.

\section{Extended Statistical Analysis Details}\label{app:statistical-details}
\subsection{Multiple Testing Correction Strategies}
\subsubsection{Family Definitions}
Correction families are defined as:

\begin{itemize}
    \item \textbf{Per-metric families:} All pairwise comparisons within one metric (e.g., all Writing Style comparisons: $en_b$ vs. $en_f$, $en_f$ vs. $en_{npo}$, etc.)
    \item \textbf{Per-category families:} Differential effect comparisons within categories (e.g., Computational vs. Stylistic vs. Content vs. Exceptions)
    \item \textbf{Independent families:} Across research questions (RQ2 and RQ3 comparisons use separate families)
\end{itemize}

\subsection{Reporting Standards}
All statistical results report:
\begin{enumerate}
    \item \textbf{Test identification:} Name and type (e.g., "Mann-Whitney U test", "Wilcoxon signed-rank test")
    \item \textbf{Test statistic:} Numerical value with symbol (e.g., $U = 234.5$, $W = 120$, $H = 45.32$)
    \item \textbf{Degrees of freedom:} Where applicable (e.g., $df = 78$ for paired t-test)
    \item \textbf{$p$-value:} Exact when $p > 0.001$, otherwise "$p < 0.001$"
    \item \textbf{Effect size:} Point estimate with $95\%$ CI (e.g., "$d = -2.34$ [$95\%$ CI: $-2.89$, $-1.79$]")
    \item \textbf{Interpretation:} Explicit significance statement (e.g., "significant at $\alpha = 0.05$" or "non-significant")
\end{enumerate}

Non-significant results are explicitly reported rather than omitted to ensure transparency about null findings.

\section{RQ1: Detailed Validation Statistics}\label{app:evaluator-performance}
This appendix summarises the core supplementary findings from the controlled validation experiments for RQ1, using the supplementary data for RQ1. The aim is to document the main numerical patterns that are not already discussed in subsection~\ref{sec:controlled-validation}.

\subsection{Evaluator-level performance}

Table~\ref{tab:rq1:category_summary_compact} gives a compact category-level view of the evaluator statistics. For each group of evaluators it reports the range of mean absolute error (MAE) and of the strict alignment rate (EXACT+CLOSE, i.e.\ predictions within $\pm 0.2$ of the expected score), computed directly from Online Resource~1.

\begin{table}[htbp]
  \centering
  \caption[Category-level summary of RQ1 validation]{Category-level summary of RQ1 validation results derived from Online Resource~1.  Strict alignment is the proportion of runs in EXACT or CLOSE bands}
  \label{tab:rq1:category_summary_compact}
  \begin{tabular}{p{5cm}p{1cm}p{2cm}p{2cm}}
    \toprule
    Category & Evaluators & MAE range & Strict alignment range \\
    \midrule
    Stylistic (Narrative Voice, Writing Style)
      & 2 & 0.028--0.048 & 0.92--0.96 \\
    Content (Character, Plot, Scene, World-building)
      & 4 & 0.028--0.056 & 0.92--0.96 \\
    Exceptions (Parody/Satire, Pastiche, Quotation/Citation)
      & 3 & 0.028--0.056 & 0.96--0.96 \\
    Other (Scènes à faire)
      & 1 & 0.044        & 0.96       \\
    \bottomrule
  \end{tabular}
\end{table}

Across all ten evaluators, MAE lies between $0.028$ and $0.056$, with RMSE between $0.082$ and $0.136$. Strict alignment is either $0.92$ or $0.96$ for every evaluator, and the overall distribution derived from Online Resource~1 shows $211$ EXACT and $26$ CLOSE alignments out of $250$ runs.  The remaining $13$ runs are POOR, with no ACCEPTABLE-only cases.  Thus, $237/250$ evaluations fall within $\pm 0.2$ of the intended similarity level and $13/250$ fall more than one full category away.

Within categories, the two stylistic evaluators differ only modestly: Narrative Voice has the lowest MAE ($0.028$) and highest strict alignment ($0.96$), while Writing Style has slightly higher MAE ($0.048$) and strict alignment of $0.92$.  The content evaluators share MAE values between $0.028$ and $0.056$; Character Similarity is the least accurate of this group with MAE $0.056$ and two POOR alignments, while Plot Structure, Scene Sequence and World-building each attain MAE $0.028$ and a single POOR run.  Among the exception evaluators, Quotation/Citation matches the best-performing content metrics (MAE $0.028$, strict alignment $0.96$), Parody/Satire sits in the middle (MAE $0.036$), and Pastiche mirrors Character Similarity (MAE $0.056$, two POOR runs). The Scènes à faire evaluator, categorised as \emph{Other} in the CSVs, has MAE $0.044$ and strict alignment $0.96$, situated between the most and least accurate evaluators.

The last column of Online Resource~1 reports the coefficient of variation of absolute error, which ranges from approximately $2.08$ to $2.83$ across evaluators.  This combination of small MAE and relatively high coefficients of variation reflects the skewed error distribution already described in the main text: most runs have zero error, with occasional deviations of $0.3$ or $0.5$.

\subsection{Test-case behaviour}
Online Resource~2 records, for each evaluator and each of the five test cases, the expected score, the mean and standard deviation of the actual scores across the five replications, and the minimum and maximum values observed.

For all evaluators, the extreme conditions behave as intended.  When the expected score is $0.0$ (TC1, \say{clearly different or opposite}) the mean actual score is exactly $0.0$ with zero standard deviation and all runs classified as EXACT.  When the expected score is $1.0$ (TC5, \say{identical or near-identical}), the mean actual score is $0.96$ for every evaluator, with standard deviation $0.089\ldots$, minimum $0.8$ and maximum $1.0$; the resulting mean absolute error is $0.04$ and strict alignment remains EXACT.

The mid-range conditions show the small systematic deviations that are only qualitatively described in the main text.  For the \say{very similar} test cases (TC4, expected score $0.8$), the mean actual score lies between $0.64$ and $0.74$ depending on the evaluator.  Character Similarity and Pastiche are the most conservative, with mean $0.64$ and standard deviation $0.230\ldots$, whereas Narrative Voice, Plot Structure, Scene Sequence, World-building, Writing Style, Parody/Satire and Quotation/Citation all have mean $0.74$ and standard deviation $0.134\ldots$.  In all these cases the mean absolute error is at most $0.16$, and the summary file labels them as EXACT overall because the deviations remain within the $\pm 0.2$ band.

For the \say{moderately similar} cases (TC3, expected $0.5$), most evaluators have mean actual scores close to $0.46$ with standard deviation $0.089\ldots$, again leading to a mean absolute error of $0.04$ and EXACT alignment at the test-case level.  Writing Style is slightly different: its “moderately similar” case shows mean actual $0.52$, standard deviation $0.178\ldots$, and mean absolute error $0.10$, indicating more spread and a small upward bias relative to $0.5$.  For the \say{somewhat different} cases (TC2, expected $0.3$), mean actual scores are either exactly $0.3$ (e.g.\ Narrative Voice, Plot Structure, Scene Sequence, World-building, Quotation/Citation and several exception metrics) or $0.34$ with standard deviation $0.089\ldots$ (Character Similarity, Parody/Satire, Pastiche, Scenes à faire, Writing Style).  These values correspond to the alignment label EXACT in the summary CSV because they stay within $0.04$ of the target.

\subsection{Alignment distribution and failure modes}
The raw per-run data (Online Resource~3) allow the $13$ POOR alignments reported in Online Resource~1 to be localised to particular evaluators and test cases.  All POOR cases occur in the middle or upper-middle of the scale and none appear in the “clearly different” or “identical” conditions.

For Character Similarity, both POOR runs arise in the “very similar characters with only minor differences” test case.  The reasoning fields attribute these errors to “weighting noise” and to over-emphasis on “lexical and environmental divergence” when structural and arc-level features were near-identical.  Narrative Voice has a single POOR case, also in the “very similar” condition, where the judge model reports that a collective voice pattern was not recognised and the verdict was demoted to “moderately similar”.  Parody/Satire’s only POOR alignment occurs in the “good parody” test case, with the explanation that subtle humour was under-recognised.

Pastiche has two POOR alignments in the “good pastiche” condition.  In one run the reasoning notes that the Artistic Skill node was down-weighted; in the other, tonal warmth was interpreted as divergence from the source rather than as part of a respectful homage.  For Plot Structure, Scene Sequence and World-building, the single POOR alignment in each case is again attached to the “very similar” test case; the reasoning refers to omitted crisis–repair links, pacing variance and phrasing simplification in specific sub-nodes.  Quotation/Citation has one POOR run in the “good quotation” test case, where prior disclosure or identification is under-weighted.  Writing Style has two POOR cases, one where a change in rhythm leads to a downgrade from “very similar” to “moderately similar”, and one where rhythmic correspondence is over-weighted and a “moderately similar” pair is pushed up to “very similar”.

Across all evaluators, these POOR cases share two features that are not explicitly spelled out in the main text.  First, they always concern borderline distinctions between adjacent verbal bands, such as “very similar” versus “moderately similar”, and never represent confusions between opposite ends of the scale.  Secondly, the reasoning strings consistently point to shifts in the relative salience of sub-dimensions—lexical versus structural features in Character Similarity, rhythmic versus syntactic features in Writing Style, or artistic execution versus tribute character in Pastiche—rather than to complete misinterpretations of the task instructions.  This supports the view that the residual errors are primarily boundary and weighting effects within otherwise well-calibrated evaluators.

\subsection{Test-case Difficulty and Boundary Effects}
\begin{figure}[!hb]
    \centering
    \includegraphics[width=\textwidth]{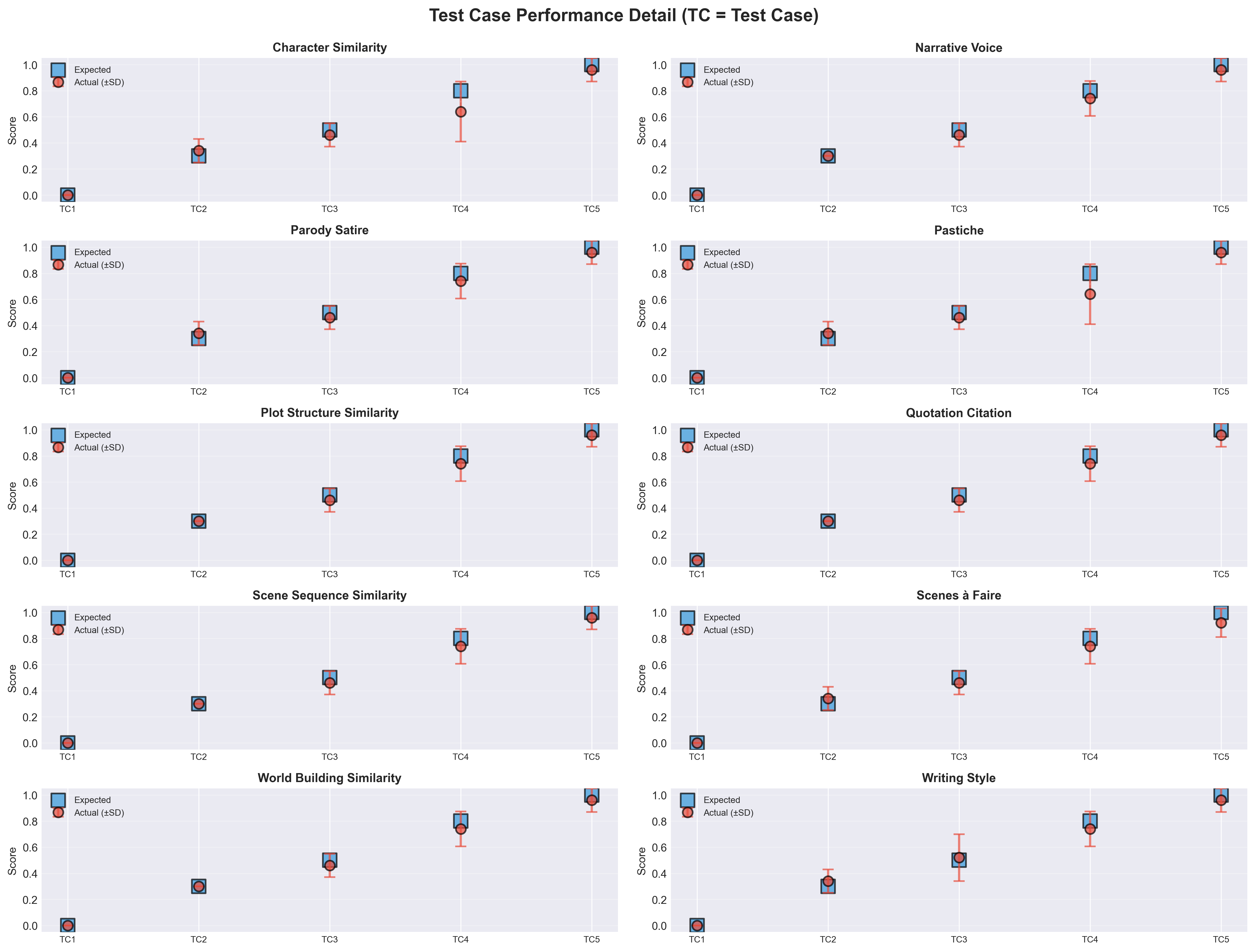}
    \caption[Test case performance across evaluators]{
    Test case performance across evaluators where for each evaluator blue bars denote expected scores for test cases TC1 through TC5 and red circles with error bars show mean and standard deviation of actual scores over five replications}
    \label{fig:rq1:test_case_performance}
\end{figure}

Figure~\ref{fig:rq1:test_case_performance} disaggregates performance by test case. The extreme conditions behave as intended: all evaluators return scores at or indistinguishable from \(0\) for TC1 (unrelated texts) and near \(1\) for TC5 (near-identical texts), confirming PSALM reliably separates obviously non-infringing from near-verbatim pairs.

The more interesting behaviour emerges in the intermediate test cases. The intermediate cases (TC2–TC4) show noticeably higher replication variance, with scores occasionally encroaching on adjacent bands — TC4 pairs pushed into TC3, or TC3 pairs lifted slightly. This is most pronounced for Writing Style and Character Similarity, where subtle gradations are inherently less discrete than, for example, the presence or absence of a quotation.

This behaviour justifies the five-level rubric: evaluators reliably distinguish the ends of the spectrum and generally separate \say{moderate} from \say{very} similarity, but finer partitioning would amplify instability in the central region. Mid-range spread is better understood as a property of the underlying construct (borderline cases are genuinely ambiguous) than as a measurement defect.

\section{RQ2: Detailed Statistical Analyses}\label{app:rq2_details}
This appendix summarises the core supplementary findings for RQ2, using the supplementary data. The aim is to document the main numerical patterns that are not already discussed in subsection~\ref{sec:stylistic-appropriation}.

\subsection{Literal Reproduction}\label{sec:rq2:computational}
\begin{figure}[!hb]
  \centering

  \begin{subfigure}[t]{0.48\textwidth}
    \centering
    \includegraphics[width=\textwidth]{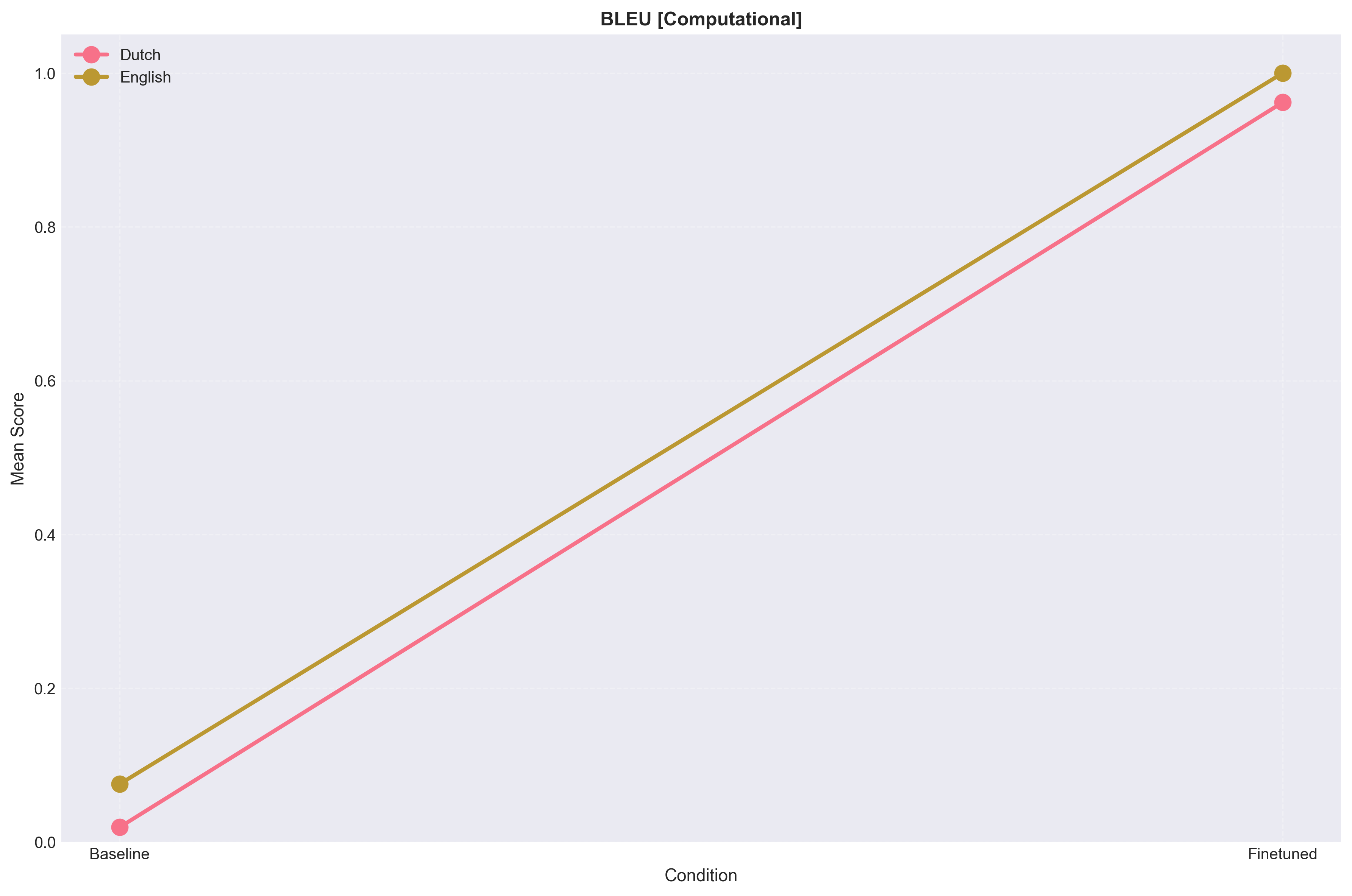}
    \caption{BLEU Interaction of Forget Set}
  \end{subfigure}
  \begin{subfigure}[t]{0.48\textwidth}
    \centering
    \includegraphics[width=\textwidth]{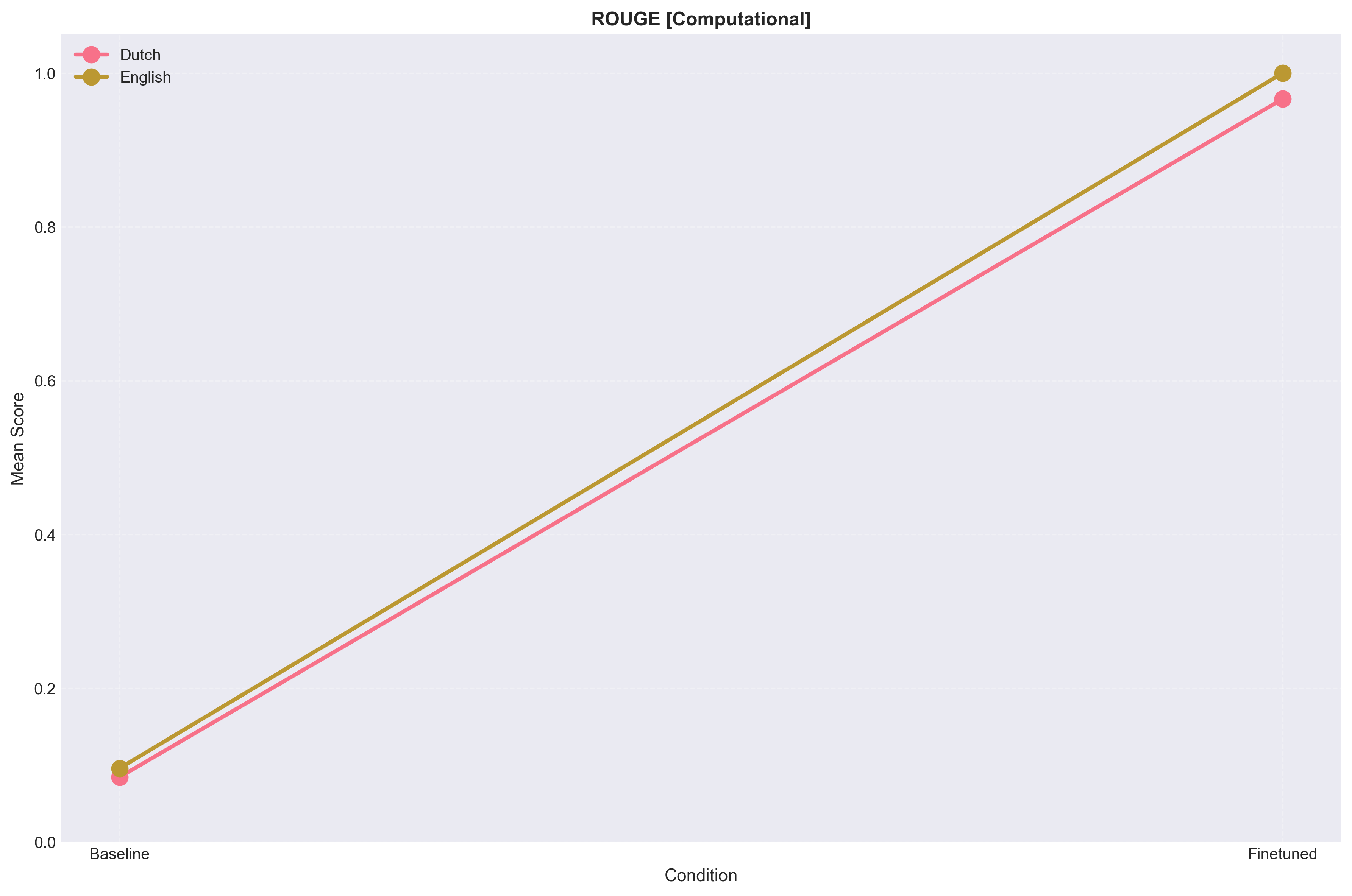}
    \caption{ROUGE-L Interaction of Forget Set}
  \end{subfigure}
  \begin{subfigure}[t]{0.48\textwidth}
    \centering
    \includegraphics[width=\textwidth]{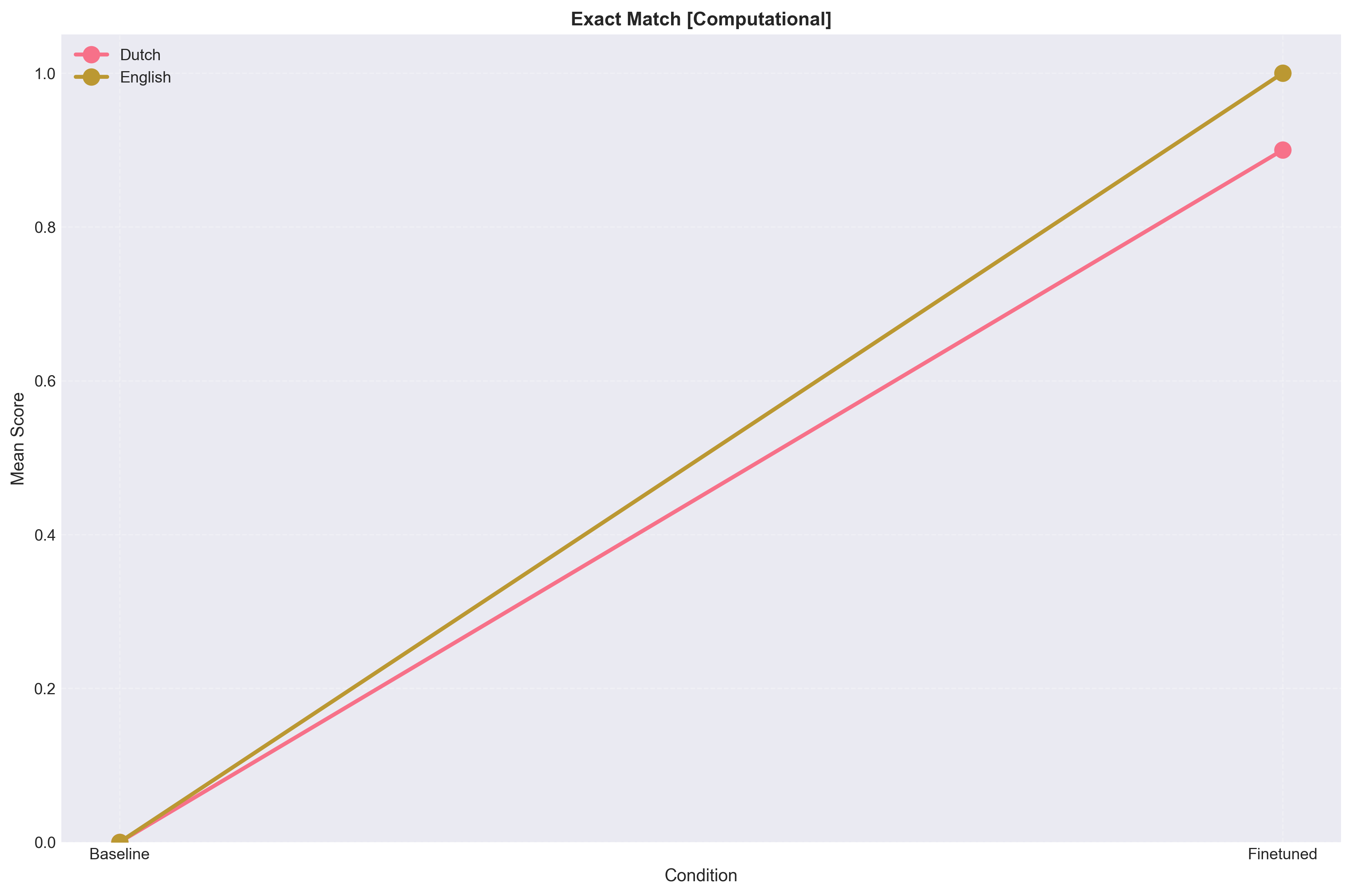}
    \caption{Exact Match Interaction of Forget Set}
  \end{subfigure}
  \begin{subfigure}[t]{0.48\textwidth}
    \centering
    \includegraphics[width=\textwidth]{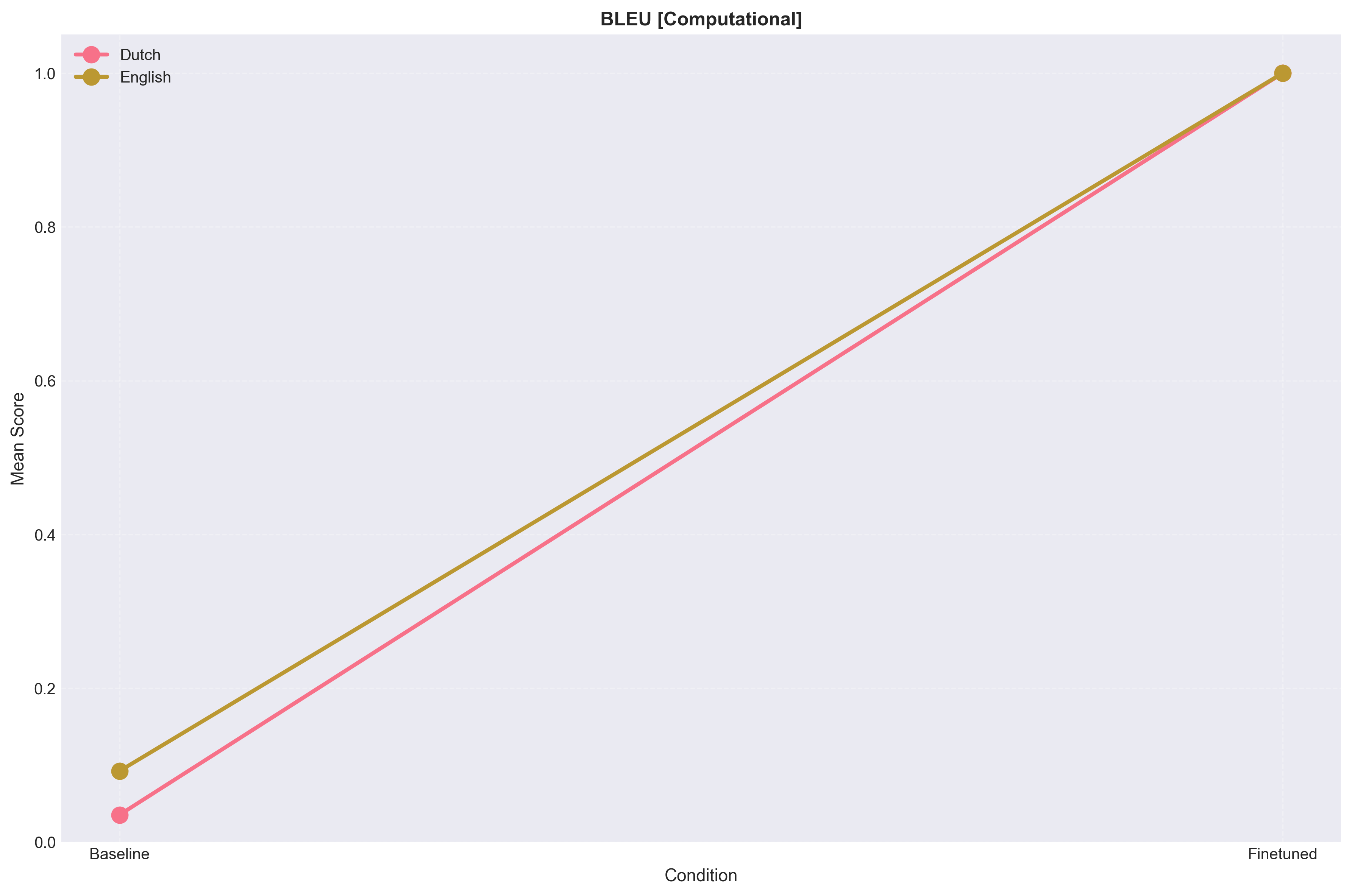}
    \caption{BLEU Interaction of Retain Set}
  \end{subfigure}
  \begin{subfigure}[t]{0.48\textwidth}
    \centering
    \includegraphics[width=\textwidth]{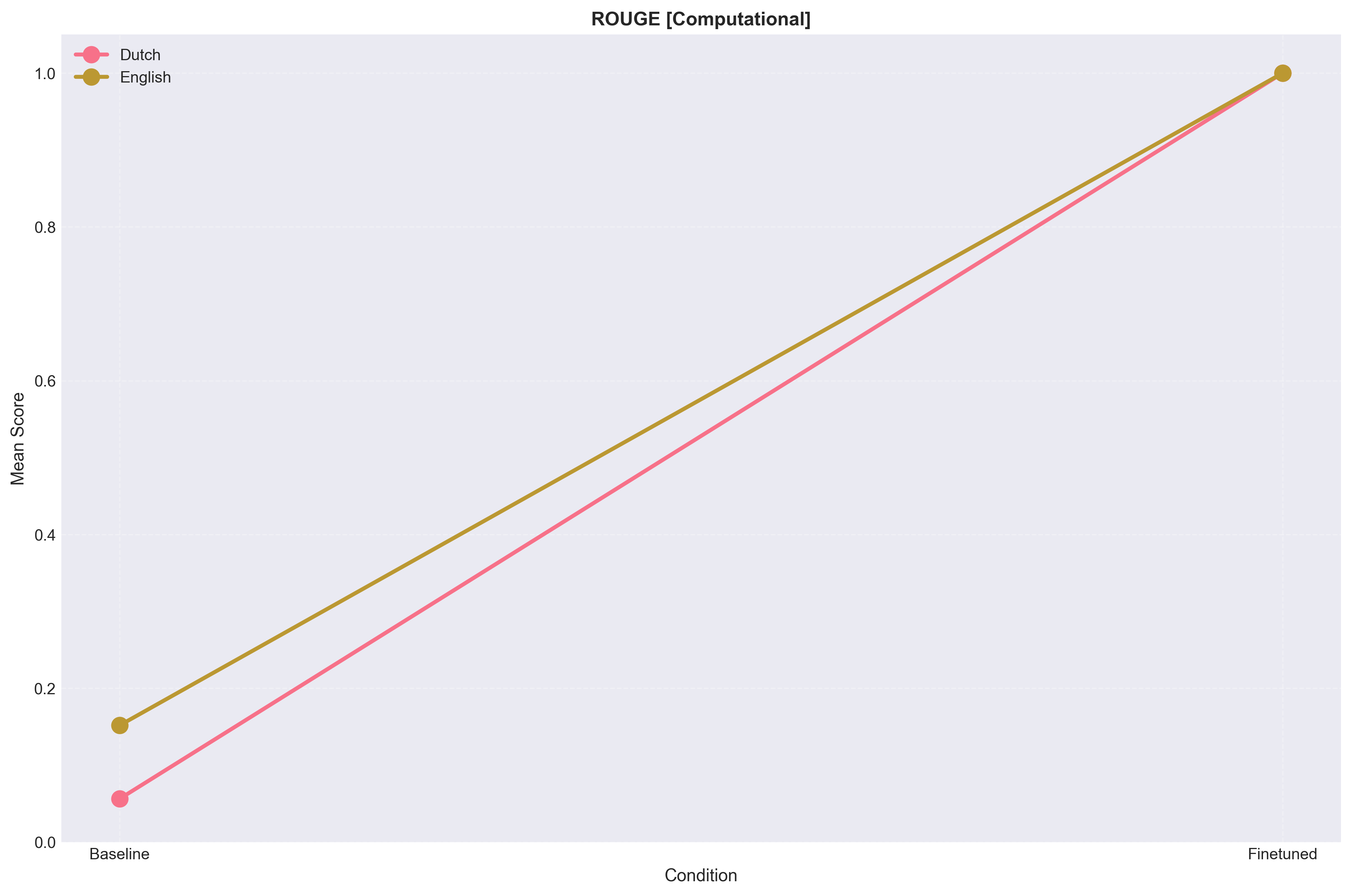}
    \caption{ROUGE-L Interaction of Retain Set}
  \end{subfigure}
  \begin{subfigure}[t]{0.48\textwidth}
    \centering
    \includegraphics[width=\textwidth]{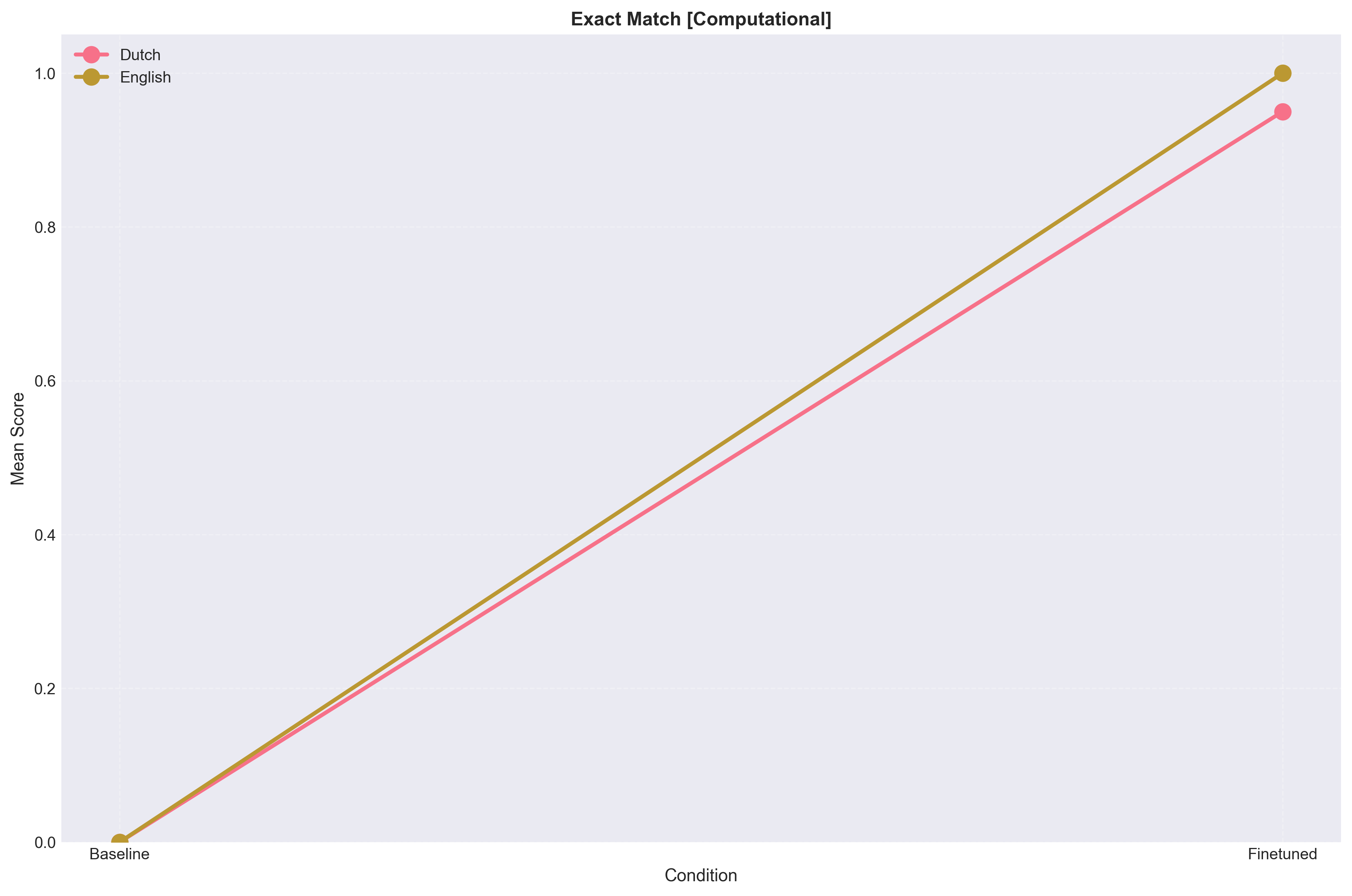}
    \caption{Exact Match Interaction of Retain Set}
  \end{subfigure}

  \caption[Computational metrics before and after fine-tuning]{
  Computational similarity metrics (Exact Match BLEU and ROUGE) for baseline and fine-tuned models where fine-tuning almost saturates all three metrics in both languages and splits indicating near-perfect literal reproduction of training passages}
  \label{fig:rq2:computational-metrics}
\end{figure}

We first examine literal reproduction as captured by Exact Match, BLEU and ROUGE. Figure~\ref{fig:rq2:computational-metrics} displays the mean scores for these metrics before and after fine-tuning, for both languages and splits. Baseline models produce near-zero scores on all three metrics, with the English baseline showing marginally higher overlap than the Dutch (consistent with LLaMA 3.2's English-centric pre-training) but both well below any threshold associated with systematic memorisation.

After fine-tuning, Exact Match and ROUGE means approach one for both languages and splits, with BLEU similarly near ceiling, implying that almost all generated answers are exact or near-exact reproductions of the corresponding corpus passages. Because Exact Match is binary at the level of individual outputs, a mean of $0.9$ for the Dutch forget set, for example, means that more than $90\%$ of outputs are exact substring copies of the source passage used as the answer in the QA pair. 

Kruskal–Wallis tests confirm that model condition has a highly significant effect on all three computational metrics for both splits, and post-hoc Mann–Whitney tests show that each fine-tuned model differs strongly from each baseline. Paired Wilcoxon tests on shared prompts show very large effect sizes when comparing each baseline with its fine-tuned counterpart.

From a copyright perspective, these results indicate that fine-tuning turns the models into near-perfect retrievers of the training passages rather than models that merely internalise style. Were the underlying works protected rather than public domain, such behaviour would present a clear risk of verbatim infringement.

\subsection{Stylistic and Structural Appropriation Extended}
\begin{figure}[!hb]
  \centering

  \begin{subfigure}[t]{0.48\textwidth}
    \centering
    \includegraphics[width=\textwidth]{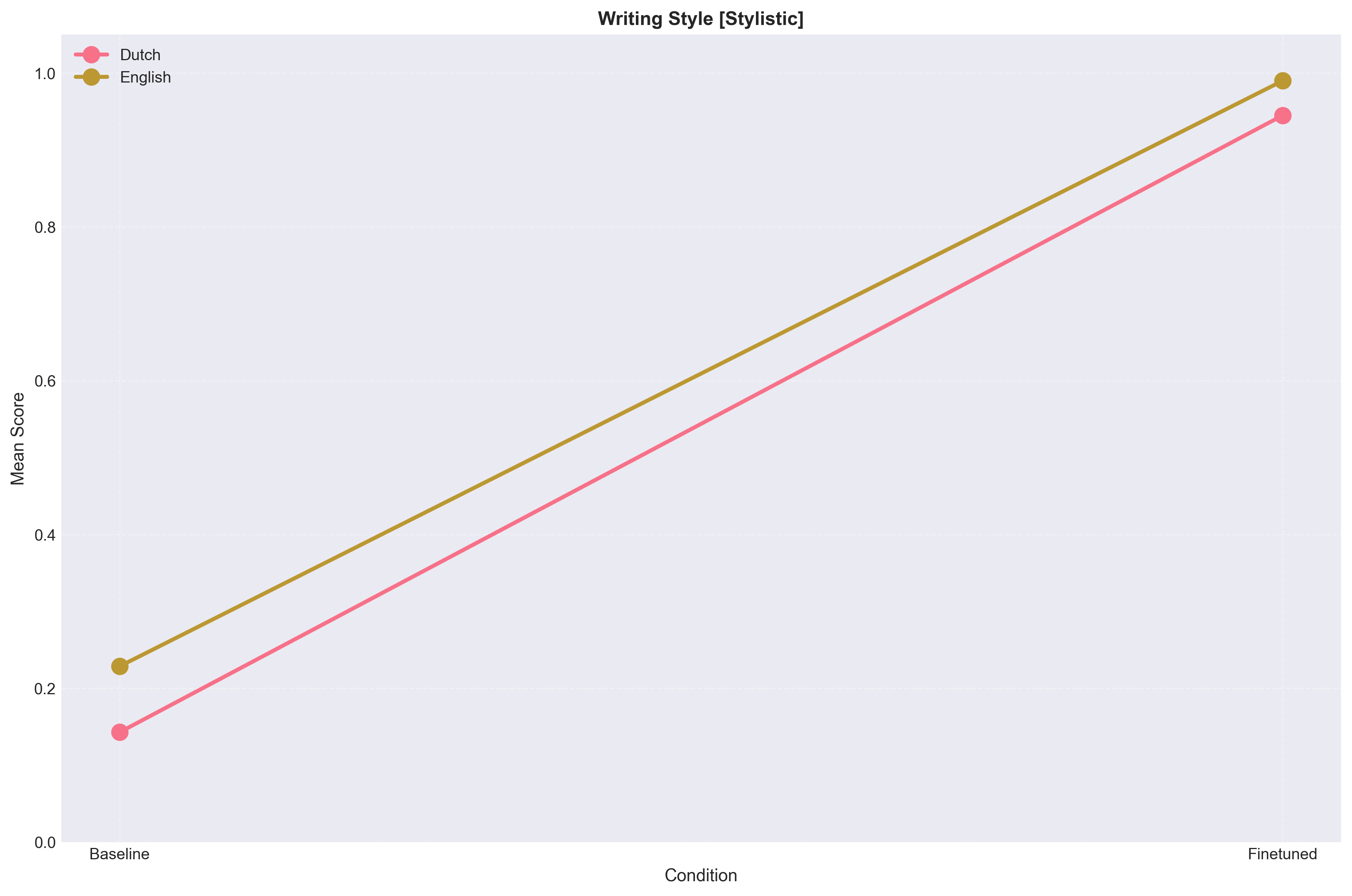}
    \caption{Writing Style Interaction on Retain Set}
  \end{subfigure}
  \begin{subfigure}[t]{0.48\textwidth}
    \centering
    \includegraphics[width=\textwidth]{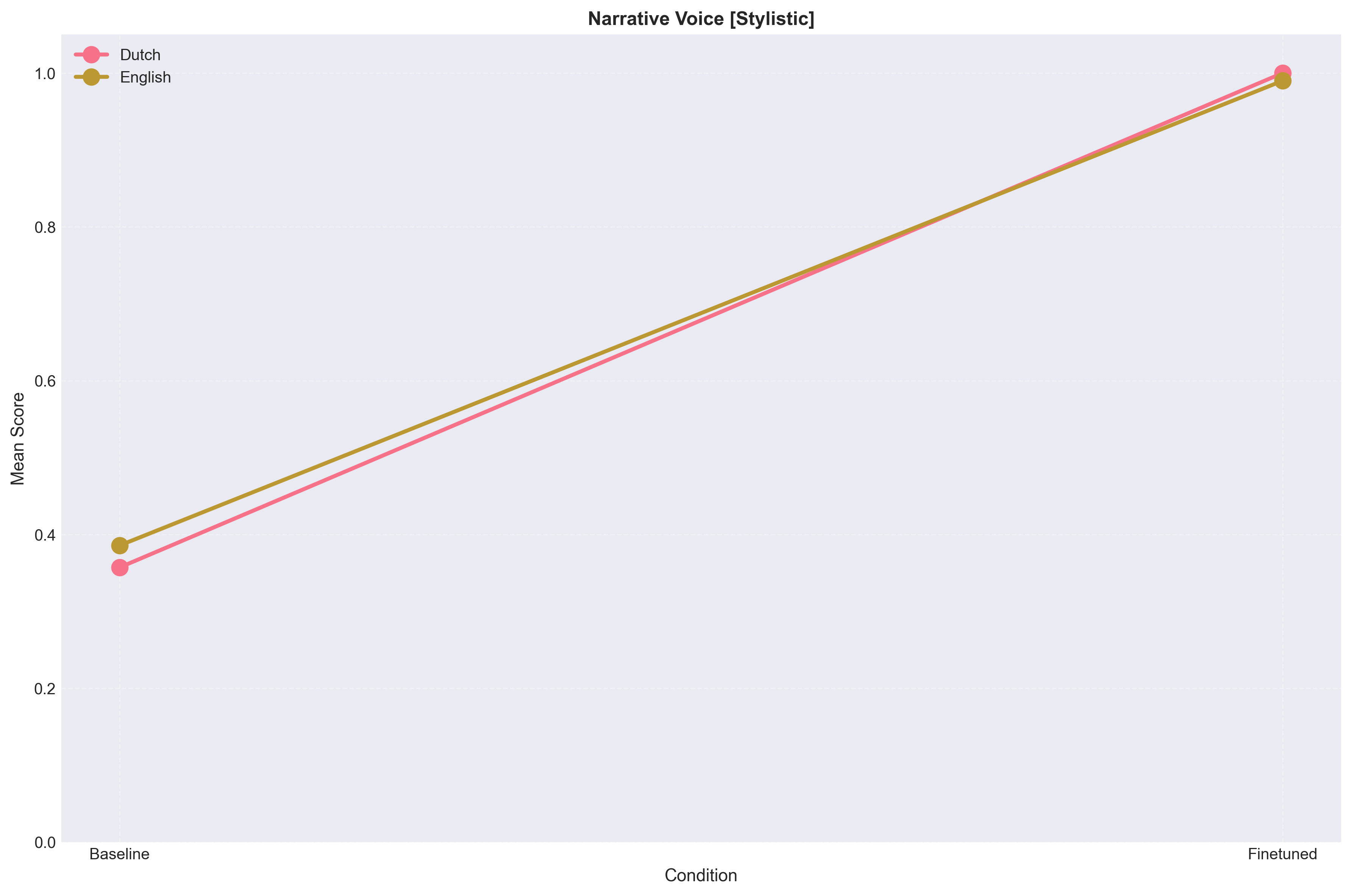}
    \caption{Narrative Voice Interaction on Retain Set}
  \end{subfigure}
  \begin{subfigure}[t]{0.48\textwidth}
    \centering
    \includegraphics[width=\textwidth]{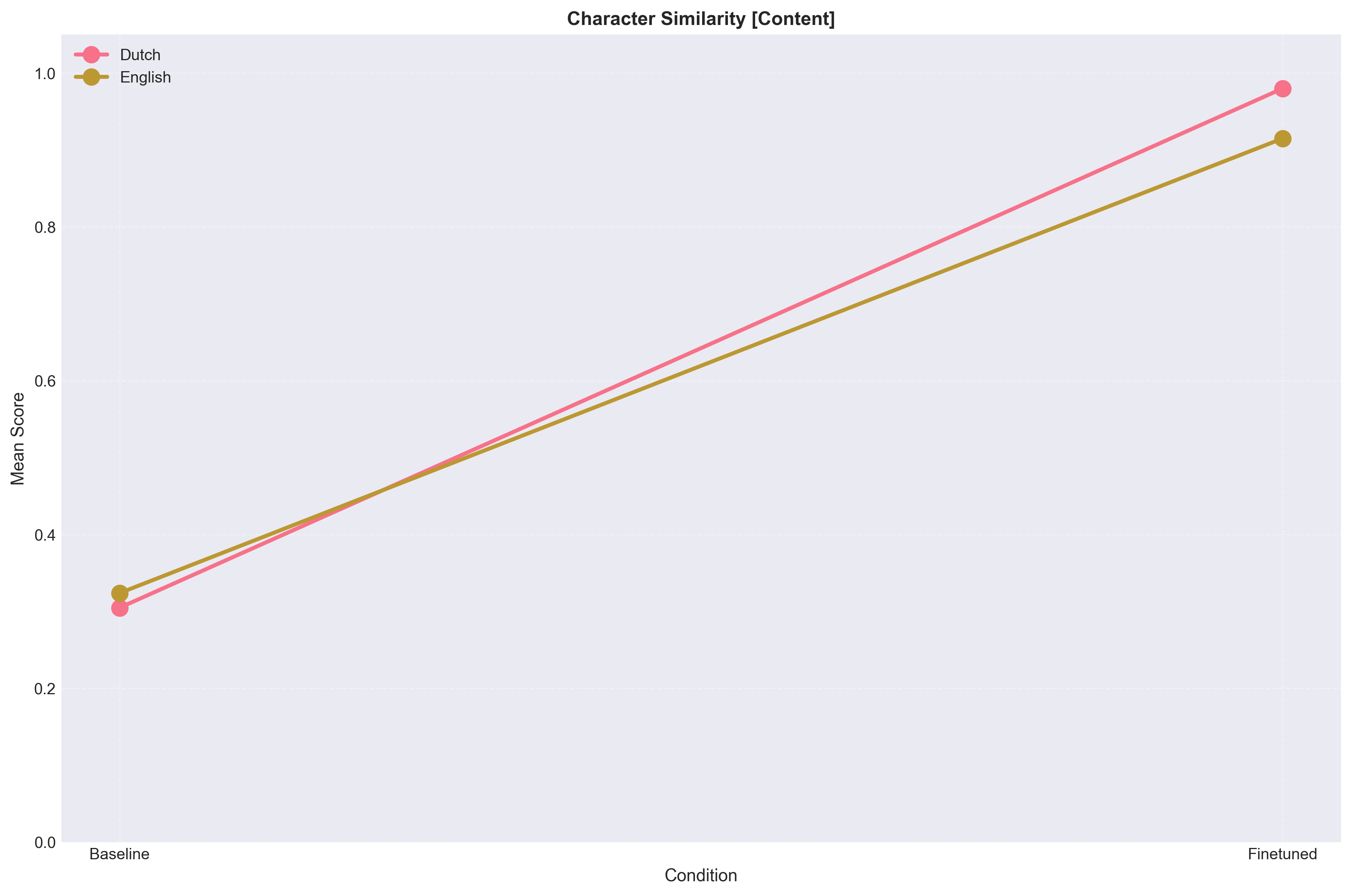}
    \caption{Character Similarity Interaction on Retain Set}
  \end{subfigure}
  \begin{subfigure}[t]{0.48\textwidth}
    \centering
    \includegraphics[width=\textwidth]{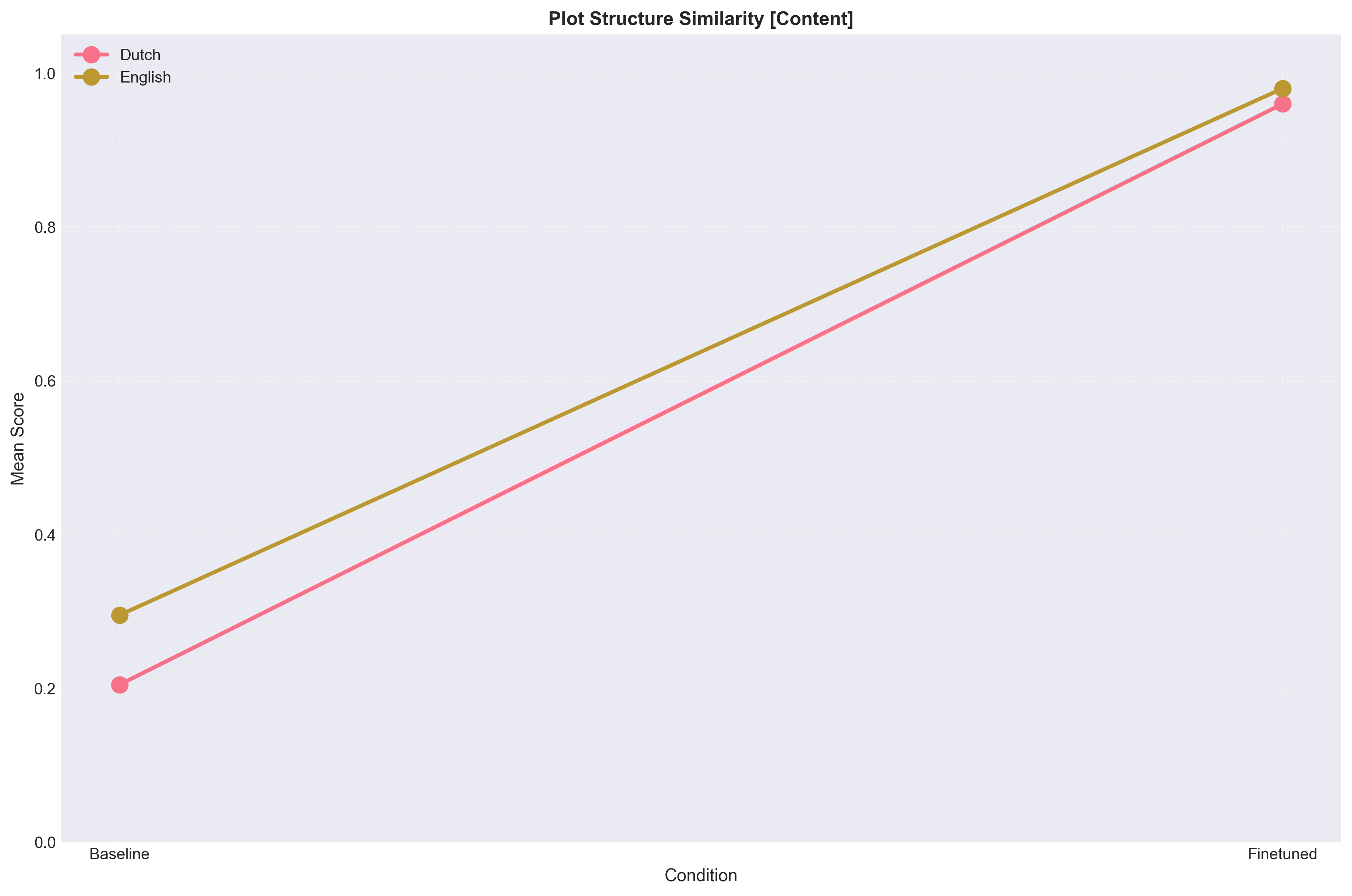}
    \caption{Plot Structure Similarity Interaction on Retain Set}
  \end{subfigure}
  \begin{subfigure}[t]{0.48\textwidth}
    \centering
    \includegraphics[width=\textwidth]{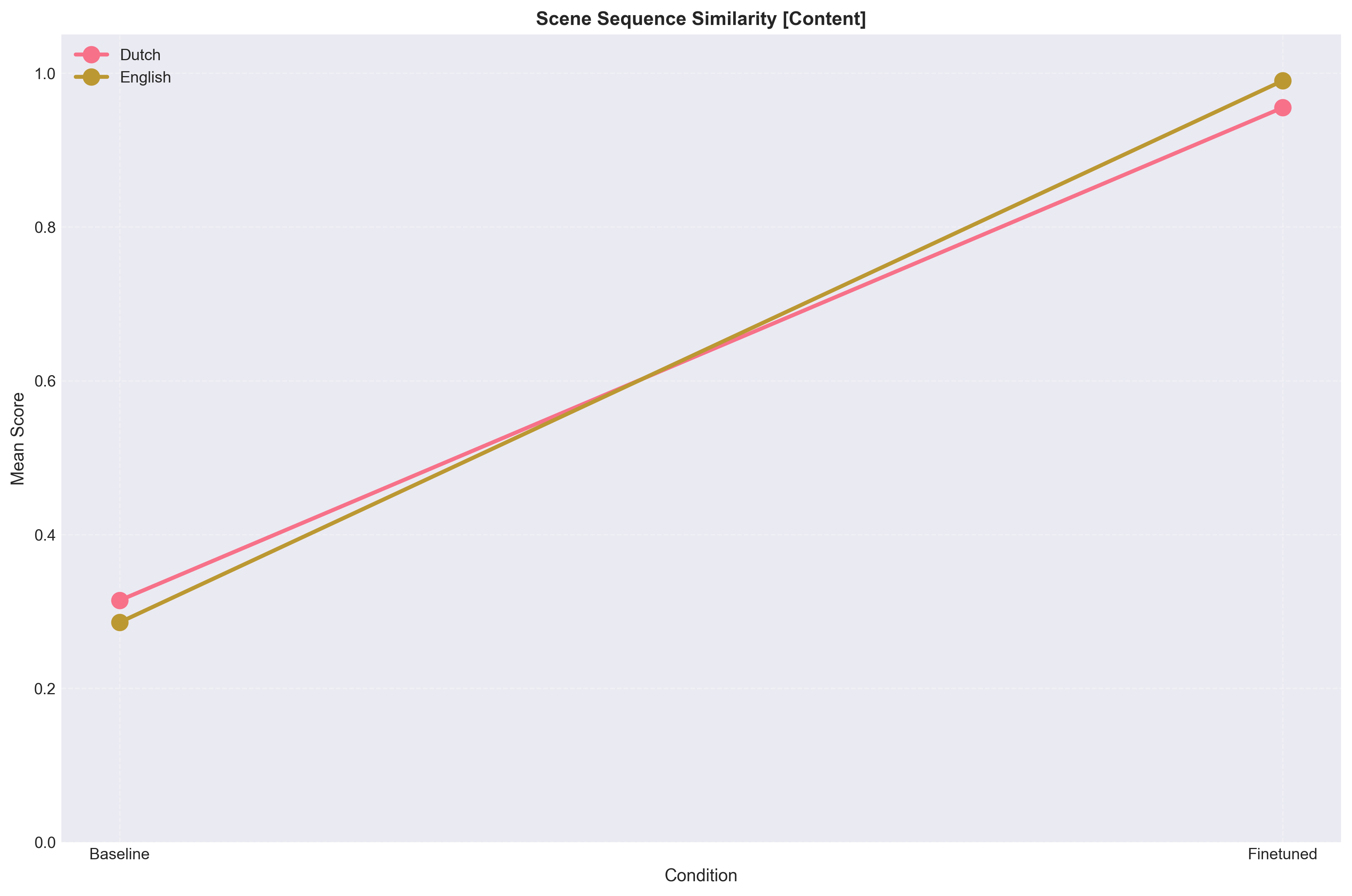}
    \caption{Scene Sequence Similarity Interaction on Retain Set}
  \end{subfigure}
  \begin{subfigure}[t]{0.48\textwidth}
    \centering
    \includegraphics[width=\textwidth]{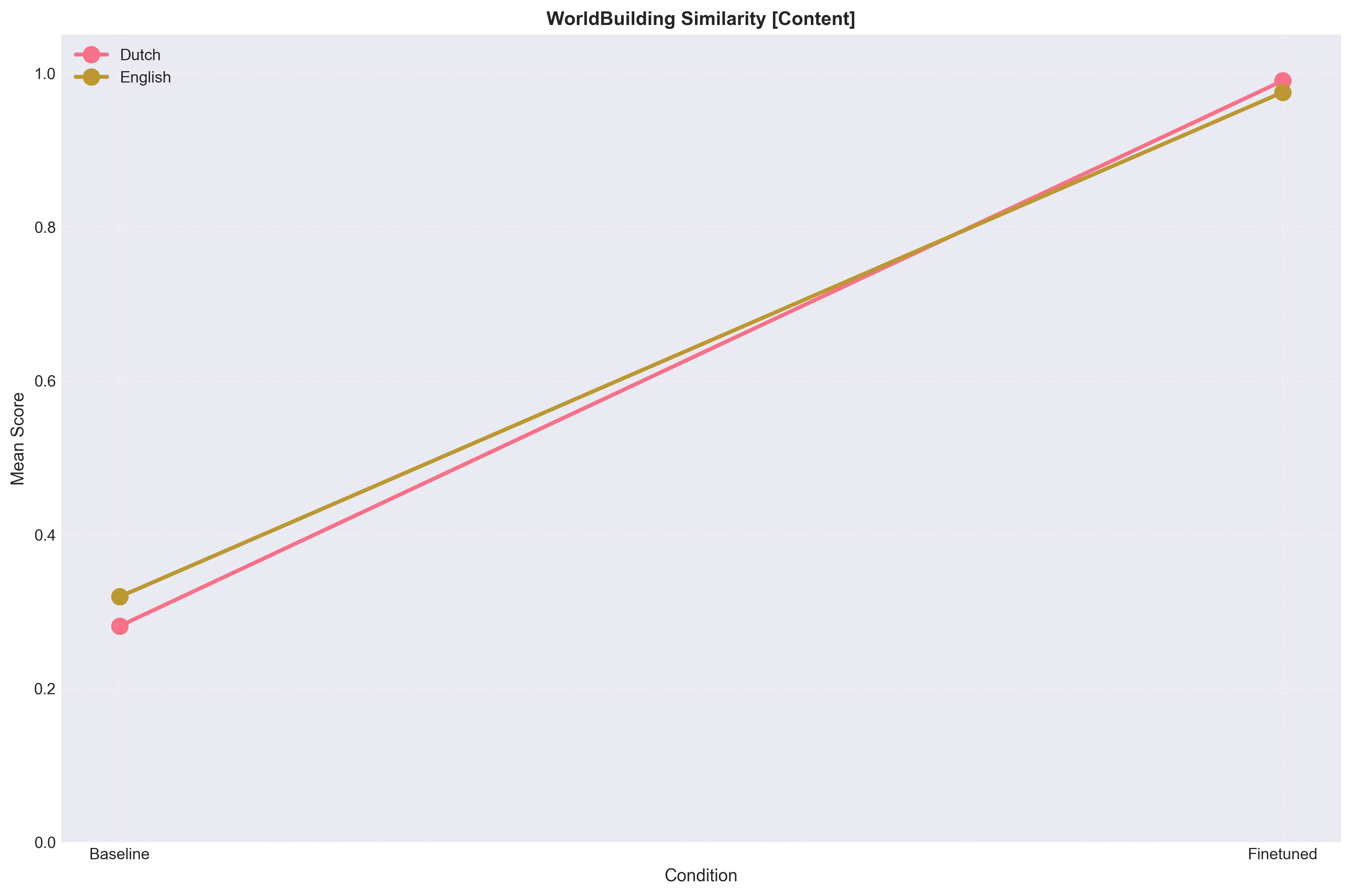}
    \caption{World Building Similarity Interaction on Retain Set}
  \end{subfigure}

  \caption[Stylistic and Content similarity before and after fine-tuning on the Retain set]{
  PSALM evaluator results for baseline and fine-tuned models where fine-tuning raises stylistic and content similarity from low to moderate levels to values close to one across both languages and splits indicating strong imprinting of the authors' style and narratorial choices}
  \label{fig:rq2:stylistic-metrics-retain}
\end{figure}

\begin{figure}[!hb]
  \centering

  \begin{subfigure}[t]{0.48\textwidth}
    \centering
    \includegraphics[width=\textwidth]{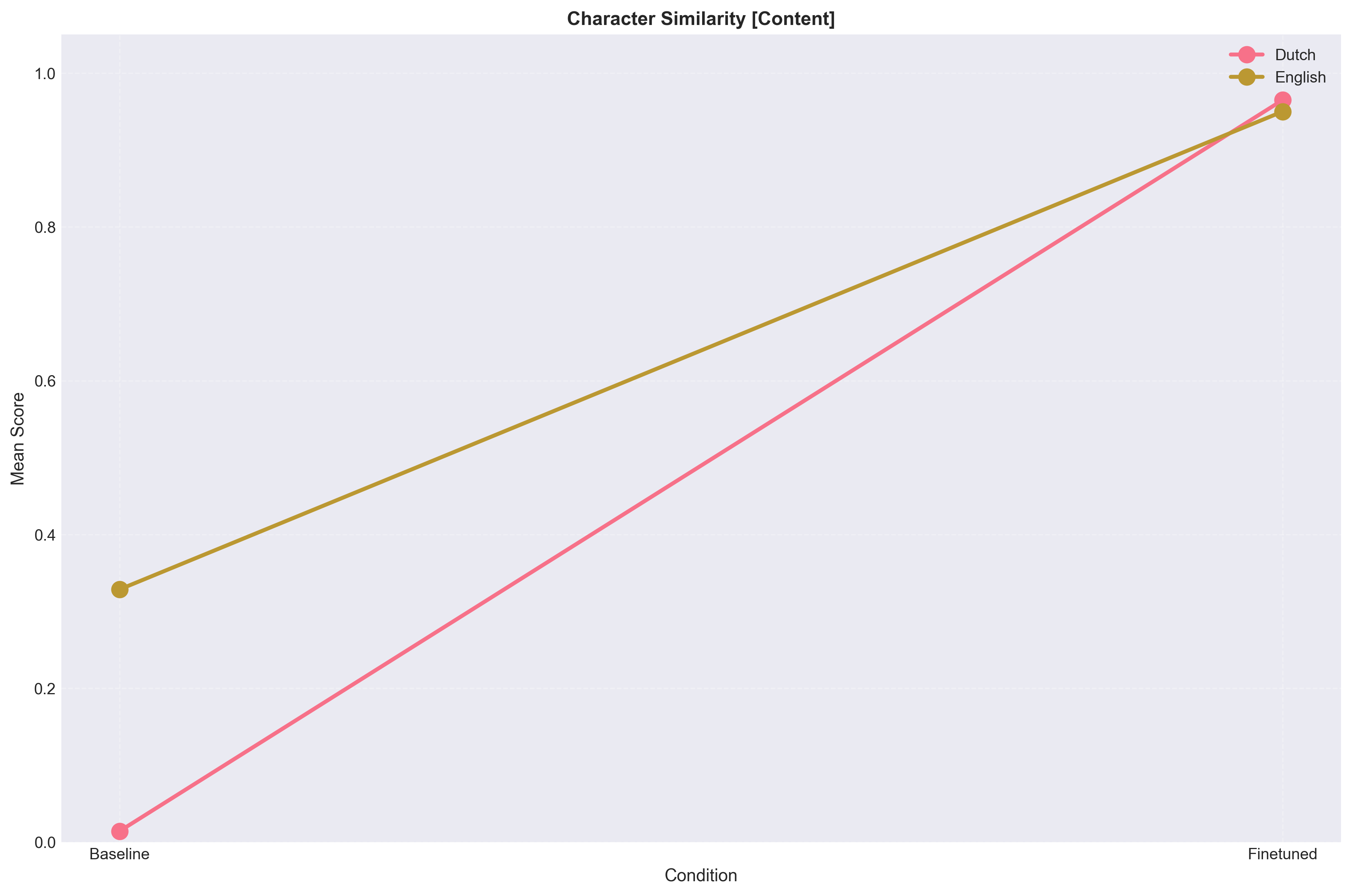}
    \caption{Character Similarity Interaction on Forget Set}
  \end{subfigure}
  \begin{subfigure}[t]{0.48\textwidth}
    \centering
    \includegraphics[width=\textwidth]{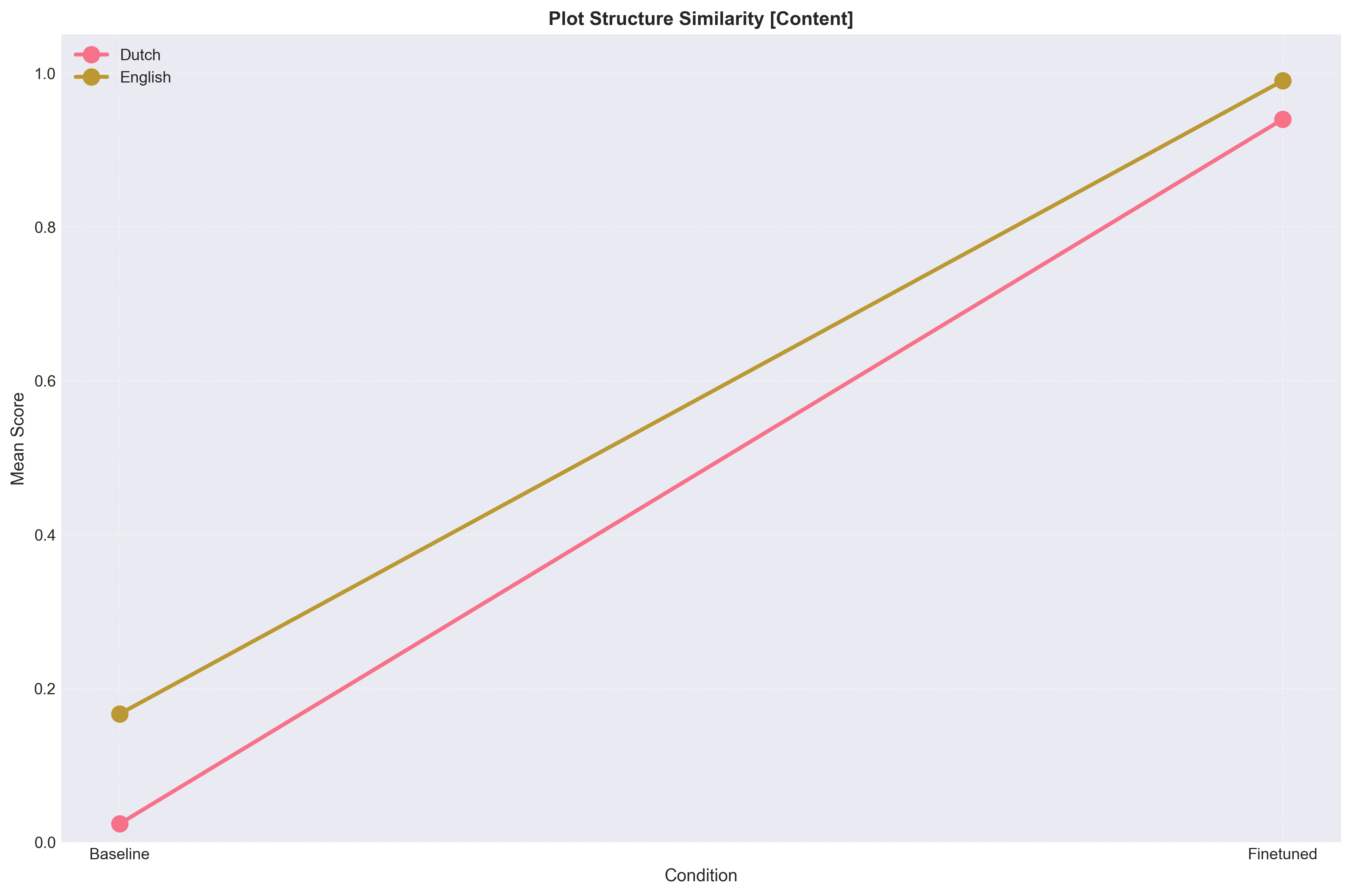}
    \caption{Plot Structure Similarity Interaction on Forget Set}
  \end{subfigure}
  \begin{subfigure}[t]{0.48\textwidth}
    \centering
    \includegraphics[width=\textwidth]{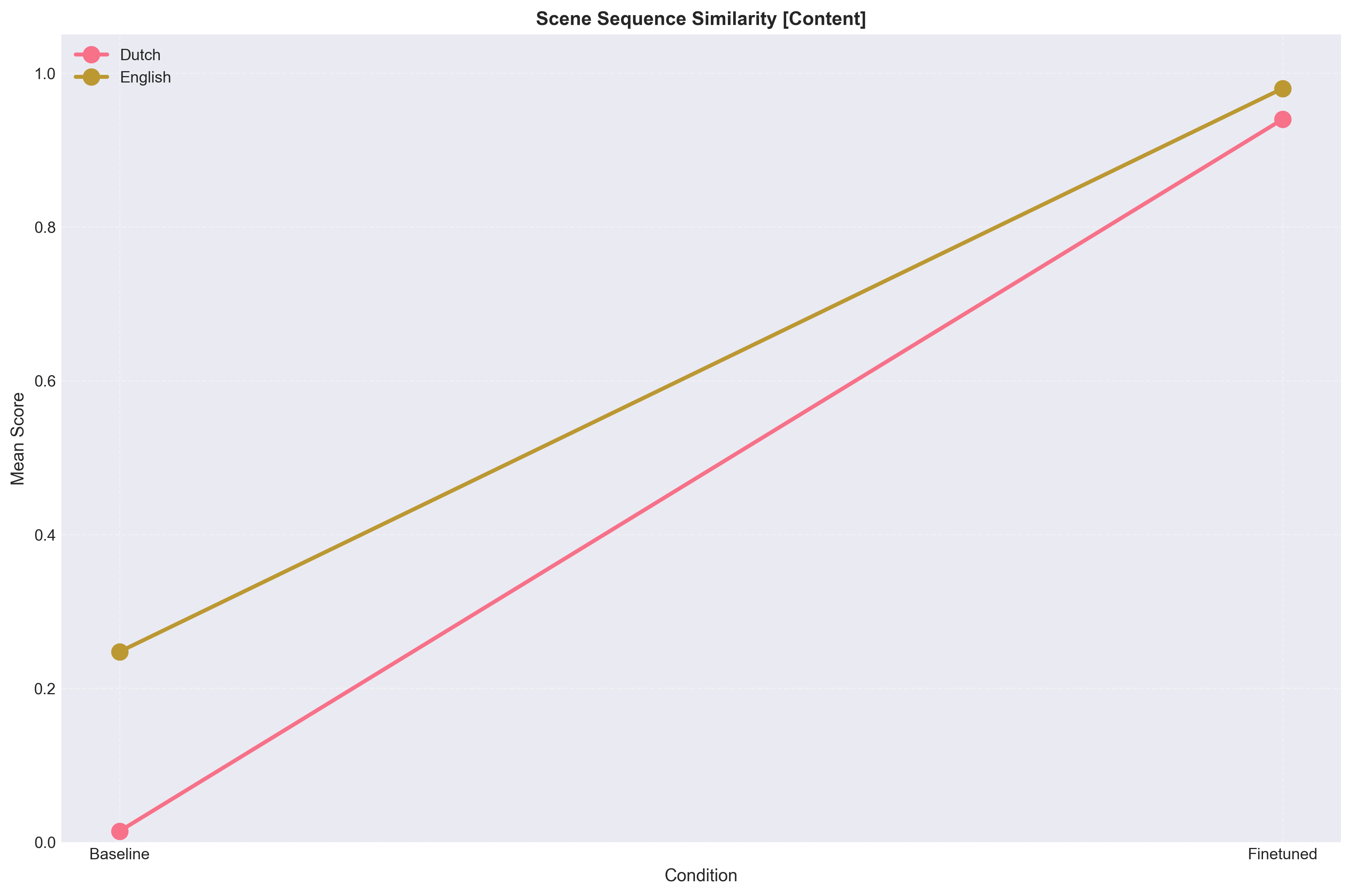}
    \caption{Scene Sequence Similarity Interaction on Forget Set}
  \end{subfigure}
  \begin{subfigure}[t]{0.48\textwidth}
    \centering
    \includegraphics[width=\textwidth]{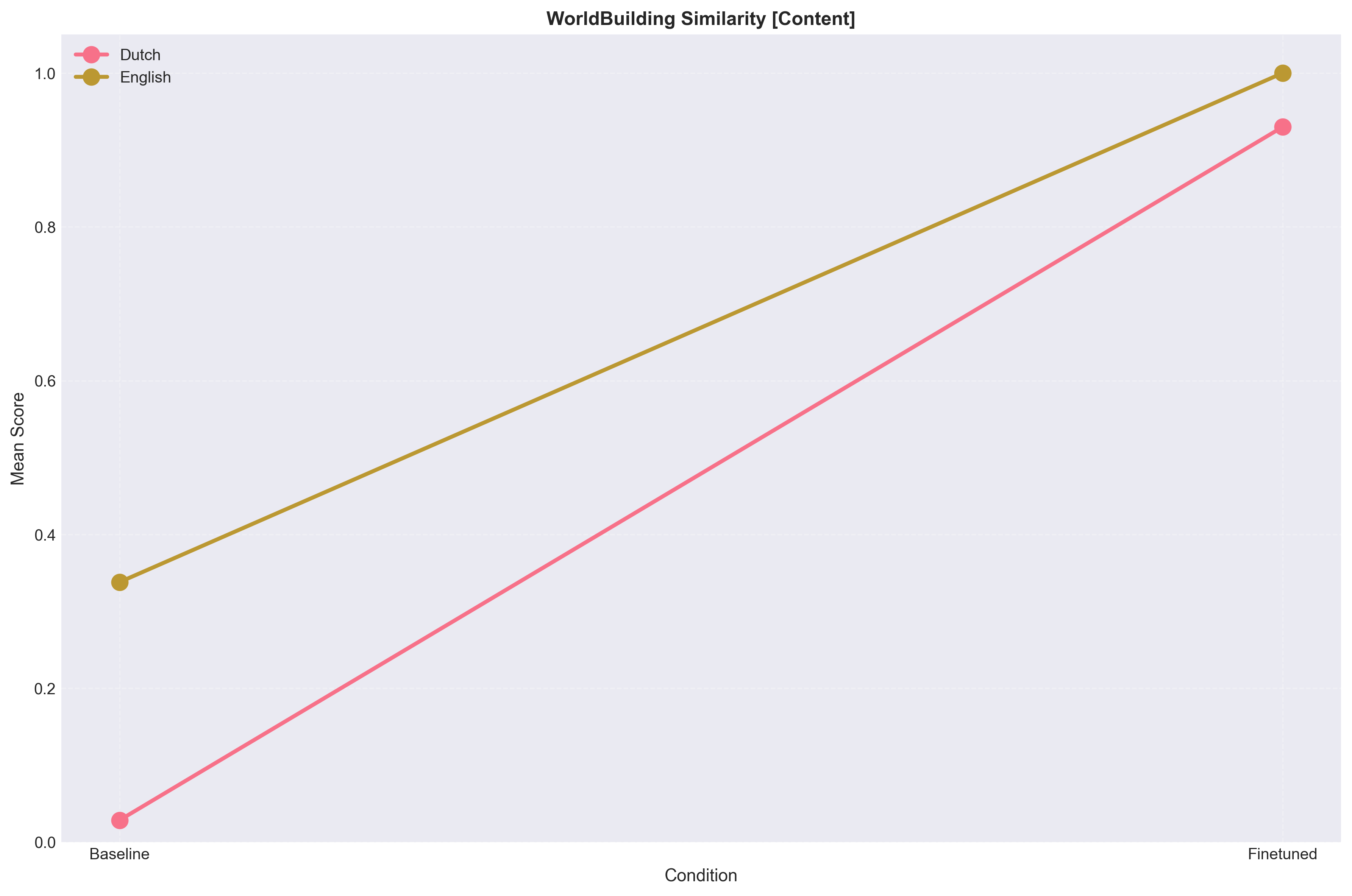}
    \caption{World Building Similarity Interaction on Forget Set}
  \end{subfigure}

  \caption[Stylistic and Content similarity before and after fine-tuning on the Forget set]{
 PSALM evaluator results for baseline and fine-tuned models where fine-tuning raises stylistic and content similarity from low to moderate levels to values close to one across both languages and splits indicating strong imprinting of the authors' style and narratorial choices}
  \label{fig:rq2:stylistic-metrics-forget-extended}
\end{figure}

The results of stylistic and content similarity before and after fine-tuning on the forget and retain sets can be seen in Figure~\ref{fig:rq2:stylistic-metrics-retain} and Figure~\ref{fig:rq2:stylistic-metrics-forget-extended}.

\subsection{Behaviour of Exceptions Extended}\label{app:rq2:behaviourofexceptions}
\begin{figure}[!hb]
  \centering

  \begin{subfigure}[t]{0.48\textwidth}
    \centering
    \includegraphics[width=\textwidth]{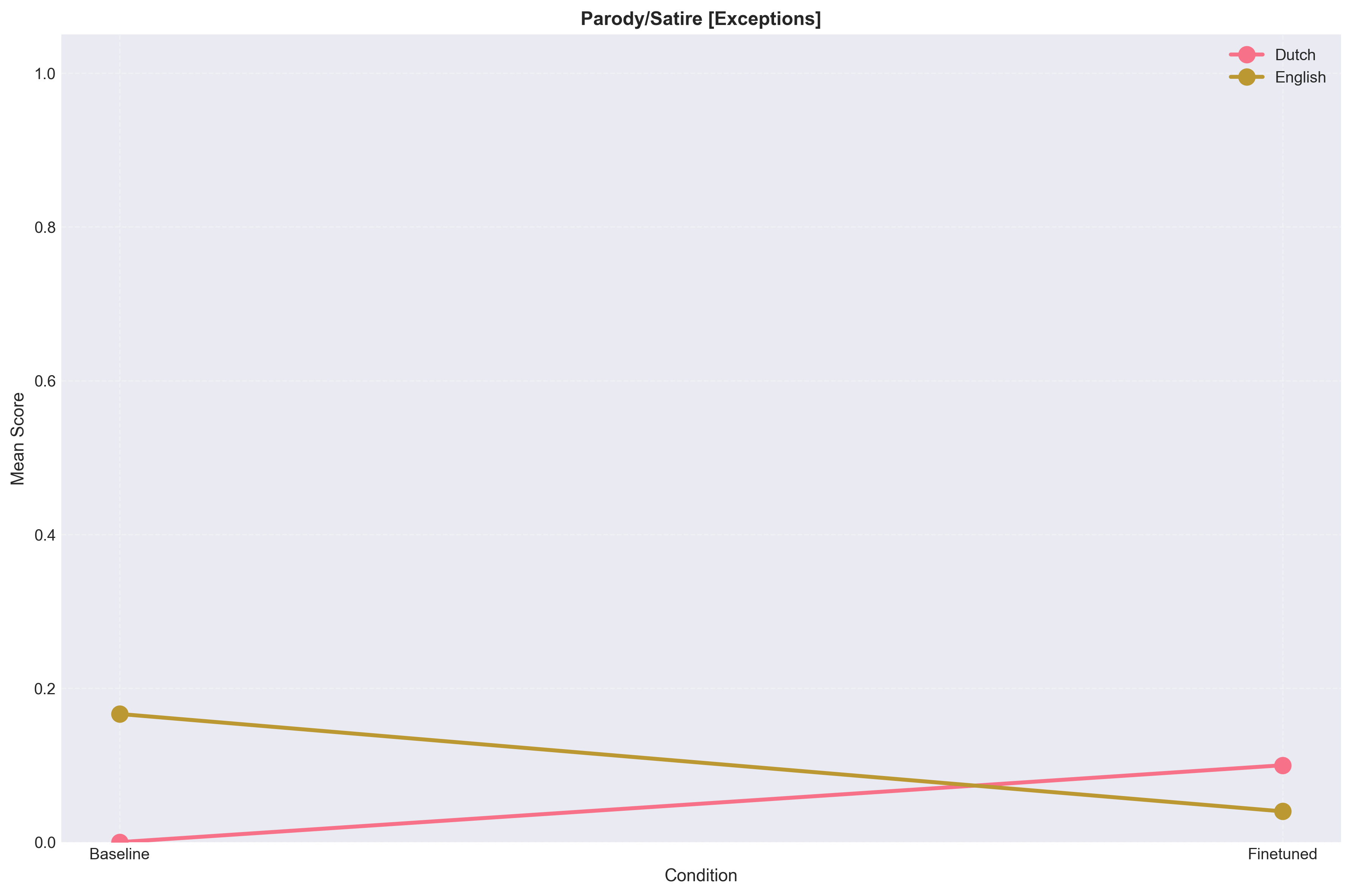}
    \caption{Parody/Satire Interaction on forget set}
  \end{subfigure}
  \begin{subfigure}[t]{0.48\textwidth}
    \centering
    \includegraphics[width=\textwidth]{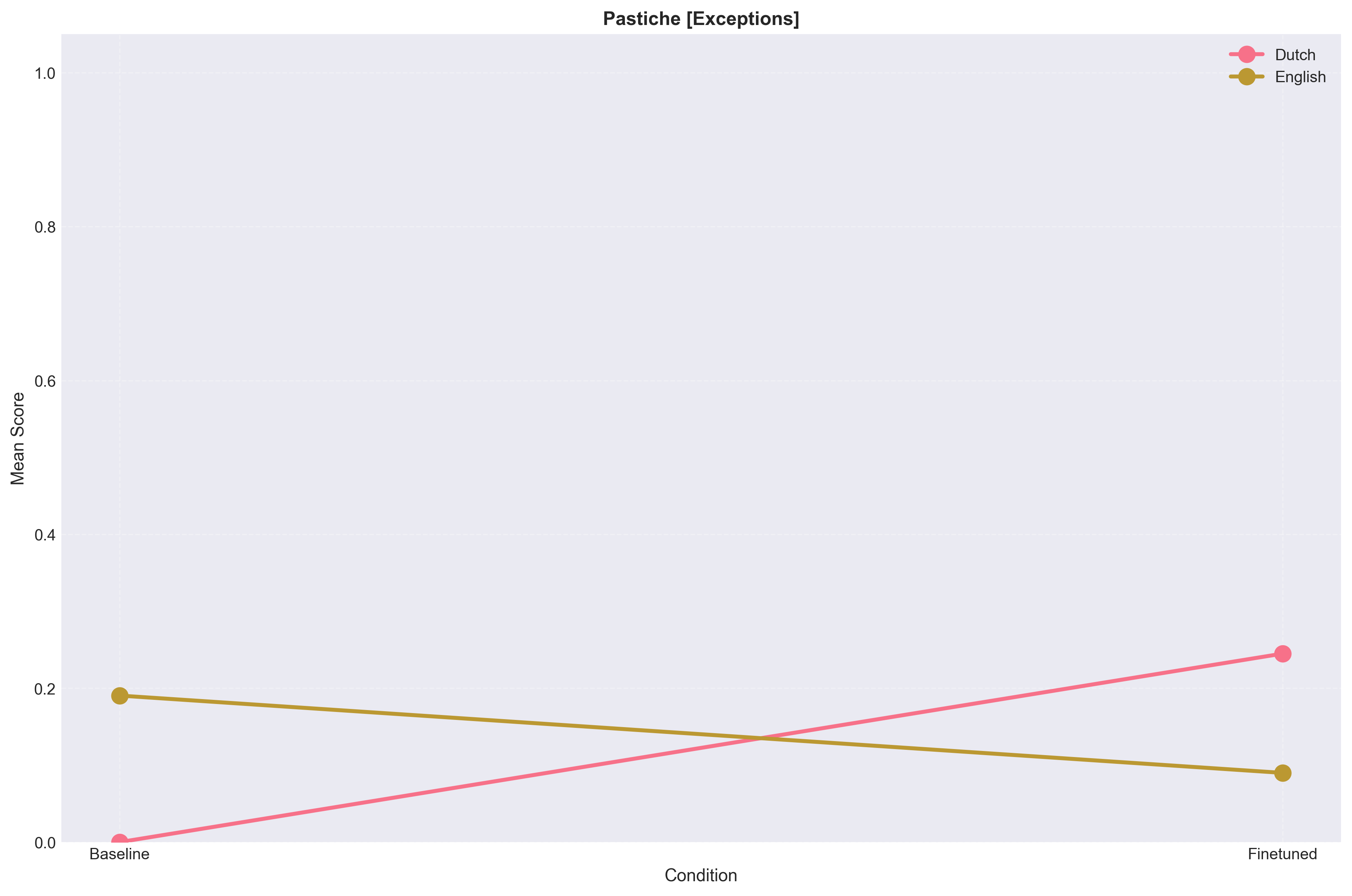}
    \caption{Pastiche Interaction on forget set}
  \end{subfigure}
  \begin{subfigure}[t]{0.48\textwidth}
    \centering
    \includegraphics[width=\textwidth]{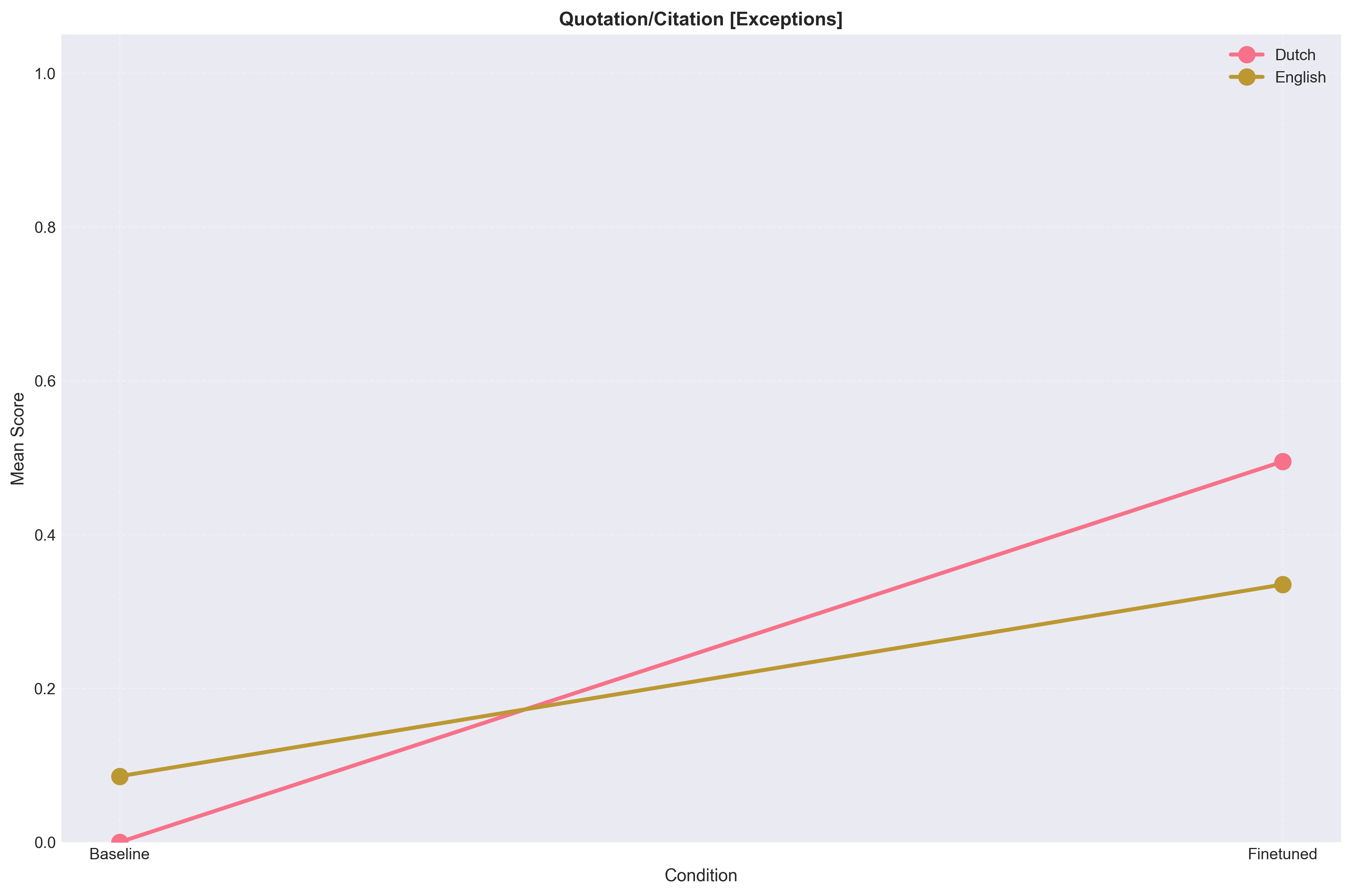}
    \caption{Quotation/Citation Interaction on forget set}
  \end{subfigure}
  \begin{subfigure}[t]{0.48\textwidth}
    \centering
    \includegraphics[width=\textwidth]{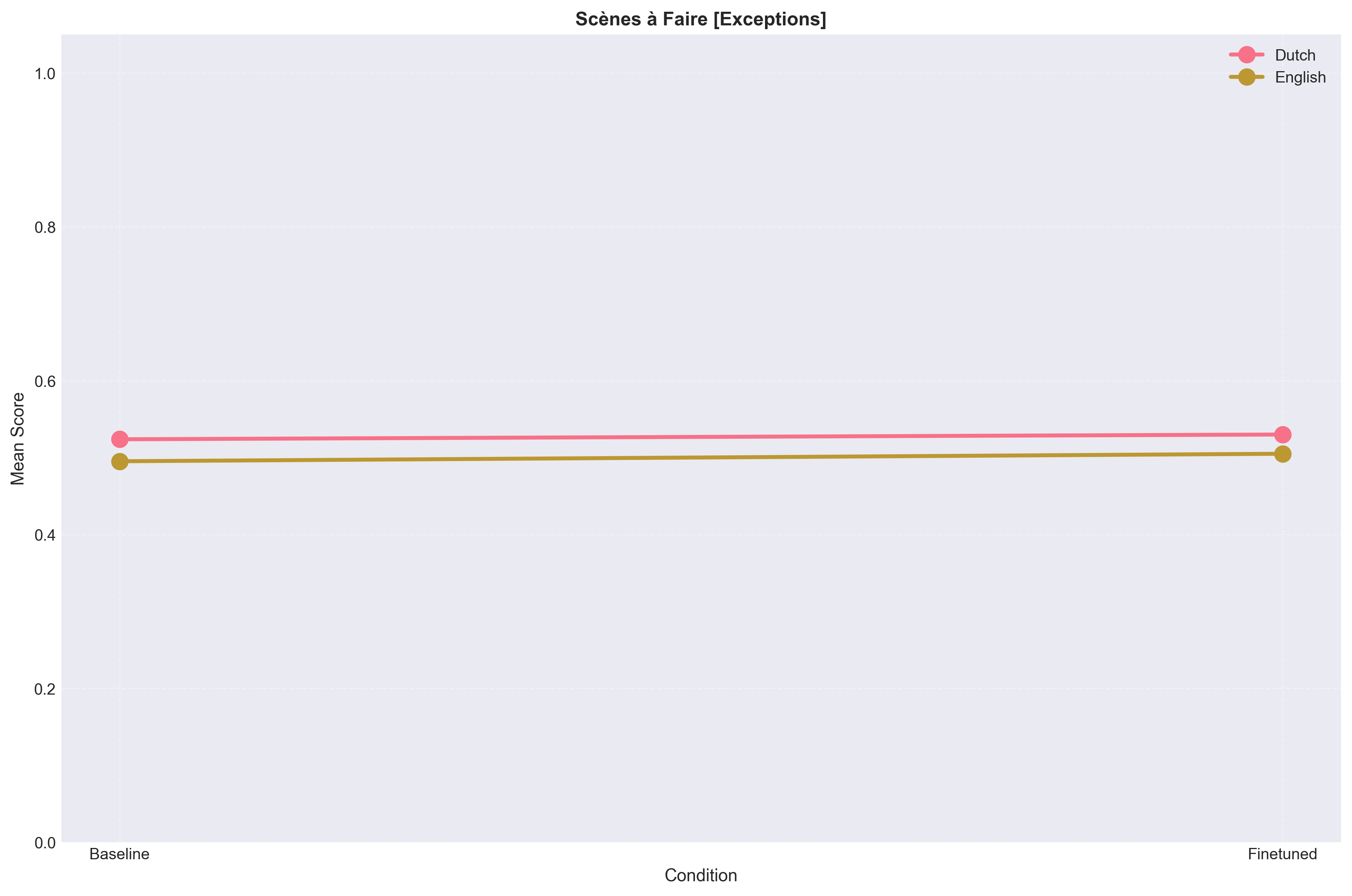}
    \caption{Scènes à Faire Interaction on forget set}
  \end{subfigure}

  \caption[Exception metrics before and after fine-tuning on forget set]{
  Exception metrics (Parody/Satire Pastiche Quotation/Citation and Scènes à Faire) for baseline and fine-tuned models on the forget set where fine-tuning modestly increases Quotation/Citation scores but leaves Parody/Satire Pastiche and Scènes à Faire largely unchanged}
  \label{fig:rq2:exception-metrics-forget}
\end{figure}

\begin{figure}[!hb]
  \centering

  \begin{subfigure}[t]{0.48\textwidth}
    \centering
    \includegraphics[width=\textwidth]{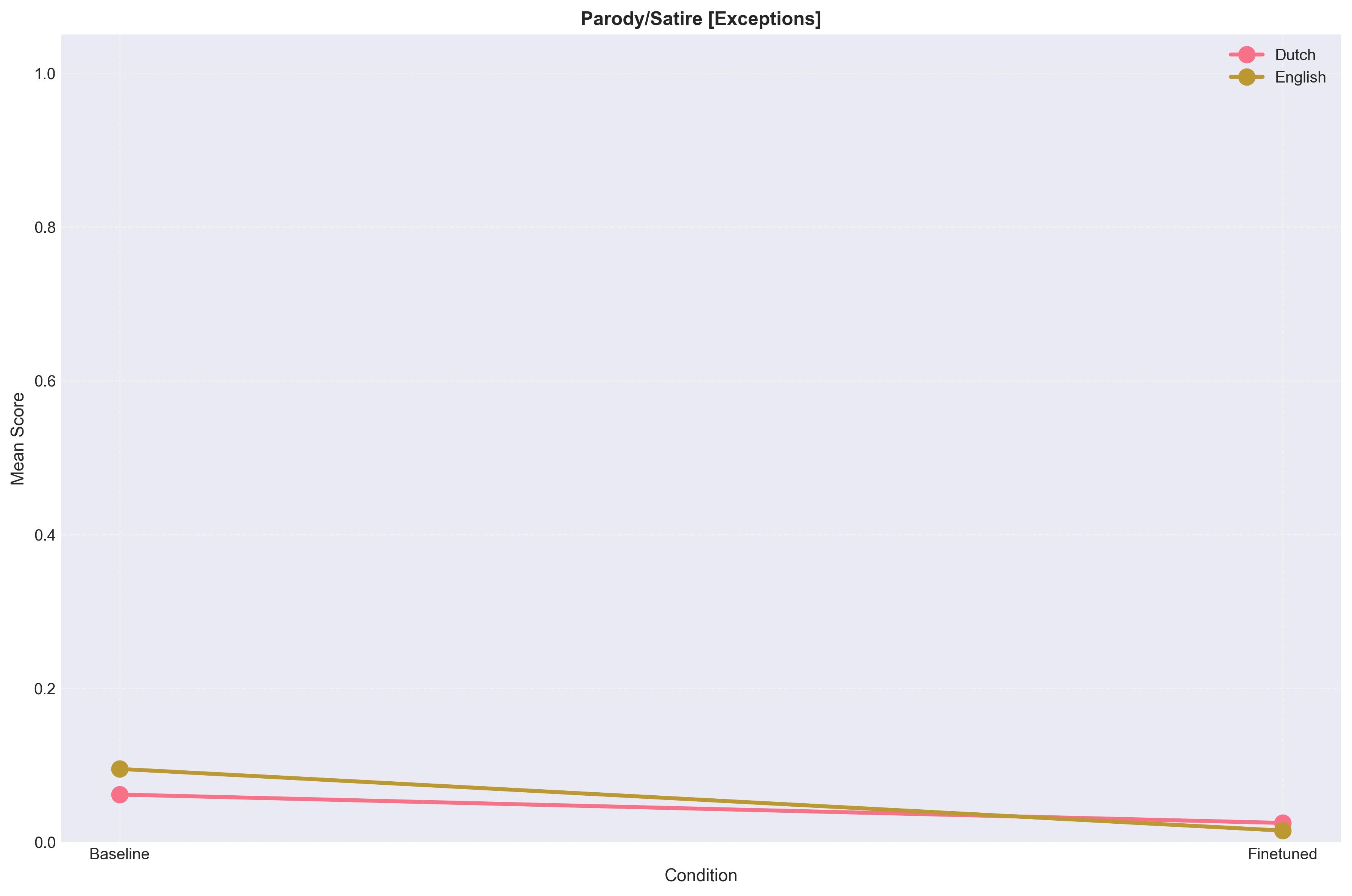}
    \caption{Parody/Satire Interaction on retain set}
  \end{subfigure}
  \begin{subfigure}[t]{0.48\textwidth}
    \centering
    \includegraphics[width=\textwidth]{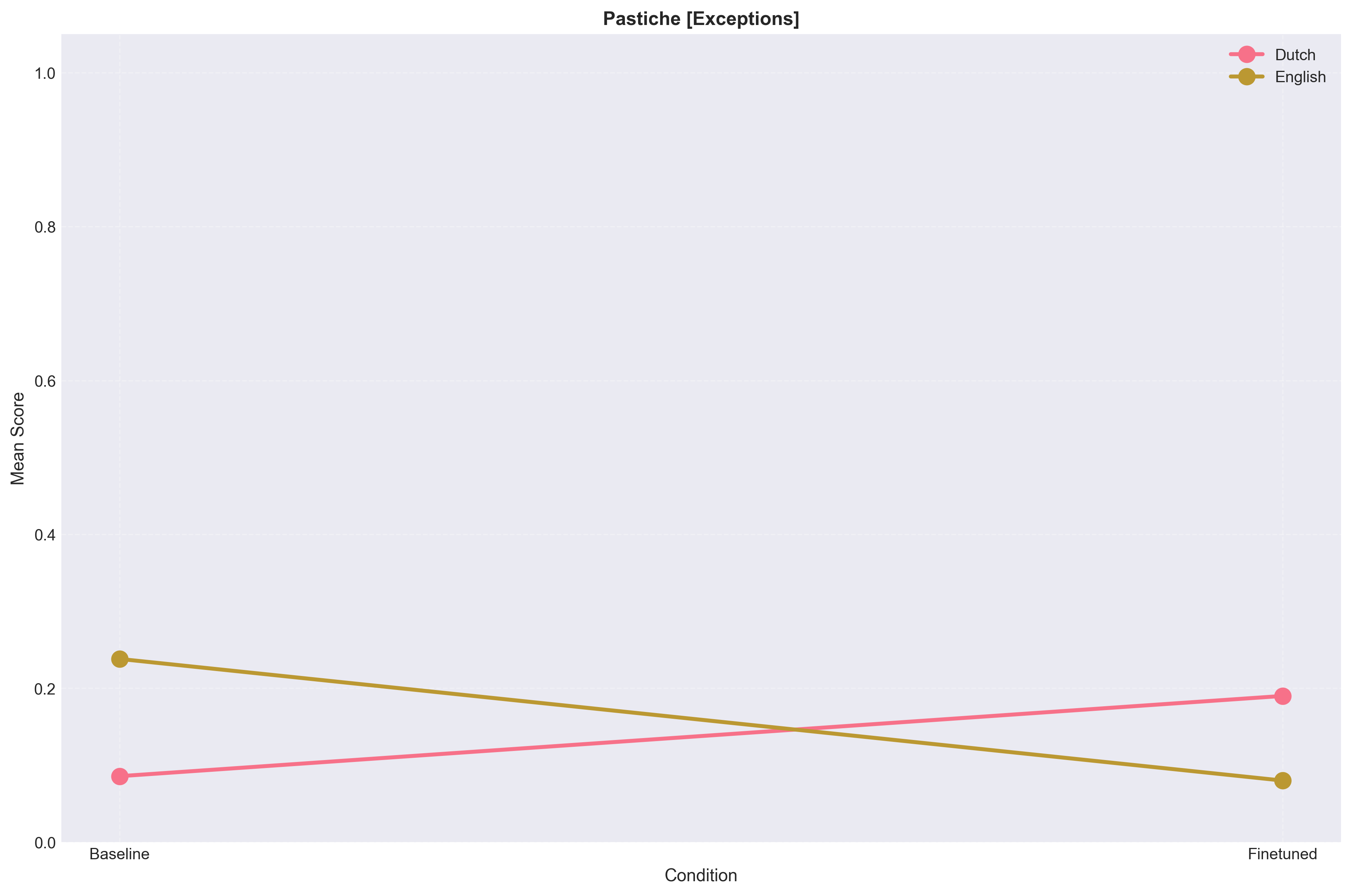}
    \caption{Pastiche Interaction on retain set}
  \end{subfigure}
  \begin{subfigure}[t]{0.48\textwidth}
    \centering
    \includegraphics[width=\textwidth]{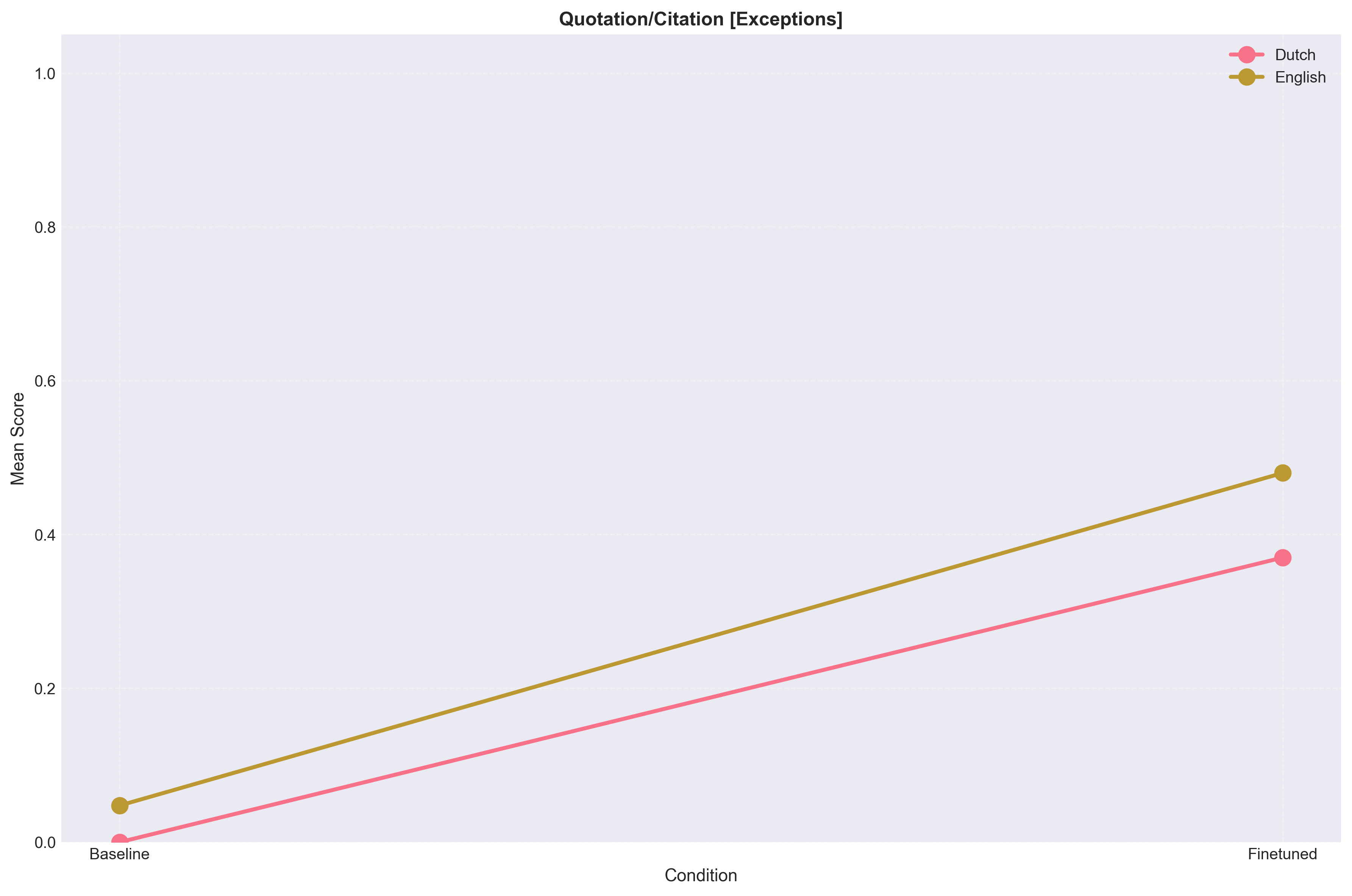}
    \caption{Quotation/Citation Interaction on retain set}
  \end{subfigure}
  \begin{subfigure}[t]{0.48\textwidth}
    \centering
    \includegraphics[width=\textwidth]{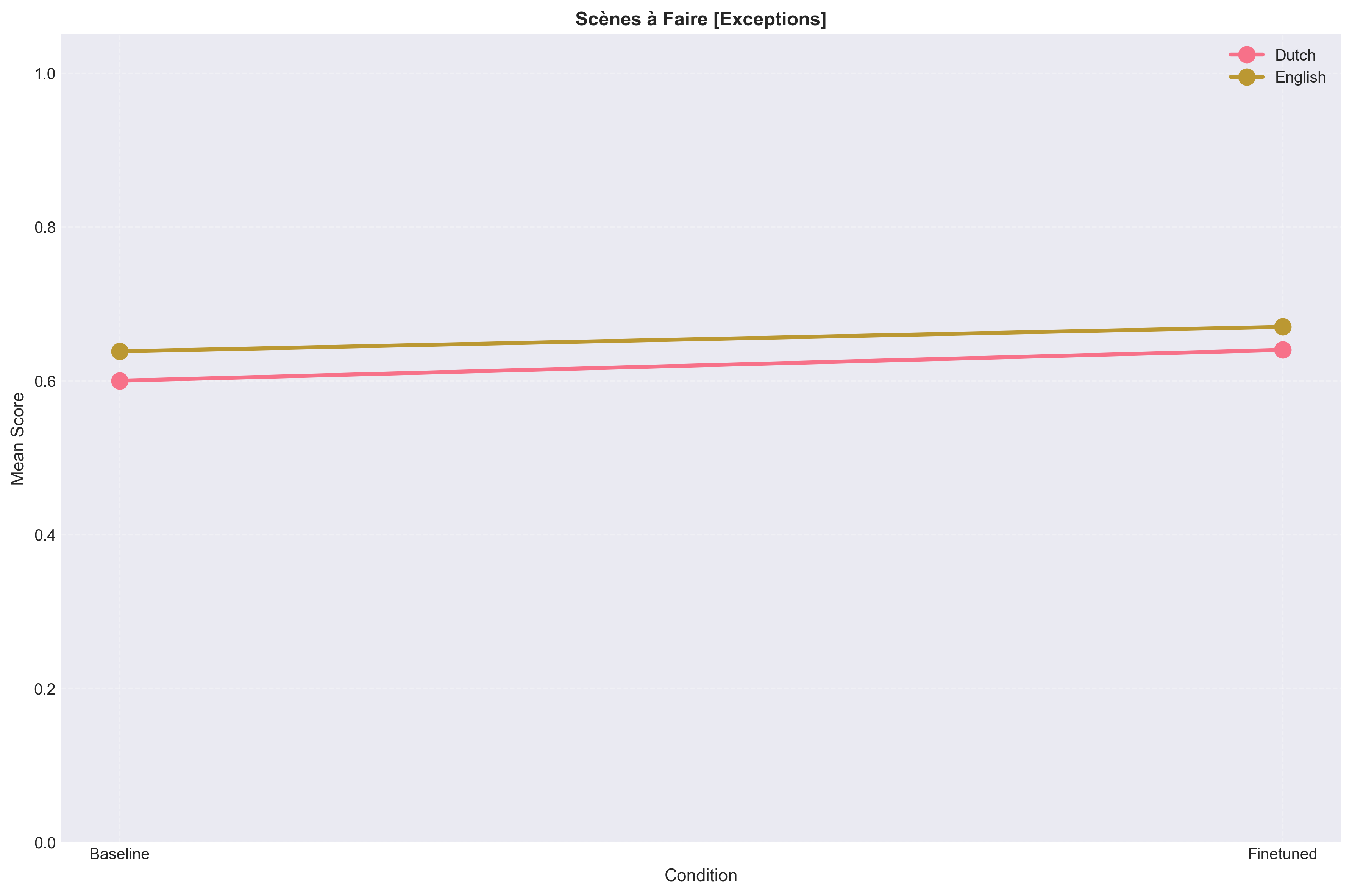}
    \caption{Scènes à Faire Interaction on retain set}
  \end{subfigure}

  \caption[Exception metrics before and after fine-tuning on retain set]{
  Exception metrics (Parody/Satire Pastiche Quotation/Citation and Scènes à Faire) for baseline and fine-tuned models on the retain set where fine-tuning modestly increases Quotation/Citation scores but leaves Parody/Satire Pastiche and Scènes à Faire largely unchanged}
  \label{fig:rq2:exception-metrics-retain}
\end{figure}

The results of exception metrics before and after fine-tuning on forget and retain sets can be seen in Figure~\ref{fig:rq2:exception-metrics-forget} and Figure~\ref{fig:rq2:exception-metrics-retain}.

\subsection{Comparing Metric Categories Extended}
\begin{figure}[!hb]
    \centering
    \adjustbox{max width=\textwidth}{
    \includegraphics[width=\textwidth]{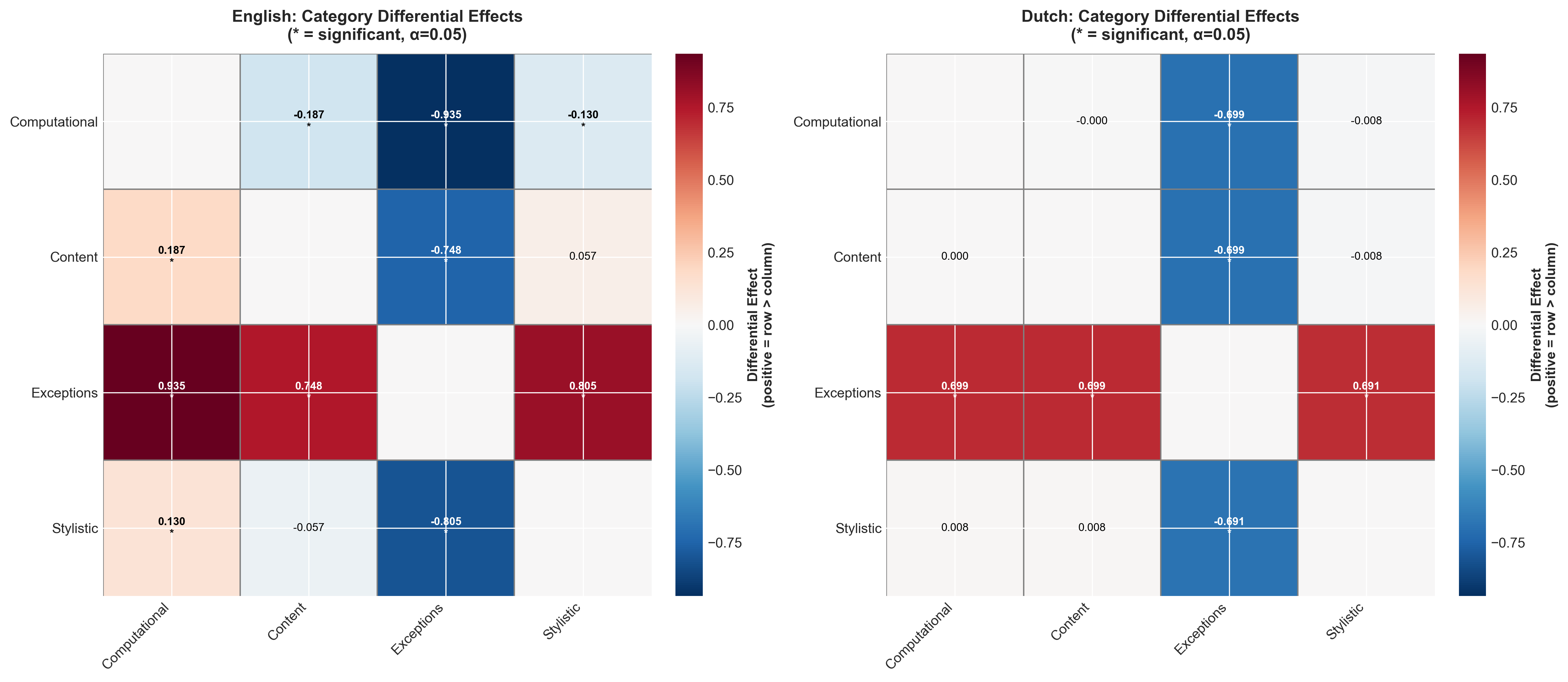}
    }
    
   \caption{Category-level pairwise contrasts for the forget split where cells show differences in reductions between categories and asterisks mark statistically significant contrasts}
    \label{fig:rq2:differential-effects-heatmap}
\end{figure}

A category-level pairwise heatmap can be seen in Figure~\ref{fig:rq2:differential-effects-heatmap}.

\subsection{Assumption Checks and Test Selection}
Shapiro–Wilk tests reported in Online Resources~4 and~5 show that none of the thirteen metrics followed a normal distribution in any model–split combination. This is expected given the strong ceiling and floor effects: baseline models often concentrate scores at or near~0, whereas fine-tuned models concentrate scores at or near~1 for infringement-related metrics. Levene’s tests indicate that variance homogeneity also fails for most metrics, especially on the retain split, where all but four metrics exhibit significant heteroscedasticity. Consequently, all main analyses use non-parametric procedures: Kruskal–Wallis tests for omnibus comparisons across the four model conditions, Mann–Whitney U-tests for independent pairwise contrasts, and Wilcoxon signed-rank tests for paired baseline–fine-tuned comparisons on shared prompts. Rank-based ART-ANOVA is used for language~$\times$~training-stage interactions.

\subsection{Omnibus Tests and Effect Size Patterns}
Kruskal–Wallis statistics in Online Resource~6 confirm that training stage has a very strong effect on all infringement-oriented metrics. For all computational, stylistic and content metrics on both forget and retain splits, the omnibus $p$-values are smaller than $10^{-11}$. Among the exception metrics, Quotation/Citation also shows highly significant omnibus differences on both splits (again with $p \ll 10^{-10}$), Parody/Satire and Pastiche are significant only on the forget split, and Scènes à Faire never reaches significance (forget: $p \approx 0.97$, retain: $p \approx 0.53$). This pattern agrees with the qualitative picture in the main text: fine-tuning produces large and systematic changes for the infringement dimensions and for Quotation/Citation, while other exceptions are only weakly affected and Scènes à Faire is essentially unchanged.

The effect size CSVs (Online Resources~7 and~8) show that almost all baseline–fine-tuned contrasts on computational, stylistic and content metrics are associated with very large Cohen’s $|d|$ and Cliff’s $|\delta|$. It is common for $|d|$ to exceed~2 for these metrics, and several computational comparisons reach values above~10 (for example BLEU and ROUGE on the retain split, where baseline means are below $0.1$ and fine-tuned means are exactly $1$). Cliff’s $\delta$ for these comparisons is typically very close to $-1$, reflecting almost complete separation between baseline and fine-tuned distributions. By contrast, most exception metrics have substantially smaller effect sizes. For Parody/Satire and Pastiche, $|d|$ values usually lie below~1, with positive and negative signs depending on the language and split. Quotation/Citation is the only exception metric that exhibits large effects, especially when moving from baseline to fine-tuned models on the retain split, where Cohen’s $d$ exceeds~2 in several comparisons. Scènes à Faire effect sizes are close to zero in both splits, which matches the non-significant omnibus tests.

\subsection{Category-Level Summaries and Ranks}
The per-category statistics in Online Resource~9 provide a compact view of these patterns. On the forget split, the Dutch baseline has mean scores of approximately $0.04$ (Computational), $0.08$ (Stylistic), $0.04$ (Content), and $0.13$ (Exceptions), whereas the English baseline is somewhat higher on all categories, particularly Content (mean $\approx 0.21$) and Exceptions (mean $\approx 0.21$). After fine-tuning, both languages converge to very similar category means: around $0.97$–$1.0$ for Computational, Stylistic and Content, and mid-range values for Exceptions (roughly $0.25$–$0.37$ depending on language and split). The retain split shows the same pattern, with baselines starting slightly higher than on the forget split for Content and Stylistic categories, and fine-tuned models again close to~1 on all infringement categories.

These differences are also reflected in the rank-based summary in Online Resource~10. When all thirteen metrics are ranked jointly per split, the two fine-tuned models always achieve the best average ranks, around $1.6$–$1.8$, while the baselines occupy the third and fourth positions, with average ranks between about $2.8$ and $3.8$. This holds on both forget and retain splits and confirms that, viewed across the full metric set, the fine-tuned models are consistently judged more similar to the sources than either baseline.

Category-level differential-effect analyses in Online Resource~11 further support the claim in the main text that fine-tuning acts as a broadly symmetric amplifier across infringement categories. For Dutch on the forget split, the mean reduction (baseline minus fine-tuned) is almost identical for Computational, Stylistic and Content categories (roughly $-0.91$ in each case), and significantly larger in magnitude than the reduction for Exceptions (around $-0.21$). For English, reductions for Computational, Stylistic and Content are again large and negative, and all three differ markedly from the Exceptions category. There is some evidence that computational metrics increase slightly more than content metrics in English (the Computational–Content contrast is statistically significant), but the absolute differences in magnitude are small compared to the gap between the infringement categories and the Exceptions category.

\begin{table}
  \centering
  \caption{Summary of baseline-to-fine-tuned changes by category (forget split). Values are mean score differences (baseline minus fine-tuned) aggregated over metrics within each category}
  \label{tab:rq2:summary_deltas}
  \begin{tabular}{lcccc}
    \toprule
    Language & Computational & Stylistic & Content & Exceptions \\
    \midrule
    English & $\approx -0.95$ & $\approx -0.82$ & $\approx -0.76$ & $\approx -0.01$ \\
    Dutch   & $\approx -0.91$ & $\approx -0.90$ & $\approx -0.91$ & $\approx -0.21$ \\
    \bottomrule
  \end{tabular}
\end{table}

Table~\ref{tab:rq2:summary_deltas} synthesises these differential effects. The precise values come directly from Online Resource~11. In both languages the three infringement categories show reductions close to $-1$, while the Exceptions category shows only small shifts, particularly in English.

\subsection{Language–Stage Interactions}

The rank-based ART-ANOVA results in Online Resources~12 and~13 provide a systematic view of interactions between language (English vs.\ Dutch) and training stage (baseline vs.\ fine-tuned). On the forget split, significant language main effects appear for several metrics, including BLEU, Writing Style, Narrative Voice, Character Similarity, Scene Sequence, World Building, Parody/Satire, Pastiche and Quotation/Citation. In each of these cases, inspection of the per-metric descriptive statistics shows the same qualitative pattern: the English baseline starts with higher similarity than the Dutch baseline, but the two languages converge after fine-tuning, with English and Dutch fine-tuned models attaining similar means. For BLEU, Character Similarity, Scene Sequence, World Building, Parody/Satire, Pastiche and Quotation/Citation, the ART-ANOVA also reports significant language~$\times$~training-stage interactions, which again reflect this “convergence” pattern rather than a reversal of direction.

On the retain split, interactions are sparser but still present. BLEU and ROUGE exhibit significant language effects and significant interactions. Here, English baselines are higher than Dutch baselines, and both fine-tuned models reach the same ceiling (BLEU~$=1$, ROUGE~$=1$), so the interaction is driven by larger English-to-English changes than Dutch-to-Dutch changes. Pastiche also shows a significant interaction on the retain set; the underlying means indicate that English models experience a small decrease in Pastiche scores after fine-tuning, while Dutch models show a small increase, but the absolute magnitudes are low in both cases and remain well below the levels associated with strong pastiche characteristics.

These interaction results do not contradict the main conclusion that fine-tuning largely erases pre-training language differences. Rather, they make that conclusion more precise: for several metrics, English models begin closer to the target works, especially on content dimensions, whereas Dutch models start further away; fine-tuning then aligns both languages to similar similarity levels. Exceptions behave differently: Quotation/Citation shows strong increases in both languages, Parody/Satire and Pastiche change only modestly, and Scènes à Faire remains stable.

\subsection{Paired Analyses on Shared Prompts}
The paired Wilcoxon tests in Online Resources~14 and~15 compare baseline and fine-tuned models on exactly the same prompts within each language. For the English forget split, the baseline means are well below $0.5$ on all infringement metrics and the fine-tuned means lie very close to~1. The resulting mean differences range from about $0.51$ (Character Similarity) to $0.90$–$0.93$ (computational metrics), and all $p$-values are well below $10^{-5}$. Cohen’s $d$ values for these paired comparisons reach extremely large magnitudes, for example $d \approx 15.6$ for BLEU and $d \approx 11.1$ for ROUGE. Dutch paired comparisons on the forget split show the same pattern with similar effect sizes. On the retain split, paired differences are slightly smaller but remain large for all infringement metrics; for example, English BLEU increases by about $0.89$ on average, and Dutch BLEU by about $0.97$. For several exception metrics, the paired tests reveal small but significant changes (e.g.\ a decrease in English Parody/Satire means on the retain split), whereas others such as Scènes à Faire show no significant paired change.

The paired analyses therefore confirm, at the level of individual prompts, the aggregate picture provided by the Kruskal–Wallis and effect-size summaries: fine-tuning consistently increases similarity across computational, stylistic and content dimensions, both for forget and retain examples, while producing more modest and metric-specific changes in exception dimensions.

\subsection{Distributional Characteristics and Atypical Cases}
The descriptive statistics in Online Resource~6 also reveal how the shape of the score distributions changes across model conditions. For most infringement metrics, baseline medians on the forget split are exactly~0.0 for both languages, whereas fine-tuned medians are exactly~1.0. The standard deviations for fine-tuned models are small (often less than $0.05$ on the retain split and slightly larger on the forget split), indicating that almost all outputs are scored at or very near the maximum similarity level once fine-tuning is applied. Baseline standard deviations are much larger and medians sometimes fall in the mid-range on the retain split, especially for English content metrics, reflecting the fact that the pre-trained model already exhibits some generic similarity to the historical fiction corpus.

Infrequent cases where fine-tuned scores fall below the ceiling can be seen, for example, in Dutch Exact Match on the forget split (mean $0.90$, standard deviation $\approx 0.31$), where a few prompts do not lead to exact reproductions, and in Dutch Character Similarity and World Building on the forget split, where the means are above $0.93$ but standard deviations around $0.18$–$0.23$ indicate some dispersion below~1. Manual inspection (reported in the main RQ2 discussion) attributes these deviations to translation noise, sampling variance, or prompts referring to story elements that are only weakly represented in the selected passages. These atypical cases do not alter the overall pattern, but they are relevant when interpreting ceiling effects: the near-1.0 means are not artefacts of a single extreme example, but reflect a general shift of the entire distribution towards very high similarity.

\section{RQ3: Detailed Statistical Analyses}\label{app:rq3_details}
This appendix summarises the core supplementary findings for RQ3, using the supplementary data. The aim is to document the main numerical patterns that are not already discussed in subsection~\ref{sec:unlearning-effectiveness}.

\subsection{Stylistic and Structural Similarity Extended}
\begin{figure}[!hb]
  \centering

  \begin{subfigure}[t]{0.48\textwidth}
    \centering
    \includegraphics[width=\textwidth]{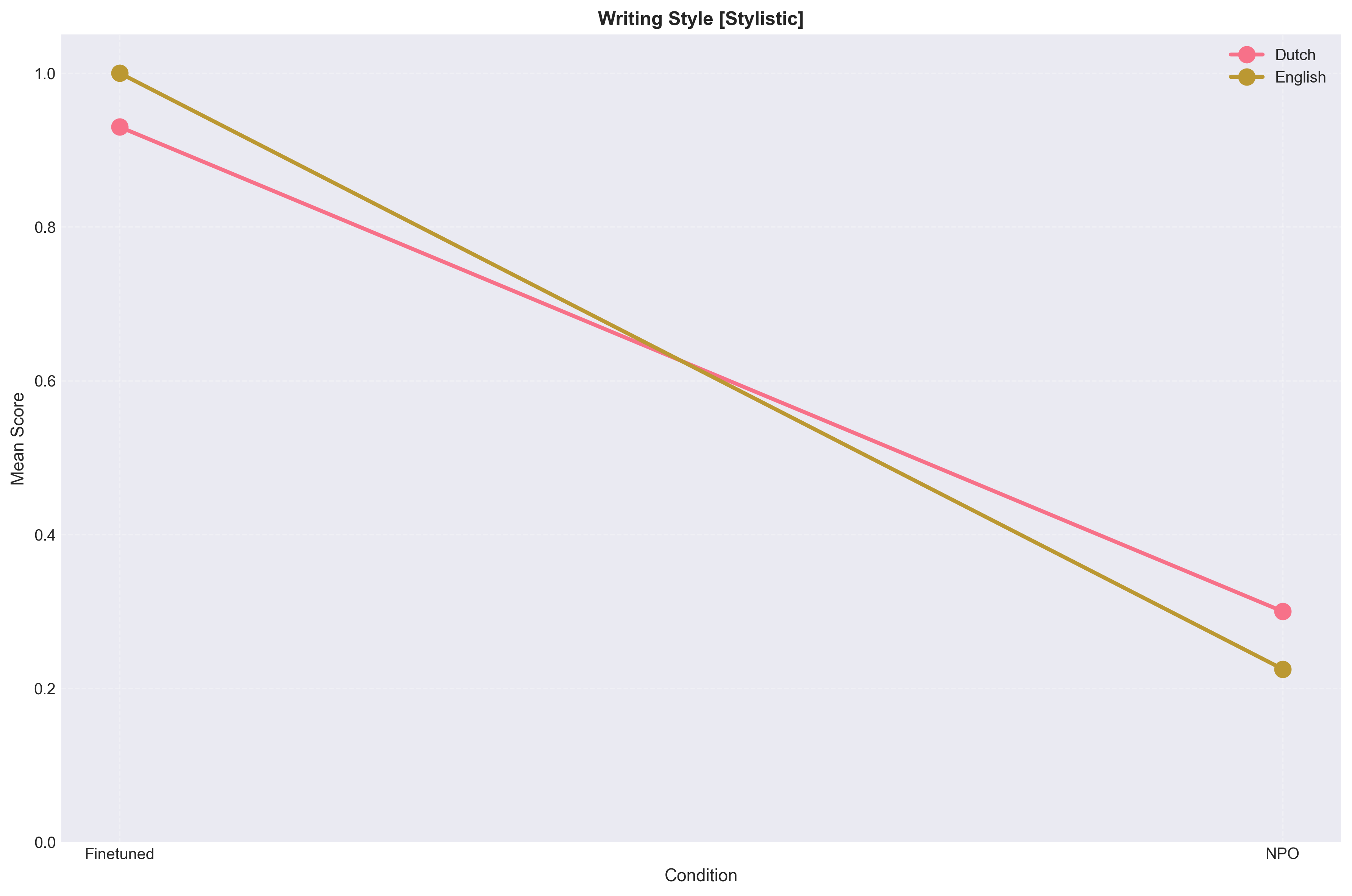}
    \caption{Writing Style Interaction on Forget Set}
  \end{subfigure}
  \begin{subfigure}[t]{0.48\textwidth}
    \centering
    \includegraphics[width=\textwidth]{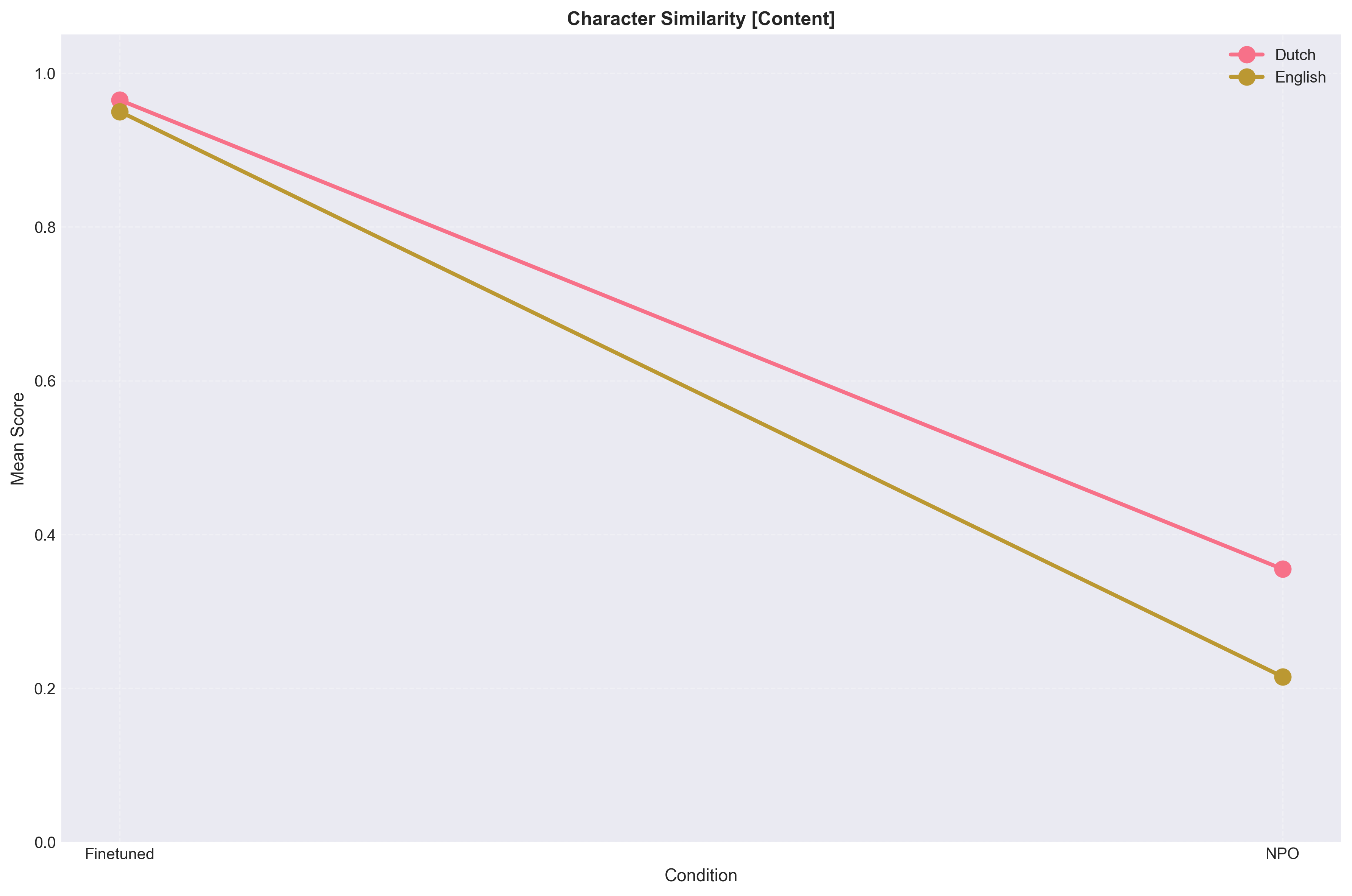}
    \caption{Character Similarity Interaction on Forget Set}
  \end{subfigure}

  \begin{subfigure}[t]{0.48\textwidth}
    \centering
    \includegraphics[width=\textwidth]{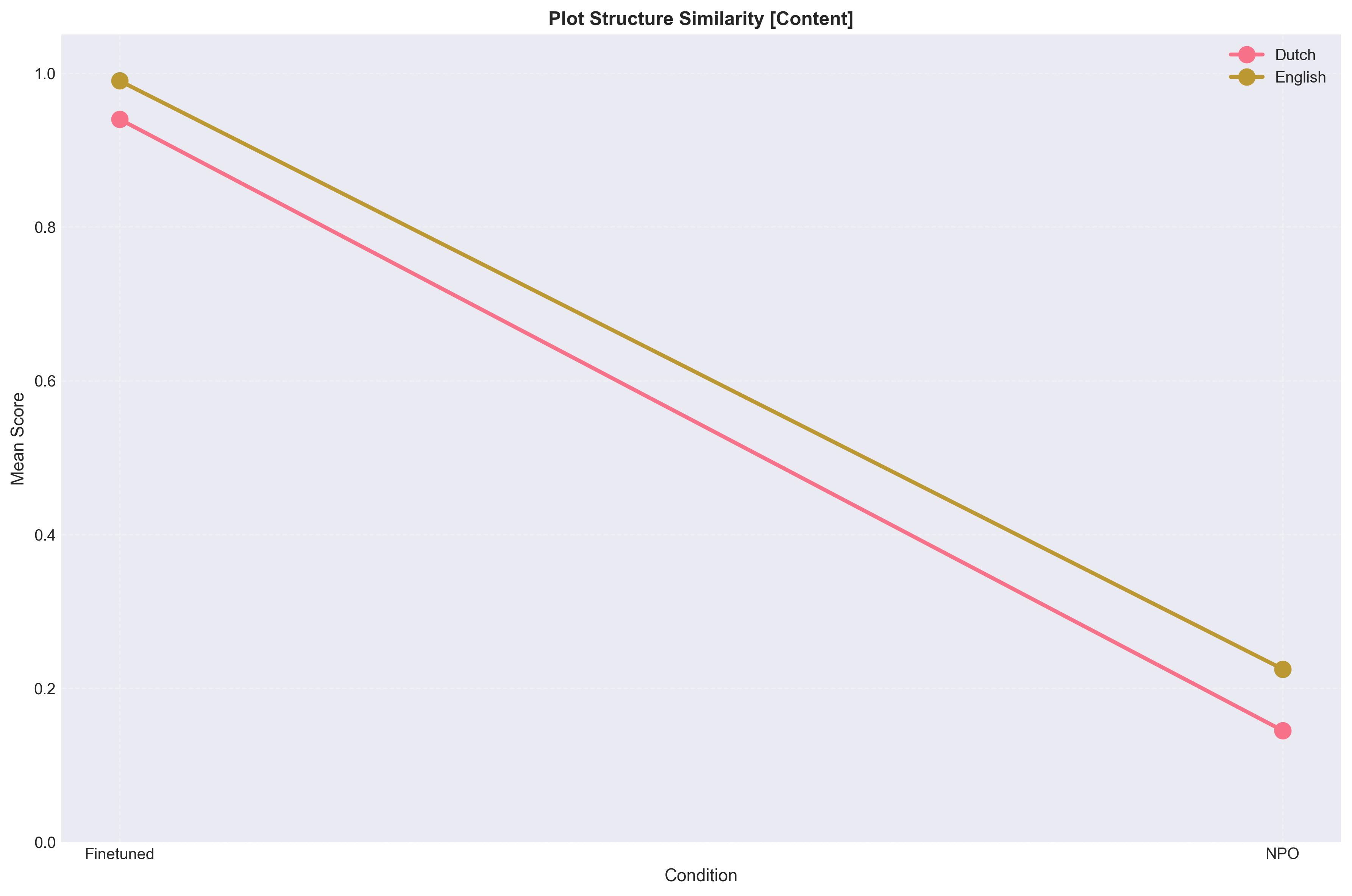}
    \caption{Plot Structure Similarity Interaction on Forget Set}
  \end{subfigure}
  \begin{subfigure}[t]{0.48\textwidth}
    \centering
    \includegraphics[width=\textwidth]{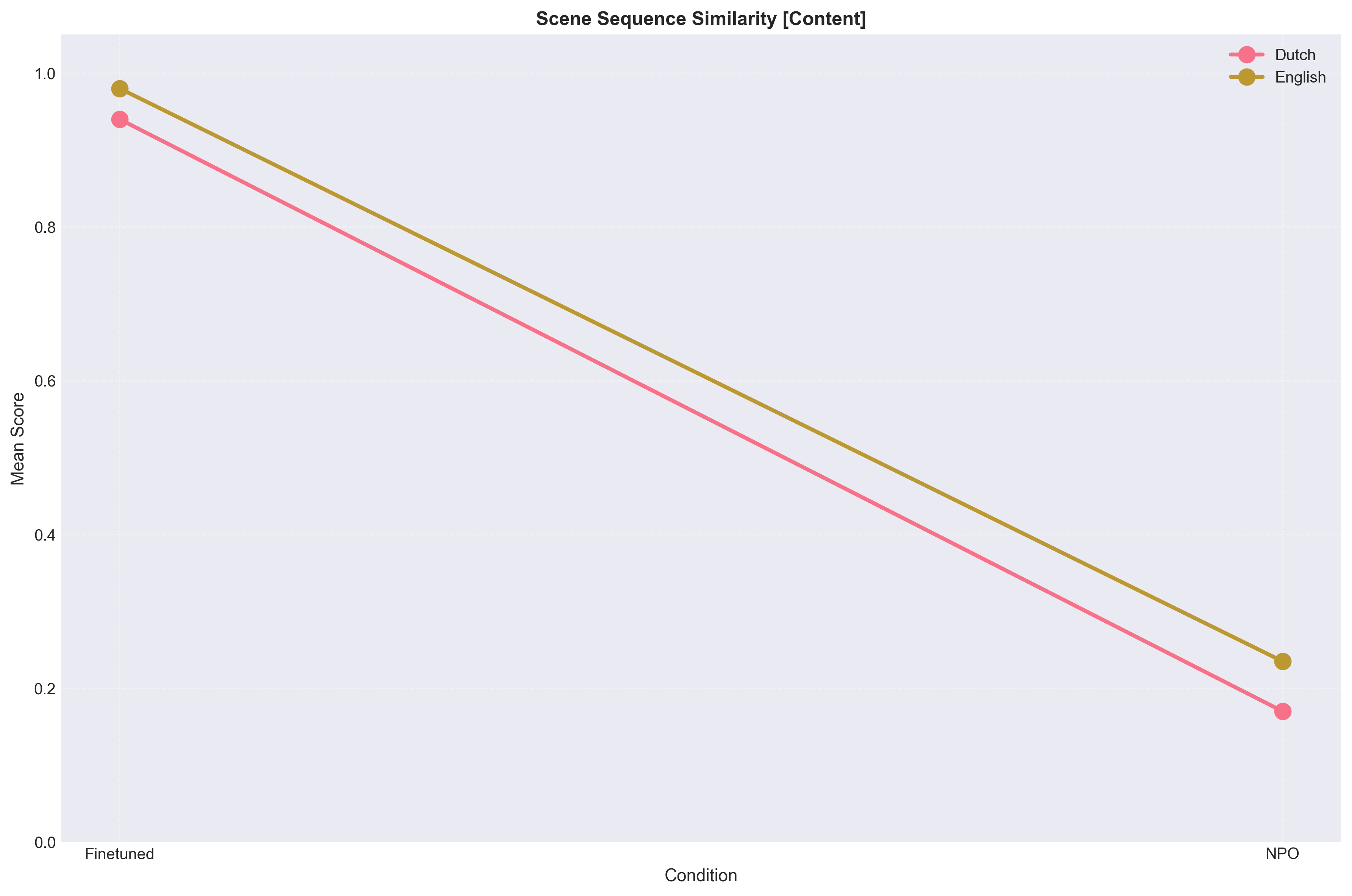}
    \caption{Scene Sequence Similarity Interaction on Forget Set}
  \end{subfigure}

  \caption[Stylistic and Content similarity before and after unlearning on the Forget set]{
  PSALM evaluator results for fine-tuned and NPO models where fine-tuning raises stylistic and content similarity from low to moderate levels to values close to one across both languages and splits indicating strong imprinting of the authors' style and narratorial choices}
  \label{fig:rq3:stylistic-metrics-forget-extended}
\end{figure}

\begin{figure}[!hb]
  \centering

  \begin{subfigure}[t]{0.48\textwidth}
    \centering
    \includegraphics[width=\textwidth]{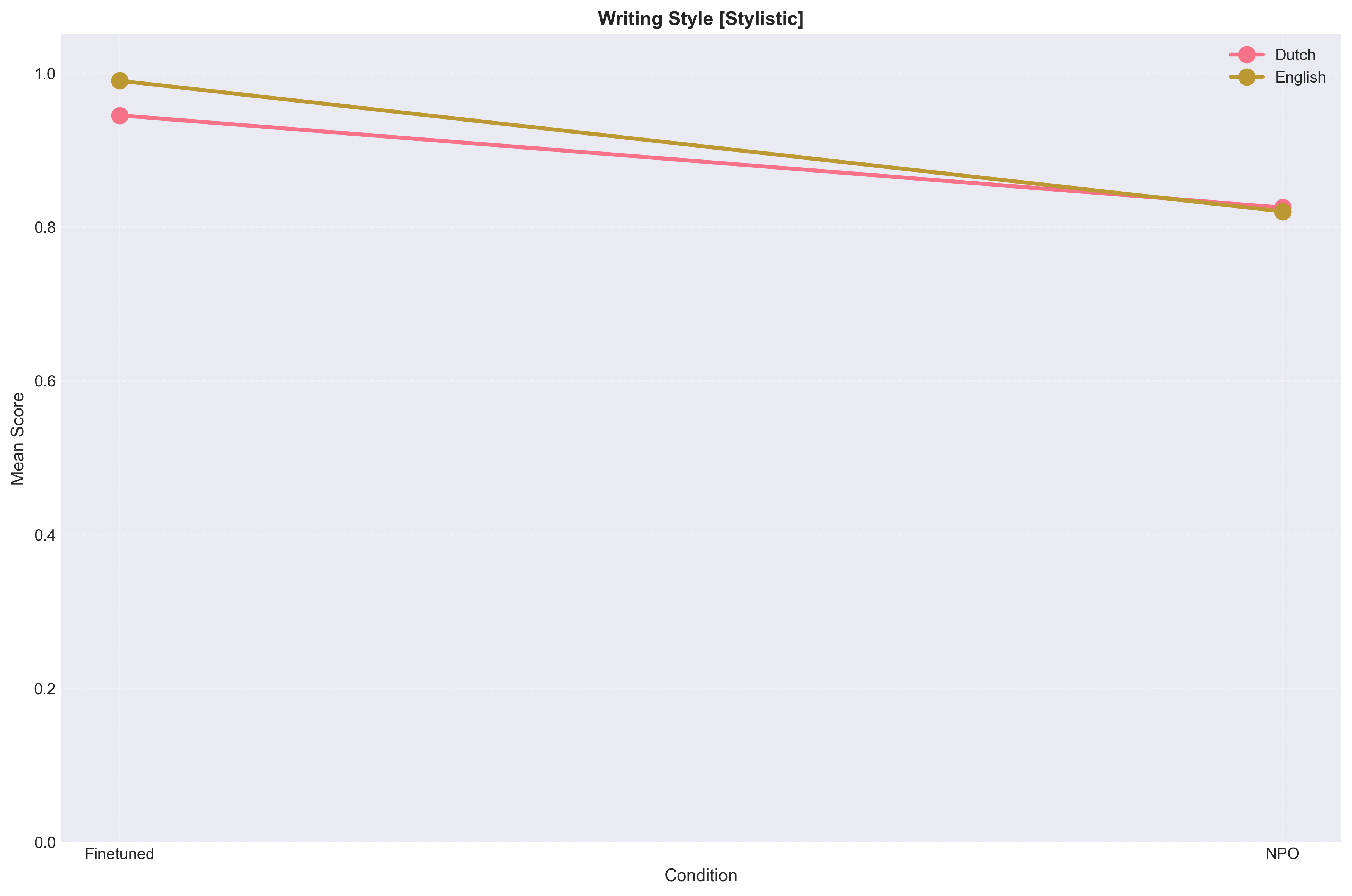}
    \caption{Writing Style Interaction on Retain Set}
  \end{subfigure}
  \begin{subfigure}[t]{0.48\textwidth}
    \centering
    \includegraphics[width=\textwidth]{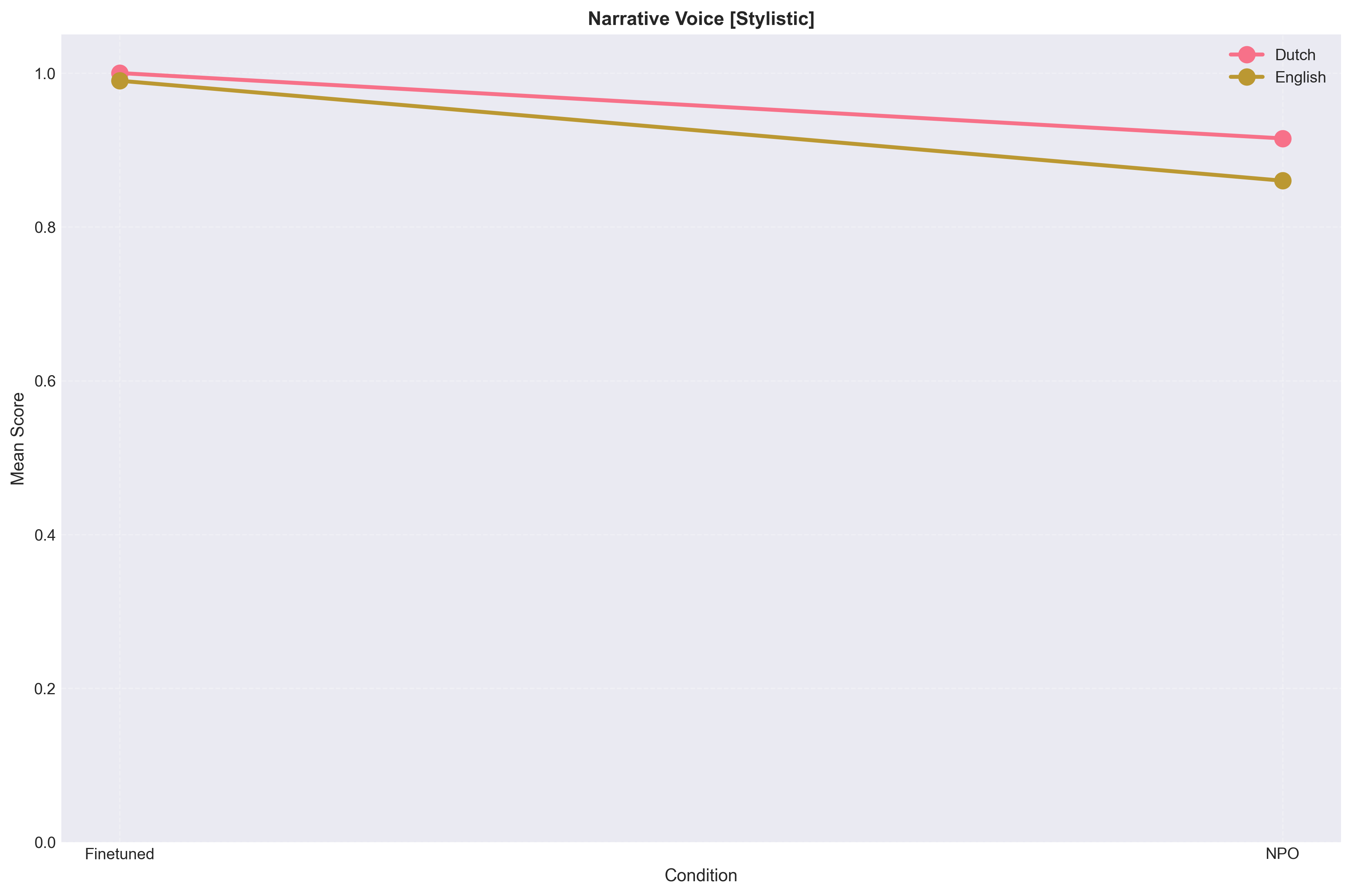}
    \caption{Narrative Voice Interaction on Retain Set}
  \end{subfigure}
  \begin{subfigure}[t]{0.48\textwidth}
    \centering
    \includegraphics[width=\textwidth]{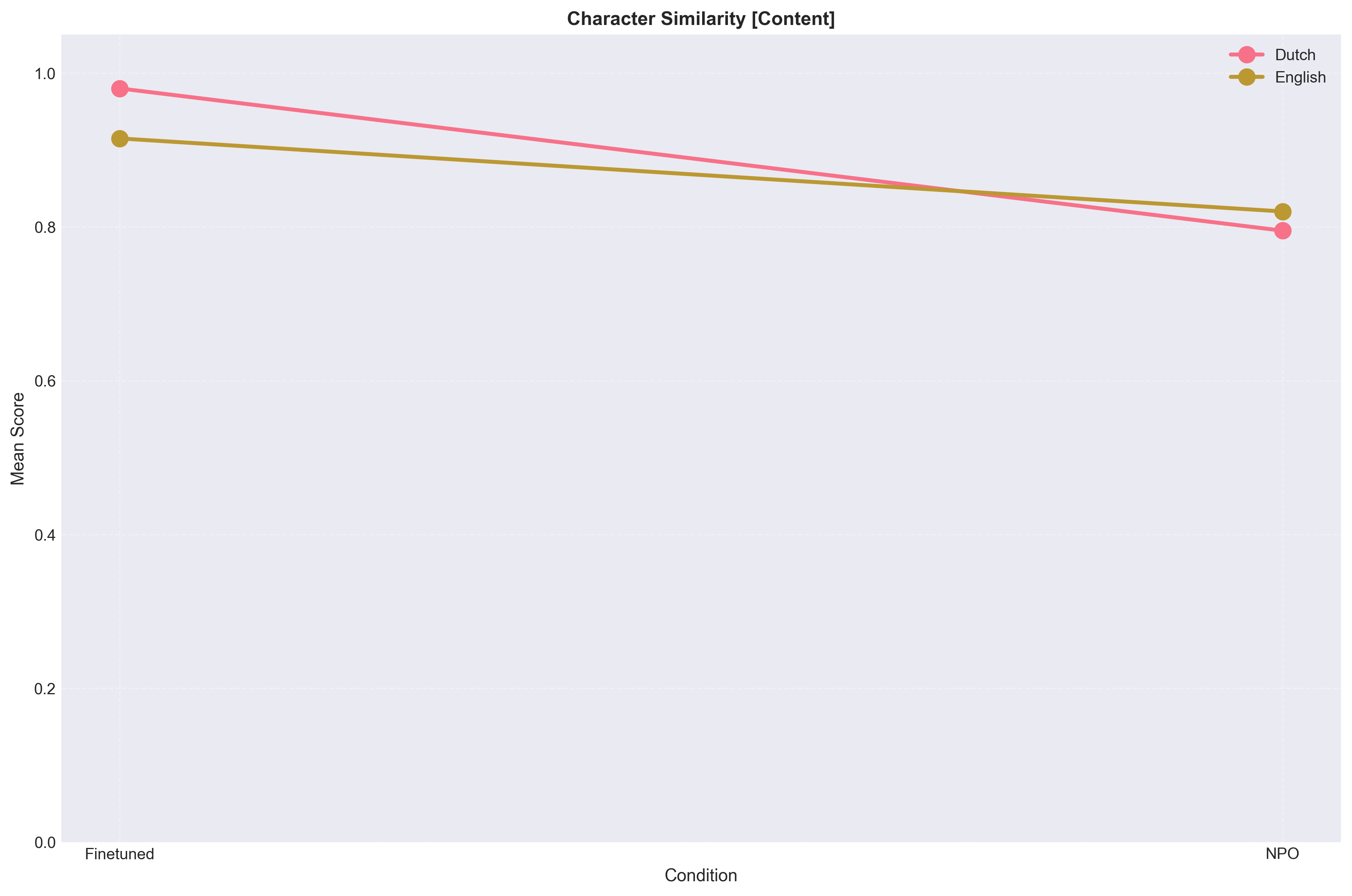}
    \caption{Character Similarity Interaction on Retain Set}
  \end{subfigure}
  \begin{subfigure}[t]{0.48\textwidth}
    \centering
    \includegraphics[width=\textwidth]{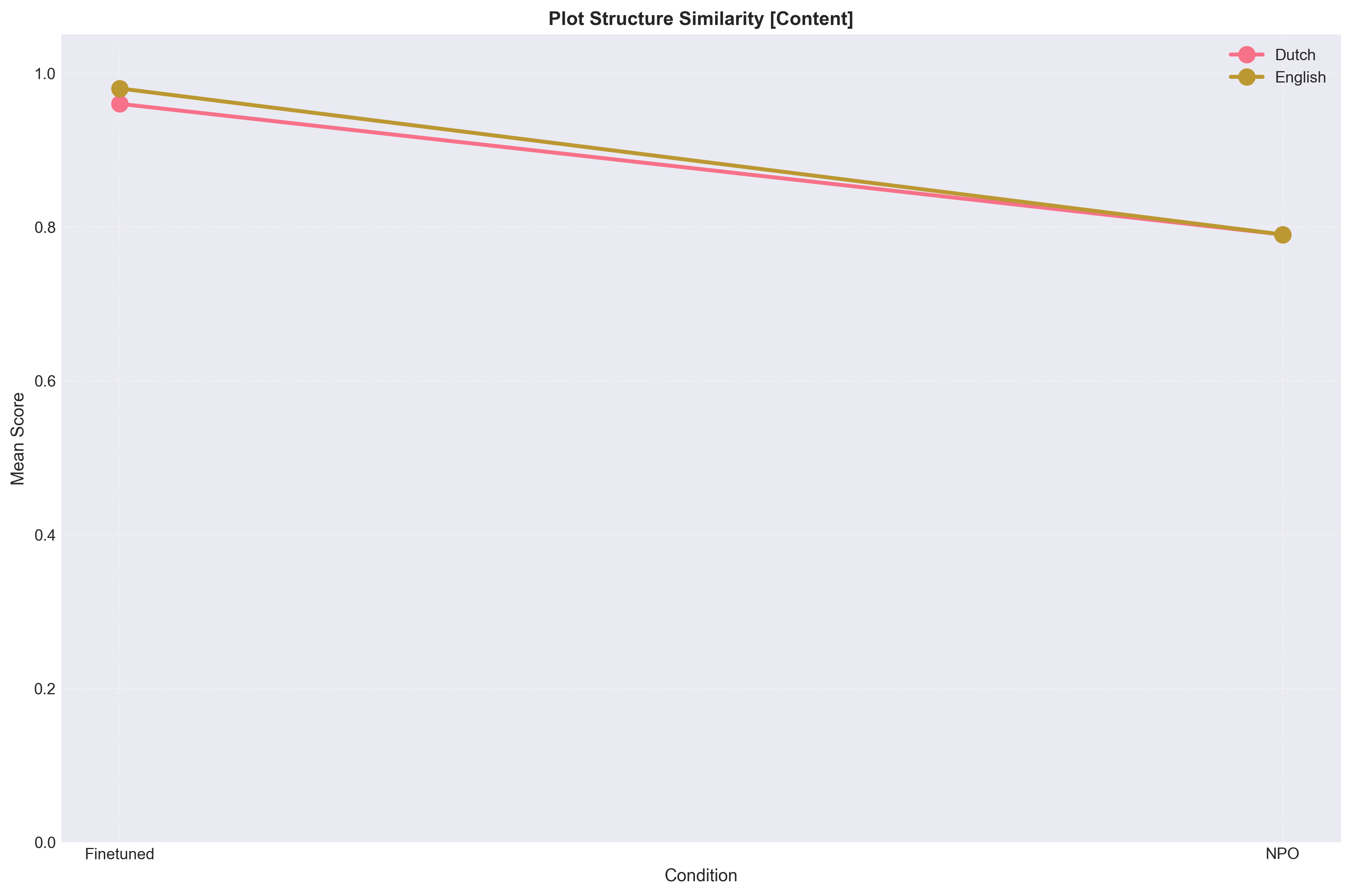}
    \caption{Plot Structure Similarity Interaction on Retain Set}
  \end{subfigure}
  \begin{subfigure}[t]{0.48\textwidth}
    \centering
    \includegraphics[width=\textwidth]{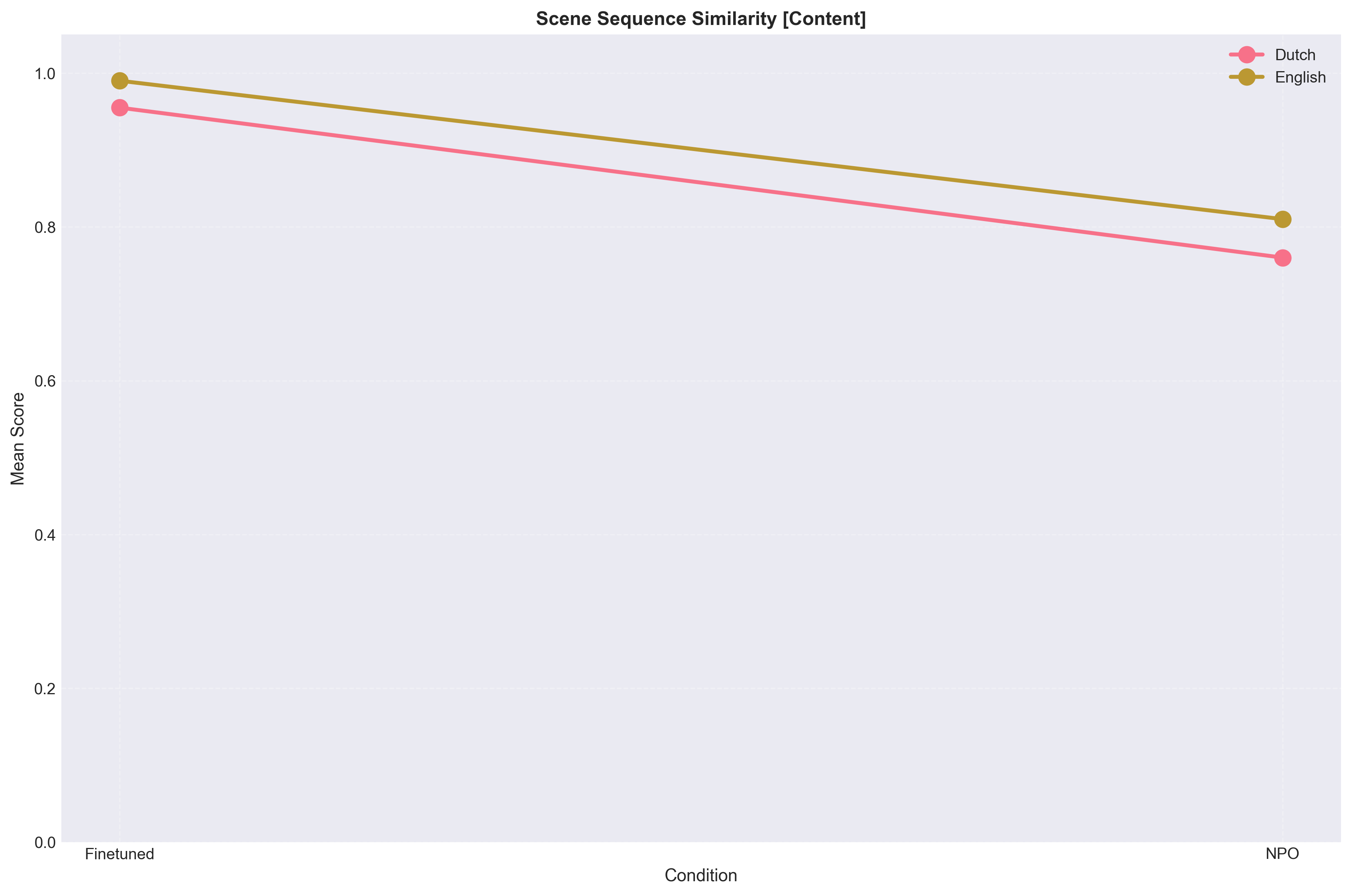}
    \caption{Scene Sequence Similarity Interaction on Retain Set}
  \end{subfigure}
  \begin{subfigure}[t]{0.48\textwidth}
    \centering
    \includegraphics[width=\textwidth]{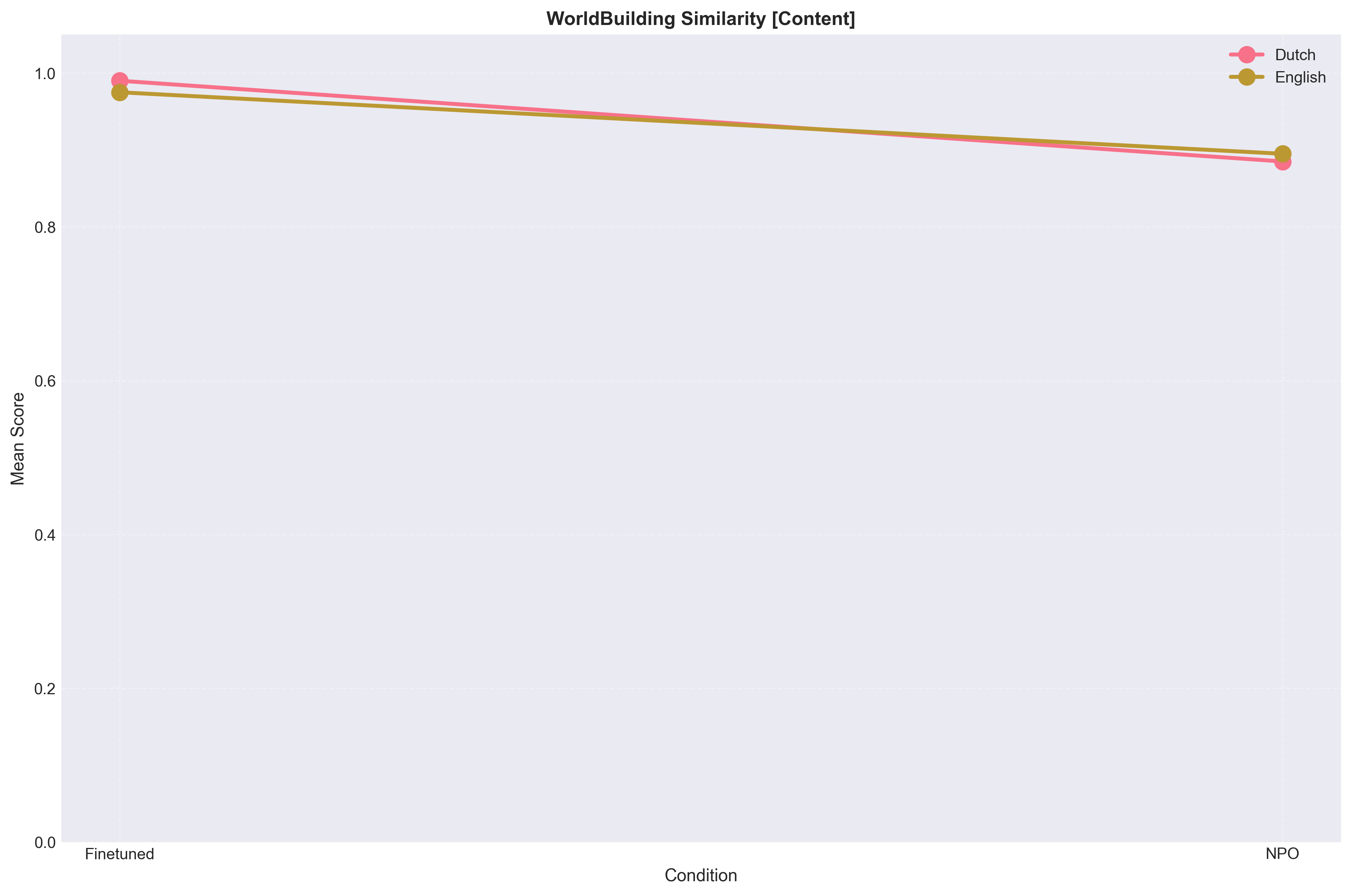}
    \caption{World Building Similarity Interaction on Retain Set}
  \end{subfigure}

  \caption[Stylistic and Content similarity before and after unlearning on the Retain set]{
  PSALM evaluator results for fine-tuned and NPO models}
  \label{fig:rq3:stylistic-metrics-retain}
\end{figure}

The results of stylistic and content similarity before and after unlearning on the forget and retain sets can be seen in Figure~\ref{fig:rq3:stylistic-metrics-forget-extended} and Figure~\ref{fig:rq3:stylistic-metrics-retain}.


\subsection{Behaviour of Exceptions Extended}
\begin{figure}[!hb]
  \centering

  \begin{subfigure}[t]{0.48\textwidth}
    \centering
    \includegraphics[width=\textwidth]{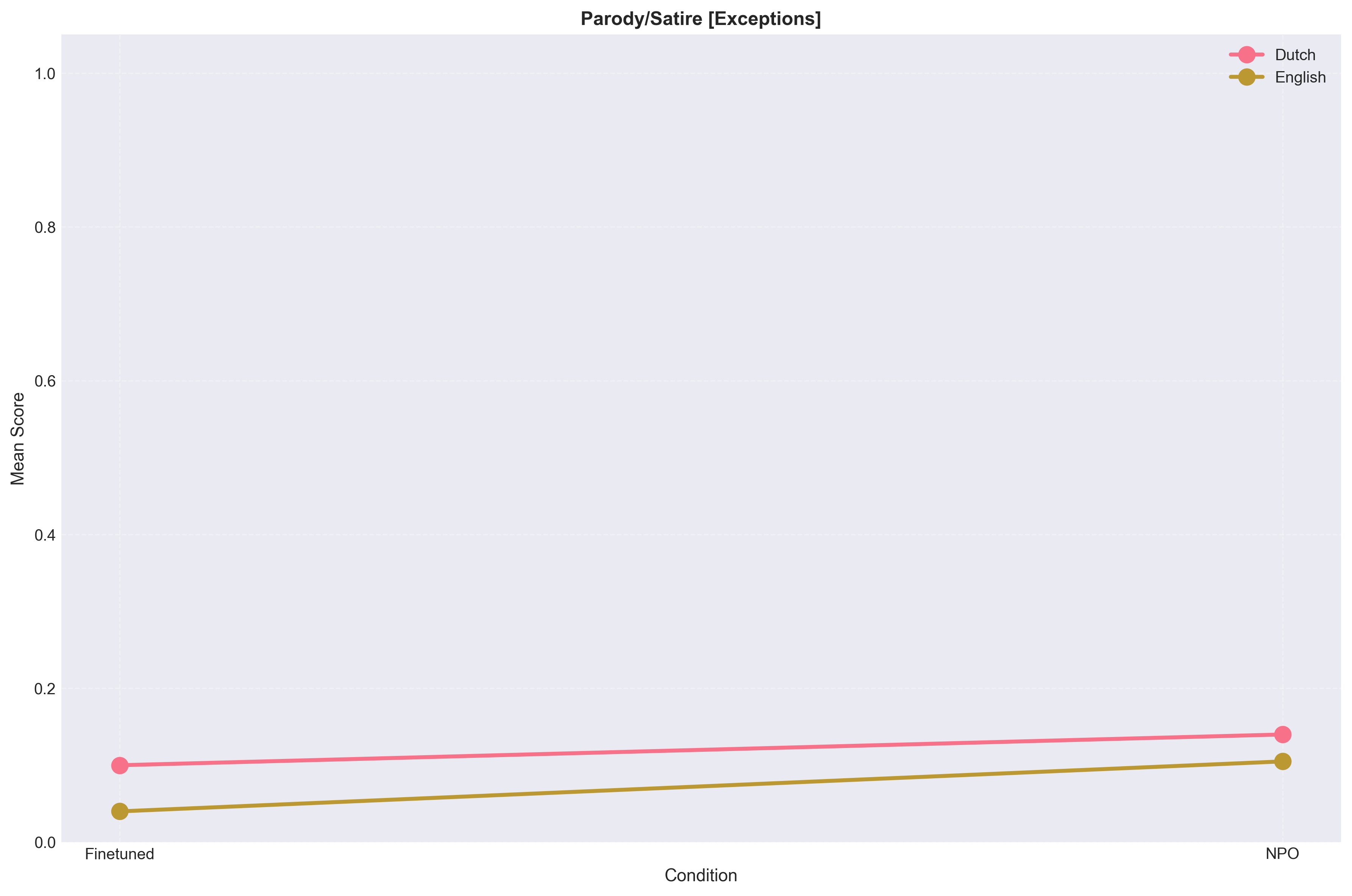}
    \caption{Parody/Satire Interaction on forget set}
  \end{subfigure}
  \begin{subfigure}[t]{0.48\textwidth}
    \centering
    \includegraphics[width=\textwidth]{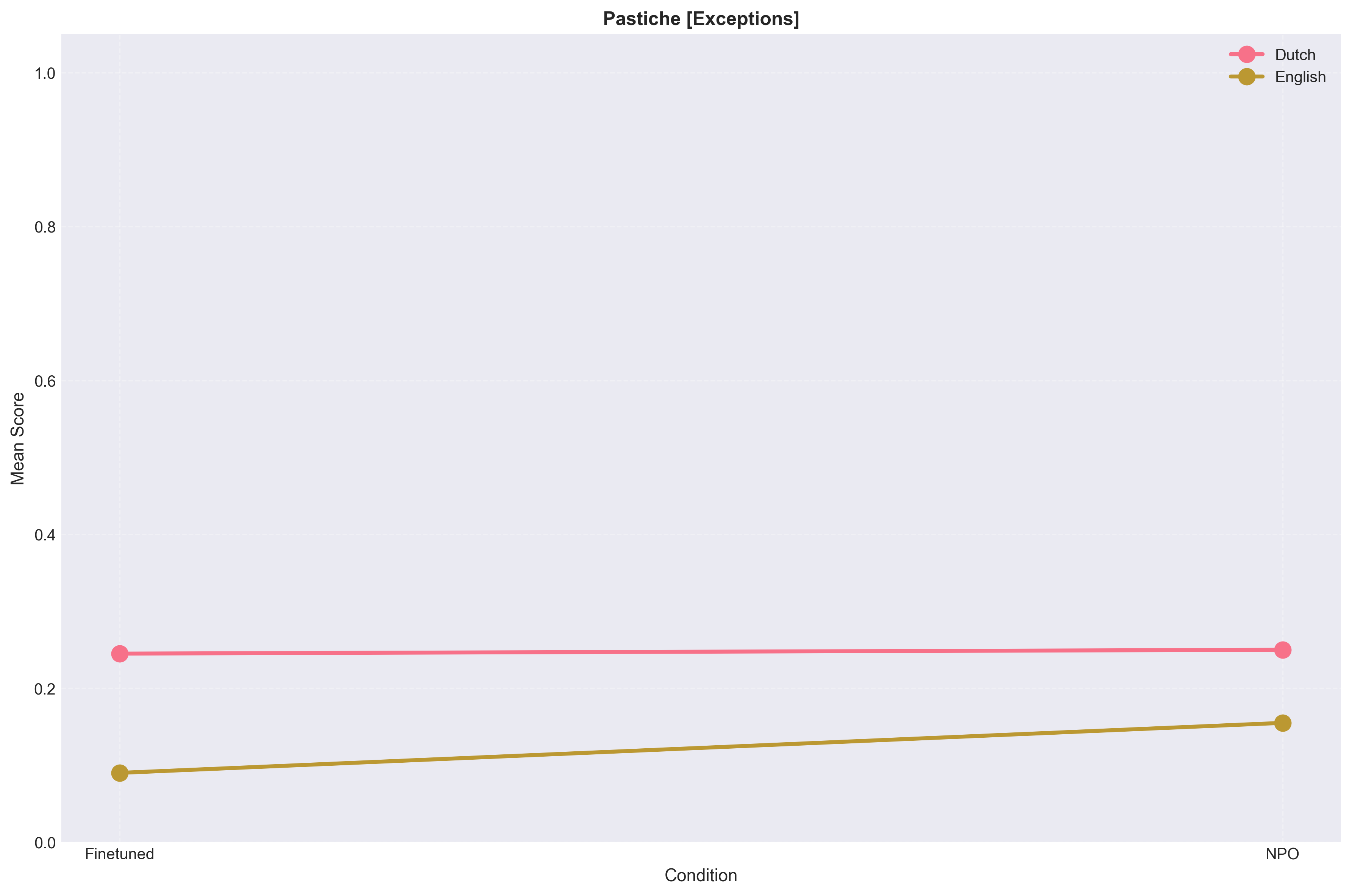}
    \caption{Pastiche Interaction on forget set}
  \end{subfigure}
  \begin{subfigure}[t]{0.48\textwidth}
    \centering
    \includegraphics[width=\textwidth]{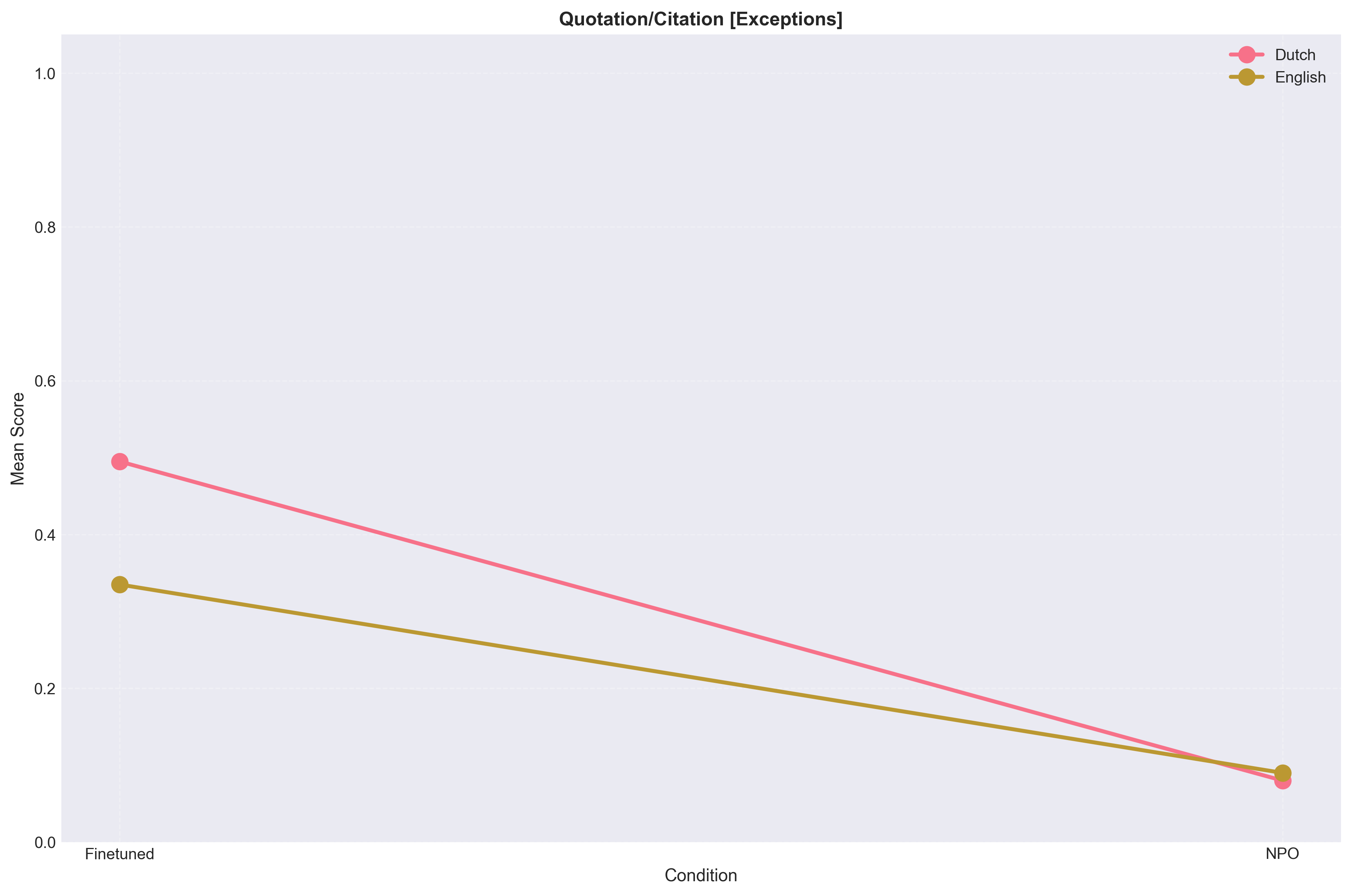}
    \caption{Quotation/Citation Interaction on forget set}
  \end{subfigure}
  \begin{subfigure}[t]{0.48\textwidth}
    \centering
    \includegraphics[width=\textwidth]{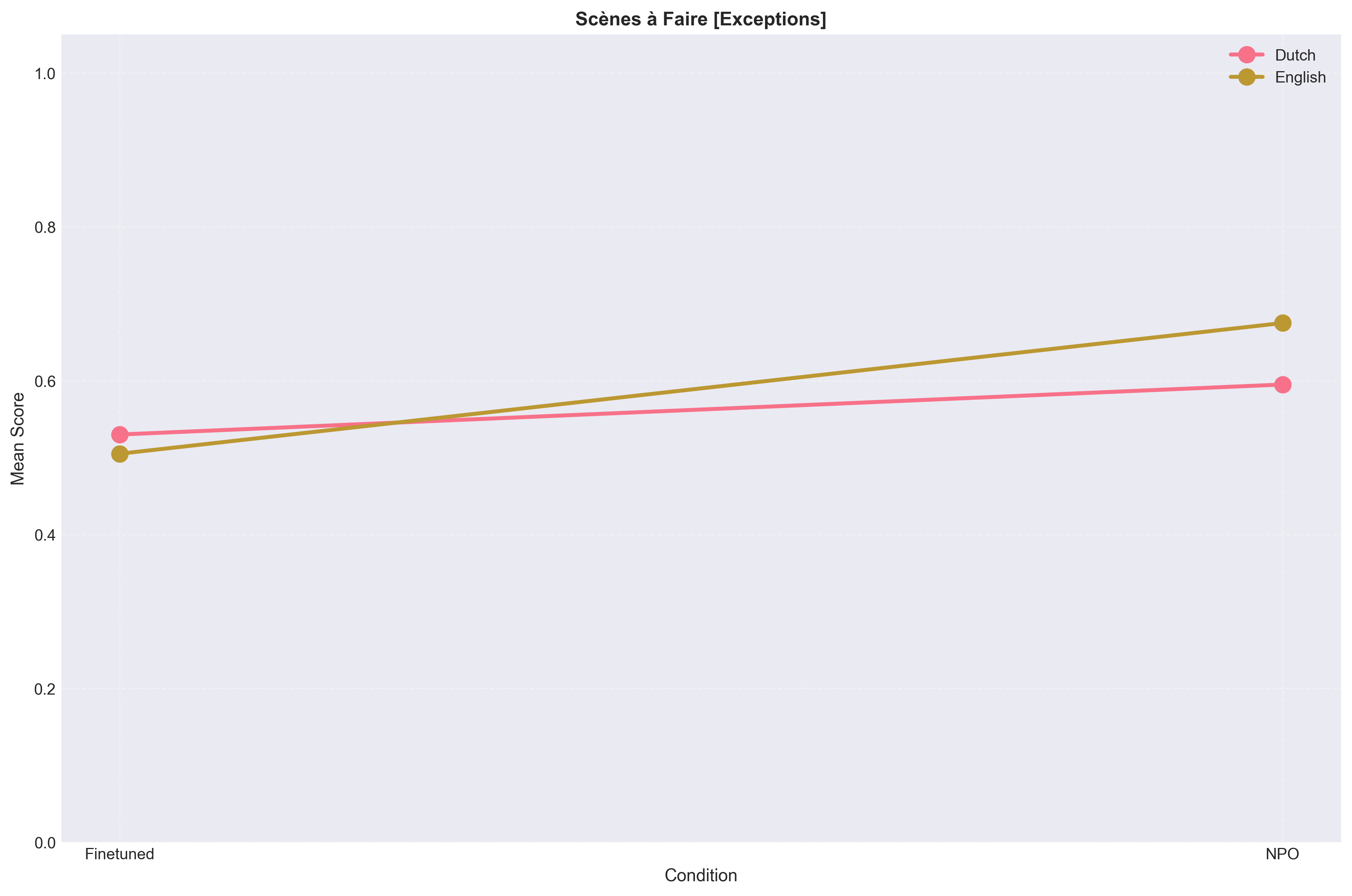}
    \caption{Scènes à Faire Interaction on forget set}
  \end{subfigure}

  \caption[Exception metrics before and after unlearning on forget set]{
  Exception metrics (Parody/Satire Pastiche Quotation/Citation and Scènes à Faire) for fine-tuned and NPO models on the forget set}
  \label{fig:rq3:exception-metrics-forget}
\end{figure}

\begin{figure}[!hb]
  \centering

  \begin{subfigure}[t]{0.48\textwidth}
    \centering
    \includegraphics[width=\textwidth]{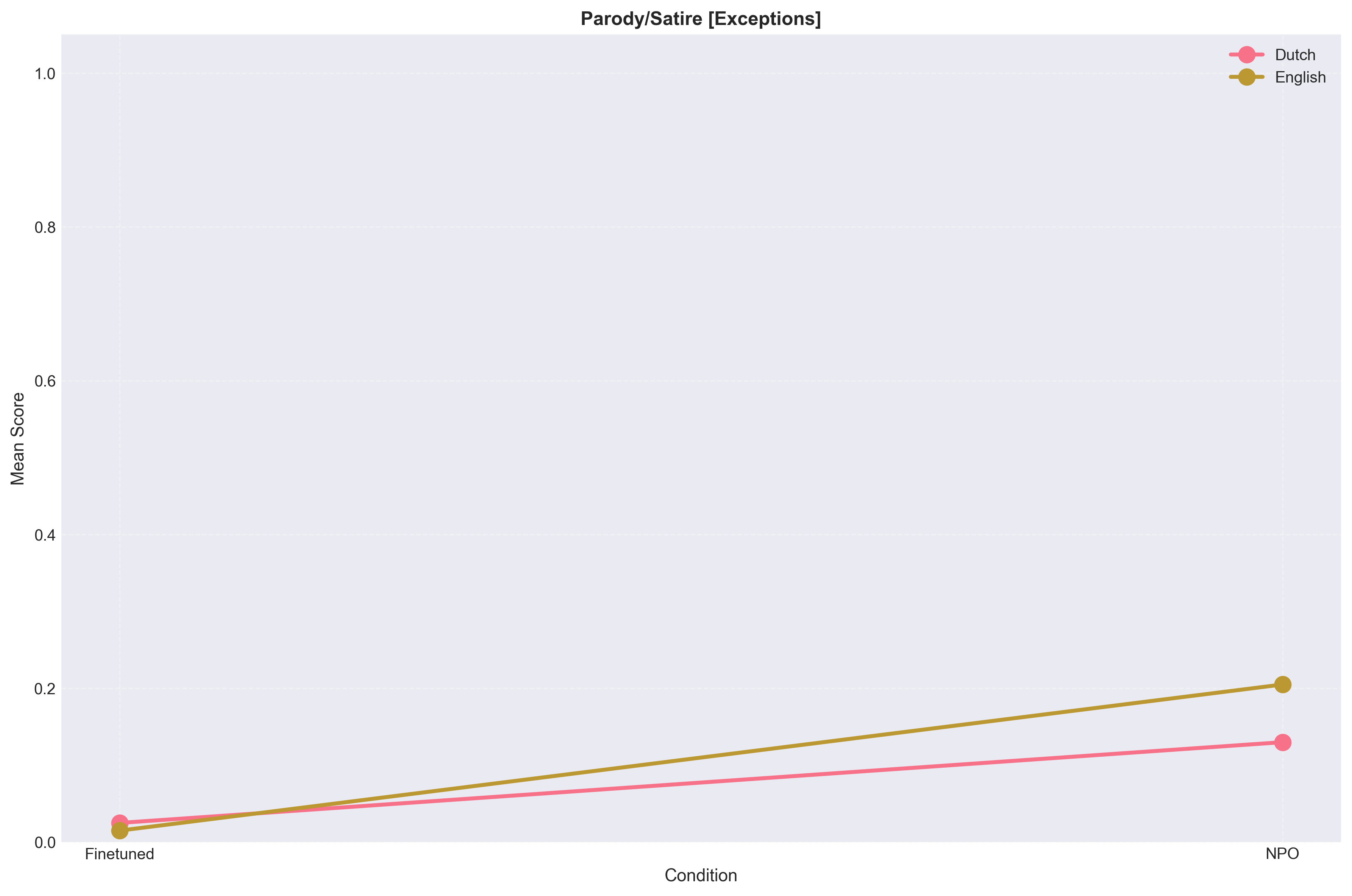}
    \caption{Parody/Satire Interaction on retain set}
  \end{subfigure}
  \begin{subfigure}[t]{0.48\textwidth}
    \centering
    \includegraphics[width=\textwidth]{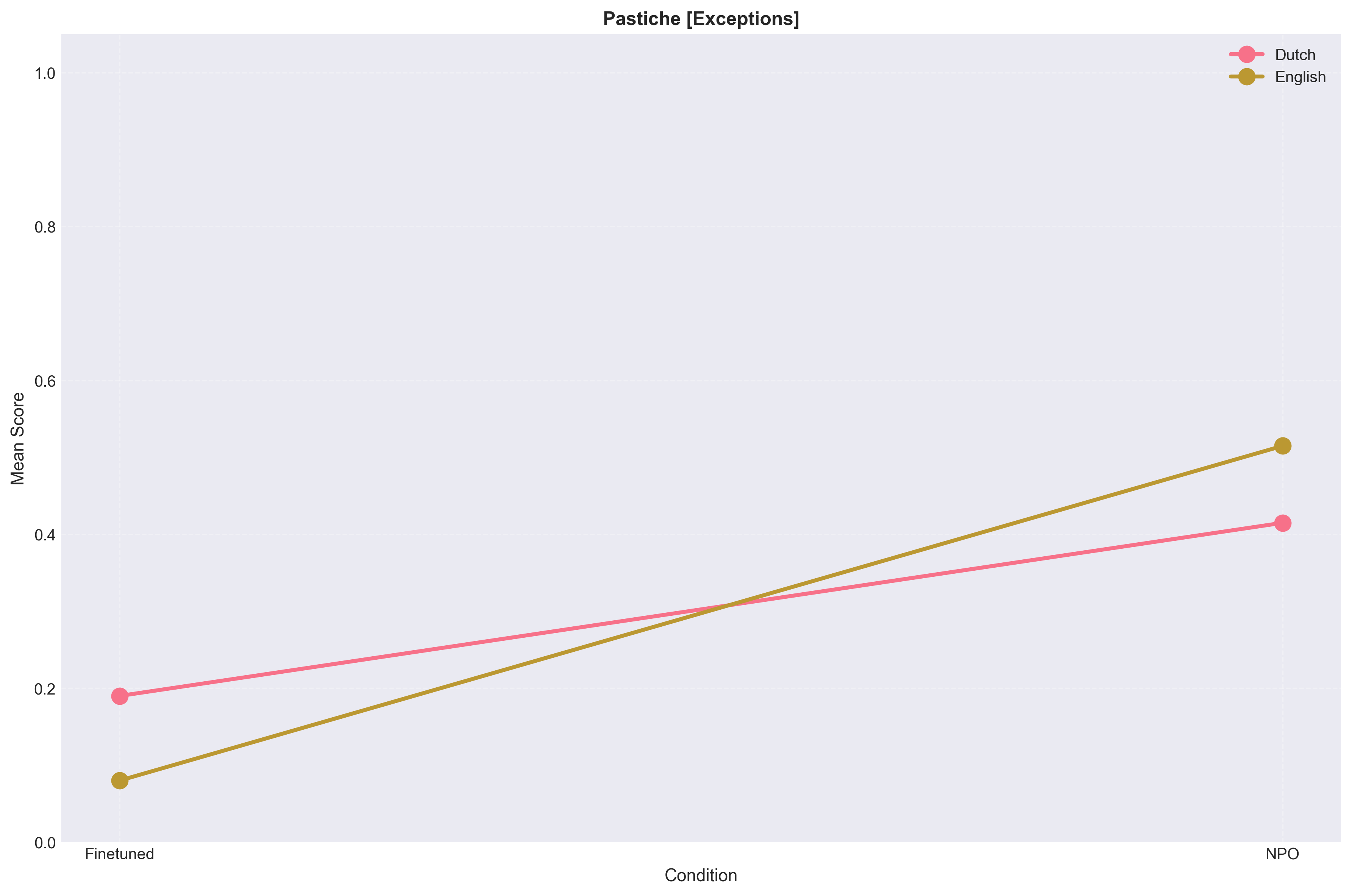}
    \caption{Pastiche Interaction on retain set}
  \end{subfigure}
  \begin{subfigure}[t]{0.48\textwidth}
    \centering
    \includegraphics[width=\textwidth]{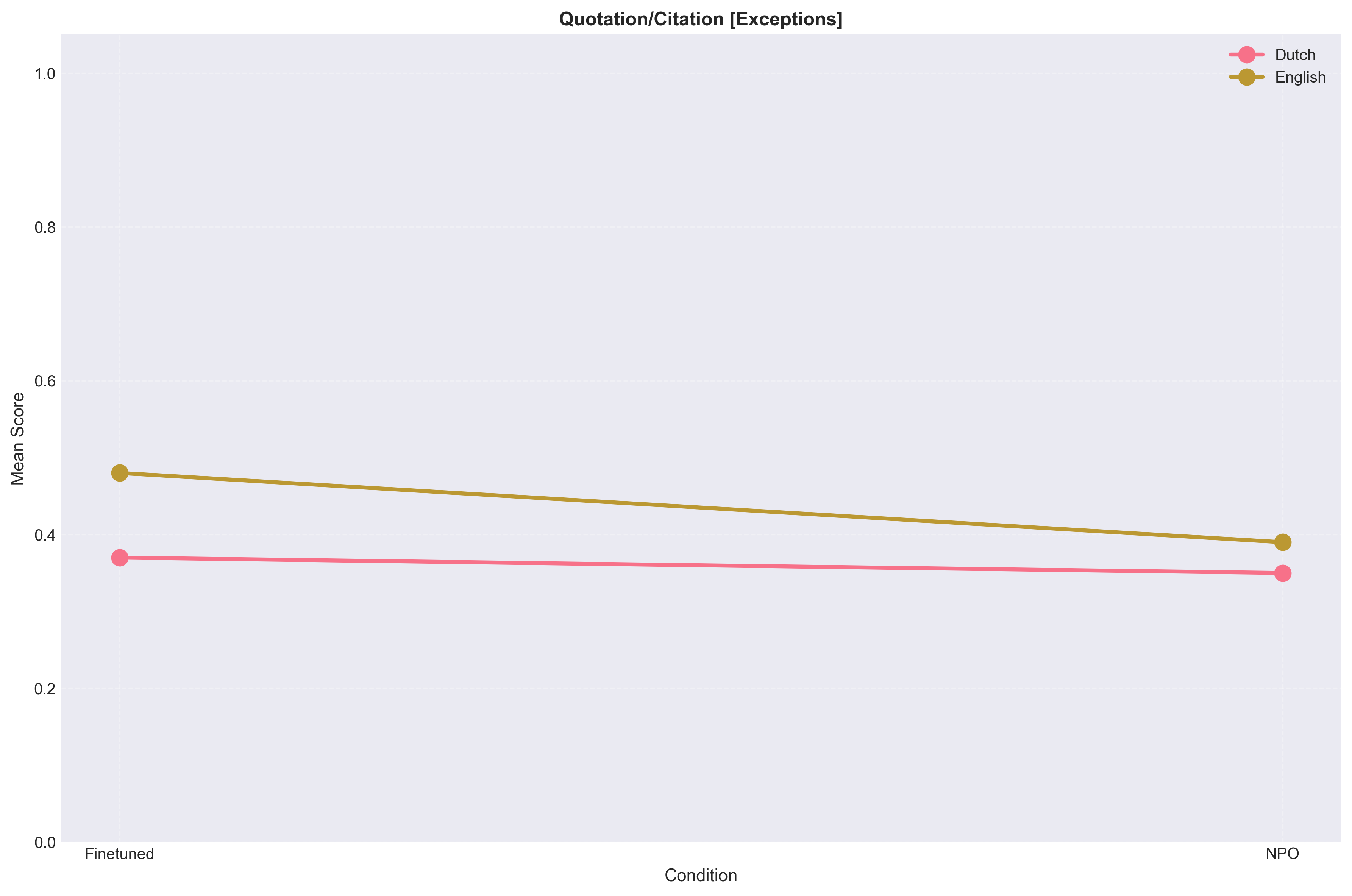}
    \caption{Quotation/Citation Interaction on retain set}
  \end{subfigure}
  \begin{subfigure}[t]{0.48\textwidth}
    \centering
    \includegraphics[width=\textwidth]{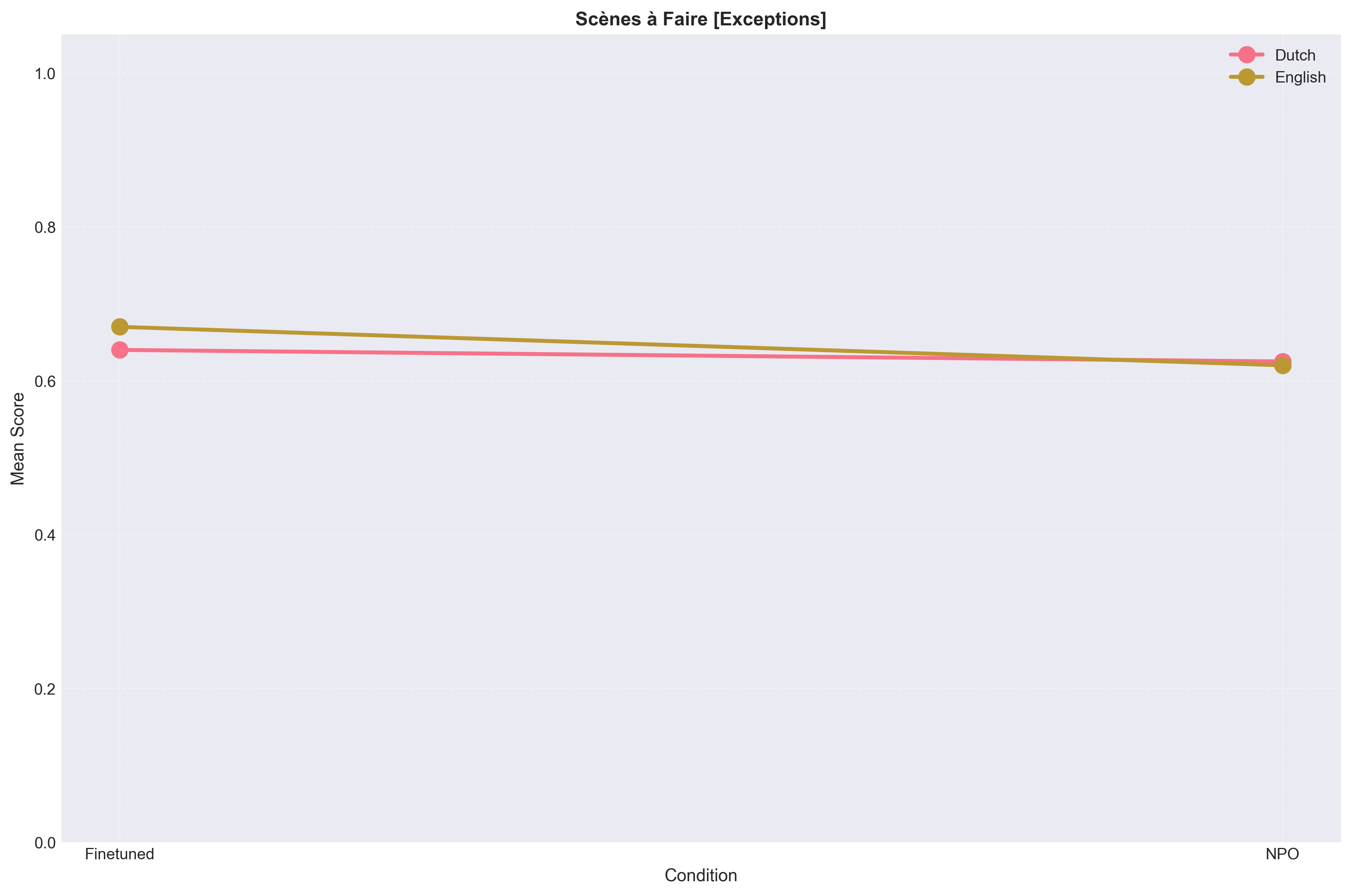}
    \caption{Scènes à Faire Interaction on retain set}
  \end{subfigure}

  \caption[Exception metrics before and after unlearning on retain set]{
  Exception metrics (Parody/Satire Pastiche Quotation/Citation and Scènes à Faire) for fine-tuned and NPO models on the retain set}
  \label{fig:rq3:exception-metrics-retain}
\end{figure}

The results of the exception metrics before and after unlearning on the forget and retain sets can be seen in Figure~\ref{fig:rq3:exception-metrics-forget} and Figure~\ref{fig:rq3:exception-metrics-retain}.

\subsection{Suppression of Literal Memorisation}
\begin{figure}[!hb]
  \centering

  \begin{subfigure}[t]{0.48\textwidth}
    \centering
    \includegraphics[width=\textwidth]{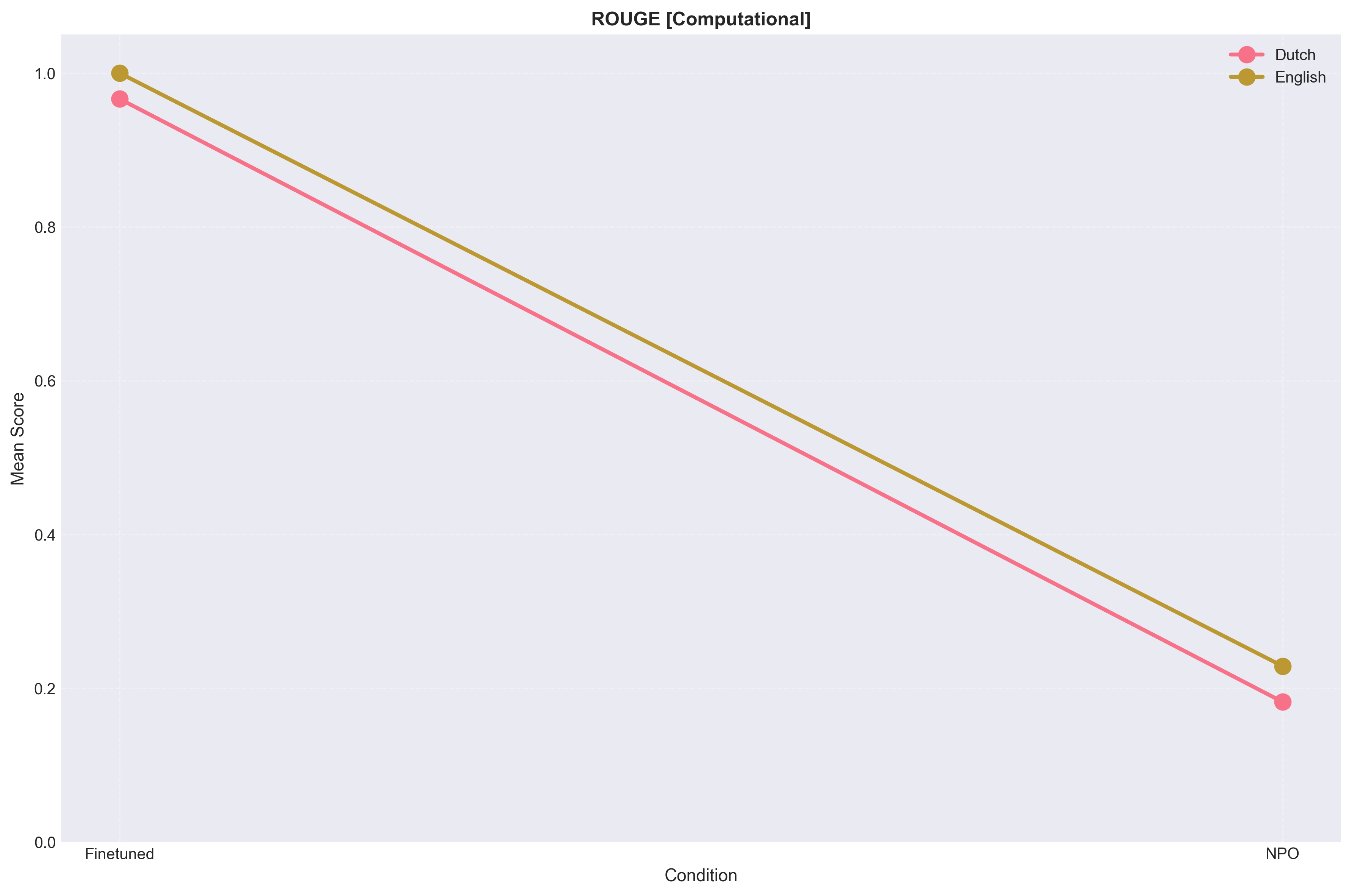}
    \caption{ROUGE-L Interaction of Forget Set}
  \end{subfigure}
  \begin{subfigure}[t]{0.48\textwidth}
    \centering
    \includegraphics[width=\textwidth]{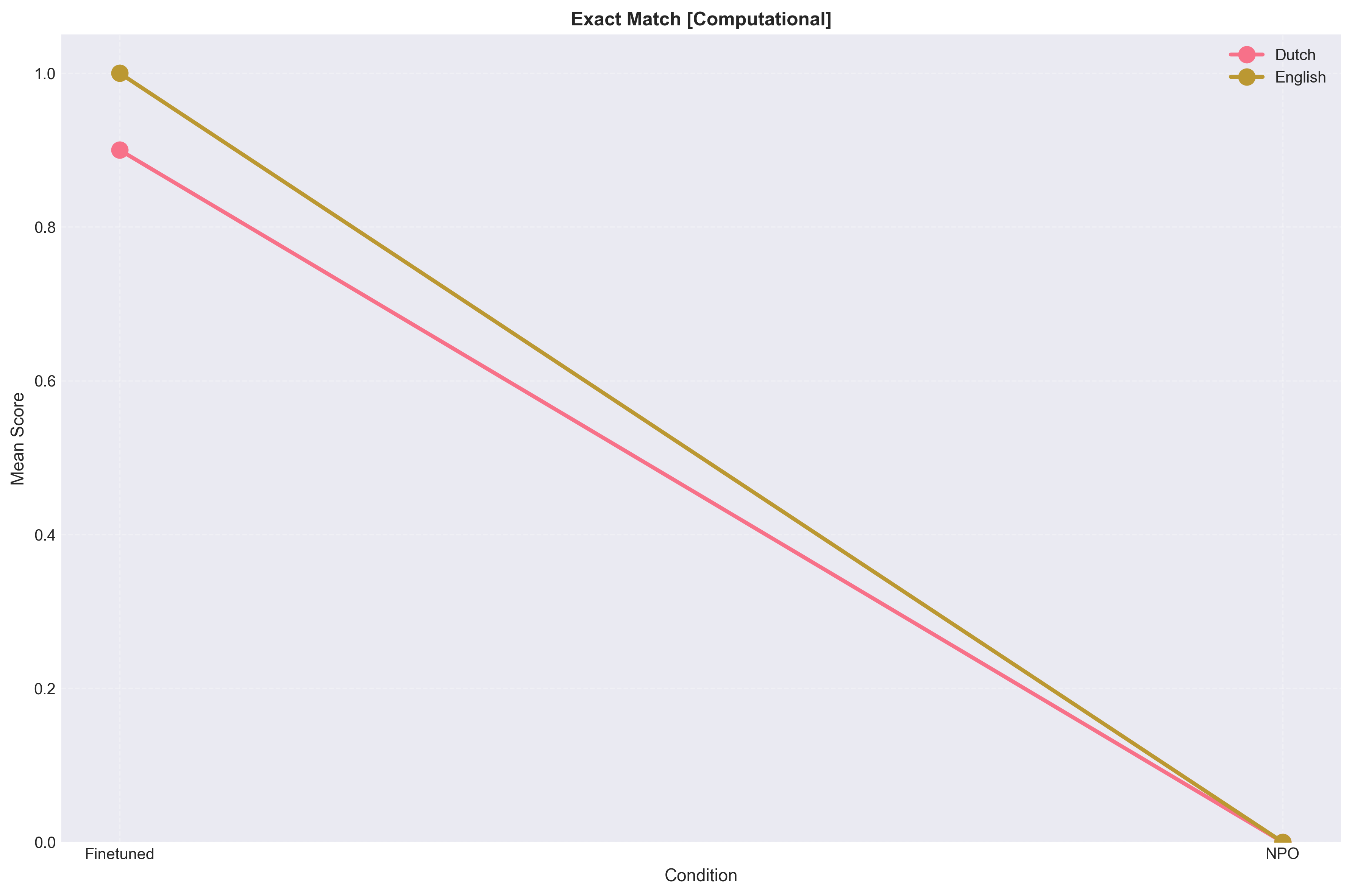}
    \caption{Exact Match Interaction of Forget Set}
  \end{subfigure}
  
  \begin{subfigure}[t]{0.48\textwidth}
    \centering
    \includegraphics[width=\textwidth]{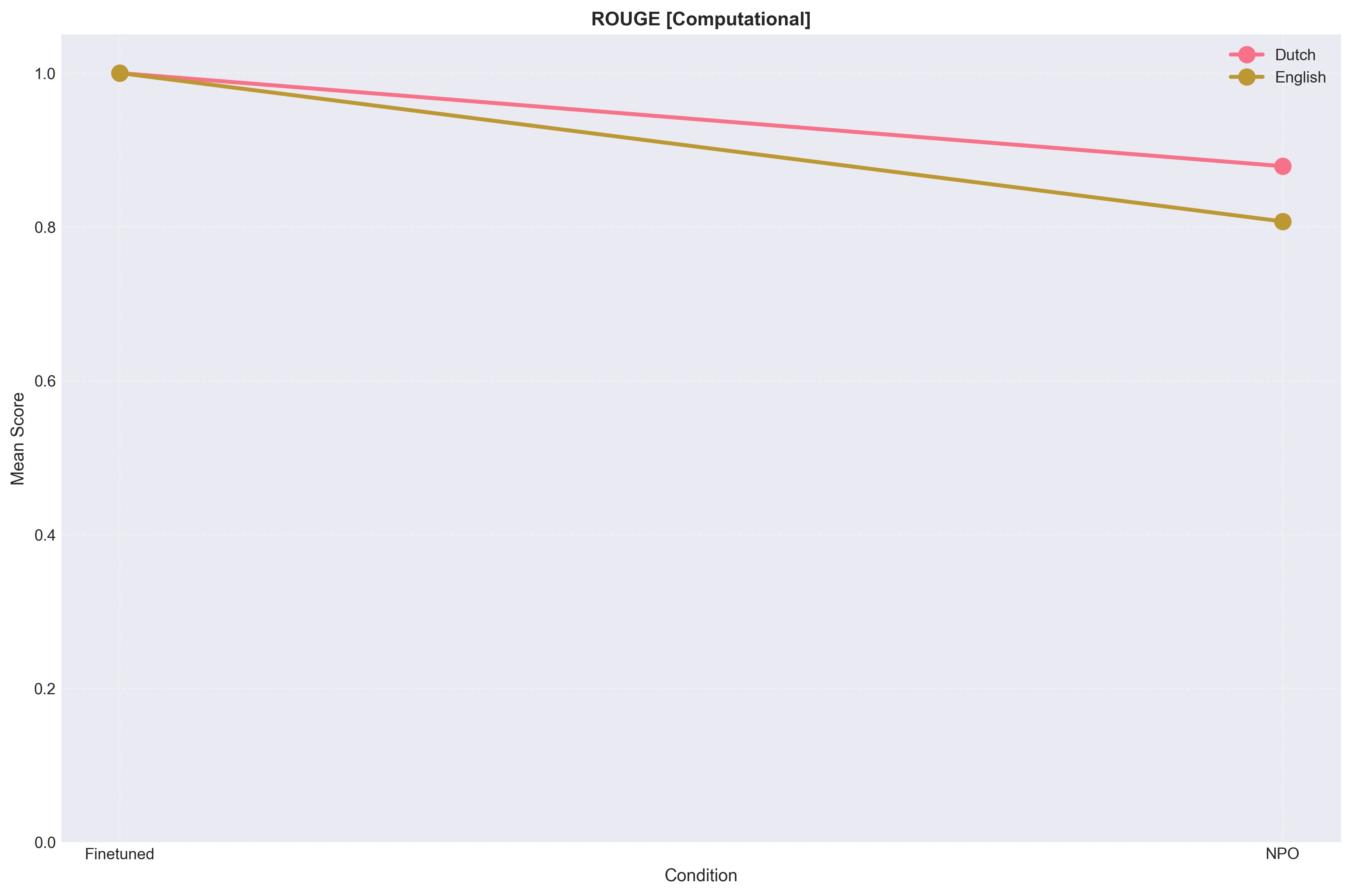}
    \caption{ROUGE-L Interaction of Retain Set}
  \end{subfigure}
  \begin{subfigure}[t]{0.48\textwidth}
    \centering
    \includegraphics[width=\textwidth]{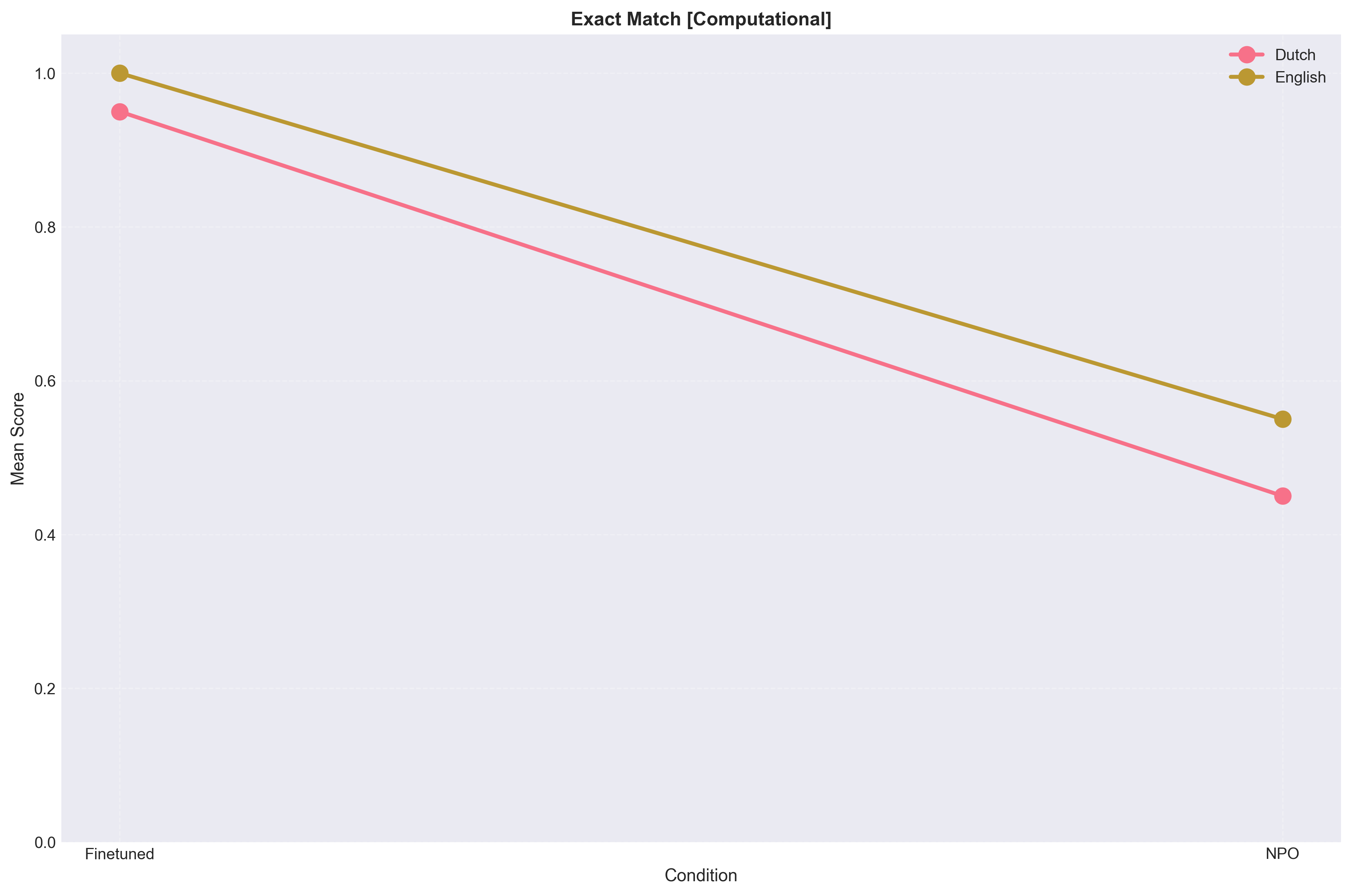}
    \caption{Exact Match Interaction of Retain Set}
  \end{subfigure}

  \caption[Computational metrics before and after unlearning]{
  Computational similarity metrics (Exact Match BLEU and ROUGE) for fine-tuned and NPO models where NPO almost eliminates verbatim reproduction on forget sets while leaving substantial overlap on retain sets}
  \label{fig:rq3:computational-metrics-extended}
\end{figure}

The results of the computational metrics before and after unlearning on the forget and retain sets can be seen in Figure~\ref{fig:rq3:computational-metrics-extended}.

\subsection{Comparing Metric Categories Extended}
\begin{figure}[!hb]
    \centering
    \adjustbox{max width=\textwidth}{
    \includegraphics[width=\textwidth]{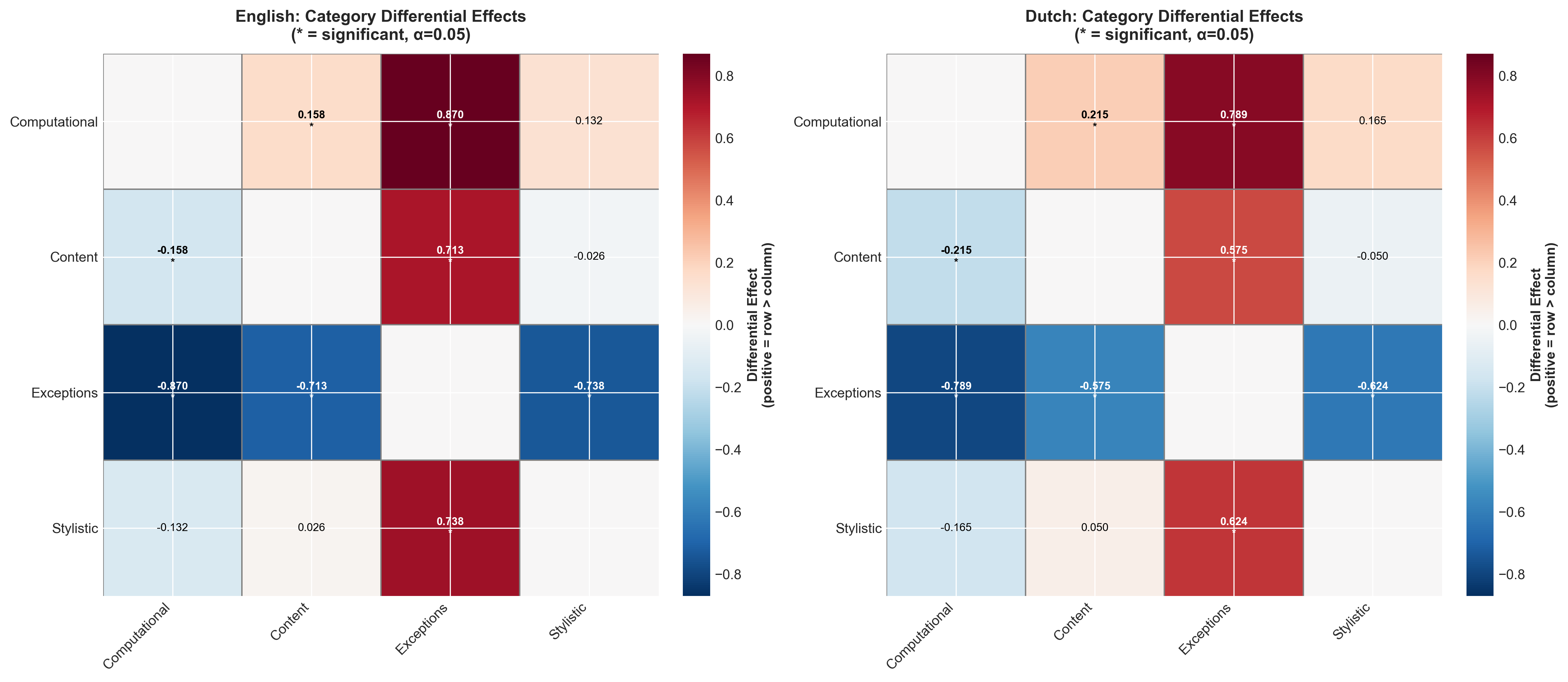}
    }
    
   \caption{Category-level pairwise contrasts for the forget split where cells show differences in reductions between categories and asterisks mark statistically significant contrasts}
    \label{fig:rq3:differential-effects-heatmap}
\end{figure}

The category-level pairwise heatmap can be seen in Figure~\ref{fig:rq3:differential-effects-heatmap}.

\subsection{Residual Similarity and Distance to Baseline}
\begin{figure}[!hb]
  \centering
  \includegraphics[width=\textwidth]{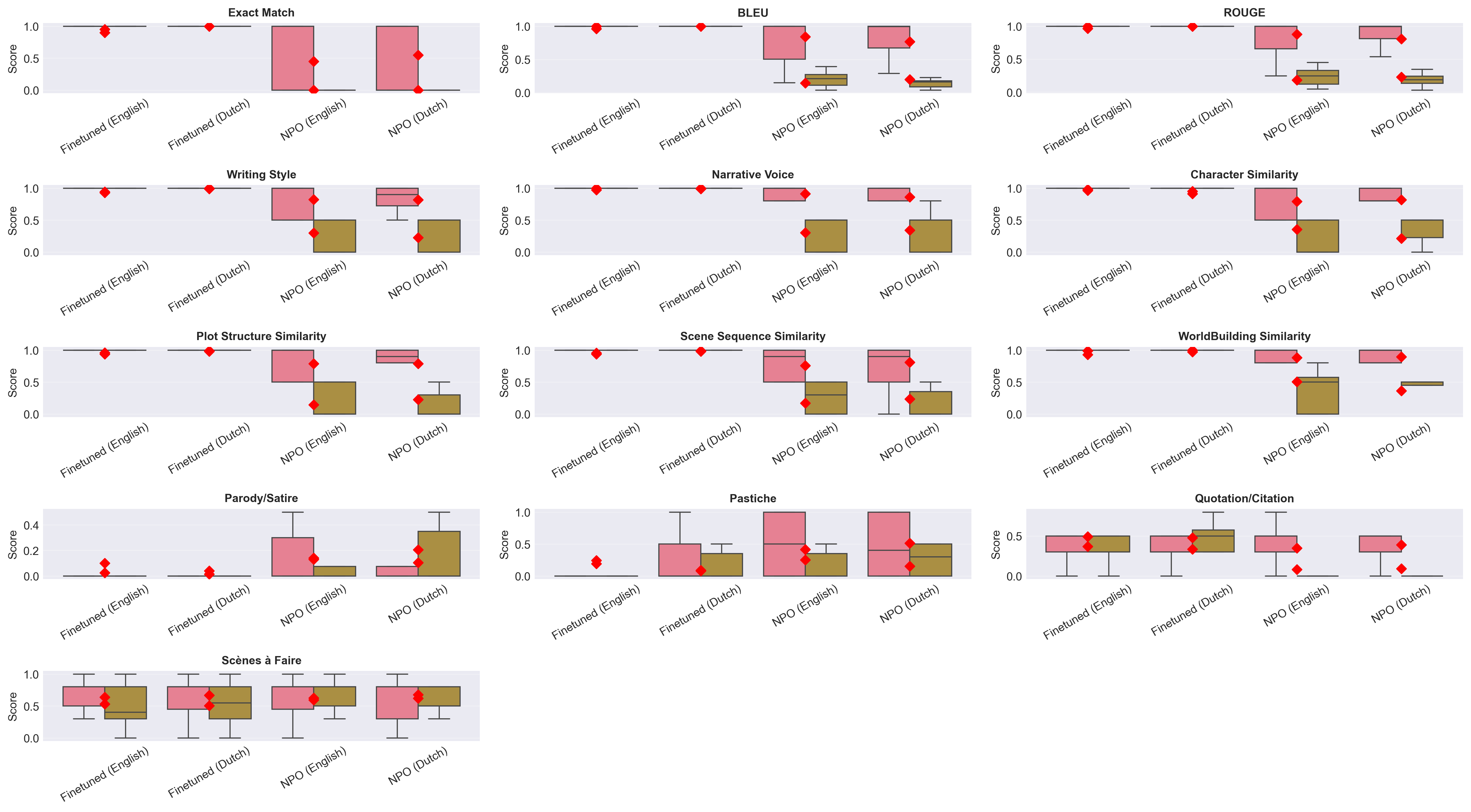}
  \caption[Score distributions after unlearning]{
  Score distributions for all PSALM metrics across fine-tuned and NPO models,
  separated by forget and retain splits. Fine-tuned scores concentrate near
  the maximum; NPO spreads the distributions and lowers their medians, but
  many retain-set scores remain high}
  \label{fig:rq3:boxplot}
\end{figure}

The score distribution after unlearning can be found in Figure~\ref{fig:rq3:boxplot}.

\subsection{Diagnostic tests and omnibus effects}
Normality and homogeneity checks (Shapiro–Wilk and Levene; Online Resources~16 and~17) indicate that none of the metrics are safely modelled with standard Gaussian assumptions. For both forget and retain splits, all metrics fail normality tests, and most also violate homoscedasticity. Consequently, the main analyses rely on rank-based methods throughout.

Kruskal–Wallis omnibus tests for each metric and split (Online Resource~18) show highly significant effects of model condition for all computational, stylistic and content metrics on both splits (all \(p \leq 5\times 10^{-3}\)). Among the exception metrics, Parody/Satire and Pastiche show non-significant or only weakly significant omnibus results on the forget split, and Quotation/Citation and Scènes à Faire show non-significant omnibus effects on the retain split. This confirms that unlearning substantially affects the infringement-oriented metrics, while its impact on exception-related dimensions is smaller and more heterogeneous.

Rank-based ART-ANOVA models (Online Resources~19 and~20) provide consistent evidence that unlearning has a strong main effect on all metrics. For the forget split, the unlearning factor attains \(F\)-statistics well above \(60\) for most infringement-oriented metrics, with \(p\)-values below \(10^{-11}\). On the retain split, unlearning remains significant for every metric except Quotation/Citation and Scènes à Faire. Language main effects and language\(\times\)unlearning interactions are generally absent; world-building similarity on the forget split is the only metric where an interaction reaches significance, matching the qualitative discussion in subsection~\ref{sec:rq3:stylistic-content}. No interaction is detected on the retain split.

\subsection{Overall and category-level patterns}
Average ranks across all thirteen metrics (Online Resource~21) confirm that the fine-tuned models remain closest to the source passages, and that NPO consistently moves models away from the training works. On the forget split, the English and Dutch fine-tuned models have average ranks of approximately \(1.85\) and \(2.00\), respectively, whereas the corresponding NPO models have average ranks of about \(3.04\) and \(3.12\). On the retain split, the fine-tuned models again occupy the top positions (around \(1.77\)–\(1.92\)), with both NPO variants clustered near rank \(3.0\) or higher. The rank differences between English and Dutch within the same training regime are modest compared to the differences between fine-tuned and unlearned models.

Category-level means (Online Resource~22) give additional context to the figures reported in the main text. On the forget split, fine-tuned models in both languages show category means above \(0.94\) for computational, stylistic and content metrics, while exception means are substantially lower (between roughly \(0.25\) and \(0.37\)). After NPO, forget-set means fall to about \(0.11\)–\(0.13\) for computational metrics and to the \(0.23\)–\(0.31\) range for stylistic and content metrics, whereas exception means remain in the \(0.24\)–\(0.28\) range. On the retain split, the fine-tuned models again have category means close to one for computational, stylistic and content dimensions, while exceptions sit around \(0.32\). The unlearned models keep retain-set means high for stylistic and content categories (roughly \(0.83\)–\(0.86\) for English and \(0.79\)–\(0.79\) for Dutch), whereas computational retain means drop to approximately \(0.69\)–\(0.71\). Exception means on the retain split increase slightly for both languages after NPO (to around \(0.39\)–\(0.43\)), but remain well below the infringement-oriented categories.

Differential-effect contrasts between categories on the forget split (Online Resource~23) show that, in both languages, reductions are large and similar for computational, stylistic and content categories, whereas exception metrics change substantially less. For English, the mean reduction in the computational category exceeds that in the content category by about \(0.16\) (paired \(t \approx 3.29\), \(p \approx 0.017\)), with no meaningful difference between stylistic and content reductions. For Dutch, the computational category is reduced more than content by approximately \(0.21\), with a highly significant contrast (\(t \approx 7.15\), \(p < 10^{-4}\)), while stylistic and content reductions are statistically indistinguishable. In both languages, all three infringement-oriented categories are reduced markedly more than exceptions.

\subsection{Pairwise comparisons and effect sizes}

Wilcoxon signed-rank tests comparing fine-tuned and NPO models on the same prompts (Online Resources~24 and~25) refine the picture at the metric level. On the forget split, NPO produces statistically significant changes for every computational, stylistic and content metric in both languages (all \(p < 10^{-6}\)), with paired Cohen's \(d\) values well above \(1.5\). Among the exception metrics, Quotation/Citation and Scènes à Faire show significant changes on the forget split, whereas Parody/Satire and Pastiche do not. On the retain split, English models exhibit significant changes for all computational, stylistic and content metrics except Character Similarity, and for Parody/Satire and Pastiche; however, Quotation/Citation and Scènes à Faire on the English retain split show no statistically significant change. For Dutch retain data, only the three computational metrics differ significantly between fine-tuned and NPO models; all stylistic, content and exception metrics have Wilcoxon \(p\)-values above \(0.09\), indicating that the Dutch unlearning configuration leaves most retain-set behaviour statistically similar to the fine-tuned state.

Effect-size summaries from independent-sample comparisons (Online Resources~26 and~27) underline the magnitude of unlearning on the forget sets. For computational metrics, Cohen's \(d\) comparing fine-tuned and NPO models on the forget split ranges from about \(4.1\) (Exact Match, Dutch) to over \(11\) (BLEU, English), with Cliff's \(\delta\) at or near \(\pm 1\), indicating almost complete separation of the distributions. Stylistic and content metrics on the forget split also show very large effects, with Cohen's \(d\) typically between \(2.0\) and \(5.5\). Exception metrics exhibit smaller and more variable effects: Quotation/Citation differences on the forget split reach \(d\) values between roughly \(1.35\) and \(2.07\), while Scènes à Faire and Pastiche are associated with \(d\) values around \(0.3\)–\(0.7\), and Parody/Satire effects are close to zero or modest in magnitude.

On the retain split, effect sizes are more moderate. For English models, Cohen's \(d\) comparing fine-tuned and NPO models is around \(1.2\) for Exact Match and approximates \(1.1\) for BLEU and ROUGE, and sits near \(1.0\) for Writing Style and Scene Sequence Similarity. For Dutch retain data, effect sizes for computational metrics are again large (around \(1.3\) for Exact Match and approximately \(0.9\)–\(1.0\) for BLEU and ROUGE), whereas stylistic and content metrics generally show medium or small effects, and many exception metrics produce negligible effect sizes. These patterns align with the Wilcoxon results: computational metrics are altered substantially on both splits, while many non-computational metrics on the retain split change only modestly or not at all, especially for Dutch.

\subsection{Equivalence to baseline models}

Two one-sided tests for equivalence (TOST) between baseline and NPO models on the retain split (Online Resource~28) assess whether unlearning returns the models to a state practically indistinguishable from the original instruction-tuned models. Equivalence bounds are set at \(\pm 0.1\) on the normalised \(0\)–\(1\) scale, corresponding to roughly half a PSALM category. For all computational, stylistic and content metrics in both languages, the combined TOST \(p\)-values exceed \(0.05\), and equivalence is therefore rejected. For example, the English Plot Structure Similarity mean drops from \(0.98\) in the baseline to \(0.79\) in the unlearned model, with a TOST \(p\)-value of about \(0.94\); similarly, the Dutch World-Building Similarity mean moves from \(0.99\) to \(0.89\), with a TOST \(p\) around \(0.54\). Even in cases where numerical differences are smaller, such as Dutch Narrative Voice on the retain split (\(1.0\) vs. \(0.979\)), the tests do not establish equivalence under the chosen bounds.

For exception metrics, the picture is mixed. Some comparisons, such as English Quotation/Citation and both languages' Scènes à Faire on the retain split, also fail the equivalence tests, despite relatively modest mean differences. Others, particularly where baseline and unlearned means are very close, produce TOST \(p\)-values that are not directly informative because of the small sample sizes and high variance. Overall, across all metrics evaluated, none of the tested retain-set comparisons between baseline and NPO models meet the predefined equivalence criterion, supporting the conclusion in subsection~\ref{sec:rq3:residual} that unlearning does not restore models to their pre-fine-tuning similarity levels.

\end{appendices}

\end{document}